\pdfoutput=1 %
\documentclass{mvdocument}

\usepackage{pdfpages}
\usepackage{hypcap}

\usepackage[accsupp]{axessibility}
\usepackage{times}
\usepackage{epsfig}
\usepackage{amsmath}
\usepackage{amssymb}
\usepackage{gensymb}

\usepackage{mathtools}  
\usepackage{url}            
\usepackage{booktabs}       
\usepackage{amsfonts}       
\usepackage{nicefrac}       
\usepackage{wrapfig}
\usepackage{bm}

\usepackage{subcaption} 
\usepackage{bbm}
\usepackage{multirow}
\usepackage{makecell}

\usepackage[frozencache,cachedir=.]{minted}
\usepackage[minted,most]{tcolorbox}
\usepackage{xcolor} 
\definecolor{LightGray}{rgb}{1,0.98,0.85}
\usepackage{listings}
\usepackage[T1]{fontenc} 

\newcommand{\glsshort}[1]{%
  \glslink{#1}{\glsentryname{#1}}%
}

\newcommand{\python}[3]{
\noindent\rule{\linewidth}{0.3pt}
\vspace{-2.25em}
\inputminted[linenos, breaklines, fontsize=\footnotesize, numbersep=5pt, fontsize=\scriptsize, xleftmargin=12pt]{python}{#1}
\vspace{-1.75em}
\noindent\rule{\linewidth}{0.3pt}

\noindent\begin{minipage}{\linewidth}
\captionof{listing}{#3}
\label{#2}
\end{minipage}
\vspace{-8pt}
}

\newcommand{\prompt}[3]{
\noindent\rule{\linewidth}{0.3pt}
\vspace{-2.25em}
\inputminted[linenos, breaklines, fontsize=\footnotesize, numbersep=5pt, fontsize=\scriptsize, xleftmargin=15pt]{text}{#1}
\vspace{-1.75em}
\noindent\rule{\linewidth}{0.3pt}

\noindent\begin{minipage}{\linewidth}
\captionof{listing}{#3}
\label{#2}
\end{minipage}
\vspace{-8pt}
}

\newcommand{\Afinished}{D}  

\newcommand{\Ssuccess}{W}  
\newcommand{\Sincorrect}{F} 
\newcommand{\Rincorrect}{R_F}  

\newcommand{\tpm}[1]{\tiny$\pm$#1}

\newcommand{\Apred}{\widehat{A}}
\newcommand{\Ppred}{\widehat{P}}
\newcommand{\Rpred}{\widehat{R}}

\newcommand{\Apredfunc}{\Apred(s,a)}
\newcommand{\Ppredfunc}{\Ppred(s'-s|a)}
\newcommand{\Rpredfunc}{\Rpred(a, s'-s)}
\newcommand{\Rpredincorrect}{\widehat{R}_F}

\newcommand{\Apredvalid}{\Apred_\text{logit}}  

\usepackage[english]{babel}
\usepackage[utf8]{inputenc}
\makeglossaries
\newglossaryentry{mdp}{
    name={MDP},
    description={Markov Decision Process},
    first={Markov Decision Process (MDP)},
    firstplural={Markov Decision Processes (MDPs)},
    plural=MDPs
}

\newglossaryentry{pomdp}{
    name={POMDP},
    description={Partially Observable Markov Decision Process},
    first={Partially Observable Markov Decision Process (POMDP)},
    firstplural={Partially Observable Markov Decision Processes (POMDPs)},
    plural=POMDPs
}

\newglossaryentry{il}{
    name={IL},
    description={Imitation Learning},
    first={Imitation Learning (IL)}
}

\newglossaryentry{rl}{
    name={RL},
    description={Reinforcement Learning},
    first={Reinforcement Learning (RL)}
}

\newglossaryentry{bc}{
    name={BC},
    description={Behavioural Cloning},
    first={Behavioural Cloning (BC)}
}

\newglossaryentry{irl}{
    name={IRL},
    description={Inverse Reinforcement Learning},
    first={Inverse Reinforcement Learning (IRL)}
}

\newglossaryentry{hrl}{
    name={HRL},
    description={Hierarchical Reinforcement Learning},
    first={Hierarchical Reinforcement Learning (HRL)}
}

\newglossaryentry{gail}{
    name={GAIL},
    description={Generative Adversarial Imitation Learning},
    first={Generative Adversarial Imitation Learning (GAIL)}
}

\newglossaryentry{meta-rl}{
    name={Meta-RL},
    description={Meta Reinforcement Learning},
    first={Meta Reinforcement Learning (Meta-RL)}
}

\newglossaryentry{vin}{
    name={VIN},
    description={Value Iteration Network},
    first={Value Iteration Network (VIN)},
    firstplural={Value Iteration Networks (VINs)},
    plural=VINs
}

\newglossaryentry{llm}{
    name={LLM},
    description={Large Language Model},
    first={Large Language Model (LLM)},
    firstplural={Large Language Models (LLMs)},
    plural=LLMs
}

\newglossaryentry{em}{
    name={EM},
    description={Expectation Maximisation},
    first={Expectation Maximisation (EM)},
}

\newglossaryentry{hmm}{
    name={HMM},
    description={Hidden Markov Model},
    first={Hidden Markov Model (HMM)},
    firstplural={Hidden Markov Model (HMMs)},
    plural=HMMs
}

\newglossaryentry{ppo}{
    name={PPO},
    description={Proximal Policy Optimisation},
    first={Proximal Policy Optimisation (PPO)},
}

\newglossaryentry{calvin}{
    name={CALVIN},
    description={Collision Avoidance Long-term Value Iteration Network},
    first={Collision Avoidance Long-term Value Iteration Network (CALVIN)},
}

\newglossaryentry{lstm}{
    name={LSTM},
    description={Long Short-Term Memory},
    first={Long Short-Term Memory (LSTM)},
    plural=LSTMs
}

\newglossaryentry{gae}{
    name={GAE},
    description={Generalised Advantage Estimate},
    first={Generalised Advantage Estimate (GAE)}
}

\newglossaryentry{goa}{
    name={GOA},
    description={Generalised Option Advantage},
    first={Generalised Option Advantage (GOA)}
}

\newglossaryentry{td}{
    name={TD},
    description={Temporal Difference},
    first={Temporal Difference (TD)}
}

\newglossaryentry{gru}{
    name={GRU},
    description={Gated Recurrent Unit},
    first={Gated Recurrent Unit (GRU)},
    firstplural={Gated Recurrent Units (GRUs)},
    plural=GRUs
}

\newglossaryentry{rnn}{
    name={RNN},
    description={Recurrent Neural Network},
    first={Recurrent Neural Network (RNN)},
    firstplural={Recurrent Neural Networks (RNNs)},
    plural=RNNs
}

\newglossaryentry{cnn}{
    name={CNN},
    description={Convolutional Neural Network},
    first={Convolutional Neural Network (CNN)},
    firstplural={Convolutional Neural Networks (CNNs)},
    plural=CNNs
}

\newglossaryentry{gpu}{
    name={GPU},
    description={Graphical Processing Unit},
    first={Graphical Processing Unit (GPU)},
    firstplural={Graphical Processing Units (GPUs)},
    plural=GPUs
}

\newglossaryentry{ppoem}{
    name={PPOEM},
    description={Proximal Policy Optimisation via Expectation Maximisation},
    first={Proximal Policy Optimisation via Expectation Maximisation (PPOEM)},
}

\newglossaryentry{soap}{
    name={SOAP},
    description={Sequential Option Advantage Propagation},
    first={Sequential Option Advantage Propagation (SOAP)},
}

\newglossaryentry{ppoc}{
    name={PPOC},
    description={Proximal Policy Option-Critic},
    first={Proximal Policy Option-Critic (PPOC)},
}

\newglossaryentry{ale}{
    name={ALE},
    description={Arcade Learning Environment},
    first={Arcade Learning Environment (ALE)},
}

\newglossaryentry{ppo_lstm}{
    name={PPO-LSTM},
    description={Proximal Policy Optimisation with Long Short-Term Memory},
    first={Proximal Policy Optimisation with Long Short-Term Memory (PPO-LSTM)},
}

\newglossaryentry{ai}{
    name={AI},
    description={Artificial Intelligence},
    first={Artificial Intelligence (AI)},
}

\newglossaryentry{vi}{
    name={VI},
    description={Value Iteration},
    first={Value Iteration (VI)},
}

\newglossaryentry{mcts}{
    name={MCTS},
    description={Monte Carlo Tree Search},
    first={Monte Carlo Tree Search (MCTS)}
}

\newglossaryentry{idm}{
    name={IDM},
    description={Inverse Dynamics Model},
    first={Inverse Dynamics Model (IDM)},
}

\newglossaryentry{vpt}{
    name={VPT},
    description={Video PreTraining},
    first={Video PreTraining (VPT)},
}

\newglossaryentry{prm}{
    name={PRM},
    description={Probabilistic Roadmap},
    first={Probabilistic Roadmap (PRM)},
    firstplural={Probabilistic Roadmaps (PRM)},
    plural={PRM}
}

\newglossaryentry{rrt}{
    name={RRT},
    description={Probabilistic Roadmap},
    first={Rapidly-exploring Random Tree (RRT)},
    firstplural={Rapidly-exploring Random Trees (RRT)},
    plural={RRT}
}

\newglossaryentry{gan}{
    name={GAN},
    description={Generative Adversarial Network},
    first={Generative Adversarial Network (GAN)},
    firstplural={Generative Adversarial Networks (GANs)},
    plural={GANs}
}

\newglossaryentry{a2c}{
    name={A2C},
    description={Advantage Actor-Critic},
    first={Advantage Actor-Critic (A2C)},
}

\newglossaryentry{trpo}{
    name={TRPO},
    description={Trust Region Policy Optimisation},
    first={Trust Region Policy Optimisation (TRPO)},
}

\newglossaryentry{kl}{
    name={KL},
    description={Kullback–Leibler},
    first={Kullback–Leibler (KL)},
}

\newglossaryentry{ddpg}{
    name={DDPG},
    description={Deep Deterministic Policy Gradients},
    first={Deep Deterministic Policy Gradients (DDPG)},
}

\newglossaryentry{td3}{
    name={TD3},
    description={Twin-Delayed Deep Deterministic Policy Gradients},
    first={Twin-Delayed DDPG (TD3)},
}

\newglossaryentry{sac}{
    name={SAC},
    description={Soft Actor-Critic},
    first={Soft Actor-Critic (SAC)},
}

\newglossaryentry{vae}{
    name={VAE},
    description={Variational Auto-Encoder},
    first={Variational Auto-Encoder (VAE)},
    firstplural={Variational Auto-Encoders (VAEs)},
    plural={VAEs}
}

\newglossaryentry{slam}{
    name={SLAM},
    description={Simultaneous Localisation and Mapping},
    first={Simultaneous Localisation and Mapping (SLAM)},
}

\newglossaryentry{vlm}{
    name={VLM},
    description={Vision-Language Model},
    first={Vision-Language Model (VLM)},
}

\newglossaryentry{avd}{
    name={AVD},
    description={Active Vision Dataset},
    first={Active Vision Dataset (AVD)},
}

\newglossaryentry{cmp}{
    name={CMP},
    description={Cognitive Mapper and Planner},
    first={Cognitive Mapper and Planner (CMP)},
}

\newglossaryentry{gppn}{
    name={GPPN},
    description={Gated Path Planning Network},
    first={Gated Path Planning Network (GPPN)},
    firstplural={Gated Path Planning Networks (GPPNs)},
    plural={GPPNs}
}

\newglossaryentry{lpn}{
    name={LPN},
    description={Lattice PointNet},
    first={Lattice PointNet (LPN)},
}

\newglossaryentry{relu}{
    name={ReLU},
    description={Rectified Linear Unit},
    first={Rectified Linear Unit (ReLU)},
}

\newglossaryentry{gpt}{
    name={GPT},
    description={Generative Pre-trained Transformer},
    first={Generative Pre-trained Transformer (GPT)},
}

\newglossaryentry{ucb}{
    name={UCB},
    description={Upper Confidence Bound},
    first={Upper Confidence Bound (UCB)},
}

\newglossaryentry{dqn}{
    name={DQN},
    description={Deep Q-Network},
    first={Deep Q-Network (DQN)},
}

\newglossaryentry{ppg}{
    name={PPG},
    description={Phasic Policy Gradient},
    first={Phasic Policy Gradient (PPG)},
}

\newglossaryentry{iic}{
    name={IIC},
    description={Invariant Information Clustering},
    first={Invariant Information Clustering (IIC)},
}

\newglossaryentry{awm}{
    name={AWM},
    description={Abstract World Model},
    first={Abstract World Model (AWM)},
}

\newglossaryentry{dag}{
    name={DAG},
    description={Directed Acyclic Graph},
    first={Directed Acyclic Graph (DAG)},
}

\newglossaryentry{dac}{
    name={DAC},
    description={Double Actor-Critic},
    first={Double Actor-Critic (DAC)},
}

\newglossaryentry{smdp}{
    name={Semi-MDP},
    description={Semi-Markov Decision Process},
    first={Semi-Markov Decision Process (Semi-MDP)},
    firstplural={Semi-Markov Decision Processes (MDPs)},
    plural=Semi-MDPs    
}

\newglossaryentry{ml}{
    name={ML},
    description={Machine Learning},
    first={Machine Learning (ML)},
}

\title      {Spatial Reasoning and Planning \\for Deep Embodied Agents}
\author     {Shu Ishida}
\college    {Christ Church}
\universitylogo{oxford-logo}       
\supervisor {Dr. Jo\~ao F. Henriques}

\degreename {A thesis submitted for the degree of \\ \textit{Doctor of Philosophy}}

\labname {Visual Geometry Group}
\department {Department of Engineering Science}
\university {University of Oxford}
\degreedate {Hilary 2024}

\newcommand{\ie}{i.e. }
\newcommand{\eg}{e.g. }

\newcommand{\etal}{et al. }
\newcommand{\etals}{et al.'s}
\newcommand{\defeq}{\vcentcolon=}
\DeclareMathOperator*{\expect}{\mathbb{E}}
\DeclareMathOperator*{\argmax}{argmax}

\DeclareMathOperator*{\avg}{avg}
\preto\chapter{\glsresetall}

\begin{document}

\frontmatter

\maketitle

\begin{abstract}
    Humans can perform complex tasks with long-term objectives by planning, reasoning, and forecasting outcomes of actions. For embodied agents (e.g. robots) to achieve similar capabilities, they must gain knowledge of the environment transferable to novel scenarios with a limited budget of additional trial and error. 
    Learning-based approaches, such as deep reinforcement learning, can discover and take advantage of inherent regularities and characteristics of the application domain from data, and continuously improve their performances, however at a cost of large amounts of training data. 
    This thesis explores the development of data-driven techniques for spatial reasoning and planning tasks, focusing on enhancing learning efficiency, interpretability, and transferability across novel scenarios. 
    
    Four key contributions are made. Firstly, \glsshort{calvin}, a differential planner that learns interpretable models of the world for long-term planning. It successfully navigated partially observable 3D environments, such as mazes and indoor rooms, by learning the rewards (goals and obstacles) and state transitions (robot dynamics) from expert demonstrations.
    
    Secondly, \glsshort{soap}, a reinforcement learning algorithm that discovers macro-actions (options) unsupervised for long-horizon tasks. Options segment a task into subtasks and enable consistent execution of the subtask. \glsshort{soap} showed robust performances on history-conditional corridor tasks as well as classical benchmarks such as Atari.
    
    Thirdly, LangProp, a code optimisation framework using Large Language Models to solve embodied agent problems that require reasoning by treating code as learnable policies. The framework successfully generated interpretable code with comparable or superior performance to human-written experts in the CARLA autonomous driving benchmark. 
    
    Finally, Voggite, an embodied agent with a vision-to-action transformer backend that solves complex tasks in Minecraft. It achieved third place in the MineRL BASALT Competition by identifying action triggers to segment tasks into multiple stages. 
    
    These advancements provide new avenues for applications of learning-based methods in complex spatial reasoning and planning challenges.
\end{abstract}

{
  \vspace{5pt}
  \small	
  \textbf{\textit{Keywords~---}} machine learning, neural networks, deep reinforcement learning, imitation learning, hierarchical reinforcement learning, policy optimisation, robotics, autonomous driving, embodied agents, option discovery, skill learning, navigation, planning, computer vision, large language models, multi-modal foundation models.
} 
\begin{declaration}
This thesis is submitted to the Department of Engineering Science, University of Oxford, in fulfilment of the requirements for the degree of Doctor of Philosophy.
I declare that this thesis is entirely my own work and, except where stated, describes my own research.

\bigbreak
\bigbreak

\noindent Shu Ishida

\noindent Christ Church

\end{declaration}

\begin{acknowledgements}

Words cannot fully express my sincere gratitude towards my supervisor, Dr. João F. Henriques, for his support and guidance throughout my journey as a DPhil student. He has never hesitated to share with me his wisdom and insight on a wide variety of disciplines, including very helpful advice on programming, debugging, visualisation, and mathematical techniques. I have always admired his exceptional intelligence, hands-on skills, deep conceptual understanding, and attention to technical details, as well as his humility and approachability. 
He has always believed in me, even in times when I lost confidence in myself. I am immensely grateful for João's unwavering trust and encouragement, and for this fortunate opportunity of working with him throughout my DPhil research.

I would like to thank the other Principal Investigators at the Visual Geometry Group (VGG) --- Prof. Andrew Zisserman, Prof. Andrea Vedaldi, Dr. Christian Rupprecht, and Dr. Iro Laina --- for making the lab a vibrant place for research. I would also like to thank all my fellow and former research students in the lab who helped make the lab a welcoming and friendly environment. A special thanks to the following people: Max Bain, who introduced me to my supervisor João and encouraged me to join VGG; Oliver Groth, Gül Varol, and João for organising the Big Picture Debates online during the COVID-19 lockdown when I was joining the lab, which helped me virtually meet people and have meaningful discussions with them during those challenging times; Shangzhe (Elliot) Wu, Luke Melas-Kyriazi, Paul Engstler, Ragav Sachdeva, Vladimir Iashin along with many others who have regularly organised lab socials, lunches and dinners for us.

This work has been made possible by the Autonomous Intelligent Machines and Systems Centre for Doctoral Training programme (AIMS CDT), funded by EPSRC (Engineering and Physical Research Council). I am extremely grateful for having Mrs. Wendy Poole as our Centre Administrator of AIMS CDT, who has dedicated so much thought and effort into running the course smoothly, and providing us with the best experience that the CDT could offer. I deeply respect Wendy for her planning and organisational skills, and her prompt and accurate responses to enquiries. I have made valuable friends in the AIMS CDT throughout the years, thanks to the annual meetings and regular social events across cohorts. My thanks also go to Ivan Kiskin, Yuki Asano, and Robert McCraith for kindly offering their knowledge, wisdom and guidance as experienced members/alumni of the CDT.

I would like to thank Stuart Golodetz for supervising me for one of my AIMS CDT mini-projects in collaboration with Five AI. This was the first research collaboration I undertook, and it was a valuable experience working with him and learning from his insight and good coding practices. I would also like to thank the organisers of the MineRL BASALT Competition at NeurIPS 2022, in particular Anssi Kanervisto, who was incredibly helpful and patient with all my technical enquiries during the competition.

I thank all the people at Wayve Technologies, where I conducted a part of my research as an intern. In particular, I would like to thank Anthony Hu, who has been an amazing mentor and supervisor during my internship. I have learnt so much both from his research and team communication skills, as well as his kindness and care. He offered me substantial guidance when I was setting up my simulation and experiments, and have always provided me with academic and emotional support. I also thank Gianluca Corrado for his valuable and helpful advice as my manager, and George Fedoseev, Hudson Yeo, Lloyd Russell, and Corina Gurau for offering their insight, support and company during my internship in the world modelling team.

I would also like to thank my supervisors and colleagues at Microsoft Research Cambridge, where I spent the last few months of my DPhil degree as a research science intern in the Game Intelligence team. My special thanks go to Sergio Valcarcel Macua, Raluca Georgescu, Abdelhak Lemkhenter, and Tabish Rashid, who have spent a considerable amount of time mentoring me throughout the project, as well as to Katja Hofmann, Sam Devlin, Dave Bignell, Tim Pearce, Yuhan Cao, Shanzheng Tan, and Linda Yilin Wen. Everyone was always happy to help me with any issues with the code and infrastructure, and I really appreciated their inclusiveness in their activities and their generosity with their time. I would also like to express my immense gratitude towards my fellow forty interns, all of whom have made my internship experience special. Everyone has been very friendly, sociable and inclusive, and created a wholesome welcoming community. Special mention to Marko Tot, Chentian Jiang, Daniel Clark, Lucia Adams, and Dmitrii Usynin.

This research would not have been possible without the generous support by the Ezoe Memorial Recruit Foundation. I would like to express my sincere gratitude to the foundation for partially funding my undergraduate studies and fully funding my CDT and DPhil studies via the Recruit academic scholarship.

Over the course of my studies at Oxford, I had the privilege of being taught and advised by brilliant researchers, lecturers and supervisors. I would especially like to thank my undergraduate college tutor and postgraduate college advisor Prof. Malcolm McCulloch, my Master's thesis supervisor Prof. Nick Hawes, my thesis mentor Marc Rigter, and lectures on machine learning and optimisation Prof. Philip Torr and Dr. M. Pawan Kumar for their valuable teaching and insights, and for their kind and generous support of my research and career. I would also like to thank my thesis examinors, Prof. Ioannis Havoutis and Dr. Markus Wulfmeier, for their thorough and insightful feedback.

Through my extra-curricular activities, I had the privilege of working with research students with great leadership and skills. My thanks go to Jonas Beuchert and Augustinas Malinauskas for teaming up at the Oxford Hackathon, Lynn Hirose and Naho Tomiki for collaborating on the University of Tokyo Project Sprint, Mark Finean, Charlie Street, and Ricardo Cannizzaro for leading Team ORIon for the RoboCup competitions, and Timothy Seabrook, Ding Shin Huang, Chuxuan (Jessie) Jiang, and Siobhan Mackenzie Hall for leading initiatives at the Oxford Artificial Intelligence Society, as well as to all the team and committee members who made the initiatives possible. I would also like to thank Prof. Paul Newman, Prof. Nick Hawes, Prof. Ioannis Havoutis, Dr. Lars Kunze and Dr. Bruno Lacerda for their support for the RoboCup competition, as well as Dr. Natalia Efremova for her mentorship and insights for one of the OxAI Labs research projects.

I was fortunate to make many dear friends and meet countless kindhearted and wonderful people during my studies, for which I am immensely thankful. These people have helped shape who I am today, and I continue to learn from their philosophies. While I would not be able to express my gratitude in this limited space for everyone who deserves them, I would like to mention the following friends who have deeply impacted me and supported me.

My flatmates: Andreas Schmid, Richard Csaky, and Danilo Jr Dela Cruz --- my time in Oxford would not have been the same without the company of my amazing flatmates. I have learnt a lot from them, both in terms of work and life. I fondly remember our co-working days and board game nights, the bike trips with Andreas, the road trip with Richard, and many geeky and philosophical evening chats with Danilo.

My Christ Church friends and Japan Society friends --- having your company, enjoying meals, events, and celebrations together, and being part of the wonderful community were the things that made my life at Oxford special, and I cannot thank you all enough.

Kengo Shibata, Hiroto Takahashi, Fabrice Wunderlich, Zihan (Amanda) Zhu, Aya Fujita, Catherine Choi --- some of my DPhil student friends who have been with me throughout most of my journey and have helped create wonderful memories at Oxford. I cherish all the deep conversations we have had and the time we have spent together.

Bingqing Liu --- my study partner throughout my undergraduate degree. Thank you for always providing inspiration and motivation, taking the initiative to organise study sessions, and proactively finding events and competitions that we could participate in.

Nader Raafat, Megan Craig --- my friends who have supported me since my undergraduate days, during the lockdown and through thick and thin, and have continued to keep in touch with me despite the long distances. You have always been there for me, which means a fortune. 

Ximei Liu --- thank you for always believing in me and encouraging me. Your enthusiasm and courage are only outshone by your kindness and care, and I am blessed to have you in my life and share the adventures together. 

Last but not least, I would like to thank my mother, father, and grandparents for their unconditional love, and for always looking out for me. Knowing that they will be there for me and that I can count on their moral support has given me courage and strength, and is the reason why I can keep pursuing my goals and dreams. 

\end{acknowledgements}

\tableofcontents
\listoffigures
\listoftables
\listofabbreviations

\mainmatter


\chapter{Introduction}
\label{chapter:introduction}

\section{Motivation}
%

The ability to plan, reason and forecast outcomes of actions, even in novel environments, are remarkable human capabilities that are instrumental in performing complex tasks with long-term objectives. Whenever we encounter a novel scenario, whether that be a new game, sport or location, even though we have never experienced that specific case, we can still strategise by extrapolating from our prior experiences, taking advantage of transferable knowledge and skills. 

With modern planning algorithms, it is possible to find a near-optimal solution to a planning problem, if the environment dynamics (specifically the state transition and reward dynamics) are fully known, the states and actions can be enumerated, and unlimited compute is available. Unfortunately, it is often the case that all three of the assumptions do not hold. The agent often only has access to a local or partial observation of the environment, and has to estimate the underlying environment state and dynamics based on this. States and actions are often continuous rather than discrete, so an estimator is required that can map continuous inputs to meaningful representations that generalise to novel inputs. Finally, since compute is finite and enumeration of states and actions is often infeasible, an efficient strategy is necessary to explore the state-action space within limited computational resources and agent lifetime.

Many real-world problems involving strategic decision making require the agent to learn transferable knowledge of the environment that can be applied to novel scenarios with a limited budget of additional trial and error. 
Conceiving an algorithm that achieves the same level of performance and efficiency as humans in the open domain remains an open question. Autonomous driving~\cite{yurtsever2020survey_autonomous}, for instance, is still an ongoing and unsolved area of research, due to the high complexity of the dynamic environment in a multi-agent problem setting, together with the challenge of imperfect information and noisy sensor inputs. This is in stark contrast to industrial robots which have been in effective operation for many preceding decades, helped by the fact that the environment is controlled, predictable, and in many cases fully known. Combined with the repetitiveness of the task, this allows humans to hard-code the system to handle commonly anticipated scenarios. 

\gls{mdp} and \gls{rl} are powerful frameworks that formulate decision-making as a learnable problem with a mathematically defined objective~\cite{sutton_reinforcement_2018}. These frameworks capture the sequential and time-evolving nature of interacting with an environment.

Advances in neural networks and their successful integration into \gls{rl}~\cite{mnih_playing_2013,mnih_human-level_2015,silver_mastering_2016} have transformed the field of computer vision and robotics, giving rise to learning-based approaches for problems traditionally solved by humans manually implementing expert systems. Learning-based approaches have two major advantages. Firstly, learning-based algorithms can keep improving and adapting to the application domain with more availability of data, whereas manually implemented methods are fixed and do not learn to adapt. Secondly, learning-based methods are capable of automatically discovering inherent regularities and characteristics of the application domain and exploiting them to improve their performances without having such strategies hardcoded. 

While \gls{rl} is highly effective in solving complex strategic problems~\cite{mnih_playing_2013,silver2017mastering_alphazero,vinyals_grandmaster_2019,badia_agent57_2020,baker2022_openai_vpt}, sample efficiency and generalisability are challenges that still need to be addressed. Current state-of-the-art \gls{rl} algorithms are highly performant in tasks they have been trained on or can solve with a reactive policy, but do not explicitly learn easily transferable skills~\cite{ase_large_scale_reusable,pertsch2020spirl,nam2022skillbased,raolearning2022,shi2023skill}.
Unlike games or tasks in a simulation where samples can be drawn easily, collecting samples can be expensive and possibly unsafe in real-world problems. Humans can work around these issues by learning transferable knowledge and skills that can applied to novel situations, thereby increasing the chances of success with less trial-and-error and avoiding catastrophic failure, \eg falling off a cliff or being run over by a car. This research aims to suggest ways of acquiring skills that allow agents to learn to perform tasks more efficiently and effectively.

\section{Research objectives}
\label{sec:intro/objectives}

This research addresses the challenge of solving tasks involving spatial reasoning, planning, and decision making in a data-driven manner, while simultaneously making the learning more efficient, interpretable and transferable. This research objective can be further broken down into five research goals, which are described in detail in the following.

\subsection{Learn a generalisable planner}
One of the core objectives of this research is to develop learnable planners that generalise to novel scenarios. A distinction between a \emph{reactive} Markovian policy and a policy with a \emph{plan} is that a reactive policy makes immediate decisions given a current state or local observation, whereas planning involves a more long-term analysis of the given situation to propose a spatially and temporally coherent solution. 

The differences between the two approaches are analogous to the System 1 (fast, unconscious, and automatic decisions) and System 2 (slow, conscious, and rigorous decisions) thinking presented in \cite{Kahneman11_thinking_fast_slow}. Both decision processes are important, since reactive policies are useful for making many decisions in real-time, whereas planning is important to ensure that the decisions made are consistent and coherent. 
For example, \gls{mcts}-based algorithms~\cite{silver_mastering_2016,silver2017mastering_alphazero} alternate between learning a reactive policy and using them for long-term planning; rollouts of a Monte Carlo tree~\cite{coulom2006efficient_mcts} are simulated and the return estimates are back-propagated using a light-weight reactive policy, which is then updated according to the rollout results. 

While the dynamics of games such as Go and simulated environments are known, this is not the case for many real-world problems. Model-based \gls{rl} approaches~\cite{ha2018worldmodels,hafner2019dreamer,schrittwieser2020mastering_muzero} address this by learning models of the environment that could be used for simulated rollouts. \Cref{chapter:calvin} explores related alternative avenues to learn a differentiable planner that solves navigation tasks in novel environments that are not effectively solvable with reactive policies. \Cref{chapter:langprop} proposes a novel paradigm of learning algorithmic decision-making from data by treating code as learnable policies with the use of \glspl{llm}. By making algorithms learnable, high-level and long-term plans that were hitherto too complex for \gls{rl} agents to learn can now be learnt using \gls{il} and \gls{rl} techniques. In addition, \Cref{chapter:ppoem} and \Cref{chapter:minecraft} demonstrate how temporal abstraction using options~\cite{sutton1999between_options, Precup2000TemporalAI} can help agents make informed long-term decisions, discussed in \Cref{sec:intro/objective/reusable_skills} and \Cref{intro/objective/pomdp_policies}. 

\subsection{Discover reusable skills}
\label{sec:intro/objective/reusable_skills}
Skill learning is another important component for efficient exploration, decision making and task-solving. With skills, it is possible to conceive a high-level plan that combines and orchestrates low-level skill policies. These skills are specialised to solve a subset of the task so that the agent may learn to solve complex novel tasks from fewer training samples by composing these skills together. Ways in which these skills can be learnt in an unsupervised way, using rewards from the environment as a learning signal, are explored in \Cref{chapter:ppoem}. The agent trajectory is segmented into \emph{options}~\cite{sutton1999between_options, Precup2000TemporalAI} that correspond to skill-specific sub-policies. 

\subsection{Solve POMDP environments with memory-augmented policies}
\label{intro/objective/pomdp_policies}
In relation to \Cref{sec:intro/objective/reusable_skills}, options can be used not just to learn skills, but also to learn temporally consistent behaviour. It functions as a memory carried forward as a discrete latent variable, allowing the agent to perform tasks in a \gls{pomdp} environment where the underlying state of the environment cannot be determined from current observations alone. The true environment state can be better determined by maintaining a history of the agent trajectory, since past observations are often correlated with future observations by hidden variables. \Cref{chapter:ppoem} examines the effectiveness and robustness of options discovered by algorithms with different training objectives, demonstrating the advantage of the proposed solution over classical recurrent policies and Option-Critic policies~\cite{optioncritic,optioncritic_ppo}.

In \Cref{chapter:minecraft}, the concepts of skills and trajectory segmentation are employed to make the agent change its policy for different stages of task completion. Breaking a complex task down to subcomponents and performing them stage-wise allowed the agent to perform temporally consistent behaviour that adheres to a high-level plan. 

\subsection{Explain the behaviour of experts and agents}
Another theme explored in this research is the explainability of the learnt policy. Skill learning discussed above is one approach that ensures better explainability, given the options segment the agent trajectory in a semantically interpretable way. Another approach to interpretability is explored in \Cref{chapter:calvin}; a differentiable planner learns targets, obstacles, and motion dynamics from expert trajectories of robot navigation. It also computes a reward map and value map during its decision-making process, similarly to \gls{irl}~\cite{ng_algorithms_2000,ziebart_maximum_2008,ziebart_modeling_2010,arora2021survey}. An even more explicit way of representing the policy as human-readable code is suggested in \Cref{chapter:langprop}. Performance issues of the policies can be directly diagnosed by reading the code, making this approach a valuable technique in explainable \gls{ai} research.

\subsection{Train embodied agents to perform complex tasks}

Finally, the aim of this research is to apply developed techniques to problems relevant to embodied agents, e.g. robotics. In \Cref{chapter:calvin}, \Cref{chapter:langprop} and \Cref{chapter:minecraft}, the challenges of robot navigation, autonomous driving and task execution in the virtual world of Minecraft~\cite{minecraft} are addressed. These challenges all have navigation and spatial reasoning as key elements to accomplishing the tasks. 
Navigation is a real-world problem that has traditionally been solved by expert-designed systems, but could be made more efficient by leveraging data-driven learning. 
For instance, lane changing and cooperation with other vehicles are tasks for autonomous vehicles which require complex planning. The problem is made especially difficult, since human cooperative behaviour is difficult to model due to compounding factors and subtle cues, and there is not always a deterministic strategy to follow. Learning cooperative behaviour from real-world data could be beneficial to optimising these tasks. 

\section{Main contributions}
\label{sec:intro/contributions}

The contributions in this thesis can be summarised as follows.

\begin{enumerate}
  \item Developed a differentiable planner named \gls{calvin}, which learns to navigate unseen 3D environments by performing differentiable value iteration. State transitions and reward models are learnt from expert demonstrations, similarly to \gls{vin}. However, \gls{vin} struggles to penalise invalid actions leading to collisions with obstacles and walls, making the value estimate inaccurate. \gls{calvin} resolves this issue by learning action affordance to constrain the agent transitions and rewards. \gls{calvin} can navigate novel 2D and 3D environments and significantly outperforms other learnable planners based on \gls{vin}. Published at the IEEE/CVF Conference on Computer Vision and Pattern Recognition (CVPR) 2022~\cite{ishida2022towards}. Details are in \Cref{chapter:calvin}.
  
  \item Based on analysis of the Options Framework and the forward-backward algorithm~\cite{baum72}, algorithms were developed to learn temporally-consistent options and associated sub-policies in order to solve \gls{pomdp} tasks that require long-term memory.
  Two learning objectives for unsupervised option discovery were proposed and studied: \gls{ppoem} and \gls{soap}. 
  \gls{ppoem} applies the forward-backward algorithm~\cite{baum72} to optimise the expected returns for an option-augmented policy. However, it was shown that this learning approach is unstable for learning causal policies without the knowledge of future trajectories, since option assignments are optimised for the entire episode. 
  As an alternative approach, \gls{soap} evaluates the policy gradient for an optimal option assignment. 
  It extends the concept of the \gls{gae} to propagate \emph{option advantages} through time, which is an analytical equivalent to performing temporal back-propagation of option policy gradients. 
  With this approach, the option policy is only conditional on the history of the agent.
  Evaluated against competing baselines, \gls{soap} exhibited the most robust performance, correctly discovering options for \gls{pomdp} corridor environments, as well as on standard benchmarks including Atari~\cite{bellemare13arcade} and MuJoCo~\cite{todorov2012mujoco}. The paper is available on \textit{arXiv}~\cite{ishida2024soaprl}. Details are in \Cref{chapter:ppoem}. 

  \item Proposed LangProp. a framework for iteratively optimising code generated by \glspl{llm}. LangProp automatically evaluates the code performance on a dataset of input-output pairs, catches any exceptions, and feeds the results back to the \gls{llm} in the training loop, so that the \gls{llm} can iteratively improve the code it generates. The LangProp training module can be used in both supervised and reinforcement learning settings. LangProp successfully solves Sudoku and CartPole, as well as generates driving code with comparable or superior performance to human-implemented expert systems in the CARLA driving benchmark~\cite{dosovitskiy17a_carla}. LangProp can generate interpretable and transparent policies that can be verified and improved in a metric- and data-driven way. Accepted at the International Conference on Learning Representations (ICLR) 2024 Workshop on \glsfirst{llm} Agents~\cite{ishida2024langprop}. This work was conducted during an internship at Wayve Technologies. Details are in \Cref{chapter:langprop}. 
  
  \item Developed Voggite, an embodied agent that performs tasks in Minecraft, an open-ended virtual world. As a backbone, Voggite uses OpenAI \gls{vpt}~\cite{baker2022_openai_vpt}, a transformer-based agent pre-trained on online videos labelled by a supervised \gls{idm}. The \gls{vpt} policy takes in $128$ frames of past observations, equivalent to $6.4~s$ of history. While effective for many reactive tasks, the \gls{vpt} agent struggles to disambiguate different stages of task execution. Voggite resolves this issue by dividing the task into separate stages. Voggite achieved 3rd place out of 63 teams in the MineRL BASALT Competition on Fine-Tuning from Human Feedback at NeurIPS 2022. In the competition, the agents are tasked to find caves and make waterfalls, farms and buildings in Minecraft. A co-authored retrospective of the competition is available on \textit{arXiv}~\cite{milani2023solving}. Details are in \Cref{chapter:minecraft}.
\end{enumerate}

Work not included in this thesis: ``You are what you eat? Feeding foundation models a regionally diverse food dataset of World Wide Dishes''~\cite{magomere2024worldwidedishes}.

\section{Outline}
\Cref{chapter:literature_review} introduces background material and key concepts relevant to this thesis. 
\Cref{chapter:calvin} proposes \gls{calvin}, a differentiable planner that learns to plan for long-term navigation in unseen 2D and 3D environments.
\Cref{chapter:ppoem} introduces \gls{ppoem} and \gls{soap} for unsupervised option discovery. 
\Cref{chapter:langprop} proposes LangProp, a framework for iteratively optimising code generated by \glspl{llm}.
\Cref{chapter:minecraft} discusses data-driven decision-making for embodied agents with composite tasks, and develops Voggite, an embodied agent that performs tasks in Minecraft.
Finally, \Cref{chapter:conclusion} concludes this thesis with discussions of findings and future research directions.

\chapter{Background on planning and data-driven decision making}
\label{chapter:literature_review}

\section{Planning algorithms}
Planning concerns strategically inducing changes to an environment state with the means of actions to achieve a certain objective~\cite{lavalle_planning_2006}. In robotics, motion and trajectory planning are essential to translating high-level specifications of tasks to low-level descriptions of motion. In the field of \gls{ai} and \gls{rl}, early successes were achieved in puzzle solving and competing in games~\cite{tesauro1994td_gammon,campbell2002deep,mnih_playing_2013,silver_mastering_2016}. 

While there are diverse ranges of planning problems, they share some commonalities. 
Firstly, they have a notion of a \emph{state} $s$ in \emph{state space} $\mathcal{S}$ that captures all details of a situation relevant to a particular problem. Secondly, the environment state can change over time, and the only way for the agent to influence its transition to another state is by taking an \emph{action} $a \in \mathcal{A}(s)$. The transition from the current state $s$ to the next state $s'$ is conditional on $a$. The plan is expected to meet some criteria and/or optimise a certain objective. In robotics, it is typical to formulate this as a path-finding problem to a target that minimises the total cost, with additional constraints so that illegal configurations and obstacles are not encountered. In \gls{rl}, the convention is to maximise the cumulative discounted rewards (returns), defined in \Cref{sec:background/mdp}.

\subsection{Markov Decision Process}
\label{sec:background/mdp}
\gls{mdp} \cite{bellman_markovian_1957,sutton_reinforcement_2018} is a standard formulation for sequential decision-making and planning. 
An \gls{mdp} consists of states $s \in \mathcal{S}$, actions available at each state $a \in \mathcal{A}(s)$, a transition probability $P(s'|s, a)$ (the probability next state $s'$ given the current state $s$ and action $a$), and a reward function $R(s,a,s')$, which (either deterministically or stochastically) determines the reward $r$ given to the agent during the transition.\footnote{It is common to assume discrete planning time steps denoted with $t$ (unitless integer), and denote the current state and next state as either $s_t$ and $s_{t+1}$, or with a shorthand $s$ and $s'$. The same convention applies to actions, rewards, etc. In this thesis, these two notations are used interchangeably.}
An agent starts at state $s_0 \sim \rho(s)$ and at each time step $t$, takes an action $a_t \in \mathcal{A}(s_t)$, collects a reward $r_t$ and moves to the next state $s_{t+1}$ where $(r_t, s_{t+1}) \sim P(\cdot | s_t, a_t)$ until episode termination at timestep $T$ when a boolean termination indicator $d_t$ is set to $1$.

In the \gls{rl} paradigm, it is useful to think of the decision-making component (the agent) separately from the environment. The environment has an inner state which defines its configuration, but the agent often can only measure the state indirectly through observations given by the environment. The agent can induce changes to the state by taking actions at every time step, and it receives a reward and a new observation. The objective is to learn a policy $\pi(a|s)$ that chooses an action that maximises the expected return at every time step $t$. 
A return is a sum of discounted rewards, $\mathcal{R}_t(\tau) = \sum_{t'=t}^{\infty}{\gamma^{t'-t} r_{t'}}$, where $\tau$ denotes the trajectory (a set of states, actions and rewards visited in an episode), and a discount factor $\gamma \in (0, 1]$ is applied to mitigate the infinite horizon problem encountered in continuous tasks, in which the sum of rewards can diverge to infinity.\footnote{Some literature uses $G_t$ to denote returns.}

\subsection{Partially Observable Markov Decision Process}
\gls{pomdp} is a special case of an \gls{mdp} where the observation available to the agent only contains partial information of the underlying state. 
In this thesis, $s$ is used to denote the (partial) state given to the agent, which may or may not contain the full information of the environment (which shall be distinguished from state $s$ as the underlying state~$\mathfrak{s}$).\footnote{In other literature, $o$ is used to denote the partial observation to distinguish from the underlying state $s$. While this makes the distinction explicit, many works on standard \gls{rl} algorithms assume a fully observable \gls{mdp} for their formulation, leading to conflicting notations.} This implies that the ``state'' transitions are no longer fully Markovian in a \gls{pomdp} setting, and may be correlated with past observations and actions. Hence, $p(r_t, s_{t+1} | s_{0:t}, a_{0:t})$ describes the full state and reward dynamics in the case of \glspl{pomdp}, where $s_{t_1:t_2}$ is a shorthand for $\{s_t | t_1 \leq t \leq t_2\}$, and similarly with $a_{t_1:t_2}$.

\subsection{Classical approaches to planning}
\label{sec:background/planning}

Before the rise of data-driven learning methods, the assumption in planning was that a model of the system or the environment is available. Planning in fully known state spaces with deterministic transitions has partially been solved by graph-search algorithms such as Dijkstra's algorithm~\cite{dijkstra} and the A* algorithm~\cite{a_star}, mainly in the domain of navigation~\cite{ishida2019_robot}.
However, many assumptions exist for such algorithms to work, such as that the environment is static, the state space is relatively low-dimensional, and the state space is enumerable. For instance, manipulation tasks are high dimensional and can be computationally expensive to solve with graph search algorithms.
The D* algorithm \cite{stentz_optimal_1994,stentz_focussed_1995,koenig_d_2002} mitigated the static environment assumption by incrementally replanning when the agent's knowledge of the environment is updated.
Sampling methods such as Probabilistic Roadmaps (PRM) \cite{prm} and Rapidly-exploring Random Trees (RRT) \cite{rrt1,rrt2} presented practical solutions that can work in high-dimensional state spaces, which are asymptotically optimal. 
For problems such as Chess and Go where the state space is large and expensive to perform brute-force search, \gls{mcts}~\cite{coulom2006efficient_mcts} is used to selectively explore states and evaluate the expected returns. 

For many real-world applications, however, the environment model cannot be accessed directly, and has to be inferred by interacting with the environment and collecting experiences. Moreover, the state transition and reward dynamics can be probabilistic. The need to accommodate these requirements gave rise to \gls{rl}.

\section{Reinforcement Learning}
\label{sec:background/rl}

\gls{rl} is a framework that formalises sequential decision-making as cumulative future reward maximisation in an \gls{mdp}~\citep{bellman_markovian_1957,sutton_reinforcement_2018}. \gls{rl} addresses problems that involve agents that interact with the environment to accomplish tasks when either the full \gls{mdp} is unknown or it is not feasible for classical planning algorithms~\cite{lavalle_planning_2006} to solve. 

The objective of an agent is to learn a policy $\pi(a_t|s_t)$ that maximises the expectation of return $\mathcal{R}_t$, where the policy specifies the probability of choosing action $a_t$ in state $s_t$. Approaches in \gls{rl} can be broadly classified into policy gradient methods (\Cref{sec:background/policy_gradient}) that directly optimises the policy based on on-policy feedback, value-based methods (\Cref{sec:background/value_methods}) that learn a separate value function to estimate the expectation of returns (can be either on-policy or off-policy), and Actor-Critic methods (\Cref{sec:background/actor_critic}) that combines the advantages of policy learning and value learning. The distinction between on-policy and off-policy methods is explained in \Cref{sec:background/on_off_policy}.

Advances in \gls{rl} and their effective integration with neural networks as high-dimensional policy and value function approximators have transformed the fields of computer vision, robotics and games. Deep \gls{rl} has outperformed humans in numerous strategic games~\cite{mnih_human-level_2015,silver_mastering_2016,vinyals_grandmaster_2019}.

\subsection{On-policy and off-policy}
\label{sec:background/on_off_policy}

An \gls{rl} agent needs to interact with the environment and collect experiences to learn about the \gls{mdp} (\emph{exploration}), while also trying to optimise the return (\emph{exploitation}). A recurring theme in \gls{rl} is to balance exploration and exploitation. This relates to an important distinction in \gls{rl} algorithms: \emph{on-policy} vs \emph{off-policy}. An algorithm is on-policy if it can only be trained on experiences generated by the current policy $\pi$. While this often results in a simpler algorithm, this implies that the learnt policy has to retain a level of stochasticity, since it has to explore and experience sub-optimal behaviour as well. Off-policy learning, enabled by techniques such as importance sampling, allows decoupling of the target policy that is optimised from the behaviour policy which the agent uses to collect experiences. This also allows training upon all experiences that are collected in the past, rather than clearing the rollout buffer after every policy update.

\subsection{Value-based methods}
\label{sec:background/value_methods}

A value function \cite{sutton_reinforcement_2018} is a fundamental concept in \gls{rl}, used to estimate the expected return following a policy $\pi$.
The (state-)value function $V^{\pi}(s) \doteq \expect_{\pi}[\mathcal{R}_t | s_t = s]$ evaluates the expected return starting from the current state $s$, while the action-value function $Q^{\pi}(s, a) \doteq \expect_{\pi}[\mathcal{R}_t | s_t = s, a_t = a]$ considers both the current state $s$ and the action to be taken $a$. 
An optimal policy $\pi^*$ is a policy that maximises the expected return for all states, i.e. $\forall s \in \mathcal{S}, V^*(s) = \max_{\pi}{V^{\pi}(s)}$. 
Computing a value function for a given policy is called policy evaluation, and improving the policy to obtain a better return is called policy improvement.

\subsubsection{Value Iteration}
\label{sec:background/value_iteration}
\gls{vi}~\cite{bellman_markovian_1957,sutton_reinforcement_2018} is an algorithm to obtain an optimal policy when the environment model (state transition $P(s'|s, a)$ and reward function $R(s, a, s')$) is known and the state-action space is relatively small and discrete so that they can be repeatedly enumerated (i.e. a tabular setting). It performs policy evaluation and policy improvement for each update step. Updates at step $k$ can be described as follows:
\begin{equation}
\begin{aligned}
\label{eq:background/value_iteration}
Q^{(k)}(s, a) &= \sum_{s'}{P(s'|s, a)[R(s, a, s') + \gamma V^{(k-1)}(s')]}, \qquad \forall a \in \mathcal{A}(s),\\
V^{(k)}(s) &= \max_{a \in \mathcal{A}(s)}{Q^{(k)}(s, a)},\\
\pi^{(k)}(s) &= \argmax_{a \in \mathcal{A}(s)}{Q^{(k)}(s, a)}.
\end{aligned}
\end{equation}

This update is repeated until the value function $V^{(k)}$ converges. The final policy $\pi$ is optimal, which means that the policy gives the maximum expected return.

\subsubsection{Value function approximation}

While \gls{vi} provides valuable insight into how value functions can be used to obtain an optimal policy, it is seldom the case that the full \gls{mdp} is known and fully expandable. Value-based methods aim to approximate this value function from experiences without having to know or expand the full \gls{mdp}.

Neural networks are often used as value function approximators. There are several approaches to computing the target value to regress towards. Monte-Carlo methods \cite{singh_monte_carlo_1996, sutton_reinforcement_2018} use the returns of fully rolled out episodes to estimate the expectation of returns (see \Cref{sec:background/monte_carlo}), whereas \gls{td} learning \cite{watkins_learning_1989, rummery_online_1994, sutton_reinforcement_2018} is at the opposite end of the spectrum, taking a bootstrapped approach with a one-step lookahead to estimate the values (see \Cref{sec:background/td_learning}). $n$-step \gls{td}~\cite{watkins_learning_1989} generalises \gls{td} learning to $n$-step lookahead (here, Monte Carlo methods can be considered as $\infty$-step \gls{td}), and TD($\lambda$)~\cite{sutton1988learning_td} takes an exponentially weighted average of $n$-step \gls{td} with varying $n$ (from $1$-step \gls{td} to a Monte Carlo estimate) to provide a return estimate with reduced variance.

\subsubsection{Monte Carlo methods}
\label{sec:background/monte_carlo}

Monte Carlo methods \cite{singh_monte_carlo_1996, sutton_reinforcement_2018} use the returns of the episode $\mathcal{R}_t(\tau) = \sum_{t'=t}^{\infty}{\gamma^{t'-t} r_{t'}}$ as a target value for a value function approximator. While Monte Carlo methods can update all values predicted by the approximator at each time step along a trajectory with actual returns, the returns cannot be computed until the end of each episode, and have high variance due to different trajectories having varying returns as a result of aggregating all rewards along the trajectory, making it difficult to solve the problem of credit assignment (i.e. which specific actions contributed to improving/reducing the rewards).

\subsubsection{Temporal Difference learning}
\label{sec:background/td_learning}

\gls{td} learning performs a Bellman update by bootstrapping values of the next state. In the simplest one-step on-policy form, the update of a value function $V^\pi(s)$ can be written as:
\begin{equation}
    V^\pi(s) \leftarrow V^\pi(s) + \alpha \left[ r + \gamma V(s') - V(s) \right],
\end{equation}
where the reward-state transitions are sampled by the current policy $\pi$. Here, $V^\pi(s)$ is the value estimator to be learnt, $\alpha$ is the learning rate, and $r + \gamma V(s')$ is the target values to regress $V^\pi(s)$ towards. $\delta_t \doteq r_t + \gamma V(s_{t+1}) - V(s_t)$ is a quantity called \gls{td} error.
\gls{td} learning can also be applied to incomplete episodes or infinite-horizon continual problems in which an episode may have infinite time steps. 

\gls{td} learning can also be applied to learning action-values $Q(s, a)$. SARSA \cite{rummery_online_1994} is an on-policy method, for which the current state $s$, current action $a$, reward $r$, next state $s'$ and next action $a'$ must be known. The \gls{td} target is $r + \gamma Q(s', a')$. Q-learning \cite{watkins_learning_1989} is an off-policy method, since it only requires $s$, $a$ and $r$, and the \gls{td} target is $r + \gamma \max_{\overline{a}}{Q(s', \overline{a})}$.
\gls{dqn}~\cite{mnih_playing_2013} was a major breakthrough for deep reinforcement learning. Several improvements have been made to stabilise training. A target network \cite{mnih_human-level_2015} uses a lagged copy of the Q-network as the target so that the target is approximately fixed during training. Double Q-learning \cite{hasselt2010double,hasselt_double_dqn_2015} uses two Q-networks, one to select the actions and one to estimate the next step Q-value, to counter the problem of systematic overestimation of Q-values. Prioritised Experience Replay \cite{schaul2015prioritized} sample informative experiences more frequently to accelerate learning.

\subsection{Policy gradient methods}
\label{sec:background/policy_gradient}
Policy gradient methods \cite{williams_reinforce_1988,Williams2004SimpleSG,sutton_policy_gradient_2000} directly learn and improve a policy $\pi(a|s)$ to maximise the expected returns. The key idea is to reinforce actions that resulted in high returns, while suppressing actions with low returns. 

The objective $J(\pi_{\theta})$ is the expected return over all completed trajectories generated by the agent following policy $\pi_{\theta}(a|s)$:
\begin{equation}
J(\pi_{\theta}) = \expect_{\tau \sim \pi_\theta}\left[\mathcal{R}(\tau)\right] = \expect_{\tau \sim \pi_\theta}\left[\sum_{t=0}^{T}{\gamma^t r_t} \right],
\end{equation}
where $\theta$ is a set of learnable parameters. Neural networks are suitable to model $\pi_{\theta}(a|s)$.

The policy gradient algorithm maximises the objective $J(\pi_\theta)$ by performing gradient ascent on the policy parameters $\theta$ according to $\nabla_\theta J(\pi_{\theta})$, which can be shown to be \Cref{eq:background/policy_gradient}~\cite{williams_reinforce_1988,sutton_reinforcement_2018}:
\begin{equation}
\label{eq:background/policy_gradient}
\nabla_\theta J(\pi_{\theta}) = \expect_{\tau \sim \pi_\theta}\left[\sum_{t=0}^T{\mathcal{R}_t(\tau)\nabla_\theta \log{\pi_\theta (a_t|s_t)}}\right].
\end{equation}

The probability of the action $\pi_\theta (a_t|s_t)$ corresponding to a positive $\mathcal{R}_t(\tau)$ is increased, while that corresponding to a negative $\mathcal{R}_t(\tau)$ is decreased. 

\subsubsection{REINFORCE}

The REINFORCE algorithm \cite{williams_reinforce_1988,Williams2004SimpleSG} is a direct application of the policy gradient algorithm, which uses Monte-Carlo approximation for the expectation in \Cref{eq:background/policy_gradient}. 

While value-based methods tend to struggle with continuous action spaces, since they have to find an argmax action over the values (see \Cref{eq:background/value_iteration}), policy gradient methods work with both discrete and continuous action spaces. The disadvantage is that they tend to have high variance because returns vary significantly from trajectory to trajectory, and credit assignment for long trajectories is non-trivial.\footnote{Credit assignment is the process of determining how individual actions contributed to the overall success or failure of an agent's strategy, particularly in problems with delayed rewards. Since REINFORCE uses the returns of full trajectories, actions that were irrelevant to the overall success or failur of the agent may still be reinforced or penalised.}

\subsection{Actor-Critic methods}
\label{sec:background/actor_critic}
Actor-Critic methods have two components that are learnt jointly: an \emph{actor} -- a parameterised policy used to take actions, and a \emph{critic} -- a value function used to estimate returns by bootstrapping.

\subsubsection{Advantage}
\gls{a2c}~\cite{mnih_asynchronous_2016} combines policy gradient methods and value-based methods. It is similar to REINFORCE~\cite{williams_reinforce_1988,Williams2004SimpleSG}, but instead of using the returns from Monte Carlo sampled experiences, it uses the advantage function $A(s_t, a_t) = \mathcal{R}_t - V(s_t)$ that measures the relative return of taking an action $a_t$ in a given state $s_t$.
It subtracts the baseline term $V(s)$ to stabilise the learning of the policy. In practice, the advantage function is trained by regressing towards the \gls{td} error $\delta_t = r + \gamma V^\pi(s') - V^\pi(s)$ as a reinforcing signal.
The gradient of the \gls{a2c} objective can be written as:
\begin{equation}
\nabla_\theta J(\pi_{\theta}) = \expect_{s, a \sim \pi_\theta}\left[A(s, a)\nabla_\theta \log{\pi_\theta (a|s)}\right].
\end{equation}

\subsubsection{Generalised Advantage Estimate}
\label{sec:background/gae}
\gls{gae}~\cite{gae_schulman_2015} provides a robust and low-variance estimate of the advantage function.
\gls{gae} can be expressed as a sum of exponentially weighted multi-step \gls{td} errors:
\begin{equation}
    A_t^\text{GAE} = \sum_{t'=t}^{T} (\gamma \lambda)^{t'-t} \delta_{t'},
\end{equation}
where $\delta_t = r_t + \gamma (1 - d_t) V(s_{t+1}) - V(s_t)$ is the \gls{td} error at time $t$, and $\lambda$ is a hyperparameter that controls the trade-off of bias and variance. When $\lambda=0$, the \gls{gae} is equivalent to the \gls{td} error $\delta_t$. When $\lambda=1$, it is equivalent to a Monte Carlo estimate of the advantage.

\subsubsection{Policy Optimisation}
A direct implementation of policy gradient algorithms such as \gls{a2c} is susceptible to policy collapses, in which the policy optimisation overshoots and the resulting new policy significantly deviates from the old policy. To counter this, \gls{trpo}~\cite{schulman2015trust} updates policies while ensuring that the old and new policies are within a certain proximity, measured in terms of \gls{kl} divergence. \gls{ppo}~\cite{schulman_proximal_2017} addresses the same problem, using a first-order method where \gls{trpo} uses a more complex second-order method. Furthermore, \gls{ppo} proposes to replace the \gls{kl} divergence regularisation with a simpler clipped objective, which has a similar performance. 
\gls{ppo}'s clipped objective is:
\begin{equation}
\label{eq:background/ppo}
\mathcal{L}_\text{PPO}(\theta) = \expect_{s, a \sim \pi} \left[\min \left(\frac{\pi_{\theta}(a|s)}{\pi_{\theta_\text{old}}(a|s)}A_t^{\text{GAE}}, \text{clip}\left(\frac{\pi_{\theta}(a|s)}{\pi_{\theta_\text{old}}(a|s)}, 1 - \epsilon, 1 + \epsilon \right) A_t^{\text{GAE}} \right)\right],
\end{equation}
where $A_t^{\text{GAE}}$ is the \gls{gae} at time step $t$, and $\frac{\pi_{\theta}(a|s)}{\pi_{\theta_\text{old}}(a|s)}$ is a ratio of probabilities of taking action $a$ at state $s$ with the new policy against that with the old policy.

\subsubsection{Q-learning-based Actor Critic}
\gls{ddpg}~\cite{silver2014deterministic, lillicrap2015continuous} adapts Deep Q-learning to work in continuous action spaces. Instead of evaluating the maximum Q-value over actions, which is a challenge in continuous action spaces, \gls{ddpg} learns a target policy network $\mu_\theta(s)$ to deterministically predict an action that maximises $Q_\phi(s, a)$, and use its result to evaluate the \gls{td}-error. This is done by solving $\max_\theta{\expect[Q_\phi(s,\mu_\theta(s))]}$.

\gls{td3}~\cite{fujimoto2018addressing} improved upon \gls{ddpg} by introducing delayed policy updates, policy smoothing, and clipped double-Q learning that learns two Q-value estimates and take their minimum to reduce the effect of over-estimation of the Q-values, while \gls{sac}~\cite{haarnoja2018soft} replaces \gls{ddpg}'s deterministic policy with a stochastic policy and adds entropy regularisation to the \gls{rl} objective, rewarding the agent for visiting states where the agent's policy has high entropy, since having high entropy may be an indication that the policy is uncertain for this state input and has room for improvement.

\subsection{Hierarchical Reinforcement Learning}
\label{sec:background/hrl}
\gls{hrl}~\citep{vezhnevets2017feudal,hiro2018,wulfmeier2019compositional,pateria2021hierarchical,zhang2021hierarchical} is a branch of \gls{rl} that involves decomposing a complex long-horizon task into a hierarchy of subproblems or subtasks so that the lower-level policies can learn more primitive actions with a shorter-term focus, while the higher-level policies select temporally abstracted macro-actions with a longer-term focus. Where multiple tasks share many subtasks, an agent may learn a reusable lower-level policy with a higher-level policy learning to optimise the task-specific objective.
This allows the agent to learn more efficiently and generalise better to novel situations.

In \gls{hrl}, subgoals~\cite{goel2003subgoal}, credit assignment~\cite{feudalnet}, and options~\cite{sutton1999between_options, Precup2000TemporalAI,optioncritic,optioncritic_ppo,zhang2019dac} are key building blocks in achieving this decomposition.
Subgoals are intermediate goals that the agent needs to achieve in order to complete the overall task. The agent uses subgoals to incentivise the agent to visit state regions which brings the agent closer to achieving its ultimate objective. Credit assignment is the process of assessing the level of each action's contribution to the overall performance of the agent. \gls{hrl} introduces temporal abstraction to task execution, which naturally allows long-term credit assignment (e.g. advantages computed for higher-level actions). 

\subsubsection{The Options Framework}
\label{sec:background/options_framework}

Options~\cite{sutton1999between_options, Precup2000TemporalAI} are temporally extended actions that allow the agent to make high-level decisions in the environment. Each option corresponds to a specialised low-level policy that the agent can use to achieve a specific subtask. In the Options Framework, the inter-option policy $\pi(z_t|s_t)$ and an option termination probability $\varpi(s_{t+1}, z_t)$ govern the transition of options, where $z_t$ is the current option, and are chosen from an $n$ number of discrete options $\{\mathcal{Z}_1, ..., \mathcal{Z}_n\}.$\footnote{The notations have been adapted to match the convention in this thesis.} 
Options are especially valuable when there are multiple stages in a task that must be taken sequentially (e.g. following a recipe) and the agent must obey different policies given similar observations, depending on the stage of the task. 

Earlier works have built upon the Options Framework by either learning optimal option selection over a pre-determined set of options~\cite{mcp_peng} or using heuristics for option segmentation~\cite{NIPS2016_f442d33f,hiro2018}, rather than a fully end-to-end approach. While effective, such approaches constrain the agent's ability to discover useful skills automatically. 

The Option-Critic architecture~\cite{optioncritic} proposed end-to-end trainable systems which learn option assignment. It formulates the problem such that inter-option policies and termination conditions are learned jointly in the process of maximising the expected returns. 
\gls{dac}~\cite{zhang2019dac} is an \gls{hrl} approach to discovering options. \gls{dac} reformulates the \gls{smdp}~\cite{sutton1999between_options} of the option framework as two hierarchical \glspl{mdp} (high-\gls{mdp} and low-\gls{mdp}). The low-\gls{mdp} concerns the selection of primitive actions within the currently active option. The high-\gls{mdp} handles high-level decision-making, such as selecting and terminating options. 

\subsubsection{Applications of the forward-backward algorithm}
The forward-backward algorithm, also known as the Baum-Welch algorithm~\cite{baum72}, is an algorithm that finds an optimal assignment of latent variables in a \gls{hmm} given sequential observations.
Several works~\cite{option_ml_2016,fox2017multi,Zhang2020ProvableHI,online_baum_welch_2021} have explored the possibility of applying the forward-backward algorithm to the options framework in a \gls{td} learning setting. 

\subsection{Transfer and generalisation in Reinforcement Learning}
\label{sec:background/transfer}

Many findings and techniques for standard deep learning in terms of transfer learning~\cite{tan2018survey}, multi-task learning~\cite{crawshaw2020multi}, and Meta Learning~\cite{hospedales2021meta} are applicable or translatable in the context of deep \gls{rl}~\cite{wulfmeier2023foundations}. 
Policy and value networks, which often have shared parameters, can be pre-trained on general tasks before being fine-tuned on a more specific task~\cite{baker2022_openai_vpt} or by learning better representations from auxiliary tasks such as learning the environment dynamics and reward models~\cite{ha2018worldmodels,nagabandi2018neural,hafner2019dreamer,hafner2023mastering}, sometimes referred to as world modelling (see \Cref{sec:background/world_models}). 
Recent approaches model the \gls{rl} problem as a more general sequential modelling task, motivating applications of large transformer models~\cite{vaswani2017attention,parisotto2020stabilizing,chen2021decision,baker2022_openai_vpt,reed2022generalist}. This formulation allows foundation models~\cite{bommasani2021opportunities} and large pre-trained models to be repurposed or fine-tuned on specific tasks~\cite{baker2022_openai_vpt,schubert2023generalist,driess2023palme,yang2023foundation}.

\gls{meta-rl}~\cite{wang2016learning,duan2016rl,pearl_2019,metarl_survey,pong2022offline,yuan2022robust} focuses on developing methods that enable agents to learn a general policy that can efficiently be adapted to solve new tasks with a small amount of data. The aim is to learn a shared representation across many different tasks in order to learn transferable skills. Large transformer policies are used in recent work such as AdA~\cite{pong2022offline} so that the agent can infer and retain knowledge about the current task, which may be used to exhibit task-specific behaviour. 

Similarly, Skill-based \gls{rl}~\cite{pertsch2020spirl,nam2022skillbased,shi2023skill} aims to develop agents that can learn a set of reusable skills that can be composed to solve new tasks. These skills are designed to be generalisable and can be often learned from offline data~\cite{ase_large_scale_reusable,raolearning2022,salter2022mo2}. This approach allows agents to solve new tasks with few-shot learning and adapt to changes in the environment more efficiently. These approaches are compatible with \gls{hrl}, outlined in \Cref{sec:background/hrl}.

\subsection{Model-based methods}
Algorithms considered so far fall under the category of model-free methods, which do not require explicit modelling of the environment's state transition dynamics. In some problems, especially in the case of simulations, games and robotics, the transition dynamics may be readily accessible as system definitions, in which case it could be beneficial to take advantage of that knowledge. In other problems where the transition dynamics are not known, modelling this has its pros and cons. The advantage is that, once a good model is learnt, it can be used to forecast moves without having to act in the environment. This increases sample efficiency, which is highly beneficial when collecting experiences is costly. On the other hand, it is difficult to learn an accurate and comprehensive model, and long-term planning using inaccurate models can lead to compounding errors, making the problem intractable. Moerland \etal\cite{moerland2023model} performs a detailed analysis of model-based methods.

\subsubsection{Monte Carlo Tree Search}
\gls{mcts}~\cite{coulom2006efficient_mcts} is a widely used model-based method for problems with deterministic discrete state spaces with known transition functions. It samples candidate actions from the current policy and performs Monte Carlo rollouts to explore states and estimate their values. 
\gls{mcts} was used as a core planning component in AlphaGo \cite{silver_mastering_2016} to solve Go. It used neural networks to learn a policy to guide the search and a value function to estimate the values of leaf nodes. A small and fast policy network is used for rapid \gls{mcts} rollouts, while a large and more expressive network is used as the main strong policy network to be trained. In AlphaZero~\cite{silver2017mastering_alphazero} that solved Go, Chess and Shogi, rollout simulations with a small network were replaced with a greedy strategy of expanding a child node with the highest \gls{ucb} score.\footnote{In AlphaZero, a variant of PUCT~\cite{rosin2011multi} was used to evaluate the \gls{ucb} score for Monte-Carlo Trees.}

\subsubsection{Efficient sampling from learnt models}
It is possible to learn the transition and reward models $P(s'|s, a)$ and $R(s, a, s')$ from experiences, even when they are not readily available. The earliest of this approach has been demonstrated in Dyna \cite{sutton1991dyna}, where samples are used both to update the policy function directly and to learn a transition model used to generate additional imagined samples.
Model-based Value Expansion \cite{feinberg2018model} performs imagined rollouts up to a fixed depth, and from there uses the value function. 
Model-Based Policy Optimisation (MBPO)~\cite{janner2019trust_mbpo} chooses rollout length based on estimated model generalisation capacity to attain the best of both worlds of model-free and model-based methods.
Model-Based Reinforcement Learning via Meta-Policy Optimisation (MB-MPO)~\cite{clavera2018model_mbmpo} meta-learns a policy that can quickly adapt to any model in the ensemble with one policy gradient step. This approach builds on the gradient-based meta-learning framework called Model-Agnostic Meta-Learning (MAML)~\cite{finn2017model_maml}. 

\subsubsection{Jointly learnt latent models}
\label{sec:background/world_models}
Simulated Policy Learning (SimPLe) \cite{kaiser2019model_simple} uses similar techniques to stochastic video prediction to learn a stochastic world model with discrete latent space, used as a simulator, and directly applies \gls{ppo}~\cite{schulman_proximal_2017} to acquire the policy.

TreeQN and ATreeC \cite{farquhar2017treeqn} incorporate recursive tree-structured neural networks. They learn a transition function that, given a state representation and an action, predicts the next state representation and corresponding rewards. This is applied recursively for all possible sequences of actions up to some predefined depth, at which a value function approximator is used to estimate values at each leaf node. Finally, TD($\lambda$) \cite{sutton1988learning_td} is used to refine the estimate of $Q(s,a)$.  While TreeQN is based on DQN~\cite{mnih_playing_2013}, ATreeC is an actor-critic variant.

Imagination-Augmented Agent (I2A) \cite{racaniere2017imagination_i2a} learns an environment model, a recurrent architecture to predict next-step observations, which can be trained in an unsupervised fashion from agent trajectories. The environment model is used to roll out imagined trajectories, which are augmented with real samples. The policy network is trained with A3C \cite{mnih_asynchronous_2016}.
World Models \cite{ha2018worldmodels} similarly learn to generate states for simulation with a variational autoencoder and a generative recurrent network.
MuZero \cite{schrittwieser2020mastering_muzero} extends the work of AlphaZero \cite{silver2017mastering_alphazero} so that it can learn the transition models for Atari, Go, chess and shogi using one architecture, without any knowledge of the game dynamics.

Deep Planning Network (PlaNet) \cite{hafner2019learning_planet} learns the environment dynamics from images and chooses actions through fast online planning in latent space. It uses a Recurrent State Space Model (RSSM), a form of a sequential \gls{vae} that relates transitions, observations and rewards. Model Predictive Control \cite{richards2005robust_mpc} is used to adapt its plan based on new observations, and Cross-Entropy-Method (CEM) \cite{rubinstein1997optimization_cem} is used to search for the best action sequence. Unlike model-free and hybrid \gls{rl} algorithms, no policy or value networks are used.

Dreamer \cite{hafner2019dreamer} leverages the PlaNet world model, and allows the agent to imagine the outcomes of potential action sequences without executing them in the environment. Unlike PlaNet, it uses an actor-critic approach to learn a policy that predicts actions that result in state trajectories with high value estimates. This is similar to \gls{ddpg}~\cite{silver2014deterministic, lillicrap2015continuous} and \gls{sac}~\cite{haarnoja2018soft}, but is applied to multi-step state transitions rather than maximising the immediate Q-values. DreamerV3~\cite{hafner2023mastering} improves the latent representation and reward/value activation function of Dreamer to solve a variety of tasks including finding diamonds in Minecraft, a complex task involving navigating a virtual terrain, gathering materials, and crafting increasingly more powerful tools for mining minerals (see \Cref{sec:background/virtual_world}). 

With the rise of diffusion models~\cite{ho2020denoising,rombach2021stablediffusion,karras2022elucidating,ramesh2022dalle2}, recent works have shown that diffusion world models~\cite{alonso2024diffusion,ding2024diffusion} can also used effectively to train \gls{rl} agents.

\section{Other learning methods}
\subsection{Imitation Learning}
\gls{il} learns a policy from expert trajectories rather than from trial and error. Perhaps the most intuitive and straightforward way of learning a policy, \gls{bc} is a simple form of \gls{il} that uses supervised learning to train an agent policy $\pi(a|s)$ given a dataset of state and action pairs as expert demonstrations. A policy trained only with \gls{bc} is prone to failure, however, since compounding errors during rollout can take the agent to states that are not encountered during training. 
DAgger~\cite{ross2011reduction_dagger} is a method to overcome the limitation of \gls{bc} by iteratively collecting expert labels for online rollouts and adding these samples to the training dataset, ensuring better coverage of the state space likely to be visited by the agent policy. 

\gls{irl}~\cite{ng_algorithms_2000,ziebart_maximum_2008,ziebart_modeling_2010} is a class of methods that aims to recover an underlying reward function that best explains the expert trajectories. Once the reward function is estimated, standard \gls{rl} techniques can be applied. This approach generalises better than \gls{bc} in tasks in which inferring the reward mechanism is easier than learning the policy directly. However, \gls{irl} is an ill-posed problem since there are many possible reward functions that can explain the given trajectories. Hence, additional assumptions must be introduced to constrain the learning problem. 

Generative Adversarial Imitation Learning (GAIL)~\cite{ho_generative_2016} applies \glspl{gan}~\cite{goodfellow2014gan} to the problem of \gls{il}, training a policy to generate behaviour indistinguishable from expert demonstrations. This setup enables the model to learn complex behaviours.

Pre-training via imitation learning on demonstrations~\cite{baker2022_openai_vpt} is an effective strategy for learning representations that are transferrable to solving more general tasks (see \Cref{sec:background/transfer}). World Modelling (see \Cref{sec:background/world_models}) learns to reconstruct the observations as well as predict the next action, and is shown to be an effective training strategy~\cite{hu2022model,hu2023gaia}.

\subsection{Value Iteration Networks}
\label{sec:background/vin}
\gls{vin} proposed by Tamar \etal\cite{tamar_value_2016} is a model-based \gls{il} method that aims to learn the entire planning procedure implicitly and in an end-to-end differentiable way. 
It generates a plan online by approximating \gls{vi} (\Cref{sec:background/value_iteration}), and back-propagates errors through the plan to train estimators of the rewards.
It models the underlying \gls{mdp} with neural network parameters learnt with back-propagation by matching the evaluated policy with expert trajectories.
In the case of a gridded state space (e.g. a discretisation of a robot's position in space), the VIN takes the form of a recurrent convolutional network with cross-channel max-pooling. The convolutional kernels approximate transition and reward models that are used to generate Q-values online.
The \gls{cnn}~\cite{lecun1989cnn} representing the \gls{mdp} can leverage repeating structural patterns in the state space's connectivity.
This enables the \gls{vin} to predict an appropriate \gls{mdp} and plan without further training in novel environments for grid-based state spaces, such as those used in robot navigation, given that the observation characteristics and the local transition dynamics are similar to the training domain.

\subsection{Fine-tuning and Curriculum Learning}

\gls{rl} relies on trial and error to uncover the underlying \gls{mdp} which is often sample-inefficient, and involves risk in real-world scenarios where a wrong decision can potentially be catastrophic. On the other hand, \gls{il} is limited by the coverage of states explored in expert demonstrations for training. Hence, a policy is often pre-trained using \gls{il} and fine-tuned using \gls{rl}.

Curriculum Learning~\cite{narvekar2020curriculum} is another training technique to improve sample-efficiency. It trains the agent in stages, starting with simpler tasks and progressively introducing harder tasks. This process is designed to help the agent acquire basic skills early on in the training, which can later be applied to solve complex problems. 

\section{Embodied agents}

While embodied tasks such as navigation and manipulation can be solved with classical path planning methods (\Cref{sec:background/planning}) for a fully known environment and known motion dynamics of the agent, in many embodied agent problems the agent must learn to solve tasks without having access to such privileged information. 
\gls{slam}
~\cite{fuentes-pacheco_visual_2015, mur-artal_orb-slam2_2017} can be used to construct a map of the environment online from the agent observations if the camera parameters are known and the environment is mostly static and is embedded in Euclidean geometry. However, many real-world problems have (a) dynamic components, (b) non-Euclidean environment dynamics, and (c) unknown camera and/or agent dynamics. For instance, a coordinate space in motion, such as a train or an aeroplane, is locally Euclidean, but the entrance/exit may lead to a different global location because there is a temporal component to the planning problem. In these examples, constructing a globally consistent map of the environment from the agent observations alone is infeasible and unscalable. Humans can work with such dynamic environments from partial observations to solve a variety of real-world tasks with various characteristics. In contrast, algorithmically solving embodied task execution over a large spatial and temporal horizon is still an unsolved problem.
This motivates research on learning-based techniques for embodied navigation, planning and task execution. 

\subsection{Visual navigation}
Visual navigation is a particular application of data-driven planning, in which the observation data is an image stream captured by a mobile robot. Observations retrieved from 3D environments have spatial consistency, and state transitions are often translation-invariant in the case of spatial states. This opens up an opportunity for data-driven planning algorithms to take advantage of these regularities to make forecasts during planning, thereby increasing their performance and sample efficiency.

Several works have made advances in training deep networks for navigation.
Parisotto \etal \cite{parisotto_neural_2018} developed the Neural Map, an \gls{a2c}~\cite{mnih_asynchronous_2016} agent that reads and writes to a differentiable memory (\ie a map).
Similarly, Mirowski \etal \cite{mirowski_learning_2017} trained reactive policies which are specific to an environment.
The Value Prediction Network~\cite{oh2017value_vpn} learns an \gls{mdp} and its state space from data, and plans with a single roll-out over a short horizon.
Hausknecht \etal \cite{hausknecht2015pomdps} added recurrence to deep Q-learning to address partially-observable environments, but considered only single-frame occlusions.
Several works treated planning as a non-differentiable module, and focused on training neural networks for other aspects of the navigation system.
For instance, Savinov \etal \cite{savinov_semi-parametric_2018} composed siamese networks and value estimators trained on proxy tasks, and using an initial map built from footage of a walk-through of the environment.
Active Neural SLAM~\cite{chaplot_learning_2020} trained a localisation and mapping component (outputting free space and obstacles), a policy network to select a target for the non-differentiable planner, and another to perform low-level control.
Subsequent work~\cite{chaplot2020object} complemented this map with semantic segmentation.
The latter two works navigated successfully in large simulations of 3D-scanned environments, and were transferred to real robots.

AI2-THOR~\cite{kolve2017ai2}, VirtualHome~\cite{puig2018virtualhome}, Matterport3D~\cite{chang2017matterport3d}, Gibson~\cite{xia2018gibson}, iGibson~\cite{li2022igibson}, Habitat~\cite{habitat19iccv,szot2021habitat}, and RoboTHOR~\cite{deitke2020robothor} are some of the well-known benchmarks for embodied navigation in naturalistic simulated environments.


\subsection{Decision making in autonomous driving}
\label{sec:background/autonomous_driving}

Approaches to autonomous driving can be broadly classified into modular systems and end-to-end systems~\citep{yurtsever2020survey_autonomous}. Most systems take a modular approach~\citep{urmson2008autonomous,levinson2011towards,wei2013towards,maddern20171}, which has human-defined rules that orchestrate separately engineered components for localisation and mapping, object detection, tracking, behaviour prediction, planning, and vehicle control. Such systems allow compartmentalisation and better interpretability, but can be complex and require domain knowledge to maintain and update. Another challenge is error propagation~\citep{mcallister2017concrete}, i.e. the upstream outputs can be erroneous and must be corrected downstream. Recent work has harnessed end-to-end learning to address these issues. \gls{il}~\citep{bojarski2016end,bansal2018chauffeurnet} optimises the policy to match actions taken by experts, and is the most widely used approach. However, its performance is upper-bounded by the expert. Deep \gls{rl} has also shown successes in simulation~\citep{sallab2017deep}, on the road~\citep{kendall2019learning}, and in combination with \gls{il}~\citep{lu2022imitation}. 

\subsubsection{CARLA driving benchmark}
\label{sec:background/carla_benchmark}

CARLA~\citep{dosovitskiy17a_carla} is a widely used open-sourced 3D simulator for autonomous driving research, and provides a benchmark~\citep{carla_leaderboard} to evaluate driving performances of both privileged expert agents and sensor-only agents. Privileged agents have access to traffic light states and ground truth motions of vehicles, bicycles, and pedestrians. These are often implemented manually and are used to generate expert demonstrations for \gls{il}. Sensor-only agents, on the other hand, have to make decisions based on RGB-D images, lidar, and accelerometer measurements, which are often achieved by learnable components. 
Many prior works on CARLA have open-sourced their expert agents. Roach~\citep{zhang2021end_carla_roach} trained a \gls{ppo} agent~\citep{schulman_proximal_2017} on handcrafted reward signals with privileged information. The heavy lifting is done at the reward shaping level, where hazardous agents are identified and the desired speed and pose are computed. Roach expert is also used in MILE~\citep{hu2022model_mile} and TCP~\citep{wu2022trajectory_tcp}, where TCP has an additional emergency braking upon detecting potential collisions. TransFuser~\citep{chitta2022transfuser}, InterFuser~\citep{shao2023safety_interfuser} and TF++~\citep{jaeger2023hidden_tfplus} implement their handcrafted expert systems, either using cuboid intersections or line intersections for hazard detection. TransFuser also introduced the Longest6 benchmark~\cite{chitta2022transfuser}, which consists of longer routes compared to the official CARLA benchmark and is less saturated.

\subsection{LLMs for automating compositional tasks}
\label{background/llm_task_automation}
\gls{llm}~\cite{NEURIPS2020_gpt3,chen2021evaluating_codex,NEURIPS2022_b1efde53_instructgpt_human_feedback,openai2023gpt4}-powered agents have demonstrated sophisticated decision making and planning capabilities. Sequential prompting with the history of observation, action, and the reason for the action was proposed by ReAct~\citep{yao2022react} as an improvement to Chain-of-Thought prompting~\citep{wei2022chainofthought}, which has also been applied to autonomous driving~\citep{fu2023drive}.
Auto-GPT~\citep{autogpt} automated tasks by iteratively generating a sequence of subtasks in finer detail until they are executable. 
A similar strategy was applied to robotics \citep{huang2022language_actionable}. 
SayCan~\citep{ahn2022saycan} used \glspl{llm} to generate candidate subgoals and assessed their affordances with a value function given visual observations to ground the agent's behaviour.
Socratic Models~\cite{zeng2022socratic} demonstrated zero-shot performance on language-conditioned robot manipulation tasks by orchestrating pre-trained multimodal foundation models.
Reed \etal developed Gato~\cite{reed2022generalist}, a multi-task agent with a multi-modal foundation model backbone that can perform a variety of tasks including image captioning, chat, playing Atari and real-world robot arm manipulation.
VIMA~\citep{jiang2023vima} and PaLM-E~\citep{driess2023palme} demonstrated profound reasoning and execution capabilities on multi-modal tasks such as Visual Q\&A and robotics by fine-tuning \glspl{llm} to allow multi-modal prompting.
Inner Monologue~\citep{huang2023inner_monologue} used environment and user feedback to replan for embodied tasks. 
\citet{liang2023codeaspolicies} and \citet{singh2023progprompt} used \glspl{llm} to directly generate code for robotics, while ViperGPT~\citep{surismenon2023vipergpt} and VisProg~\citep{gupta2023visprog} composed pre-trained vision-and-language models to solve challenging vision tasks which require reasoning and domain knowledge.

\subsection{Task execution in virtual worlds}
\label{sec:background/virtual_world}

Significant innovations have been made in recent decades on research in embodied agents that can interact with video games and complex simulated 3D environments, including Atari~\cite{bellemare13arcade,mnih_human-level_2015}, Dota 2~\cite{berner2019dota}, StarCraft II~\cite{vinyals_grandmaster_2019}, and Minecraft~\cite{guss2019minerl,baker2022_openai_vpt}. More recently, SIMA~\cite{deepmind2024_sima} trained an agent to follow free-form instructions across a diverse range of virtual 3D environments, including curated research environments as well as open-ended commercial video games. In this thesis, the Minecraft environment is explored in depth.

Minecraft is an open-ended 3D virtual gaming environment where human players and agents can perform many life-like tasks to survive and live, such as exploring, mining, crafting, building, raising animals, and harvesting crops.
MineRL~\cite{guss2019minerl} provided a behaviour dataset to train embodied agents, and hosted a competition for \gls{rl} agents to obtain diamonds in Minecraft. This is a complex task, since it involves many stages of subtasks of gathering wood, cobblestones and iron, and crafting a crafting table, furnace, and pickaxes of different materials (i.e. wood, stone and iron). 

OpenAI presented \gls{vpt}~\cite{baker2022_openai_vpt}, a foundation model trained on a large collection of videos on the Internet, which successfully managed to discover diamonds using RGB pixel inputs as observations and predicting cursor movements, mouse clicks, and keyboard presses as outputs. DreamerV3~\cite{hafner2023mastering} achieved the same task but trained just on its own experiences without external datasets, by learning a world model to improve sample efficiency. 

Many of the tasks in Minecraft are compositional, and hence require long-term planning, guided by intuition from real-world experiences. 
The MineRL BASALT Competition~\cite{milani2023solving} evaluated agents on their life-like task execution capabilities in four domains: finding a cave, making a waterfall, creating an animal pen, and building a house. 

Recent works on Minecraft explore ways to use \glspl{llm} and multi-modal foundation models to interface the agent with language instructions and address such long-term planning problems. MineDojo~\cite{fan2022minedojo} provides an Internet-scale multimodal knowledge base of images, videos and text, accompanied by a benchmarking suite with thousands of diverse open-ended tasks specified in natural language prompts. The paper also proposes MineCLIP, a video-text contrastive model based on CLIP~\cite{radford2021learning}, which associates natural language subtitles with their time-aligned video segments and uses the correlation score as an open-vocabulary reward function for \gls{rl} training.
Steve-1~\cite{lifshitz2024steve} builds on MineCLIP, combining with methods from unCLIP~\cite{ramesh2022hierarchical} to generate language-guided behaviours.

\textsc{Voyager}~\citep{wang2023voyager} takes a unique approach to generating code as directly executable policies. Assuming privileged information about the environment (\ie types and coordinates of objects and materials nearby, the items in the inventory, and access to a path planner given a target coordinate), it integrates environment feedback, execution errors, and self-verification into an iterative prompting mechanism for embodied control in Minecraft. \textsc{Voyager} maintains a \emph{skill library}, a collection of verified reusable code, which can be considered as \emph{checkpoints} that can be saved, shared, loaded to the agent to determine its strategy, and used as a for further training. 

Approaches that use \gls{llm} to generate plans and subgoals are also suggested. DECKARD~\cite{nottingham2023embodied} learns modular sub-policies trained on \gls{rl} objectives to work with a high-level plan generated by the \gls{llm}, referred to as an \gls{awm}. The \gls{awm} is a \gls{dag}, where the nodes represent subgoals (e.g. crafting specific items in Minecraft), and the edges represent dependencies between these subgoals. Subgoal policies predict actions over a large multi-discrete action space from a pixel-only observation, while the overall policy initiates transitions between subgoals based on the agent’s current inventory. ``Describe, Explain, Plan and Select'' (DEPS)~\cite{wang2023describe} uses a \gls{llm} to interactively adjust the agent plan upon failure to complete a subgoal. A \gls{vlm} (e.g. CLIP~\cite{radford2021learning}) is used to provide the agent's state descriptions in natural language. Zhu \etal \cite{zhu2023ghost} similarly implements an \gls{llm} Decomposer, \gls{llm} Planner, and \gls{llm} Interface to perform interactive planning. Their method manages to obtain all items in Minecraft Overworld, although it also assumes a more privileged observation space including voxels, GPS location, and inventory information. JARVIS-1~\cite{wang2023jarvis}, a multi-modal agent that takes visual observations and human instructions, combines MineCLIP~\cite{fan2022minedojo} with an \gls{llm}, and further augments the agent with a multimodal memory, which stores successful plans in the past with corresponding scenarios to enhance the correctness and consistency of planning. JARVIS-1 employs self-instruct~\cite{wang2023selfinstruct} to generate a dynamic curriculum for the exploration of diverse tasks.

\chapter{Towards real-world navigation with deep differentiable planners}
\label{chapter:calvin} 

This chapter introduces \gls{calvin}, an embodied neural network trained to plan and navigate unseen complex 2D and 3D environments.
Rather than requiring prior knowledge of the agent or environment, the planner learns to model the state transitions and rewards.
To avoid the potentially hazardous trial-and-error of \gls{rl}, this work focuses on differentiable planners from the family of \glspl{vin}~\cite{tamar_value_2016}, which are trained offline from safe expert demonstrations. 
Although they work well in small simulations, two major limitations hinder their deployment.
Firstly, current differentiable planners struggle to plan long-term in environments with a high branching complexity.
While they should ideally learn to assign low rewards to obstacles to avoid collisions, these penalties are not strong enough to guarantee collision-free operation.
\gls{calvin} thus imposes a structural constraint on the value iteration, which explicitly learns to model impossible actions and noisy motion.
Secondly, \gls{calvin} extends the model to plan exploration with a limited perspective camera under translation and fine rotations, which is crucial for real robot deployment.
These proposed modifications significantly improve semantic navigation and exploration on several 2D and 3D environments, including the \gls{avd} that consists of real images captured from a robot, succeeding in settings that are otherwise challenging for differentiable planners.

This work was published at the IEEE/CVF Conference on Computer Vision and Pattern Recognition (CVPR) 2022~\cite{ishida2022towards}.

\begin{figure}
  \centering
  \renewcommand*{\arraystretch}{0}
  \begin{tabular}{*{3}{@{}c}@{}}
  \includegraphics[height=3.6cm]{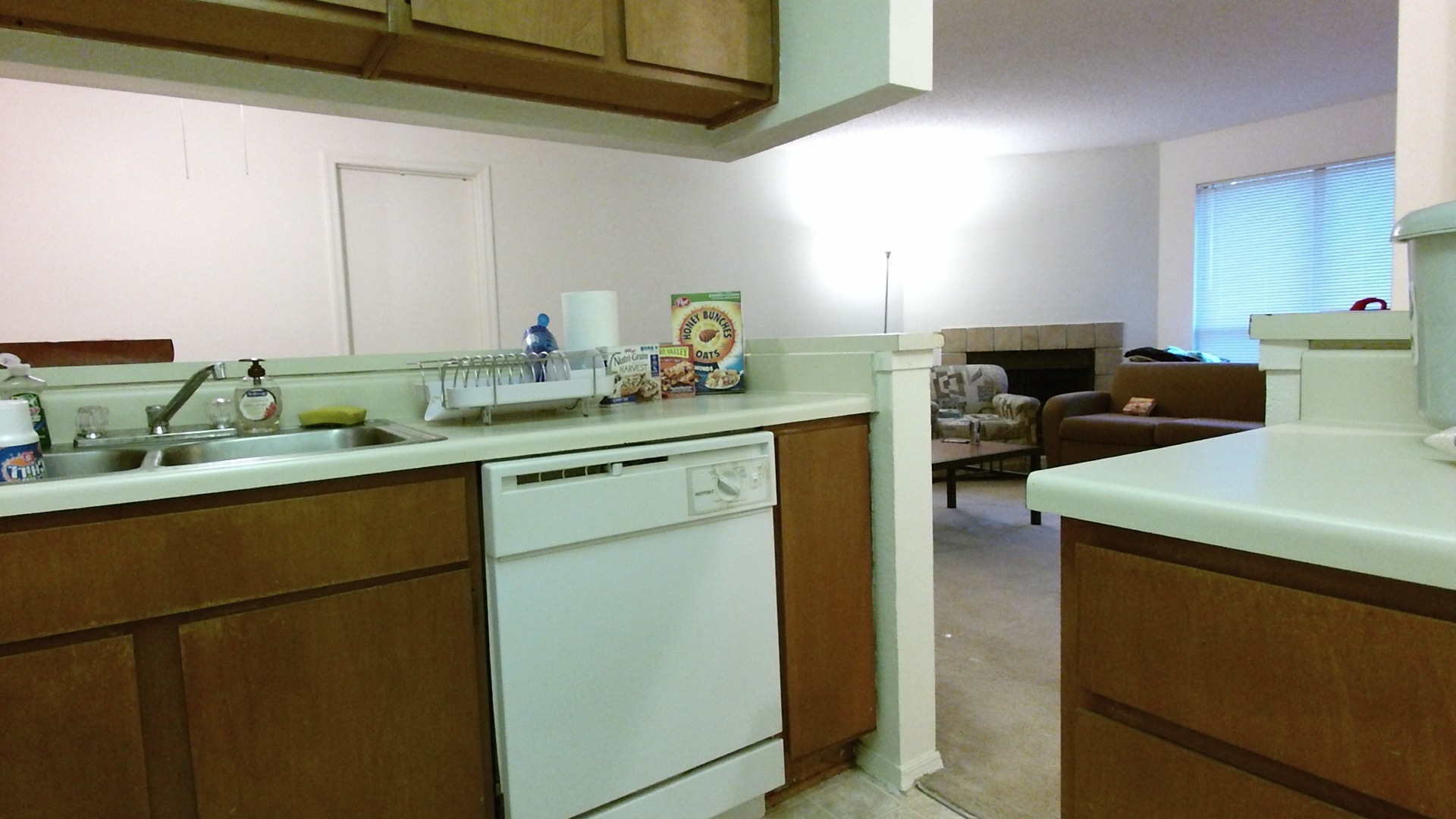} &
  \includegraphics[height=3.6cm]{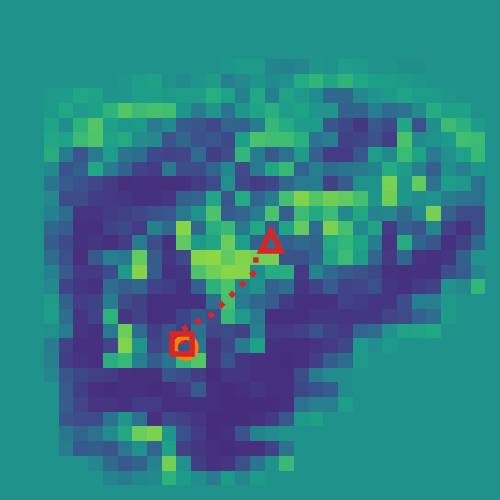} &
  \includegraphics[height=3.6cm]{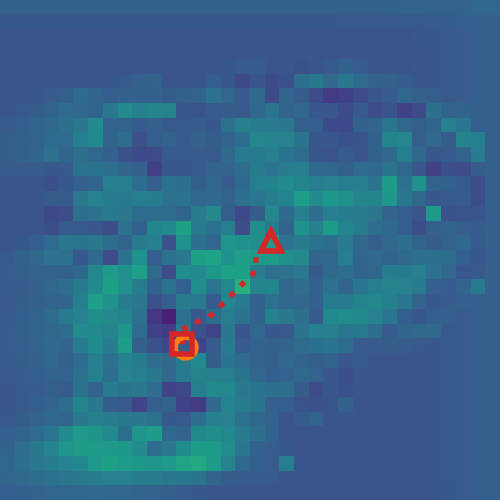} \\
  \includegraphics[height=3.6cm]{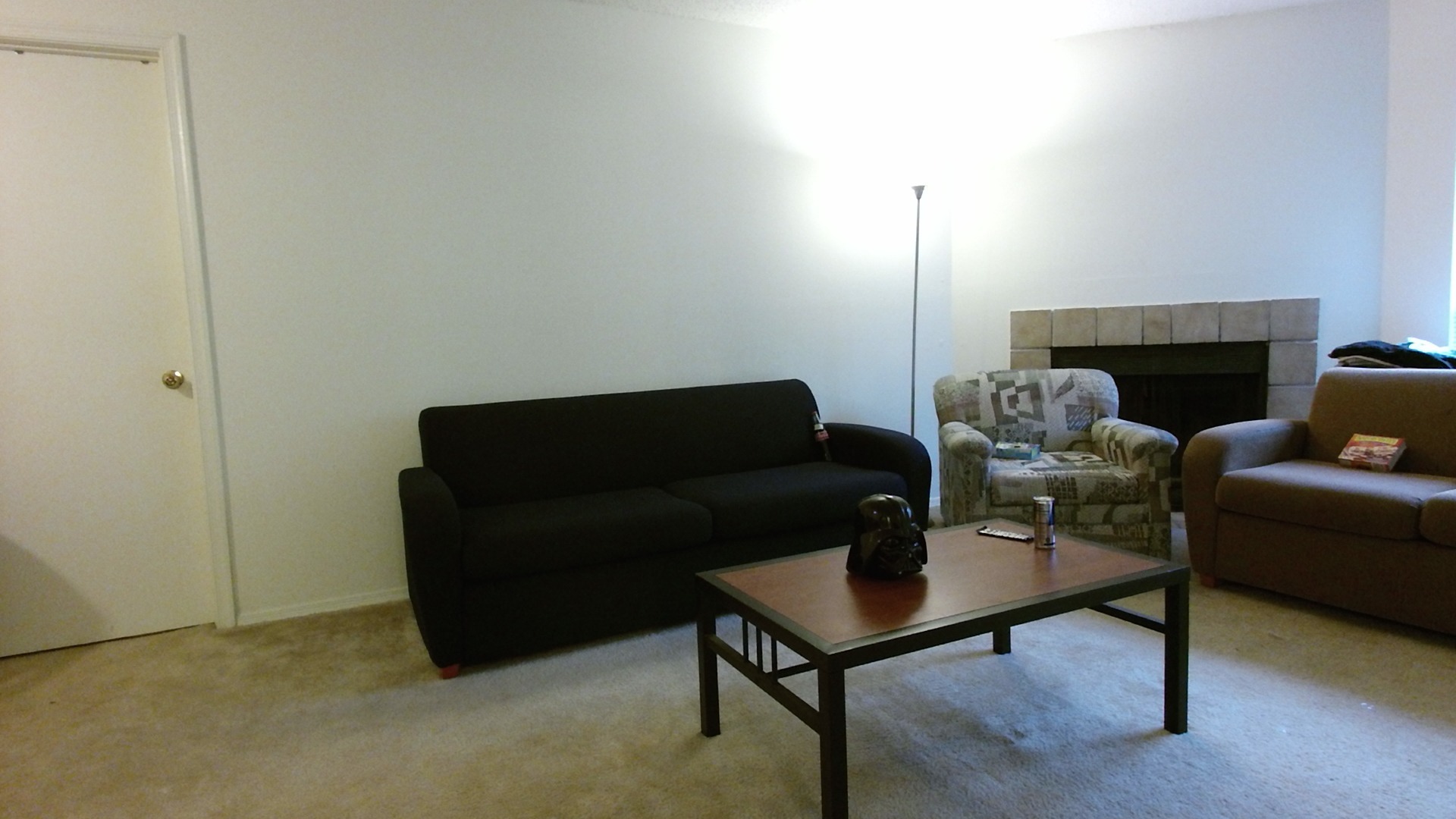} &
  \includegraphics[height=3.6cm]{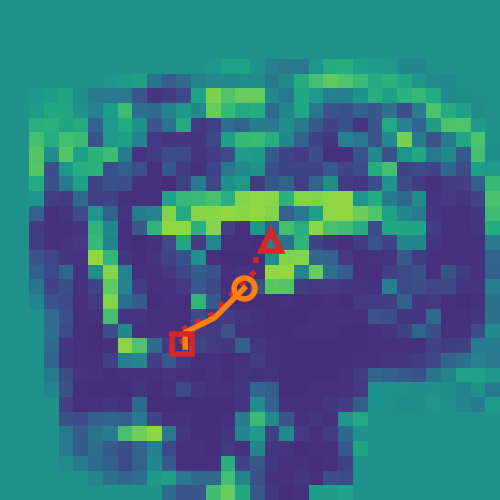} &
  \includegraphics[height=3.6cm]{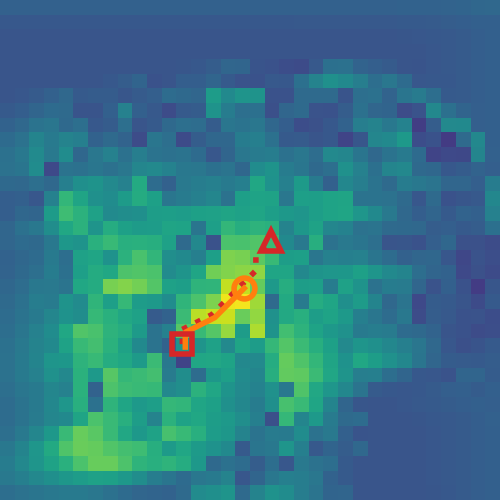} \\
  \includegraphics[height=3.6cm]{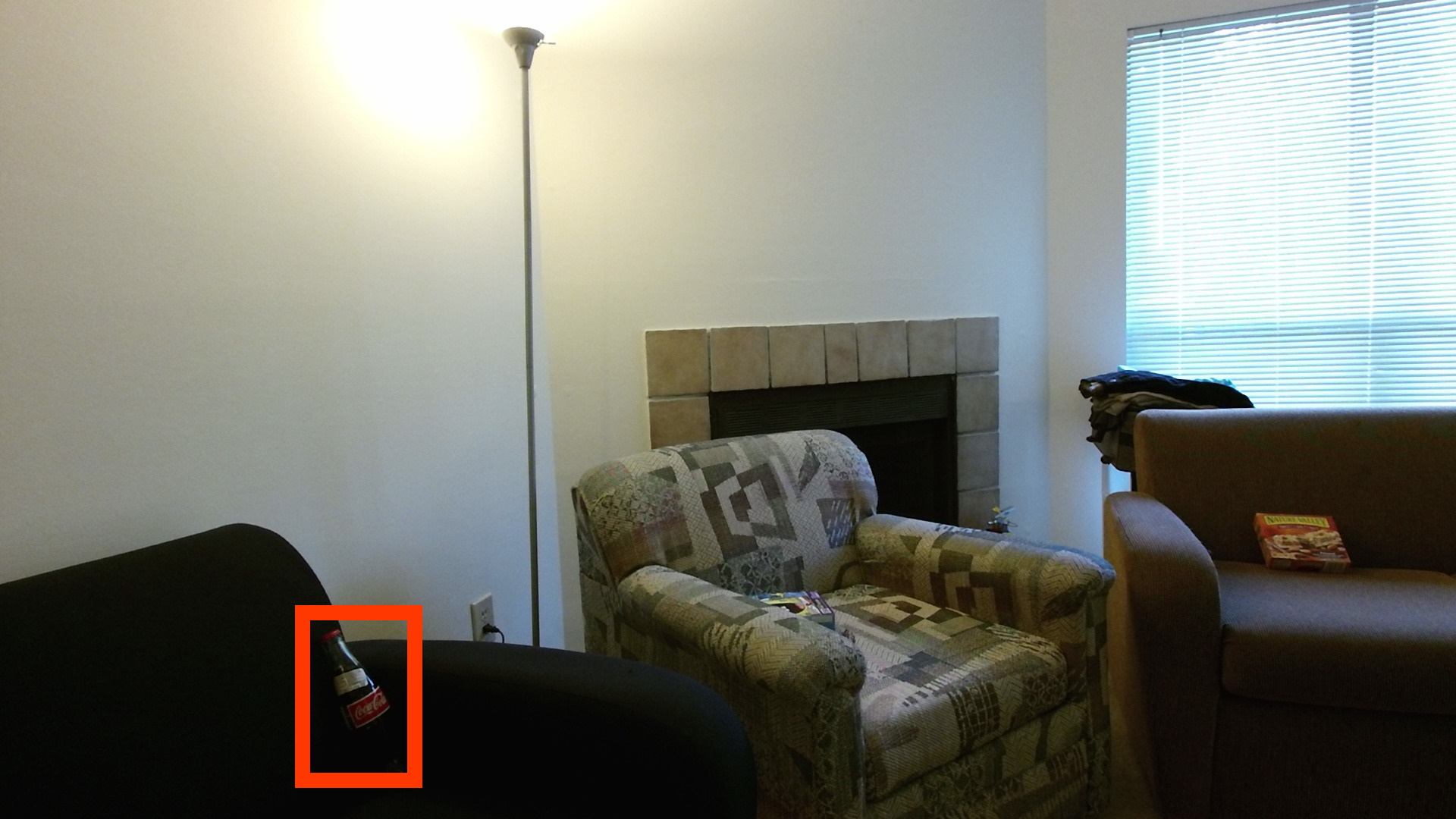} &
  \includegraphics[height=3.6cm]{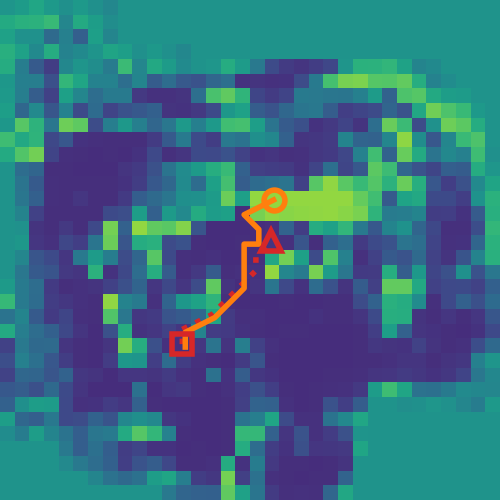} &
  \includegraphics[height=3.6cm]{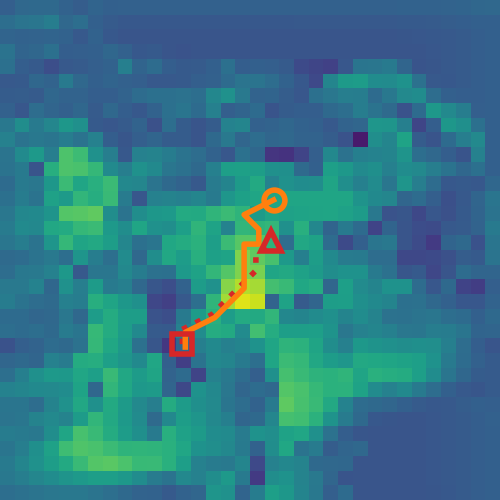}
  \end{tabular}
  \caption{
    \textbf{(1\textsuperscript{st} column)} Input images seen during a run of \glsshort{calvin} on \glsshort{avd} (\Cref{sec:calvin/experiments/avd}). This embodied neural network has learned to efficiently explore and navigate unseen indoor environments, to seek objects of a given class (highlighted in the last image). \textbf{(2\textsuperscript{nd}-3\textsuperscript{rd} columns)} Predicted rewards and values (resp.), for each spatial location (higher for brighter values). The unknown optimal trajectory is dashed, while the robot's trajectory is solid.
  }
  \label{fig:calvin/avd}
\end{figure}

\section{Introduction}
Continuous advances in robotics have enabled robots to be deployed to a wide range of scenarios, from manufacturing in factories and cleaning in households, to the emerging applications of autonomous vehicles and delivery drones~\cite{anderson2018evaluation}.
Improving their autonomy is met with many challenges, due to the difficulty of planning from uncertain sensory data.
In classical robotics, the study of planning has a long tradition~\cite{thrun2002probabilistic,lavalle_planning_2006}, using detailed knowledge of a robot's configuration and sensors, with little emphasis on learning from data.
An orthogonal approach is to use deep learning, an intensely data-driven approach~\cite{goodfellow2016deep}.
Modern deep neural networks excel at pattern recognition~\cite{sharif2014cnn}, although they do not offer a direct path to planning applications.
While one approach would be to parse a scene into pre-defined elements (\eg object classes and their poses) to be passed to a more classical planner, an end-to-end approach where all modules are learnable has the chance to improve with data and adapted to novel settings without manual tuning.
Because of the data-driven setup, a deep network has the potential to learn behaviour that leverages the biases of the environment, such as likely locations for certain types of rooms.
For example, while the best generic strategy to reach the exit in a maze may be to follow a wall on one side, a better one for office buildings could be to first exit any room and then follow the corridors to the end~\cite{anderson2018evaluation}. 

\glspl{vin}~\cite{tamar_value_2016} emerged as an elegant way to merge classical planning and data-driven deep networks, by defining a \emph{differentiable} planner.
The end-to-end differentiability of the network allows the planner to include sub-components learnable from demonstrations.
For example, it can learn to identify and avoid obstacles, and to recognise and seek classes of target objects, without explicitly labelled examples.
However, there are gaps between VIN’s idealised formulation and realistic robotics scenarios.
The \gls{cnn}-based \gls{vin} considers the full environment to be visible and expressible as a 2D grid. As such, it does not account for embodied (first-person) observations in 3D spaces, unexplored and partially visible environments, or the mismatch between egocentric observations and idealised world-space discrete states.
It is also observed that \glspl{vin} are less robust in complex environments where the degree of branching is high, \eg in a maze as opposed to an obstacle map used in \cite{tamar_value_2016}.

This work addresses these challenges, and closes the gap between current differentiable planners and realistic robot navigation applications. The main contributions of this work are:
\begin{enumerate}
    \item A constrained transition model for \gls{vi}, following a rigorous probabilistic formulation, which explicitly models illegal actions and task termination (\Cref{sec:calvin/augmented}).
    
    \item A 3D state space for embodied planning through the robot's translation and rotation (\Cref{sec:calvin/method/orientations}). Planning through \emph{fine-grained} rotations is often overlooked (\Cref{sec:calvin/background/vin_applications}), requiring better priors for transition modelling (\Cref{sec:calvin/method/transition_loss}).

    \item A trajectory reweighting scheme that addresses the unbalanced nature of navigation training distributions (\Cref{sec:calvin/method/reweighting}).

    \item An efficient representation that fuses observations of the same spatial location across time and space (different points-of-view), allowing differentiable planners to handle long time horizons and free-form trajectories (\Cref{sec:calvin/method/lpn}).

    \item Training the differentiable planner to learn to navigate in both complex 3D mazes, and the challenging \gls{avd}~\cite{ammirato2017dataset}, with images from a real robot platform (\Cref{sec:calvin/experiments/avd}), showing that the planner can be trained with limited data collected offline.
\end{enumerate}

\section{Background}
\subsection{Value Iteration}

\gls{vi}~\cite{bellman_markovian_1957,sutton_reinforcement_2018} is an algorithm to obtain an optimal policy, by alternating a refinement of both value ($V$) and action-value ($Q$) function estimates in each iteration $k$. The algorithm and the update rule are described in \Cref{sec:background/value_iteration}.

When $s$ and $a$ are discrete, $Q^{(k)}$ and $V^{(k)}$ can be implemented as simple tables (tensors).
The cells of a 2D grid are considered as states, corresponding to discretised locations in an environment, \ie $s=(i,j) \in \mathcal{S}=\{1, \ldots, X\} \times \{1, \ldots, Y\}$. Furthermore, transitions are local (only to coordinates offset by $\delta \in \{-1, 0, 1\}^2$):
\begin{align}
\begin{split}
\label{eq:calvin/background/vi_grid}
Q^{(k)}(s, a) &=
\smashoperator{\sum_{\delta\in K}}
{P(s+\delta|s, a) \left[R(s, a, \delta) +
\gamma V^{(k-1)}(s+\delta)\right]},
\qquad \forall a \in \mathcal{A}(s),\\
V^{(k)}(s) &= \max_{a \in \mathcal{A}(s)}{Q^{(k)}(s, a)},
\end{split}
\end{align}
Note that to avoid repetitive notation $s$ and $\delta$ are 2D indices, and $V$ is a 2D matrix.
The policy $\pi^{(k)}(a|s) = \argmax_{a \in \mathcal{A}}{Q^{(k)}(s, a)}$ chooses the action with the highest action-value.
While the sum in \Cref{eq:calvin/background/vi_grid} resembles a convolution, the filters ($P$) are space-varying (depend on $s=(i,j)$), so it is not directly expressible as such.
\Cref{eq:calvin/background/vi_grid} represents the ``ideal'' \gls{vi} for local motion on a 2D grid, without further assumptions.
For this reason, $P$ and $R$ are unconstrained 5-dimensional tensors.
Contrasting \Cref{eq:calvin/background/vi_grid} to any proposed modification will be instructive, by revealing explicitly any additional assumptions.

\subsection{Value Iteration Network}
While \gls{vi} guarantees the optimal policy, it requires that the functions for transition probability $P$ and reward $R$ are known.
Tamar \etal \cite{tamar_value_2016} pointed out that all \gls{vi} operations are (sub-)differentiable, and as such a model of $P$ and $R$ can be \emph{trained from data} by back-propagation.
For the case of planning on a 2D grid (navigation), they related \Cref{eq:calvin/background/vi_grid} to a \gls{cnn}, as:
\begin{align}
\begin{split}
Q^{(k)}_{s, a} &=
\smashoperator{\sum_{\delta\in K}}
{\left(P_{a,\delta}^R \widehat{R}_{s+\delta} +
P_{a,\delta}^V V_{s+\delta}^{(k-1))}\right)}, \qquad \forall a \in \mathcal{A},
\\V^{(k)}_{s} &= \max_{a \in \mathcal{A}}{Q^{(k)}_{s, a}},
\label{eq:calvin/background/vin}
\end{split}
\end{align}
where $P^R,\,P^V \in \mathbb{R}^{A \times |K|}$ are two learned convolutional filters that represent the transitions ($P$ in \Cref{eq:calvin/background/vi_grid}), and $\widehat{R}$ is a predicted 2D reward map.
Note that $\mathcal{A}$ is independent of $s$, \ie all actions are allowed in all states.
This turns out to be problematic (see \Cref{sec:calvin/background/vin_limitations}).
The reward map $\widehat{R}$ is predicted by a \gls{cnn}, from an input of the same size that represents the available observations. In Tamar \etals~experiments~\cite{tamar_value_2016}, the observations were a fully visible top-down view map of the environment, from which negative rewards such as obstacles and positive rewards such as navigation targets can be located.
Each action channel in $\mathcal{A}$ corresponds to a move in the 2D grid, typically 8-directional or 4-directional.
\Cref{eq:calvin/background/vin} is attractive because it can be implemented as a \gls{cnn} consisting of alternating convolutional layers ($Q$) and max-pooling along the actions (channels) dimension ($V$).

\subsection{Applications of Value Iteration Networks}
\label{sec:calvin/background/vin_applications}

\glspl{vin} were applied to localisation from partial observations by Karkus \etal \cite{karkus_qmdp-net_2017}, but assuming a full map of the environment.
\gls{gppn}~\cite{lee_gated_2018} replaced the maximum over actions in the \gls{vi} formula with a \gls{lstm}. 
The evaluation included an extension to rotation, but assumed a fully known four-directional view for every gridded state, which is often not available in practice.
A subtle difference is that they handle rotation by applying a linear policy layer to the hidden channels of a 2D state space grid, which is less interpretable than the 3D translation-rotation grid implemented in \gls{calvin}

Most previous work on deploying \gls{vin} to robots \cite{Schleich_2019,nie2021cin} assumed an occupancy map and goal map rather than learning the map embedding and goal themselves from data.
Gupta \etal \cite{gupta_cognitive_2017} evaluated a differentiable planner called \gls{cmp} on real-world data using map embeddings learned end-to-end.
\gls{cmp} uses a hierarchical \gls{vin}~\cite{tamar_value_2016} to plan on larger environments, and updates an egocentric map.
It handles rotation as a warping operation external to the \gls{vin} (\ie it plans in a 2D translation state space, not 3D translation-rotation space).
Warping an egocentric map per update achieves scalability at the cost of progressive blurring of the embeddings.
\gls{cmp} requires gathering new trajectories online during training with DAgger~\cite{ross2011reduction_dagger}.
In contrast, \gls{calvin} is trained solely on a fixed set of trajectories, foregoing the need for extra expert labels.

\subsection{Limitations of Value Iteration Networks}
\label{sec:calvin/background/vin_limitations}

\subsubsection{Differences between the VIN and the idealised VI on a grid.}
There are several differences between the \gls{vin} (\Cref{eq:calvin/background/vin}) and an idealised \gls{vi} on a grid (\Cref{eq:calvin/background/vi_grid}):
\begin{enumerate}
    \item  The \gls{vin} value estimate $V$ takes the maximum action-value $Q$ across \emph{all possible actions} $\mathcal{A}$, even illegal ones (e.g. moving into an obstacle).
    Similarly, the $Q$ estimate is also updated even for illegal actions.
    In contrast, \gls{vi} only considers legal actions for each state (cell), i.e. $\mathcal{A}(i,j)$.

    \item The \gls{vin} reward $\widehat{R}$ is assumed independent of the action.
    This means that, for example, a transition between two states cannot be penalised directly; a penalty must be assigned to one of the states (regardless of the action taken to enter it).

    \item The \gls{vin} transition probability is expanded into 2 terms, $P^R$ which affects the reward and $P^V$ which affects the estimated values.
    This decoupling means that they do not enjoy the physical interpretability of \gls{vi}'s $P$ (i.e. probability of state transitions), and rewards and values can undergo very different transition dynamics.

    \item Unlike the \gls{vi}, the \gls{vin} considers the state transition translation-invariant.
    This means that it cannot model obstacles (illegal transitions) using $P$, and must rely on assigning a high penalty to the reward $\widehat{R}$ of those states instead.
\end{enumerate}

\begin{figure}
  \centering
  \includegraphics[width=0.325\columnwidth]{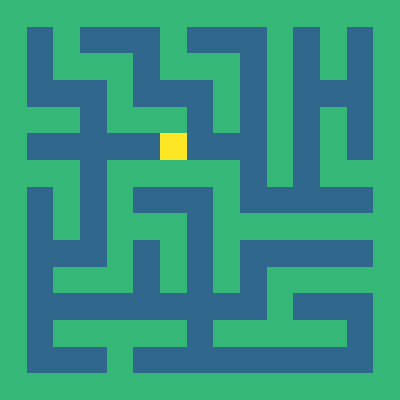}\hfill
  \includegraphics[width=0.325\columnwidth]{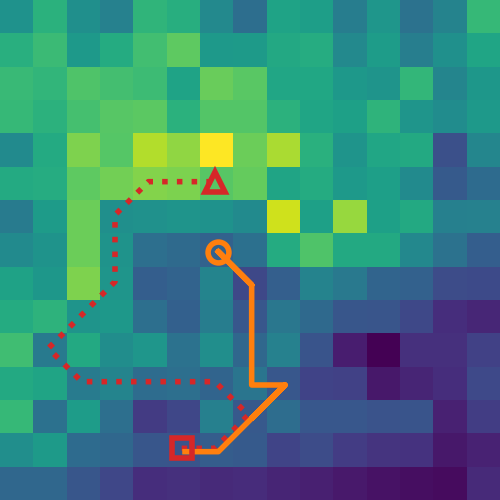}\hfill
  \includegraphics[width=0.325\columnwidth]{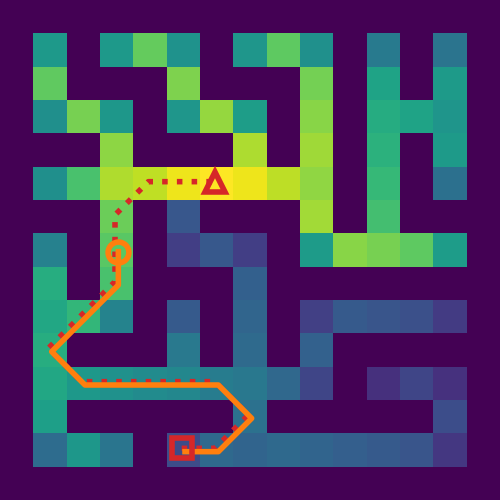}
  \caption{
    \textbf{(left)} A 2D maze, with the target in yellow.
    \textbf{(middle)} Values produced by the \glsshort{vin} for each 2D state (actions are taken towards the highest value). Higher values are brighter. The correct trajectory is dashed, and the current one is solid. The agent (orange circle) is stuck due to the local maximum below it.
    \textbf{(right)} Same values for \glsshort{calvin}. There are no spurious maxima, and the values of walls are correctly considered low (dark).
  }
  \label{fig:calvin/fully_observable}
\end{figure}

\subsubsection{Motivating experiment}
\label{sec:calvin/background/motivation}

Since the \gls{vin} allows all actions at all states ($\mathcal{A}$ does not depend on $s$), collisions must be modelled as states with low rewards.
In practice, the \gls{vin} does not learn to forbid such actions completely, resulting in the propagation of values from states which cannot be reached directly due to collision along the way.
This is verified experimentally by training a \gls{vin} according to Tamar \etal \cite{tamar_value_2016} on 4K mazes (see \Cref{sec:calvin/experiments} for details).
For each state, predicted scores are measured to identify whether the scores for all valid actions are larger than those for all invalid actions (\ie collisions).
Intuitively, this means the network always prefers free space over collisions.
Surprisingly, it turned out that this was not true for 24.6\% of the states.
For a real-world robot to work reliably, this is an unacceptably high chance of collisions.
As a comparison, for the proposed method \gls{calvin}, this rate is only 1.6\%.
In the same experiment, the \gls{vin} was often trapped in local minima of the value function and failed to move (\Cref{fig:calvin/fully_observable}), which is another failure mode.
\gls{calvin} addresses these issues, and push \glspl{vin} towards realistic scenarios.
An alternative would be to employ online retraining (DAgger)~\cite{ross2011reduction_dagger}, but this trades off the benefit of training solely on a fixed set of offline trajectories.

\section{Collision Avoidance Long-term Value Iteration Network}
\label{sec:calvin/method}

\gls{calvin} implements a transition model that accounts for illegal actions and termination. The state space is extended to rotations and incorporates distant observations into a spatial map of embeddings, both necessary for real 3D environments.

\subsection{Augmented navigation state-action space}\label{sec:calvin/augmented}
A probabilistic transition model is derived in this section from first principles, with only two assumptions and no extra hyper-parameters.
The first assumption is locality and translation invariance of the agent motion, which was introduced in the \gls{vin} to allow efficient learning with shared parameters.
Unlike with the \gls{vin}, the transition model $P(s'|s,a)$ is decomposed into two components: the agent motion model $\Ppredfunc$, which is translation invariant and shared across states (depending only on the spatial difference between states $s'-s$); and an observation-dependent predictor $\Apredfunc\in[0,1]$ that evaluates whether action $a$ is available from state $s$ to disqualify illegal actions.

In robotics, it is essential that the agent understands that the current task has been completed to move on to the next one.
Since there is a high chance that a randomly-acting agent will stumble upon the target in small environments, Anderson \etal \cite{anderson2018evaluation} suggested that an explicit termination action must be taken at the target to finish successfully.
Therefore, in addition to positional states, success and failure states are defined, where a success state $\Ssuccess$ (``win'') is reached only by triggering a termination action $\Afinished$ (``done''), and a failure state $\Sincorrect$ (``fail'') is reached upon triggering an incorrect action.
The reward for reaching $\Sincorrect$ is denoted as $\Rincorrect$, and the translation-invariant rewards is denoted as $\Rpredfunc$. For simplicity, the reward for success is considered as a special case of $\Rpredfunc$ where $a = \Afinished$ and $s' = \Ssuccess$. With these assumptions, the reward function $R(s, a, s')$ is:
\begin{equation}
R(s, a, s')=
\begin{cases}
    \Rpredincorrect, &s'= \Sincorrect,\\
    \Rpredfunc, &s'\neq \Sincorrect.
\end{cases}
\end{equation}
%
%
%
Together with the agent motion model $\Ppredfunc$, action validity predictor $\Apredfunc$ and the definition of a failure state $\Sincorrect$, the transition model $P(s' | s, a)$ can be derived as
\begin{equation}
P(s'|s,a)
=\begin{cases}
    1-\Apredfunc, &s'=\Sincorrect,\\
    \Apredfunc\Ppredfunc,  &s'\neq \Sincorrect.
\end{cases}
\end{equation}

From the above, the reward function $R(s, a)$ is evaluated by marginalising over the neighbour states $s'$:
\begin{align}
\begin{split}
\label{eq:calvin/method/reward_function}
R(s, a) &= \sum_{s'}{P(s'|s,a)R(s,a,s')}\\
&= \Rpredincorrect(1-\Apredfunc) + \Apredfunc\sum_{s'}{\Ppredfunc\Rpredfunc}.
\end{split}
\end{align}

Finally, \Cref{eq:calvin/method/reward_function} can be substituted into \Cref{eq:calvin/background/vi_grid} to obtain a proposed method for evaluating $Q(s,a)$:
\begin{align}
\begin{split}
\label{eq:calvin/method/proposed_update}
Q(s, a) &= R(s, a) + \gamma \mathbb{I}_{a\in \mathcal{A}}\sum_{s'}{P(s'|s,a)V(s')}\\
&= R(s, a) + \gamma \Apredfunc\mathbb{I}_{a\neq \Afinished}\sum_{s'}{\Ppredfunc V(s')},
\end{split}
\end{align}
where $\mathbb{I}$ is an indicator function.
\Cref{eq:calvin/method/reward_function,eq:calvin/method/proposed_update} essentially express a constrained \gls{vi}, which models the case of a \gls{mdp} on a grid with unknown illegal states and termination at a goal state.
The inputs to this model are three learnable functions ---
the motion model $\Ppredfunc$, the action validity $\Apredfunc$ (\ie obstacle predictions), and the rewards $\Rpredfunc$ and $\Rpredincorrect$.
These are implemented as \glspl{cnn} with the observations as inputs ($\Rpredincorrect$ is a single learned scalar).
All the constraints follow from a well-defined world model, with interpretable predicted quantities, unlike previous proposals~\cite{lee_gated_2018,tamar_value_2016}.
The resulting method is named \glsfirst{calvin}.

\subsubsection{Training}
\label{sec:calvin/method/train_loss}
Similarly to Tamar \etal \cite{tamar_value_2016}, the model is trained with a softmax cross-entropy loss $L$, comparing an example trajectory $\{(s_t, a^*_t):t\in T\}$ with predicted action scores $Q(s, a)$:
\begin{equation}
\label{eq:calvin/method/training}
    \min_{\Ppred,\,\Apred,\,\Rpred}\tfrac{1}{|T|}
    \textstyle\sum_{t\in T}w_t L\left(Q(s_{t}),\,a_{t}^{*}\right),
\end{equation}
where $Q(s)$ is a vector with one element per action ($Q(s,a)$), and $w_t$ is an optional weight that can be used to bias the loss if $w_t \neq 1$ (\Cref{sec:calvin/method/reweighting}).
Example trajectories are the shortest paths computed from random starting points to the target (also chosen randomly).
Note that these shortest paths are computable only during training time, since it is assumed that the ground-truth layout of the map and the obstacle/target positions are not readily available during test time.

The learned functions can be conditioned on input observations. These are 2D grids of features, with the same size as the considered state space (\ie a map of observations), which is convenient since it enables $\Ppred$, $\Apred$ and $\Rpred$ to be implemented as \glspl{cnn}.

\subsubsection{Transition modelling}
\label{sec:calvin/method/transition_loss}
Similarly to the \gls{vin} (\Cref{eq:calvin/background/vin}), the motion model $\Ppredfunc$ is implemented as a convolutional kernel $\mathbb{R}^{|\mathcal{A}|\times K_x \times K_y}$, where $K_x$ and $K_y$ are the x- and y- dimensions of the kernel, so it only depends on the relative spatial displacement $s'-s$ between the two states $s$ and $s'$.
Transitions already observed in the example trajectory can be used to constrain the model by adding a cross-entropy loss term $L(\widehat{P}(a_t^{*}),\,s_{t+1}-s_t)$ for each step in the example trajectory.
After training, the filter $\Ppredfunc$ will consist of a distribution over possible state transitions for each action (visualised in \Cref{fig:calvin/experiments/miniworld_motion_model}).

\subsubsection{Action availability}
\label{sec:calvin/method/action_availability}
Although the action availability $\Apred(s, a)$ could be learned completely end-to-end in theory, additional regularisation is necessary in practice.
Assuming a reliable log-probability of each action being taken ($\Apredvalid(s,a)$), available and unavailable actions can be distinguished by thresholding $\Apredvalid(s,a)$ at a learnt threshold $\Apred_\text{thresh}(s)$. Using a sigmoid function $\sigma$ as a soft threshold, $\Apred(s, a)$ could be written as:
\begin{equation}
    \Apred(s, a) = \sigma\left(\Apredvalid(s,a) - \Apred_\text{thresh}(s)\right).
\end{equation}

Both $\Apredvalid(s,a)$ and $\Apred_\text{thresh}(s)$ are predicted by the network, given the observations at each time step.
In order to ground the probabilities of each action being taken, the action logit $\Apredvalid(s_t)$ is trained to match the example trajectory action $a_t^{*}$ for all steps $t$, with an additional cross-entropy loss $L(\Apredvalid(s),\,a^{*})$.
Note that \emph{there is no additional ground truth supervision} -- \gls{calvin} strictly uses the same training data as a \gls{vin}.

\subsubsection{Fully vs. partially observable environments}\label{sec:calvin/method/partially_observable}
Some previous works~\cite{tamar_value_2016,lee_gated_2018} assume that the entire environment is fully observable, which is often unrealistic.
Partially observable environments, involving unknown scenes, require exploring to gather information and are thus more challenging.
This is accounted for in \gls{calvin} with a simple but significant modification.
Note that $Q(s_t)$ in \Cref{eq:calvin/method/training} depends on the learned functions (\glspl{cnn}) $\Ppred$, $\Apred$ and $\Rpred$ through \Cref{eq:calvin/method/proposed_update}, and these in turn are computed from the observations.
\gls{calvin} extends the \gls{vin} framework to the case of partial observations by ensuring that $Q(s_t| O_{\leq t})$ depends \emph{only} on the observations up to time $t$, $ O_{\leq t}$, which are aggregated into a map of observation embeddings $M_t$.
This means that the \gls{vin} recomputes a plan at each step $t$ (since the observations are different), instead of once for a whole trajectory.
Unobserved locations have their features set to zero, so in practice knowledge of the environment is slowly built up during an expert demonstration, which enables exploration behaviour to be learned.
Finally, because it is not possible to infer an optimal move in a region that was not observed yet, loss terms for unexplored cells are not considered in \Cref{eq:calvin/method/training}.

\subsubsection{Trajectory reweighting}
\label{sec:calvin/method/reweighting}
Exploration provides observations $ O_{\leq t}$ of the same locations at different times (\Cref{sec:calvin/method/partially_observable}).
Thus, \Cref{eq:calvin/method/training} can be augmented with these partial observations as extra samples.
The sum in \Cref{eq:calvin/method/training} becomes $\sum_{t'\in T} \sum_{t\in T_{1:t'}} w_t L\left(Q(s_{t}| O_{\leq t'}),\,a_{t}^{*}\right)$.
These constitute extra samples for \Cref{eq:calvin/method/training}, without needing more annotations.
For long trajectories, the training data then becomes severely imbalanced between the exploration phase (target is not visible, hence the agent explores the environment) and goal completion phase (target is visible, hence the agent directly navigates towards it), with a large proportion of the former ($90\%$ of the data in \Cref{sec:calvin/experiments/partially_observable}).
This is addressed in \gls{calvin} by reweighting the samples.
The weights are set such that $w_{t}=\beta^{d_{t}}/\max_{j\in T}\beta^{d_{j}}$ in \Cref{eq:calvin/method/training}, \ie a geometric decay scaling with the topological distance to the target $d_t$.
Since expert trajectories are shortest paths, this simplifies to $w_{t}=\beta^{|T|-t}$.

\subsection{Embodied navigation in 3D environments}
Embodied agents such as robots have a pose in 3D space. This work considers a particular category of robots that operate in Special Euclidean group 2, which means that the agent can be both translated and rotated. While holonomic robots can navigate omnidirectionally, this is not the case for non-holonomic robots, and their navigation actions are constrained relative to their orientation (\eg moving forward or rotating).
They may also have a limited sensor suite, such as a perspective camera aimed in one direction, which will limit their observations.

\subsubsection{Embodied pose states (position and orientation)}
\label{sec:calvin/method/orientations}
To address the limitation on the available action space (e.g. only able to move forward or rotate, rather than navigating omnidirectional), the 2D state space is augmented with an extra dimension, which corresponds to $\Theta$ discretised orientations: $\mathcal{S}= \{1,\ldots,\Theta\} \times \{1,\ldots,X\} \times \{1,\ldots,Y\}$.
This can be achieved by directly adding one extra dimension to each spatial tensor in \Cref{eq:calvin/method/reward_function,eq:calvin/method/proposed_update}.
Likewise, the loss (\Cref{eq:calvin/method/training}) must compute the cross-entropy over the new action space $\mathcal{A}$ of size $\Theta \times K_x \times K_y$, which includes both transitions in rotation as well as spatial translation.
\Cref{table:embodied} shows correspondences between tensor dimensions of the positional method and the embodied method for each component of the architecture.

\begin{table}[t]
\centering
\caption{Comparison of individual components in the implementation of \glsshort{calvin} for positional states and for embodied pose states. $X$ and $Y$ are the size of the internal spatial discretisation of the environment, $\Theta$ is the internal discretisation of the orientation, $A$ is the number of actions, $K_x$ and $K_y$ are the kernel dimensions for spatial locality, and $M$ is the map of embeddings given as input with channel dimension $C$.}
\begin{tabular*}{\columnwidth}{@{\extracolsep{\fill}}lll}
 \toprule
 & Positional & Position + Orientation \\ [0.5ex] 
 \midrule
 State $s$ & $(x,y)$ & $(\theta, x, y)$ \\
 Map of embeddings $M$ & $C \times X \times Y$ & $C \times X \times Y$ \\
 VI step & Conv2d & Conv3d \\
 \midrule
 $V(s)$ & $X \times Y$ & $\Theta \times X \times Y$ \\
 $Q(s,a)$ & $A \times X \times Y$ & $A \times \Theta \times X \times Y$ \\
 $\Apredfunc$ & $A \times X \times Y$ & $A \times \Theta \times X \times Y$ \\
 $\Ppred$ & $A \times K_x \times K_y$ & $A \times \Theta \times \Theta \times K_x \times K_y$ \\
 $\Rpred$ & $A \times K_x \times K_y$ & $A \times \Theta \times \Theta \times K_x \times K_y$ \\ [1ex] 
 \bottomrule
\end{tabular*}
\label{table:embodied}
\end{table}

The \gls{vi} step in \gls{calvin} performs a 2D convolution of $\Ppred$ over a 2D value map in the case of positional states and a 3D convolution over a 3D value map with orientation in the case of embodied pose states. In the embodied case, the second dimension of $\Ppred$ corresponds to the orientation of the current state, and the third dimension corresponds to that of the next state.

Note that the much larger state space makes long-term planning more difficult to learn.
It can be observed that when naively augmenting the state space in this way, the models fail to learn correct motion kernels $\Ppredfunc$.
This further reinforces the need for an auxiliary motion loss (\Cref{sec:calvin/method/transition_loss}), which overcomes this obstacle, and may explain why prior works did not plan through fine-grained rotations.

\subsubsection{3D embeddings for geometric reasoning}\label{sec:calvin/method/lpn}
While the \gls{vin} performs its operations on a grid map -- a discretisation of an absolute (world-centric) reference frame -- generally an agent only has access to observations in a local (camera-centric) reference frame.
Since the learnable functions ($\Ppred$, $\Apred$ and $\Rpred$) in \gls{calvin} (and other \gls{vin}-based methods) are \glspl{cnn}, their natural input is a grid of $m$-dimensional embeddings, denoted $e_{thij}\in\mathbb{R}^{m}$, for time $t$ and discrete world-space coordinates $(h,i,j) \in \mathbf{Z}^{Z \times X \times Y}$, where $Z$ is a discretisation of the height in 3D space. A resulting map of embeddings $M_t$ is a spatio-temporal tensor of dimensions $(m \times Z) \times X \times Y$. \gls{cnn} embeddings $\phi(I_{t})$ obtained from images $I_{t}$ in first-person view must be projected and aggregated into this 3D lattice map where \gls{vin} performs planning. A method named \gls{lpn} is proposed in this work to accomplish the above. 

Each embedding is associated with a 3D point in world-space via projective geometry, assuming that the camera position $c_{t}$ and rotation matrix $R_{t}$ are known (as assumed in prior work \cite{gupta_cognitive_2017,lee_gated_2018,tamar_value_2016,cartillier2021semantic,lenton2021endtoend}, which can be estimated from monocular vision \cite{mur-artal_orb-slam2_2017}).
Spatial projection also requires knowing (or estimating) the depths $d_{t}(p)$ of each pixel $p$ in $I_{t}$ (either with a RGB-D camera as in this work, or monocular depth estimation~\cite{fu2018deep}). 

The homogenous 3D coordinates of each pixel $p = (p_1, p_2)$ in the absolute reference frame can be written using projective geometry~\cite{hartley2003multiple}:
\begin{equation}
\left[x_{t}(p),\,y_{t}(p),\,z_{t}(p),\,1\right]=c_{t}+R_{t}K\left[p_{1},p_{2},d_{t}(p),1\right]^{\top},
\end{equation}
where $K$ is the camera's intrinsics matrix. Given these absolute coordinates of pixels $p$, for each cell $(h, i, j)$ in a 3D lattice map $M_t$, \gls{cnn} embeddings $\phi\left(I_{t}\right)(p)$ of pixels close to the lattice cell is aggregated with the current map embedding $e_{thij}$. 
Inspired by PointNet~\cite{charles_pointnet_2017}, the embeddings of 3D points \emph{from all past frames} that fall into each cell of the world-space lattice are aggregated with mean-pooling. This framework of spatial aggregation can be easily extended to work spatio-temporally, aggregating information from past frames $t'\leq t$. More formally:
\begin{align}
\begin{split}
e_{thij} =\avg_{t'\leq t} \{ \phi_{p}(I_{t'}):~
 &\tau_x i\leq x_{t'}(p)<\tau_x(i+1),\\
 &\tau_y j\leq y_{t'}(p)<\tau_y(j+1),\\
 &\tau_z h\leq z_{t'}(p)<\tau_z(h+1), p\in I_{t'}\},\label{eq:calvin/method/lpn}
\end{split}
\end{align}
where $\tau_x$, $\tau_y$ and $\tau_z$ are the absolute dimensions of each grid voxel,
``avg'' averages the elements of a set,
and $\phi_{p}\left(I_{t'}\right)$ retrieves the CNN embedding of image $I_{t'}$ for pixel $p$. 

As opposed to the PointNet's unstructured multi-layer perceptrons~\cite{charles_pointnet_2017}, the \gls{lpn} takes advantage of the spatial structure by applying \glspl{cnn} for feature extraction. It also maintains a spatially meaningful lattice output with a causal temporal constraint ($t'\leq t$), which is beneficial for downstream \gls{cnn} predictors: ($\Ppred(e_{t})$, $\Apred(e_{t})$ and $\Rpred(e_{t})$). 

A related proposal for \gls{slam} used spatial max-pooling but with more complex \glspl{lstm} / \glspl{gru} for temporal aggregation \cite{henriques_mapnet_2018,cartillier2021semantic}. Another related work on end-to-end trainable spatial embeddings uses ego-spherical memory 
\cite{lenton2021endtoend}.

\subsubsection{Memory-efficient mapping} 
The \gls{lpn} has some appealing properties in the context of navigation: 1) it allows reasoning about far away, observed but yet unvisited locations; 2) it fuses multiple observations of the same location, whether from different points-of-view or different times.

Temporal aggregation during rollout can be computed recursively as $e_{t,h,i,j} / n_{t,h,i,j}$ for
$e_{t,h,i,j}=e_{t-1,h,i,j}+e'_{t,h,i,j}$ and $n_{t,h,i,j}=n_{t-1,h,i,j}+n'_{t,h,i,j}$, where $e'_{t,h,i,j}$
is the summed embedding for the points in cell $(h,i,j)$ at time $t$, and $n'_{t,h,i,j}$ is the number of points per cell.
Only the previous map $e_{t-1,h,i,j}$ and previous counts $n_{t-1,h,i,j}$ must be kept, not all past observations.
Thus at run-time the memory cost is \emph{constant} over time, allowing unbounded operation (unlike methods that do not have an explicit map~\cite{savinov_semi-parametric_2018}).

\section{Experiment setup}
\subsection{Evaluation benchmarks}
The capabilities of \gls{calvin} are tested on increasingly challenging environments, leading up to unseen 3D environments emulating the real world. The measured metric of performance is the navigation success rate (fraction of trajectories that reach the target) in unseen environments.

\subsubsection{Randomly generated 2D maze environments}
Two types of 2D grid environments are considered: obstacle maps similar to the ones used in \cite{tamar_value_2016}, and mazes generated using Wilson's algorithm \cite{wilson_generating_1996} (examples shown in \Cref{fig:calvin/fully_observable,fig:calvin/partially_observable,fig:calvin/experiments/embodied}). 
The maze environments are more challenging due to their high branching complexity, especially with local observability, where backtracking is required. Hence, the focus of this work is on maze environments. Implements of both environments are available in an open-sourced code repository accompanying this work (\Cref{sec:calvin/implementation}).

\gls{calvin} is tested against other \gls{vin}-based baselines. Realistic constraints of limited observability and embodied navigation with orientation are progressively introduced.

\subsubsection{MiniWorld}
\label{sec:calvin/benchmark/miniworld}

MiniWorld~\cite{gym_miniworld} is a minimalistic 3D interior environment simulator for \gls{rl} and robotics research. It can be used to simulate environments with rooms, doors, hallways and various objects. In particular, randomly generated maze environments of $3\times 3$ and $8\times 8$ are used for evaluating the agents in a 3D setting with first-person view RGB-D observations from a monocular camera.

\subsubsection{Active Vision Dataset}
\label{sec:calvin/benchmark/avd}

\glsfirst{avd}~\cite{ammirato2017dataset} is a dataset of over $30K$ images of indoor environments with objects in the scene labelled with bounding boxes. Images are densely collected from a monocular RGB-D camera onboard a robot at $30$~cm intervals and at \ang{30} rotations in 19 unique household and office environments.
It could be used for benchmarking object recognition methods and simulating motion in real-world environments by creating arbitrary trajectories connecting the poses of the robot and associated observations that are present in the dataset.
This allows interactive navigation with real image streams, without synthetic rendering (as opposed to~\cite{habitat19iccv}).

\subsection{Baselines and hyper-parameter choices}
\label{sec:calvin/baselines}

\gls{calvin} is compared against the standard \gls{vin} by Tamar \etal \cite{tamar_value_2016}, as well as against \gls{gppn}~\cite{lee_gated_2018} that uses an \gls{lstm} as a recurrent operation to propagate values. 

Similarly to \gls{vin}~\cite{tamar_value_2016} which uses a 2-layer \gls{cnn} to predict the reward map, and GPPN~\cite{lee_gated_2018}, which uses a 2-layer \gls{cnn} to produce inputs to the \gls{lstm}, \gls{calvin} uses a 2-layer \gls{cnn} as an available actions predictor $\Apredfunc$. For each experiment, the size of the hidden layer was chosen from $\{40, 80, 150\}$. $150$ was used for all the grid environments, $80$ for MiniWorld and $40$ for \gls{avd}, partially due to memory constraints. 

\gls{vin} has an additional hyperparameter for the number of hidden action channels, which is set to $40$, sufficiently bigger than the number of actual actions in all the experiments.
While the kernel sizes $K$ for \gls{vin} and \gls{calvin} were set to $3$ for experiments in the grid environment, it was noted in \cite{lee_gated_2018} that \gls{gppn} works better with larger kernel size. Therefore, the best kernel size is chosen out of $\{3, 5, 7, 9, 11\}$ for \gls{gppn}. For experiments on MiniWorld and \gls{avd}, there are state transitions with step size of $2$, hence a kernel size of $K=5$ is chosen for \gls{vin} and \gls{calvin}.

The number of value iteration steps $k$ was chosen from $\{20, 40, 60, 80, 100\}$. For trajectory reweighting, $\beta$ was chosen from $\{0.1, 0.25, 0.5, 0.75, 1.0\}$. 

\subsection{Expert trajectory generation}
Expert trajectories are generated by running an A*~\cite{a_star} planner from the start state to the target state. Euclidean costs are assigned to every transition in the 2D grid environments, and a cost of $1$ per move for the MiniWorld and \gls{avd} environments. In the case of MiniWorld, an additional cost is assigned to locations near obstacles to ensure that the trajectories are not in close proximity to the walls. 

\subsection{Inference time rollouts}
The performance of the model is tested by running navigation trials (rollouts) on a randomly generated environment. At every time step, the model is queried the set of $Q$-values $\{Q(s, a): a \in \mathcal{A}\}$ for the current state $s$, and an action which gives the maximum predicted $Q$-value is taken.

While the \gls{vin} is trained with $V^{(0)}$ initialised with zeros, in a true \gls{vi} algorithm, the value function must converge for an optimal policy to be obtained.
To help the value function converge faster under a time and compute budget, the value function is initialised with predicted values from the previous time step at test time with online navigation.

A limit is set to the maximum number of steps taken by the agent, which were $200$ for the fully known $15 \times 15$ grid, $500$ for the partially known grid, $300$ for MiniWorld $(3 \times 3)$, $1000$ for MiniWorld $(8 \times 8)$, and $100$ for \gls{avd}.

\subsubsection{Architectural design of Lattice PointNet}
\label{sec:calvin/method/lpn_design}

The \gls{lpn} consists of three stages: a \gls{cnn} that extracts embeddings from observations in image-space (image encoder), a spatial aggregation step (\Cref{eq:calvin/method/lpn}) that performs mean pooling of embeddings for each map cell, and another \gls{cnn} that refines the map embedding (map encoder). The image encoder consists of two \gls{cnn} blocks, each consisting of the following layers in order: optional group normalisation, 2D convolution, dropout, \gls{relu} and 2D max pooling. The map encoder consists of 2D convolution, dropout, \gls{relu}, optional group normalisation, and finally, another 2D convolution. The number of channels of each convolutional layer are $(80, 80, 80, 40)$ for MiniWorld and $(40, 40, 40, 20)$ for \gls{avd} respectively.
The point clouds can consume a significant amount of memory for long trajectories. Hence, the most recent $40$ frames are used for the $8\times8$ MiniWorld maze.

The input to the \gls{lpn} is a 3-channel RGB image for the MiniWorld experiment, and a 128-channel embedding extracted using the first 2 blocks of ResNet18 pre-trained on ImageNet for the \gls{avd} experiment.

\subsubsection{Architectural design of the CNN backbone}
\label{sec:calvin/method/cnn_backbone_design}

A \gls{cnn} backbone is used in a control experiment in \Cref{sec:calvin/experiments/cnn_backbone} to show the effectiveness of the \gls{lpn} backbone. In contrast to \gls{lpn} which performs spatial aggregation of embeddings, the \gls{cnn} backbone is a direct application of an encoder-decoder architecture that transforms image-space observations into map-space embeddings. Gupta \etal \cite{gupta_cognitive_2017} employed a similar architecture to obtain their map embeddings. While they use ResNet50 as the encoder network, a simple \gls{cnn} is used in the control experiment in \Cref{sec:calvin/experiments/cnn_backbone} to match the result obtained with \gls{lpn} in the MiniWorld experiment (\Cref{sec:calvin/experiments/miniworld}).

The \gls{cnn} backbone consists of three stages: a \gls{cnn} encoder, two fully-connected layers with \gls{relu} to transform embeddings from image-space to map-space, and a \gls{cnn} decoder. The encoder consists of 3 blocks of batch normalisation, 2D convolution, dropout, \gls{relu} and 2D max pooling, and a final block with just batch normalisation and 2D convolution. The number of channels of each convolutional layer is $(64, 128, 128, 128)$, respectively. 

The fully-connected layers take in an encoder output of size $128 \times 5 \times 7$, reduce it to a hidden size of $128$, and output either $128 \times 5 \times 5$ for the smaller maze or $128 \times 4 \times 4$ for the larger maze, which is then passed to the decoder.

The decoder consists of 3 blocks of batch normalisation, 2D deconvolution, dropout and \gls{relu}, and a final block with just 2D deconvolution. The number of channels of each deconvolution layer is $(128, 128, 64, 20)$, respectively.
The output size of the decoder depends on the map resolution, hence appropriate strides, kernel sizes and paddings are chosen for the decoder network to match the output sizes of $30 \times 30$ and $80 \times 80$. This approach is not scalable to maps with high resolution or with arbitrary size, which is one of the drawbacks of this approach. 

\section{Experiments}
\label{sec:calvin/experiments}

\subsection{Fully observable 2D grid environment}
\label{sec:calvin/experiments/gridworld}

\gls{calvin} was first evaluated on 2D environments with positional states, where the observations are top-down views of the whole scene, and which thus do not require dealing with perspective images (\Cref{sec:calvin/method/lpn}).
Since Tamar \etal\cite{tamar_value_2016} obtain near-perfect performance in their 2D environments, after reproducing their results, evaluation was primarily performed on 2D mazes, which are much more challenging since they require frequent backtracking to navigate if exploration is required. Mazes of size $15\times 15$ mazes are used.
The allowed moves $\mathcal{A}$ are to any of the 8 neighbours of a cell.
As discussed in \Cref{sec:calvin/method}, a termination action $D$ must be triggered at the target to successfully complete the task.
The target is placed in a free cell chosen uniformly at random, with a minimum topological distance from the (random) start location equal to the environment size to avoid trivial tasks.
This was necessary to avoid trivial tasks (\ie starting close to the target).


\subsubsection{Baselines and training.}
For the first experiment, \gls{calvin} was compared with other differentiable planners: the \gls{vin}~\cite{tamar_value_2016} and the more recent \gls{gppn}~\cite{lee_gated_2018}, on fully-observed environments.
Other than using mazes instead of convex obstacles, this setting is close to Tamar \etals~\cite{tamar_value_2016}.
The \gls{vin}, \gls{gppn} and \gls{calvin} all use 2-layer \glspl{cnn} to predict their inputs, as described in \Cref{sec:calvin/baselines}.
All networks are trained with $4K$ example trajectories in an equal number of different mazes, using the Adam optimiser with the optimal learning rate chosen from $\{0.01, 0.005,0.001\}$, until convergence (up to 30 epochs). Navigation success rates (fraction of trajectories that reach the target) for epochs with minimum validation loss are reported.
Reweighted loss (\Cref{sec:calvin/method/reweighting}) is equally applicable to all differentiable planners, so results are reported both with and without it.

\begin{table}
\caption{Navigation success rate (fraction of trajectories that reach the target) on unseen 2D mazes. Partial observations (exploring an environment gradually) and embodied navigation (translation-rotation state space) are important yet challenging steps towards full 3D environments.
}
\centering\footnotesize
\setlength{\tabcolsep}{2pt}
\begin{tabular*}{\columnwidth}{@{\extracolsep{\fill}}lcccccc}
\toprule 
 & \multicolumn{3}{c}{Standard loss} & \multicolumn{3}{c}{Reweighted loss (ours)}\tabularnewline
\cmidrule{2-4}\cmidrule{5-7} Environment & VIN & GPPN & CALVIN (ours) & VIN & GPPN & CALVIN (ours)\tabularnewline
\midrule 
Fully observable & 75.6\tpm{20.6} & 91.3\tpm{8.1} & \textbf{99.0}\tpm{1.0} & 77.5\tpm{26.6} & 96.6\tpm{4.0} & \textbf{99.7}\tpm{0.5} \tabularnewline
\midrule 
Partially observable & 3.6\tpm{0.6} & 8.5\tpm{3.5} & \textbf{48.0}\tpm{5.2} & 1.7\tpm{1.7} & 11.25\tpm{3.7} & \textbf{92.2}\tpm{1.3} \tabularnewline
\midrule 
Embodied & 11.0\tpm{1.0} & 14.5\tpm{2.1} & \textbf{90.0}\tpm{7.9} & 15.2\tpm{3.6} & 28.5\tpm{3.5} & \textbf{93.7}\tpm{6.2} \tabularnewline
\bottomrule
\end{tabular*}\label{tab:calvin/results_2d}
\end{table}

\subsubsection{Results}
\Cref{tab:calvin/results_2d} (first row) shows the navigation success rate, averaged over 3 random seeds (and the standard deviation).
The \gls{vin} has a low success rate, showing that it does not scale to large mazes. 
\gls{gppn} achieves a high success rate, and \gls{calvin} performs near-perfectly.

The low performance of \gls{vin} is likely due to the value function not being learnt correctly, as outlined in \Cref{sec:calvin/background/motivation}. A comparison of values learnt by the \gls{vin} and \gls{calvin} are shown in \Cref{fig:calvin/fully_observable}.

The high performance of \gls{gppn} may be explained by its higher capacity, as it contains a \gls{lstm} with more parameters.
Nevertheless, \gls{calvin} has a more constrained architecture, so its higher performance hints at a better inductive bias for navigation.
It is interesting to note that the proposed reweighted loss has a beneficial effect on all methods, not just \gls{calvin}.
With the correct data distribution, any method with sufficient capacity can fit the objective.
This shows that addressing the imbalanced nature of the data is an important, complementary factor.


\subsection{Partially observable 2D grid environment}\label{sec:calvin/experiments/partially_observable}
Next, the same methods are compared in unknown environments, where the observation maps only contain observed features up to the current time step (\Cref{sec:calvin/method/partially_observable}).
To simulate local observations, ray-casting is performed to identify cells that are visible from the current position, up to 2 cells away. No information about the location of the target is given until it is within view of the agent -- this corresponds to object class-based navigation~\cite{anderson2018evaluation}, \ie exploring to find an object of a given class.

\begin{figure}[t]
  \centering
  \includegraphics[width=0.33\columnwidth]{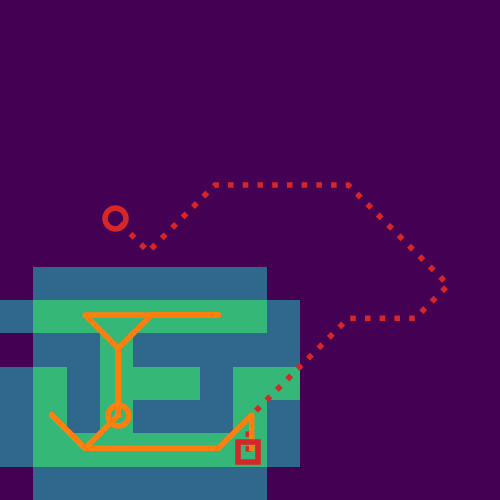}\hfill
  \includegraphics[width=0.33\columnwidth]{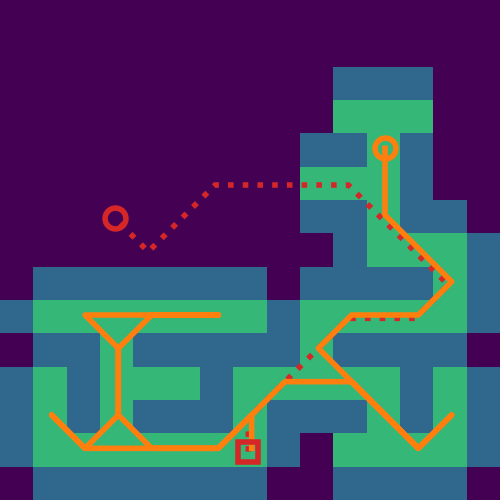}\hfill
  \includegraphics[width=0.33\columnwidth]{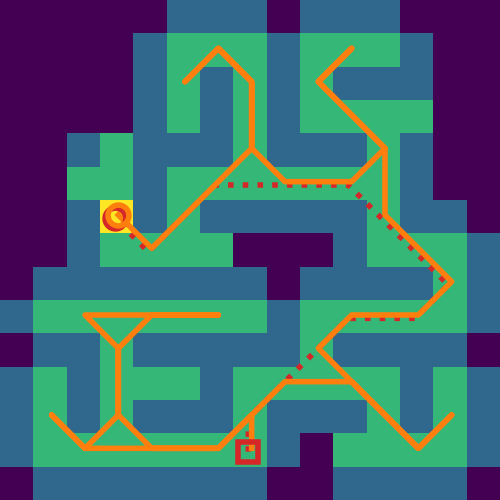}
  \includegraphics[width=0.33\columnwidth]{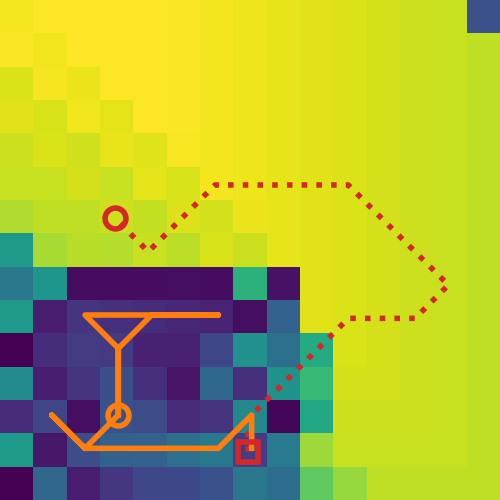}\hfill
  \includegraphics[width=0.33\columnwidth]{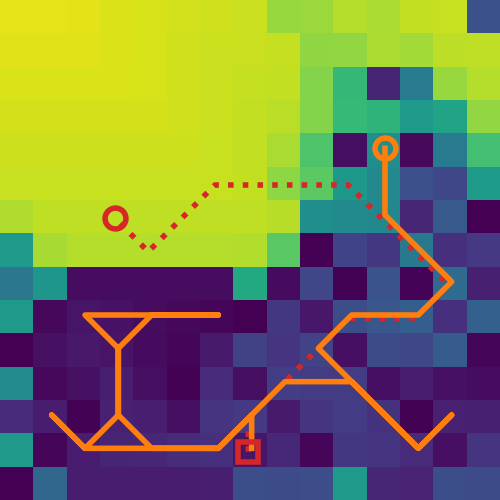}\hfill
  \includegraphics[width=0.33\columnwidth]{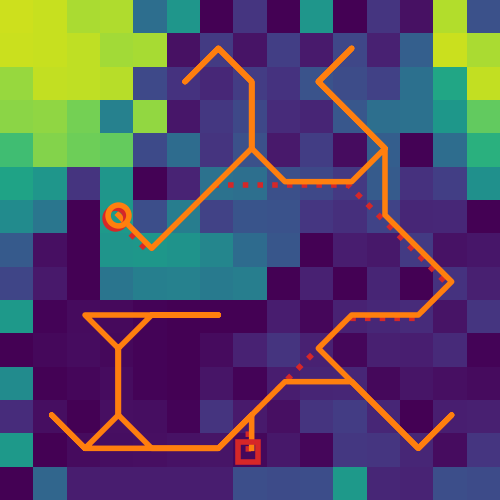}
  \caption{
    Example rollout of \glsshort{calvin} after $21$ steps (left column), $43$ steps (middle column) and $65$ steps (right column). \glsshort{calvin} successfully terminated at $65$ steps.
    \textbf{(top row)} Input visualisation: unexplored cells are dark, and the discovered target is yellow. The correct trajectory is dashed, and the current one is solid. The orange circle shows the position of the agent.
    \textbf{(bottom row)} Predicted values (higher values are brighter). Explored cells have low values, while unexplored cells and the discovered target are assigned high values. 
  }
  \label{fig:calvin/partially_observable}
\end{figure}

\subsubsection{Results}
In this case, the agent has to take significantly more steps to explore compared to a direct route to the target.
From \Cref{tab:calvin/results_2d} (2\textsuperscript{nd} row) it could be observed that partial observability causes most methods to fail catastrophically.
The sole exception is \gls{calvin} with the reweighted loss (proposed), which performs well.
Note that to succeed, an agent must acquire several complex behaviours: directing exploration to large unseen areas; backtracking from dead ends; 
and seeking the target when seen.
\gls{calvin} displays all of these behaviours.

\Cref{fig:calvin/partially_observable} shows a trajectory taken by \gls{calvin} at runtime, with corresponding observation maps and predicted values. At each rollout step, \gls{calvin} performs inference on the best action to take based on its current observation map.
No information about the location of the target is given until it is within view of the agent.
This makes the problem challenging, since the agent may have to take significantly more steps compared to an optimal route to reach the target.
In this example, the agent managed to backtrack every time it encountered a dead end, successfully reaching the target after 65 steps. 
The model initially assigns high values to all unexplored states. When the target comes into view, the model assigns a high probability to the availability of the ``done'' action at the corresponding state. The agent learns a sufficiently high reward for a successful termination so that the ``done'' action is triggered at the target.

Since only a combination of \gls{calvin} and a reweighted loss works at all, it could be inferred that a correct inductive bias and a balanced data distribution are both necessary for success.
One explanation is that the agent needs to explore, which typically is to navigate away from locations that it has previously observed, and the reweighting encourages such behaviour by placing larger weights closer to the target.
\gls{calvin} trained with the reweighted loss managed to backtrack when it encountered a dead end, successfully reaching the target.

\subsubsection{Ablation study of removing loss components}
\gls{calvin} is trained on three additive loss components: a loss term for the predicted Q-values $L_Q$ (\Cref{sec:calvin/method/train_loss}), a loss term for the transition models $L_P$ (\Cref{sec:calvin/method/transition_loss}), and a loss term for the action availability $L_A$ (\Cref{sec:calvin/method/action_availability}). The aim is to assess the contributions of each loss component to the overall performance.

Experiments are conducted on the partially observable grid environment. The results in \Cref{tab:calvin/ablation} indicate that all loss components, in particular the transition model loss, contribute to the robust performance of the network.

\begin{table}[h]
\caption{Navigation success rate of \glsshort{calvin} in partially observable 2D mazes with loss components removed.
}
\centering\footnotesize
\setlength{\tabcolsep}{2pt}
\begin{tabular*}{\columnwidth}{@{\extracolsep{\fill}}cccc}
\toprule 
Loss & $L_Q + L_P + L_A$ & $L_Q + L_P$ & $L_Q + L_A$ \tabularnewline
\midrule 
Success rate & 92.2 & 84.1 & 8.3\tabularnewline
\bottomrule
\end{tabular*}
\label{tab:calvin/ablation}
\end{table}

\subsection{Embodied navigation with orientation}\label{sec:calvin/experiments/embodied}
Next, the agents are evaluated on embodied navigation, in which transitions depend on the agent's orientation (\Cref{sec:calvin/method/orientations}).
The state space of all methods is augmented with 8 orientations at $\ang{45}$ intervals following \Cref{sec:calvin/method/orientations}, and the action space is defined as 4 move actions $\mathcal{A}$: forward, backward, turn left, and turn right.

\begin{figure}
  \centering
  \includegraphics[width=0.325\columnwidth]{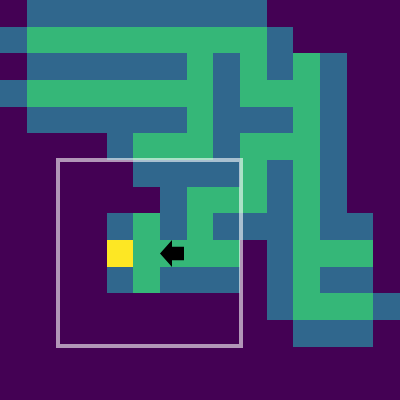}\hfill
  \includegraphics[width=0.325\columnwidth]{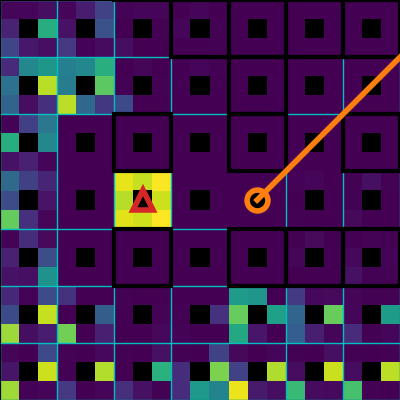}\hfill
  \includegraphics[width=0.325\columnwidth]{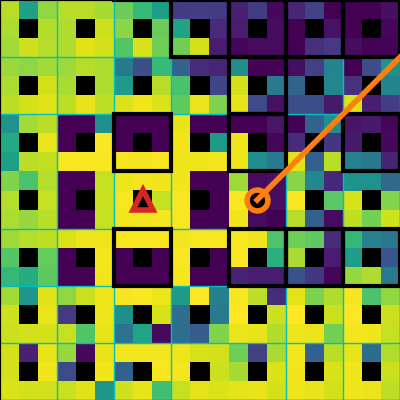}
  \caption{
    \glsshort{calvin}'s learnt rewards and values on partially observable 2D mazes with embodied navigation (\Cref{sec:calvin/experiments/embodied}).
    \textbf{(left)} Input visualisation: unexplored cells are dark, the target is yellow (just found by the agent), and a black arrow shows the agent's position and orientation.
    \textbf{(middle)} Close-up of predicted rewards (higher values are brighter) inside the white rectangle of the left panel. The 3D state space (position/orientation) is shown, with rewards for the 8 orientations in a radial pattern within each cell (position). Explored cells have low rewards, with the highest reward at the target.
    \textbf{(right)} Close-up of predicted values.
    They are higher facing the direction of the target.
    Obstacles (black border) have low values.
  }
  \label{fig:calvin/experiments/embodied}
\end{figure}

\subsubsection{Results}
\Cref{tab:calvin/results_2d} (3\textsuperscript{rd} row) shows that \gls{vin} and \gls{gppn} perform slightly better, but still have a low chance of success. \gls{calvin} outperforms them by a large margin.
A typical run is visualised in \Cref{fig:calvin/experiments/embodied} (refer to the caption for more detailed analysis).
A high reward is predicted for the (visible) target, with lower but still high rewards for unexplored regions; in this snapshot the agent just reached the target.
One advantage of \gls{calvin} displayed in \Cref{fig:calvin/fully_observable,fig:calvin/partially_observable,fig:calvin/experiments/embodied} is that values and rewards are fully interpretable and play the expected roles in \gls{vi} (\Cref{eq:calvin/background/vi_grid}).
Less constrained architectures~\cite{tamar_value_2016,lee_gated_2018} insert operators that deviate from the value iteration formulation, and thus lose their interpretability as rewards and values.

\begin{table}
\caption{Navigation success rate on unseen 3D mazes (MiniWorld).
Note that the baselines do not generalise to larger mazes.
}
\centering\footnotesize
\begin{tabular*}{\columnwidth}{@{\extracolsep{\fill}}ccccccc}
\toprule 
 & \multicolumn{2}{c}{CNN backbone} & \multicolumn{3}{c}{LPN backbone (ours)}\\
\cmidrule{2-3}\cmidrule{4-6} Size & A2C & PPO & VIN & GPPN & CALVIN (ours)\\
\midrule 
$3\times 3$ & \textbf{98.7}\tpm{1.9} & 81.0\tpm{26.9} & 90.3\tpm{3.1} & 91.3\tpm{4.7} & \textbf{97.7}\tpm{1.7} \\
\midrule 
$8\times 8$ & 23.6\tpm{4.9} & 14.7\tpm{6.2} & 41.2\tpm{9.5} & 33.3\tpm{8.6} & \textbf{69.2}\tpm{5.3} \\
\bottomrule
\end{tabular*}
\label{tab:calvin/experiments/miniworld}
\end{table}

\subsection{Synthetically-rendered 3D environments}
\label{sec:calvin/experiments/miniworld}
Having validated embodied navigation and exploration, \gls{calvin} is now integrated with the \gls{lpn} (\Cref{sec:calvin/method/lpn}) to handle first-person views of 3D environments.

\begin{figure}
  \centering
  \renewcommand*{\arraystretch}{0}
  \begin{tabular}{*{3}{@{}c}@{}}
  \includegraphics[height=3.6cm]{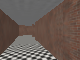} &
  \includegraphics[height=3.6cm]{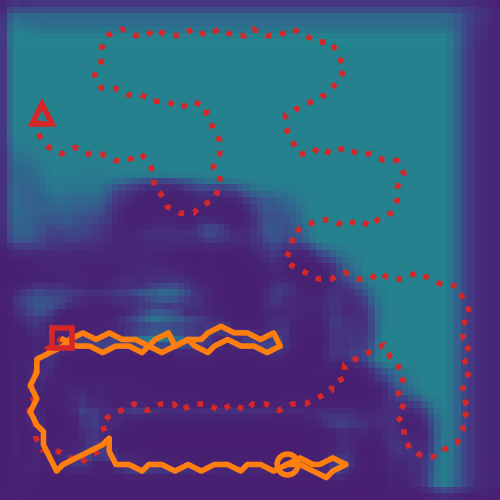} &
  \includegraphics[height=3.6cm]{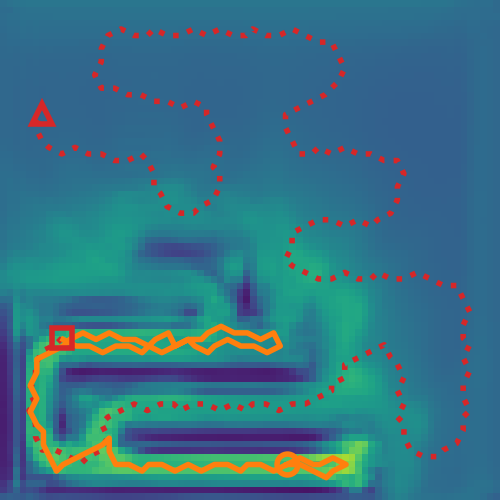} \\
  \includegraphics[height=3.6cm]{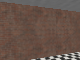} &
  \includegraphics[height=3.6cm]{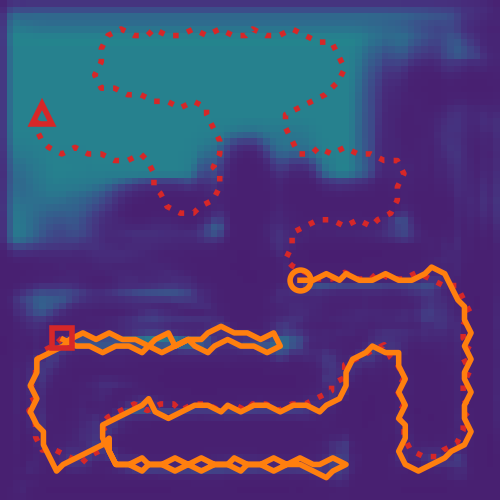} &
  \includegraphics[height=3.6cm]{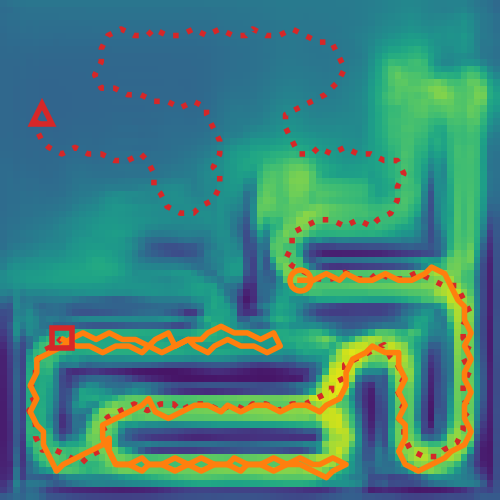} \\
  \includegraphics[height=3.6cm]{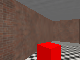} &
  \includegraphics[height=3.6cm]{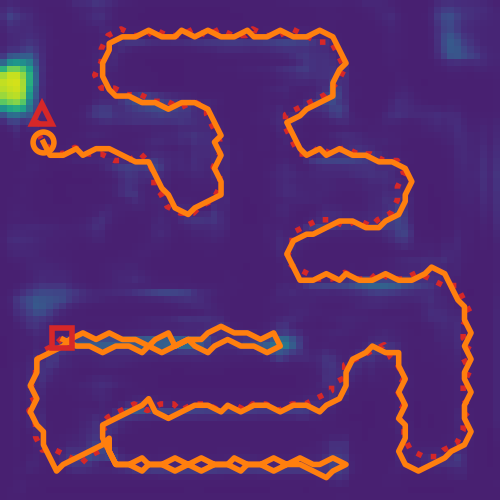} &
  \includegraphics[height=3.6cm]{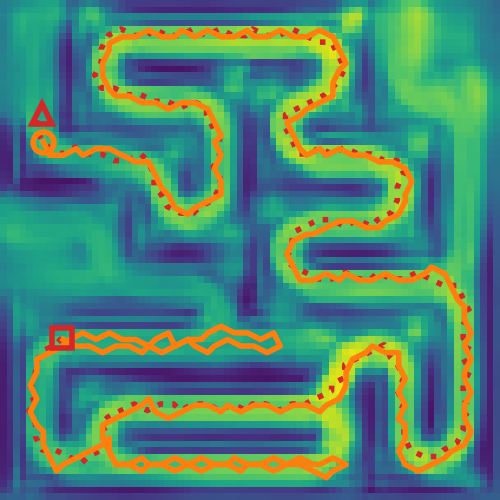}
  \end{tabular}
  \caption{
    Example results on MiniWorld (\Cref{sec:calvin/experiments/miniworld}). Left to right: input images, predicted rewards and values. The format is as in \Cref{fig:calvin/avd}. Notice the high reward on unexplored regions, replaced with a single peak around the target when it is seen (last row).
  }
  \label{fig:calvin/experiments/miniworld}
\end{figure}

The MiniWorld simulator~\cite{gym_miniworld} is used to generate 3D maze environments with arbitrary layouts (\Cref{sec:calvin/benchmark/miniworld}).
Only a monocular camera is considered (not \ang{360} views~\cite{lee_gated_2018}).
The training trajectories now consist of first-person videos of the shortest path to the target, visualised in \Cref{fig:calvin/experiments/miniworld}.
$1K$ random trajectories are generated in mazes of both small ($3\times 3$) and large ($8\times 8$) grids by adding or removing walls at the boundaries of this grid's cells.
Note that the maze's layout and the agent's location do not necessarily align with the 2D grids used by the planners (as in \cite{lee_gated_2018}).
Thus, planning happens on a fine discretisation of the state space ($30\times30$ for small mazes and $80\times80$ for large ones, with $8$ orientations). This allows smooth motions and no privileged information about the environment.
Both translation and rotation are perturbed by Gaussian noise, forcing all agents to model uncertain dynamics. 
To accommodate such probabilistic transitions, motion dynamics are modelled with a $5\times5$ kernel (see \Cref{sec:calvin/baselines}).

\subsubsection{Baselines and training}
Several baselines are compared: two popular \gls{rl} methods, \gls{a2c}~\cite{mnih_asynchronous_2016} and \gls{ppo}~\cite{schulman_proximal_2017}, as well as the \gls{vin}, \gls{gppn} and the proposed \gls{calvin}.
Since \gls{a2c} and \gls{ppo} are difficult to train if triggering the ``done'' action is strictly required, this assumption is relaxed, allowing the agent to terminate once it is near the target. 
All methods use as a first stage a simple 2-layer \gls{cnn} (details in \Cref{sec:calvin/method/cnn_backbone_design}).
Since this \gls{cnn} was not enough to get the \gls{vin}, \gls{gppn} and \gls{calvin} to work well, they all use the proposed \gls{lpn} backbone (\Cref{sec:calvin/method/lpn_design}).
It was not feasible to adopt \gls{gppn}'s strategy of taking views at all possible states as input~\cite{lee_gated_2018}, due to the high memory requirements and the environment not being fully visible (only a forward camera input of RGB-D is available).
Other training details are identical to \Cref{sec:calvin/experiments/gridworld}.

\subsubsection{Results}
Results in \Cref{tab:calvin/experiments/miniworld} indicate that although the \gls{rl} methods are successful in small mazes, their reactive policies do not scale to large mazes.
\gls{calvin} succeeds reliably even for longer trajectories, outperforming the others.
It is interesting to note that the proposed \gls{lpn} backbone is important for all differentiable planners, and it allows them to achieve very high success rates for small environments (though \gls{calvin} performs slightly better).
An example run of \gls{calvin} is shown in \Cref{fig:calvin/experiments/miniworld}. The \gls{calvin} agent found the target after an efficient exploration period, despite not knowing its location and never encountering this maze before.

\begin{figure}[h]
  \centering
  \includegraphics[width=0.12\columnwidth]{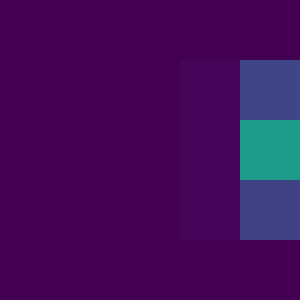}\hfill
  \includegraphics[width=0.12\columnwidth]{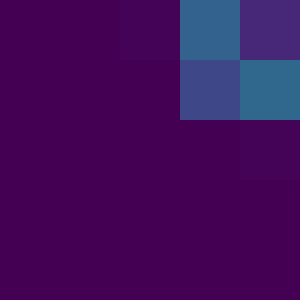}\hfill
  \includegraphics[width=0.12\columnwidth]{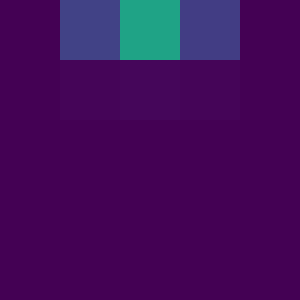}\hfill
  \includegraphics[width=0.12\columnwidth]{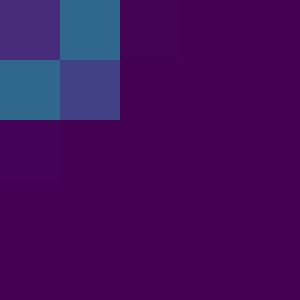}\hfill
  \includegraphics[width=0.12\columnwidth]{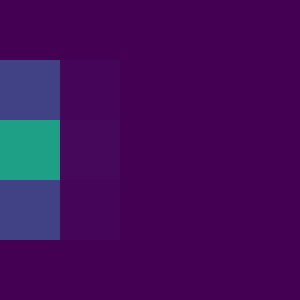}\hfill
  \includegraphics[width=0.12\columnwidth]{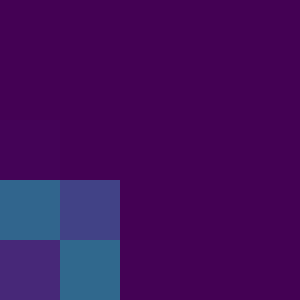}\hfill
  \includegraphics[width=0.12\columnwidth]{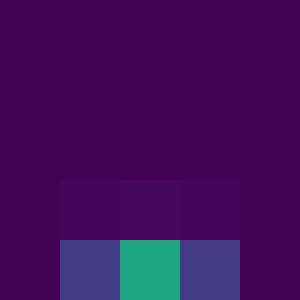}\hfill
  \includegraphics[width=0.12\columnwidth]{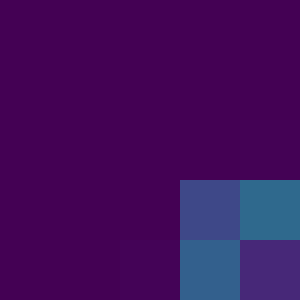}
  \caption{
    Transition model learnt from MiniWorld trajectories for the \textit{move forward} action at each discretised orientation, at $\ang{45}$ intervals. Higher values are brighter (yellow for a probability of 1), and lower values are darker (purple for a probability of 0).
  }
  \label{fig:calvin/experiments/miniworld_motion_model}
\end{figure}

\Cref{fig:calvin/experiments/miniworld_motion_model} visualises the learnt state transitions $\Ppredfunc$ for the \textit{move forward} action in CALVIN. It could be observed that the learning mechanism outlined in \Cref{sec:calvin/method/transition_loss} works even for transitions with added noise, which is the case for MiniWorld experiments. The network learns to propagate values probabilistically from the possible next states.


\subsubsection{Comparison of LPN against CNN backbone}
\label{sec:calvin/experiments/cnn_backbone}

In this ablation study, the proposed \gls{lpn} backbone is compared against a typical encoder-decoder \gls{cnn} backbone as a component that maps observations to map embeddings. The performance of the two methods is evaluated for \gls{vin}, \gls{gppn} and \gls{calvin}. It could be seen in \Cref{tab:calvin/experiments/miniworld_lpn_ablation} that the \gls{lpn} backbone is highly effective, especially for larger environments where long-term planning based on spatially aggregated embeddings is necessary.

\begin{table}[h]
\caption{Navigation success rate on unseen 3D mazes (MiniWorld), comparing the CNN backbone against the LPN backbone (also in \Cref{tab:calvin/experiments/miniworld}).
Most methods do not generalise to larger mazes.
The proposed \glsshort{lpn} demonstrates robust performance in larger unseen mazes. 
}
\centering\footnotesize
\begin{tabular*}{\columnwidth}{@{\extracolsep{\fill}}ccccccc}
\toprule 
 & \multicolumn{3}{c}{CNN backbone} & \multicolumn{3}{c}{LPN backbone (ours)}\\
\cmidrule{2-4}\cmidrule{5-7} Size & VIN & GPPN & CALVIN (ours) & VIN & GPPN & CALVIN (ours)\\
\midrule 
$3\times 3$ & 89.4 & 73.1 & 75.2 & 90.3 & 91.3 & \textbf{97.7} \\
\midrule 
$8\times 8$ & 0.6 & 18.3 & 8.6 & 41.2 & 33.3 & \textbf{69.2} \\
\bottomrule
\end{tabular*}\label{tab:calvin/experiments/miniworld_lpn_ablation}
\end{table}

\subsection{Indoor images from a real-world robot}\label{sec:calvin/experiments/avd}
Finally, \gls{calvin} is tested on real images obtained with a robotic platform.
The \gls{avd}~\cite{ammirato2017dataset} is used (\Cref{sec:calvin/benchmark/avd}).
Images can be composed to simulate any trajectory, up to some spatial granularity.
There are also bounding box annotations of object instances, which are used to evaluate semantic navigation.
The last four indoor environments in \gls{avd} are kept as a validation set, and the other 18 scenes are used for training (by sampling $1K$ shortest paths to the target).
A visualisation is shown in \Cref{fig:calvin/avd}.

\subsubsection{Tasks and training}
A semantic navigation task of seeking an object of a learnt class is considered. The most common class (``soda bottle'') is chosen as a target object.
The training follows \Cref{sec:calvin/experiments/miniworld}.

\subsubsection{Results}
Performances are reported for \gls{vin}, \gls{gppn} and \gls{calvin} after 8 epochs of training in \Cref{tab:calvin/experiments/avd}.
As in \Cref{sec:calvin/experiments/miniworld}, they also fail without the \gls{lpn}, so all results are with the \gls{lpn} backbone.
A similar conclusion to that for synthetic environments can be drawn: proper spatio-temporal aggregation of local observations is essential for differentiable planners to scale realistically.
\gls{calvin} achieves a significantly higher success rate on the training set than the other methods.
On the other hand, while it has higher mean validation success, the high variance of this estimate does not allow the result to be as conclusive as for training. This could be attributed to the small size of \gls{avd} in general, and of the validation set in particular, which contains only 3 different indoor scenes.
Nevertheless, this shows that \gls{calvin} learns effective generic strategies to seek a specific object, and that \gls{vin} and \gls{gppn} equipped with the \gls{lpn} backbone can achieve partial success in several training environments. \Cref{fig:calvin/avd} shows an example of a successful navigation sequence.

\begin{table}[t]
\caption{Navigation success rate on \glsshort{avd}, with real robot images taken in indoor spaces. The task is to navigate to an object of a learned class.
All methods use the proposed \glsshort{lpn} backbone, as they fail without it.}
\centering\footnotesize
\setlength{\tabcolsep}{2pt}
\begin{tabular*}{\columnwidth}{@{\extracolsep{\fill}}cccc}
\toprule 
Subset & VIN & GPPN & CALVIN (ours)\tabularnewline
\midrule 
Training & 61.6\tpm{4.5} & 50.6\tpm{9.2} & \textbf{70.3}\tpm{4.9} \tabularnewline
\midrule 
Validation & 45.0\tpm{1.0} & 44.0\tpm{3.5} & \textbf{47.6}\tpm{6.0} \tabularnewline
\bottomrule
\end{tabular*}\label{tab:calvin/experiments/avd}
\vspace{15pt}
\end{table}

\subsection{Implementation}
\label{sec:calvin/implementation}
The code used in this paper is open-sourced, and can be found at \url{https://github.com/shuishida/calvin}. This includes code for \gls{calvin} as well as baselines such as \gls{vin} and \gls{gppn}, and for different training environments (\ie grid world, MiniWorld and \gls{avd}).

\section{Conclusion}
This work proposed \gls{calvin}, which addresses limitations of current \gls{vin} implementations with an extended state-action space, an action availability mechanism, and additional constraints to the motion model.
Contributions include robust embodied navigation (position and orientation) in unexplored environments and 3D environments from a single camera, as well as a dense sampling scheme that achieves higher sample efficiency in some scenarios. 
The \gls{lpn} backbone was also proposed, which efficiently fuses spatio-temporal information in a differentiable way.

\gls{calvin} achieved high performance in challenging 3D navigation environments by learning transition models and rewards in a data-driven manner. All components of the model are end-to-end trainable, but they also have clear definitions of what they mathematically represent. 
This internal representation grants robustness to randomised and unknown environments, unlike reactive policies learnt by \gls{rl} agents.

\chapter{Option discovery via Expectation Maximisation and Policy Gradients}
\label{chapter:ppoem}

This chapter aims to propose ways to overcome the limitations of \gls{rl} algorithms designed for \glspl{mdp} with the means of \emph{options}. Options are temporally abstracted macro-actions that enable hierarchical planning and decision-making over multiple time steps. Each option is associated with a sub-policy which dictates how a sequence of primitive actions should be taken during the execution of the option.
Options also function as a memory that allows the agent to retain historical information beyond the policy's context window. While option assignment could be handled using heuristics and hand-crafted objectives, learning an optimal option assignment for any given task is an unsolved challenge. In this chapter, two algorithms, \gls{ppoem} and \gls{soap}, are proposed and investigated in depth to address this problem. 

The first approach, \gls{ppoem}, applies \gls{em} to a \gls{hmm} describing a \gls{pomdp} for the options framework. The method is an extension of the forward-backward algorithm, also known as the Baum-Welch algorithm~\cite{baum72}, applied to options. While this approach has previously been explored~\cite{option_ml_2016,fox2017multi,Zhang2020ProvableHI,online_baum_welch_2021}, these applications were limited to 1-step \gls{td} learning. In addition, the learnt options have limited expressivity due to how the option transitions are defined (see \Cref{sec:ppoem/background/baum_related_work}). In contrast, \gls{ppoem} augments the forward-backward algorithm with \gls{gae}, which is a temporal generalisation of \gls{td} learning, and extends the \gls{ppo}~\cite{schulman_proximal_2017} to work with options. While this approach was shown to be effective in a limited setting of a corridor environment requiring memory, the performance degraded with longer corridors. It could be hypothesised that this is due to the learning objective being misaligned with the true \gls{rl} objective, as the approach assumes access to the full trajectory of the agent for the optimal assignment of options, even though the agent only has access to its past trajectory (and not its future) at inference time. 

As an alternative approach, \gls{soap} evaluates and maximises the policy gradient for an optimal option assignment directly. With this approach, the option policy is only conditional on the history of the agent. The derived objective has a surprising resemblance to the forward-backward algorithm, but showed more robustness when tested in longer corridor environments.
The algorithms were also evaluated on the Atari~\cite{bellemare13arcade} and MuJoCo~\cite{todorov2012mujoco} benchmarks. 
Results demonstrated that using \gls{soap} for option learning is more effective and robust than using the standard approach for learning options, proposed by the Option-Critic architecture~\cite{optioncritic,optioncritic_ppo}. 

The proposed approach can improve the efficiency of skill discovery in skill-based \gls{rl} algorithms, allowing them to adapt efficiently to complex novel environments. The paper is available on \textit{arXiv}~\cite{ishida2024soaprl}. 


\section{Introduction}
While deep \gls{rl} has seen rapid advancements in recent years, with numerous real-world applications such as robotics~\citep{gu2017deep,akkaya2019solvingrubik,haarnoja2024learning}, gaming~\citep{hasselt_double_dqn_2015,arulkumaran2019alphastar,baker2022_openai_vpt}, and autonomous vehicles~\citep{kendall2019learning,lu2022imitation}, many algorithms are limited by the amount of observation history they condition their policy on, due to the increase in computational complexity. Developing learnable embodied agents that plan over a wide spatial and temporal horizon has been a longstanding challenge in \gls{rl}. 

With a simple Markovian policy $\pi(a_t|s_t)$, the agent's ability to make decisions is limited by only having access to the current state as input. Early advances in \gls{rl} were made on tasks that either adhere to the Markov assumption that the policy and state transitions only depend on the current state, or those can be solved by frame stacking~\citep{mnih_human-level_2015} that grants the policy access to a short history. 
However, many real-world tasks are better modelled as \glspl{pomdp}~\citep{ASTROM1965174_pomdp} with a long-term temporal dependency, motivating solutions that use a bounded working memory for computational scalability. 

For \gls{pomdp} tasks, the entire history of the agent's trajectory may contain signals to inform the agent to make a more optimal decision. This is due to the reward and next state distribution $p(r_t, s_{t+1} | s_{0:t}, a_{0:t})$ being conditional on the past states and actions, not just on the current state and action.

A common approach of accommodating \glspl{pomdp} is to learn a latent representation using sequential policies, typically using a \gls{lstm}~\cite{hochreiter_lstm_1997}, \gls{gru}~\cite{cho_gru_2014} or Transformer~\citep{vaswani2017attention}. This will allow the policy to gain access to signals from the past. However, this approach has an inherent trade-off between the duration of history it can retain (defined by the policy's context window size) and the compute and training data required to learn the policy. This is because the entire history of observations within the context window have to be included in the forward pass at training time to propagate useful gradients back to the sequential policy. Another caveat is that, with larger context windows, the input space is less constrained and it becomes increasingly unlikely that the agent will revisit the same combination of states, which makes learning the policy and value function sample-expensive, and potentially unstable at inference time if the policy distribution has changed during training.

Training \gls{rl} agents to work with longer working memory is a non-trivial task, especially when the content of the memory is not pre-determined and the agent also has to learn to store information relevant to each task. 
With the tasks that the \gls{rl} algorithms are expected to handle becoming increasingly complex~\citep{dulac2021challenges,milani2023solving,chen2023end}, there is a vital need to develop algorithms that learn policies and skills that generalise to dynamic and novel environments. Many real-world tasks are performed over long time horizons, which makes it crucial that the algorithm can be efficiently trained and quickly adapted to changes in the environment. This gives motivation to develop an algorithm that (a) can solve problems modelled as \gls{pomdp} using options, (b) has a bounded context length input for the policy and value function so that they can be trained more sample-efficiently, (c) only requires the current observation to be forward-passed through a neural network at training time to reduce the \gls{gpu} memory and computational requirements.

There has been considerable effort in making \gls{rl} more generalisable and efficient. Relevant research fields include \gls{hrl}~\citep{vezhnevets2017feudal,hiro2018,pateria2021hierarchical,zhang2021hierarchical}, skill learning~\citep{pertsch2020spirl,nam2022skillbased,ase_large_scale_reusable,shi2023skill}, \gls{meta-rl}~\citep{wang2016learning,duan2016rl,pearl_2019,metarl_survey} and the options framework~\cite{sutton1999between_options, Precup2000TemporalAI}, with a shared focus on learning reusable policies. In particular, this research focuses on the options framework, which extends the \gls{rl} paradigm with a \gls{hmm} that uses options to execute long-term behaviour.

Options are instrumental in abstract reasoning and high-level decision-making, since they enable temporal abstraction, credit assignment, and identification of skills and subgoals. Acquiring transferable skills and composing them to execute plans, even in novel environments, are remarkable human capabilities that are instrumental in performing complex tasks with long-term objectives. Whenever one encounters a novel situation, one can still strategise by applying prior knowledge with a limited budget of additional trial and error. One way of achieving this is by abstracting away the complexity of long-term planning by delegating short-term decisions to a set of specialised low-level policies, while the high-level policy focuses on achieving the ultimate objective by orchestrating these low-level policies. 

The Option-Critic architecture~\cite{optioncritic} presents a well-formulated solution for end-to-end option discovery. The authors showed that once the option policies are learned, the Option-Critic agent can quickly adapt when the environment dynamics are changed, whereas other algorithms suffer from the changes in reward distributions.

However, there are challenges with regard to automatically learning options. A common issue is that the agent may converge to a single option that approximates the optimal policy under a Markov assumption. Additionally, learning options from scratch can be less sample-efficient due to the need to learn multiple option policies. (The data-efficiency may be improved if the options are learned off-policy~\cite{wulfmeier2021data,salter2022mo2}. This work mostly focused on on-policy algorithms due to their robustness and ease of analysis.)

In the following sections, two candidate training objectives are proposed and derived to learn an optimal option assignment. The first objective, which will be referred to as \gls{ppoem}, is derived by using the forward-backward algorithm~\cite{baum72}, applied to a \gls{pomdp} according to the options framework. However, this objective is developed to assign latent variables for offline sequences, and is less suitable for \gls{rl} with on-policy rollouts. As an alternative training objective, \gls{soap} is proposed by adapting the policy gradient algorithm to work with options, which exhibits a resemblance to the forward-backward algorithm.

\section{Background}
\subsection{Option-Critic architecture}

As mentioned in \Cref{sec:background/options_framework}, the options framework~\cite{sutton1999between_options, Precup2000TemporalAI} formalises the idea of temporally extended actions that allow agents to make high-level decisions. Let there be $n$ discrete options $\{\mathcal{Z}_1, ..., \mathcal{Z}_n\}$ from which $z_t$ is chosen and assigned at every time step $t$. Each option corresponds to a specialised sub-policy $\pi_\theta(a_t | s_t, z_t)$ that the agent can use to achieve a specific subtask. At $t=0$, the agent chooses an option according to its inter-option policy $\pi_\phi(z_t|s_t)$ (policy over options), then follows the option sub-policy until termination, which is dictated by the termination probability function $\varpi_\psi(s_t, z_{t-1})$. Once the option is terminated, a new option $z_t$ is sampled from the inter-option policy and the procedure is repeated.

The Option-Critic architecture~\cite{optioncritic} learns option assignments end-to-end. It formulates the problem such that the option sub-policies $\pi_\theta(a_t | s_t, z_t)$ and termination function $\varpi_\psi(s_{t+1}, z_t)$ are learned jointly in the process of maximising the expected returns. 
The inter-option policy $\pi_\phi(z_t|s_t)$ is an $\epsilon$-greedy policy that takes an argmax $z$ of the option value function $Q_\phi(s, z)$ with $1-\epsilon$ probability, and uniformly randomly samples options with $\epsilon$ probability.
In every step of the Option-Critic algorithm, the following updates are performed for a current state $s$, option $z$, reward $r$, episode termination indicator $d \in \{0, 1\}$, next state $s'$, and discount factor $\gamma \in [0, 1)$:
\begingroup
\small
\begin{align}
\begin{split}
    \delta &\leftarrow r + \gamma (1 - d) \left[ \left (1 - \varpi_\psi(s', z) \right) Q(s', z) + \varpi_\psi(s', z) \max_z Q_\phi(s', z) \right] - Q_\phi(s, z),\\
    Q_\phi(s, z) &\leftarrow Q_\phi(s, z) + \alpha_\phi \delta,\\
    \theta &\leftarrow \theta + \alpha_\theta \frac{\partial \log \pi_\theta(a | s, z)}{\partial \theta} [r + \gamma Q_\phi(s', z)],\\
    \psi &\leftarrow \psi - \alpha_\psi \frac{\partial \varpi_\psi(s', z)}{\partial \psi} [Q_\phi(s', z) - \max_z Q_\phi(s', z)].
\end{split}
\end{align}
\endgroup
Here, $\alpha_\phi$, $\alpha_\theta$ and $\alpha_\psi$ are learning rates for $Q_\phi(s,z)$, $\pi_\theta(a | s, z)$, and $\varpi_\psi(s, z)$, respectively.

\gls{ppoc}~\cite{optioncritic_ppo} builds on top of the Option-Critic architecture~\cite{optioncritic}, replacing the $\epsilon$-greedy policy over the option-values with a policy network $\pi_\varphi(z | s)$ parametrised by $\varphi$ with corresponding learning rate $\alpha_\varphi$, substituting the policy gradient algorithm with \gls{ppo}~\cite{schulman_proximal_2017} to optimise the sub-policies $\pi_\theta(a | s, z)$, and introducing \gls{gae}~\cite{gae_schulman_2015} for the advantage estimate and value function updates. The policy loss function for \gls{ppo} is given as $\mathcal{L}_\text{PPO}(\theta)$ in \Cref{eq:background/ppo}. 
Extending the definition of \gls{gae} given in \Cref{sec:background/gae} to work with options,
\begingroup
\small
\begin{align}
\begin{split}
    A^\text{GAE}(s, z) &\leftarrow r + \gamma V(s', z') - V(s, z) + \lambda \gamma (1-d) A^\text{GAE}(s', z'),\\
    Q_\phi(s, z) &\leftarrow Q_\phi(s, z) + \alpha_\phi A^\text{GAE}(s, z),\\    
    \theta &\leftarrow \theta + \alpha_\theta \frac{\partial \mathcal{L}_\text{PPO}(\theta)}{\partial \theta},\\
    \psi &\leftarrow \psi - \alpha_\psi \frac{\partial \varpi_\psi(s, z)}{\partial \psi} A^\text{GAE}(s, z),\\
    \varphi &\leftarrow \varphi + \alpha_\varphi \frac{\partial \log \pi_\varphi(z | s)}{\partial \varphi} A^\text{GAE}(s, z).
\end{split}
\end{align}
\endgroup

\gls{ppoc} is used as one of the baselines in this work.

\subsection{Double Actor-Critic}
\gls{dac}~\cite{zhang2019dac} is an \gls{hrl} approach to discovering options. \gls{dac} reformulates the \gls{smdp}~\cite{sutton1999between_options} of the option framework as two hierarchical \glspl{mdp} (high-\gls{mdp} and low-\gls{mdp}). 

The low-\gls{mdp} concerns the selection of low-level actions within the currently active option.
Given the current state $s_t$, selected option $z_t$, and action $a_t$, the state and action spaces of the low-\gls{mdp} can be defined as $S_t^L = (s_t, z_t)$ and $A_t^L = a_t$. With this definition, the transition function, reward function and policy for the low-\gls{mdp} can be written as:
{
\small
\begin{align}
\begin{split}
    P_L(S_{t+1}^L | S_t^L, A_t^L) &= p((s_{t+1}, z_{t+1}) | (s_t, z_t), a_t) = P(s_{t+1} | s_t, a_t) \cdot p(z_{t+1} | s_{t+1}, z_t),\\
    R_L(S_t^L, A_t^L) &= r(s_t, a_t),\\
    \pi_L(A_t^L | S_t^L) &= \pi_L(a_t | s_t, z_t).
\end{split}
\end{align}
}

The high-\gls{mdp} handles high-level decision-making, such as selecting and terminating options. 
The state and action spaces of the high-\gls{mdp} can be defined as $S_t^H = (s_t, z_{t-1})$ and $A_t^H = z_t$. With this definition, the transition function, reward function and policy for the high-\gls{mdp} can be written as:
{
\small
\begin{align}
\begin{split}
    P_H(S_{t+1}^H | S_t^H, A_t^H) &= p((s_{t+1}, z_t) | (s_t, z_{t-1}), A_t^H) = \mathbb{I}_{A_t^H = z_t} p(s_{t+1} | s_t, z_t),\\
    R_H(S_t^H, A_t^H) &= r(s_t, z_t),\\
    \pi_H(A_t^H | S_t^H) &= \pi_H(z_t | s_t, z_{t-1}).
\end{split}
\end{align}
}

The key idea is to treat the inter-option and intra-option policies as independent actors in their respective \glspl{mdp}. Under this formulation, standard policy optimisation algorithms such as \gls{ppo} can be used to optimise the policies (high-level and low-level) for both \glspl{mdp}.

\subsection{Expectation Maximisation algorithm}

The \gls{em} algorithm~\cite{em_algorithm} is a well-known method for learning the assignment of latent variables, often used for unsupervised clustering and segmentation. The $k$-means clustering algorithm~\cite{kmeans_Forgy1965ClusterAO} can be considered a special case of \gls{em}. The following explanation in this section is a partial summary of Chapter 9 of Bishop's book~\cite{bishop_ml_book}. 

The objective of \gls{em} is to find a maximum likelihood solution for models with latent variables. Denoting the set of all observed data as $\bm{X}$, the set of all latent variables as $\bm{Z}$, and the set of all model parameters as $\Theta$, the log-likelihood function is given by:
\begin{equation}
    \log p(\bm{X} | \Theta) = \log \left\{ \sum_{\bm{Z}} p(\bm{X}, \bm{Z} | \Theta) \right\}.
\end{equation}

However, evaluating the above summation (or integral for a continuous $\bm{Z}$) over all possible latents is intractable. The \gls{em} algorithm is a way to strictly increase the likelihood function by alternating between the E-step that evaluates the expectation of a joint log-likelihood $\log p(\bm{X}, \bm{Z} | \Theta)$, and the M-step that maximises this expectation. 

In the E-step, the current parameter estimate $\Theta_\text{old}$ (using random initialisation in the first iteration, or the most recent updated parameters in subsequent iterations) is used to determine the posterior of the latents $p(\bm{Z} | \bm{X}, \Theta_\text{old})$. The joint log-likelihood is obtained under this prior. The expectation, denoted as $\mathcal{Q}(\Theta; \Theta_\text{old})$, is given by:
\begin{equation}
    \mathcal{Q}(\Theta; \Theta_\text{old}) = \expect_{\bm{Z}\sim p(\cdot | \bm{X}, \Theta_\text{old})} \left[ \log p(\bm{X}, \bm{Z} | \Theta) \right] = \sum_{\bm{Z}} p(\bm{Z} | \bm{X}, \Theta_\text{old}) \log p(\bm{X}, \bm{Z} | \Theta).
\end{equation}

In the M-step, an updated parameter estimate $\Theta_\text{new}$ is obtained by maximising the expectation:
\begin{equation}
    \Theta_\text{new} = \argmax_\Theta \mathcal{Q}(\Theta, \Theta_\text{old}).
\end{equation}

The E-step and the M-step are performed alternately until a convergence criterion is satisfied. The \gls{em} algorithm makes obtaining a maximum likelihood solution tractable.

\subsection{Forward-backward algorithm}
\label{sec:ppoem/forward_backward}
\begin{figure}[h]
  \centering
  \includegraphics[width=0.6\textwidth]{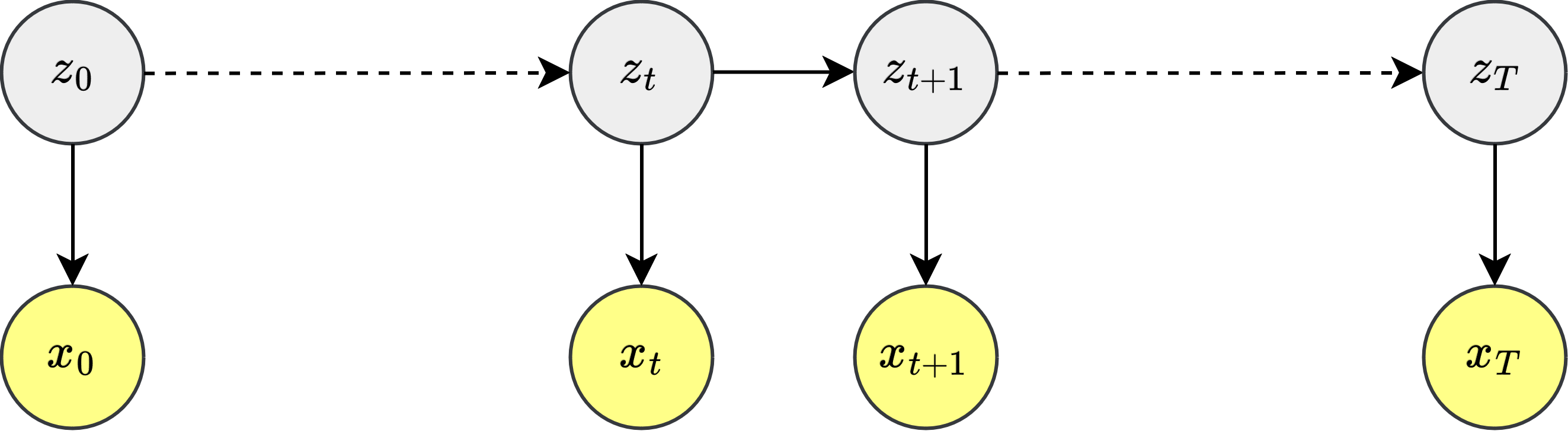}
  \caption{
    An \glsshort{hmm} for sequential data $\bm{X}$ of length $T$, given latent variables $\bm{Z}$.
  }
  \label{fig:ppoem/hmm}
\end{figure}

The \gls{em} algorithm can also be applied in an \gls{hmm} setting for sequential data, resulting in the forward-backward algorithm, also known as the Baum-Welch algorithm~\cite{baum72}. \Cref{fig:ppoem/hmm} shows the graph of the \gls{hmm} of interest. At every time step $t \in \{0, ..., T\}$, a latent $z_t$ is chosen out of $n$ number of discrete options $\{\mathcal{Z}_1, ..., \mathcal{Z}_n\}$, which is an underlying conditioning variable for an observation $x_t$. In the following derivation, $\{x_t | t_1 \leq t \leq t_2\}$ is denoted with a shorthand $x_{t_1:t_2}$, and similarly for other variables. Chapter 13 of Bishop's book~\cite{bishop_ml_book} offers a comprehensive explanation for this algorithm.

For this \gls{hmm}, the joint likelihood function for the observed sequence $\bm{X} = \{x_0, ..., x_{T}\}$ and latent variables $\bm{Z} = \{z_0, ..., z_{T}\}$ is given by:
\begin{equation}
    p(\bm{X}, \bm{Z} | \Theta) = p(z_0 | \Theta) \prod_{t=0}^T p(x_t | z_t, \Theta) \prod_{t=1}^T p(z_t | z_{t-1}, \Theta).
\end{equation}

Using the above, \gls{em} objective can be simplified as:
\begingroup
\small%
\begin{align}
\begin{split}
\label{eq:ppoem/em_objective_normal_hmm}
    \mathcal{Q}(\Theta; \Theta_\text{old}) =& \sum_{\bm{Z}} p(\bm{Z} | \bm{X}, \Theta_\text{old}) \log p(\bm{X}, \bm{Z} | \Theta)\\
    =& \sum_{z_0} p(z_0 | \Theta_\text{old}) \log p(z_0 | \Theta) + \sum_{t=0}^T \sum_{z_t} p(z_t | \bm{X}, \Theta_\text{old}) \log p(x_t | z_t, \Theta)\\
    &+ \sum_{t=1}^T \sum_{z_{t-1}, z_t} p(z_{t-1}, z_t | \bm{X}, \Theta_\text{old}) \log p(x_t, z_t | z_{t-1}, \Theta).
\end{split}
\end{align}
\endgroup

\subsubsection{E-step}
\label{sec:ppoem/e_step_normal_hmm}
In the E-step, $p(z_t | \bm{X})$ and $p(z_{t-1}, z_t | \bm{X})$ are evaluated. Note that in the following derivation, it is assumed that the probability distributions are conditioned on $\Theta$.
Defining $\alpha(z_t) \defeq p(z_t | x_{0:t})$, $\beta(z_t) \defeq \frac{p(x_{t+1:T} | z_t)}{p(x_{t+1:T} | x_{0:t})}$ and normalising constant $c_t \defeq p(x_t | x_{0:t-1})$,
\begingroup
\small%
\begin{align}
p(z_t | \bm{X}) 
&= \frac{p(x_{0:T}, z_t)}{p(x_{0:T})}= \frac{p(x_{0:t}, z_t) p(x_{t+1:T} | z_t)}{p(x_{0:t}) p(x_{t+1:T} | x_{0:t})} = \alpha(z_t)\beta(z_t),\\
p(z_{t-1}, z_t | \bm{X})
&= \frac{p(x_{0:T}, z_{t-1}, z_t)}{p(x_{0:T})}= \frac{p(x_{0:t-1}, z_{t-1}) p(x_t | z_t) p(z_t | z_{t-1}) p(x_{t+1:T} | z_t)}{p(x_{0:t-1}) p(x_t|x_{0:t-1}) p(x_{t+1:T} | x_{0:t})}\nonumber\\
&= p(x_t|z_t) p(z_t | z_{t+1}) \frac{\alpha(z_t)\beta(z_t)}{c_t}.
\end{align}
\endgroup

Recursively evaluating $\alpha(z_t)$, $\beta(z_t)$ and $c_t$,
\begingroup
\small%
\begin{align}
\alpha(z_t) &= \frac{p(x_{0:t}, z_t)}{p(x_{0:t})}
= \frac{p(x_t, z_t | x_{0:t-1})}{p(x_t | x_{0:t-1})} = \frac{\sum_{z_{t-1}} \left[ p(z_{t-1}|x_{0:t-1}) p(x_t | z_t) p(z_t | z_{t-1}) \right]}{p(x_t|x_{0:t-1})} \nonumber\\
&= \frac{p(x_t | z_t) \sum_{z_{t-1}} \left[ \alpha(z_{t-1}) p(z_t | z_{t-1}) \right]}{c_t}, \\
\beta(z_t) &= \frac{p(x_{t+1:T} | z_t)}{p(x_{t+1:T} | x_{0:t})} 
= \frac{\sum_{z_{t+1}} \left[p(x_{t+2:T} | z_{t+1}) p(x_{t+1} | z_{t+1}) p(z_{t+1} | z_t) \right]} {p(x_{t+2:T} | x_{0:t+1}) p(x_{t+1}|x_{0:t})} \nonumber\\
&= \frac{\sum_{z_{t+1}} \left[\beta(z_{t+1}) p(x_{t+1} | z_{t+1}) p(z_{t+1} | z_t) \right]} {c_{t+1}},\\
c_t &= p(x_t | x_{0:t-1}) = \sum_{z_{t-1},z_t} \left[ p(z_{t-1}|x_{0:t-1}) p(x_t | z_t) p(z_t | z_{t-1}) \right] \nonumber\\ 
&= \sum_{z_{t-1},z_t} \left[ \alpha(z_{t-1}) p(x_t | z_t) p(z_t | z_{t-1}) \right].
\end{align}
\endgroup

Initial conditions are $\alpha(z_0) = \frac{p(x_0 | z_0)p(z_0)}{\sum_{z_0}[p(x_0 | z_0)p(z_0)]}$, $\beta(z_T) = 1$. 

\subsubsection{M-step}
In the M-step, the parameter set $\Theta$ is updated by maximising $\mathcal{Q}(\Theta ; \Theta_\text{old})$, which can be rewritten by substituting $p(z_t | \bm{X})$ and $p(z_{t-1}, z_t | \bm{X})$ in \Cref{eq:ppoem/em_objective_normal_hmm} with $\alpha(z)$ and $\beta(z)$ (ignoring the constants) as derived in \Cref{sec:ppoem/e_step_normal_hmm}. 

\subsubsection{Option discovery via the forward-backward algorithm}
\label{sec:ppoem/background/baum_related_work}
The idea of applying the forward-backward algorithm to learn option assignments is first introduced in~\cite{option_ml_2016}, and has later been applied in both \gls{il} settings ~\cite{Zhang2020ProvableHI,online_baum_welch_2021} and \gls{rl} settings~\cite{fox2017multi}. However, in previous literature, the option policy is decoupled into an option termination probability $\varpi(s_t, z_{t-1})$, and an inter-option policy $\pi(z_t|s_t)$. Due to the inter-option policy being unconditional on the previous option $z_{t-1}$, the choice of a new option $z_t$ will be uninformed of the previous option $z_{t-1}$. This may be problematic for learning \gls{pomdp} tasks as demonstrated in \Cref{sec:ppoem/experiments/corridor}. Previous literature also does not address the issues of exponentially diminishing magnitudes which arise from recursively applying the formula. This is known as the scaling factor problem~\cite{bishop_ml_book}. 

This chapter presents a concise derivation of the forward-backward algorithm applied to an improved version of the options framework. The scaling factor is also built into the derivation. 

\section{Option assignment formulation}
The aim is to learn a diverse set of options with corresponding policy and value estimates, such that each option is responsible for accomplishing a well-defined subtask, such as reaching a certain state region. At every time step $t$, the agent chooses an option $z_t$ out of $n$ number of discrete options $\{\mathcal{Z}_1, ..., \mathcal{Z}_n\}$. 

\subsection{Option policy and sub-policy}

\begin{figure}[h]
    \centering
     \subbottom[Standard options framework]{%
        \includegraphics[width=0.3\textwidth]{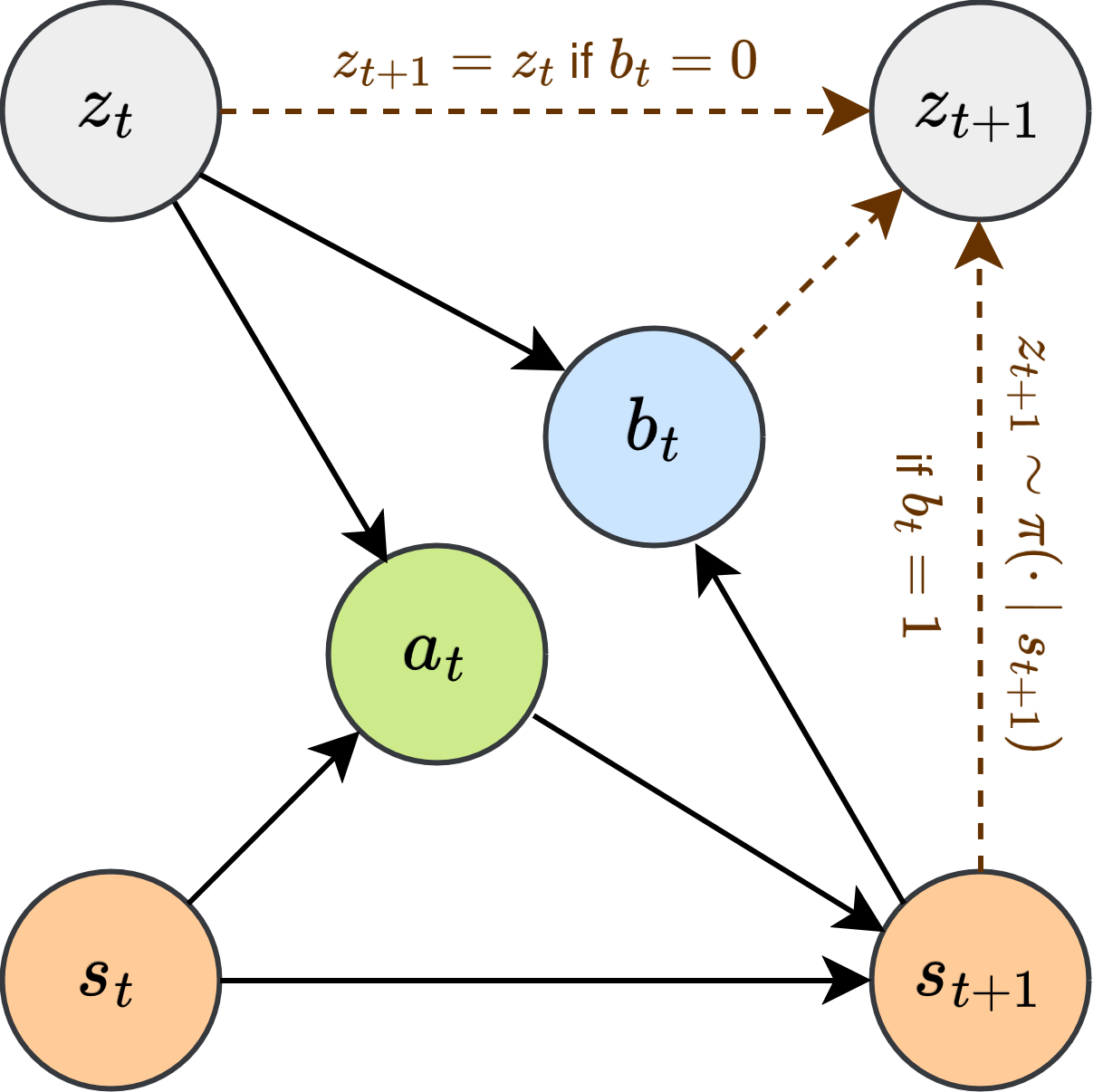}}
        \hspace{0.1\textwidth}
     \subbottom[Options used in this work]{%
        \includegraphics[width=0.3\textwidth]{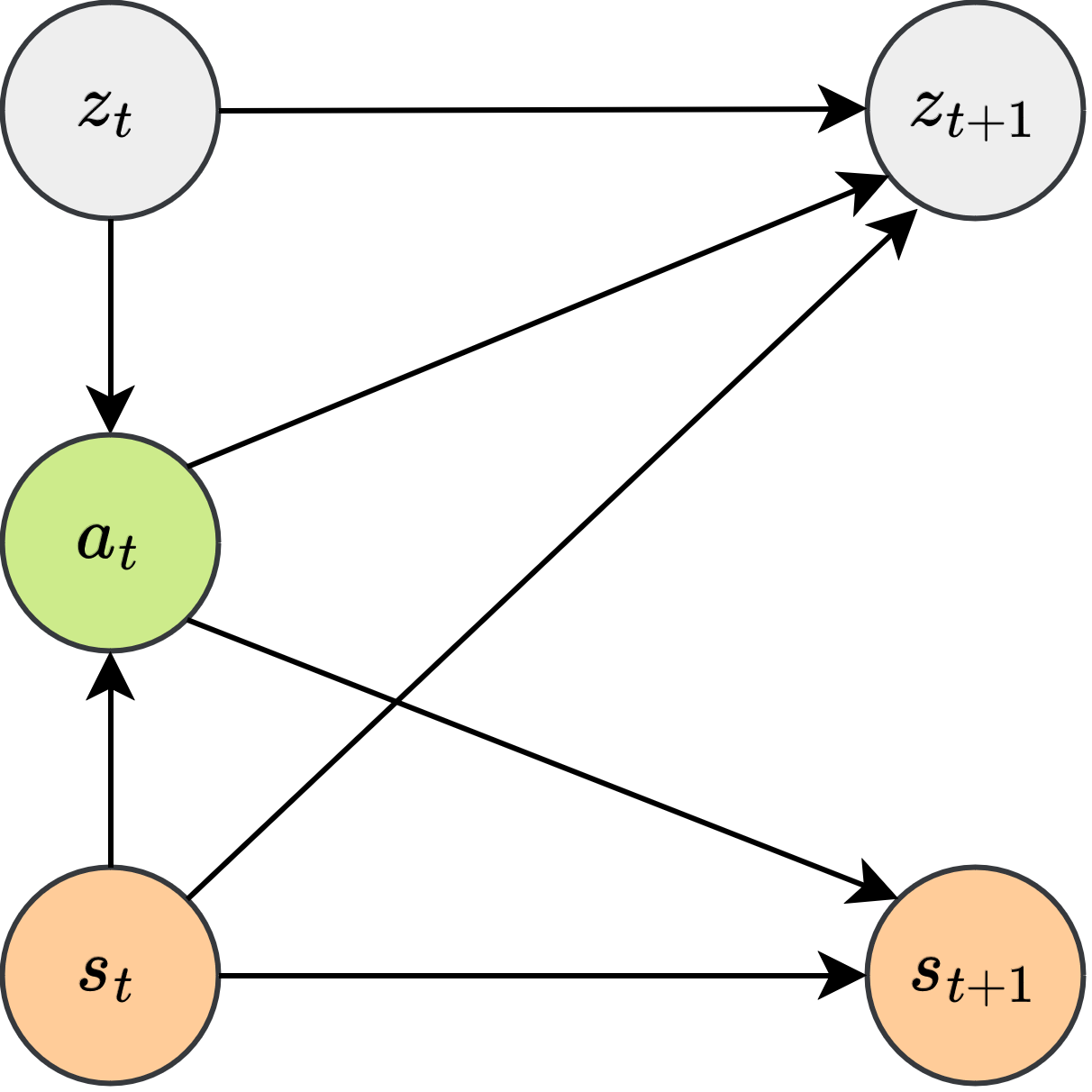}}
    \caption{Probabilistic graphical models showing the relationships between options $z$, actions $a$ and states $s$ at time step $t$. $b_t$ in the standard options framework denotes a boolean variable that initiates the switching of options when activated. This work adopts a more general formulation compared to the options framework, as defined in \Cref{eq:ppoem/joint_option_policy}.}
    \label{fig:ppoem/graphical_models_options}
\end{figure}

The goal is to learn a sub-policy $\pi_{\theta}(a | s, z)$ conditional to a latent option variable $z$, and an option policy $\pi_{\psi}(z' | s, a, z)$ used to iteratively assign options at each time step, to model the joint option policy
\begin{equation}
    p_{\Theta}(a_t, z_{t+1} | s_t, z_t) = \pi_{\theta}(a_t | s_t, z_t) \pi_{\psi}(z_{t+1} | s_t, a_t, z_t).
    \label{eq:ppoem/joint_option_policy}
\end{equation} 
Here, the learnable parameter set of the policy is denoted as $\Theta = \{\theta, \psi\}$. 

\begin{figure}[h]
  \centering
  \includegraphics[width=0.9\textwidth]{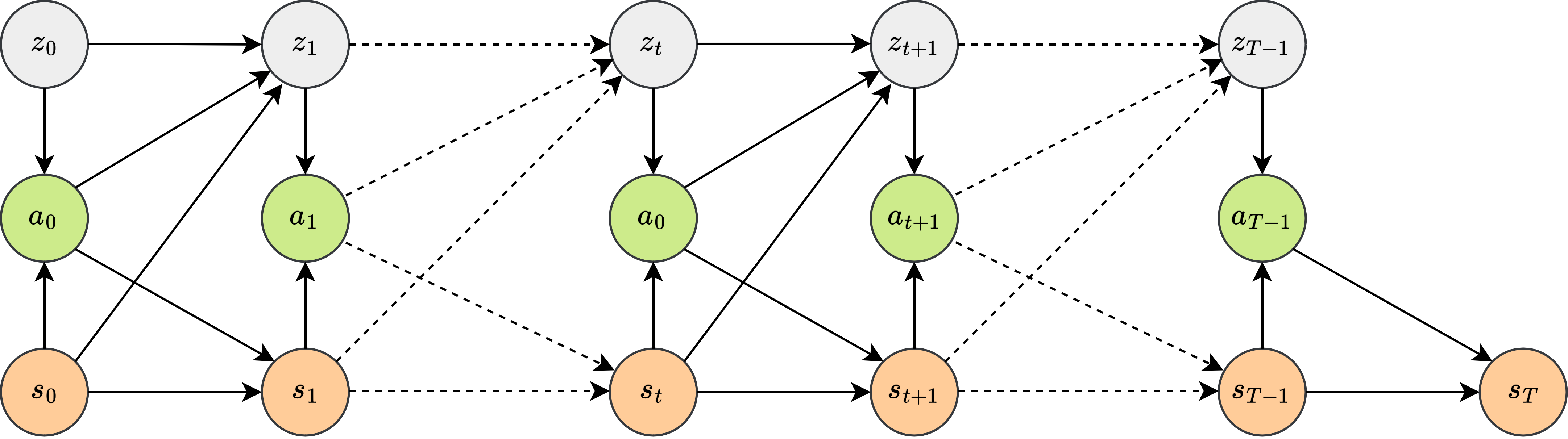}
  \caption{
    An \glsshort{hmm} showing the relationships between options $z$, actions $a$ and states $s$. The dotted arrows indicate that the same pattern repeats where the intermediate time steps are abbreviated.
  }
  \label{fig:ppoem/full_ppoem_hmm}
\end{figure}

A comparison of the option policy used in this work and the standard options framework is shown in \Cref{fig:ppoem/graphical_models_options}. Unlike the options framework, which further decouples the option policy $\pi_{\psi}$ into an option termination probability $\varpi(s_t, z_{t-1})$, and an unconditional inter-option policy $\pi(z_t|s_t)$, in this work the option policy is modelled $\pi_{\psi}$ with one network so that the inter-option policy is informed by the previous option $z_t$ upon choosing the next $z_{t+1}$. A graphical model for the full \gls{hmm} is shown in \Cref{fig:ppoem/full_ppoem_hmm}.

\subsection{Evaluating the probability of latents}
\label{sec:ppoem/eval_prob_latents}
Let us define an auto-regressive action probability $\alpha_t \defeq p(a_t | s_{0:t}, a_{0:t-1})$, an auto-regressive option forward distribution $\zeta(z_t) \defeq p(z_t | s_{0:t}, a_{0:t-1})$, and an option backward feedback $\beta(z_t) \defeq \frac{p(s_{t:T}, a_{t:T-1} | s_{t-1}, a_{t-1}, z_t)}{p(s_{t:T}, a_{t:T-1} | s_{0:t-1}, a_{0:t-1})}$. Notice that the definitions of action probability $\alpha$, option forward $\zeta(z_t)$, and option backward $\beta(z_t)$ resemble $c_t$, $\alpha(z_t)$ and $\beta(z_t)$ defined in \Cref{sec:ppoem/forward_backward}, respectively. 
While it is common practice to denote the forward and backward quantities as $\alpha$ and $\beta$ in the forward-backward algorithm (also known as the $\alpha$-$\beta$ algorithm), here $\alpha_t$ is redefined to denote the action probability (corresponding to the normalising constant $c_t$), and $\zeta(z_t)$ for the option forward distribution, to draw attention to the fact that these are probabilities of option $z_t$ and action $a_t$, respectively. 

$\alpha_t$, $\zeta(z_t)$ and $\beta(z_t)$ can be recursively evaluated as follows:
\begingroup
\small%
\begin{align}
\alpha_t &= p(a_t | s_{0:t}, a_{0:t-1}) = \sum_{z_t,z_{t+1}} p(z_t | s_{0:t}, a_{0:t-1}) p_{\Theta}(a_t, z_{t+1} | s_t, z_t) \nonumber\\ 
&= \sum_{z_t,z_{t+1}} \zeta(z_t) p_{\Theta}(a_t, z_{t+1} | s_t, z_t),\\
\zeta(z_{t+1}) &= \frac{p(z_{t+1}, s_{t+1}, a_t | s_{0:t}, a_{0:t-1})}{p(s_{t+1}, a_t | s_{0:t}, a_{0:t-1})}\nonumber\\
&= \frac{\sum_{z_t} p(z_t|s_{0:t}, a_{0:t-1}) p_{\Theta}(a_t, z_{t+1} | s_t, z_t) P(s_{t+1}|s_{0:t}, a_{0:t})}{p(a_t|s_{0:t}, a_{0:t-1}) P(s_{t+1}|s_{0:t}, a_{0:t})} \nonumber\\
&= \frac{\sum_{z_t} \zeta(z_t) p_{\Theta}(a_t, z_{t+1} | s_t, z_t)}{\alpha_t},\\
\beta(z_t) &= \frac{p(s_{t:T}, a_{t:T-1} | s_{t-1}, a_{t-1}, z_t)}{p(s_{t:T}, a_{t:T-1} | s_{0:t-1}, a_{0:t-1})} \nonumber\\
&= \frac{\sum_{z_{t+1}} \left[p(s_{t+1:T}, a_{t+1:T-1} | s_t, a_t, z_{t+1}) p_{\Theta}(a_t, z_{t+1} | s_t, z_t) P(s_t|s_{0:t-1}, a_{0:t-1}) \right]} {p(s_{t+1:T}, a_{t+1:T-1} | s_{0:t}, a_{0:t}) p(a_t|s_{0:t}, a_{0:t-1}) P(s_t|s_{0:t-1}, a_{0:t-1})} \nonumber\\
&= \frac{\sum_{z_{t+1}} \left[\beta(z_{t+1}) p_{\Theta}(a_t, z_{t+1} | s_t, z_t) \right]} {\alpha_t}.
\end{align}
\endgroup

Initial conditions are $\zeta(z_0) = p(z_0) = \frac{1}{n}$ for all possible $z_0$, indicating a uniform distribution over the options initially when no observations or actions are available, and $\beta(z_T) = \frac{p(s_T|s_{T-1}, a_{T-1}, z_{T})}{p(s_T|s_{0:T-1}, a_{0:T-1})} = 1$. 

\section{Proximal Policy Optimisation via Expectation Maximisation}
In this section, \gls{ppoem} is introduced, an algorithm that extends \gls{ppo} for option discovery with an \gls{em} objective. The expectation of the returns is taken over the joint probability distribution of states, actions and options, sampled by the policy. This objective gives a tractable objective to maximise, which has a close resemblance to the forward-backward algorithm.

\subsection{Expected return maximisation objective with options}
The objective is to maximise the expectation of returns $R(\tau)$ for an agent policy $\pi$ over a trajectory $\tau$ with latent option $z_t$ at each time step $t$. The definition of a trajectory $\tau$ is a set of states, actions and rewards visited by the agent policy in an episode. The objective $J[\pi]$ can be written as:
\begin{align}
\begin{split}
J[\pi_\Theta] = \expect_{\tau, \bm{Z} \sim\pi}[R(\tau)] = \int_{\tau, \bm{Z}}R(\tau)p(\tau, \bm{Z} | \Theta).
\end{split}
\end{align}

Taking the gradient of the maximisation objective,
\begin{align}
\begin{split}
\label{eq:ppoem/ppoem/return_max_grad}
\nabla_\Theta J[\pi_\Theta] &= \int_{\tau, \bm{Z}}R(\tau)\nabla_\Theta p(\tau, \bm{Z} | \Theta) = \int_{\tau, \bm{Z}}R(\tau)\frac{\nabla_\Theta p(\tau, \bm{Z} | \Theta)}{p(\tau, \bm{Z} | \Theta)}p(\tau, \bm{Z} | \Theta) \\
&= \expect_{\tau, \bm{Z}}[R(\tau)\nabla_\Theta \log p(\tau, \bm{Z} | \Theta)].
\end{split}
\end{align}

To simplify the derivation, let us focus on the states and actions that appear in the trajectory. The joint likelihood function for the trajectory $\tau$ and latent options $\bm{Z} = \{z_0, ..., z_{T-1}\}$ is given by:
\begin{align}
\begin{split}
p(\tau, \bm{Z} | \Theta) &= p(s_{0:T}, a_{0:T-1}, z_{0:T} | \Theta) = p(s_0, z_0) \Pi_{t=0}^{T-1}[p_{\Theta}(a_t, z_{t+1} | s_t, z_t) P(s_{t+1} | s_{0:t}, a_{0:t})],
\end{split}
\end{align}

Evaluating $\nabla_\Theta \log p(\tau, \bm{Z} | \Theta)$, the log converts the products into sums, and the terms which are constant with respect to $\Theta$ are eliminated upon taking the gradient, leaving
\begin{align}
\begin{split}
\label{eq:ppoem/ppoem/log_joint_likelihood}
\nabla_\Theta \log p(\tau, \bm{Z} | \Theta) &= \sum_{t=0}^{T-1} \nabla_\Theta \log [\pi_{\theta}(a_t | s_t, z_t) \pi_{\psi}(z_{t+1} | s_t, a_t, z_t, s_{t+1})].
\end{split}
\end{align}

Substituting \Cref{eq:ppoem/ppoem/log_joint_likelihood} into \Cref{eq:ppoem/ppoem/return_max_grad} and explicitly evaluating the expectation over the joint option probabilities,
\begingroup
\small
\begin{align}
\begin{split}
\label{eq:ppoem/ppoem/return_max_grad_full}
\nabla_\Theta J[\pi_\Theta] &= \expect_{\tau, \bm{Z} \sim\pi} \left[\sum_{t=0}^{T-1} R(\tau) \nabla_\Theta \log p_{\Theta}(a_t, z_{t+1} | s_t, z_t) \right]\\
&= \expect_{\tau\sim\pi}\int_{\bm{Z}} \left[ \sum_{t=0}^{T-1} \left[R(\tau) \nabla_\Theta \log p_{\Theta}(a_t, z_{t+1} | s_t, z_t) \right] \right] p(\bm{Z}|\tau)\\
&= \expect_{\tau\sim\pi} \left[ \sum_{t=0}^{T-1} \sum_{z_t, z_{t+1}} \left[R(\tau) p(z_t, z_{t+1}|\tau) \nabla_\Theta \log p_{\Theta}(a_t, z_{t+1} | s_t, z_t) \right] \right].
\end{split}
\end{align}
\endgroup

Using the action probability $\alpha_t \defeq p(a_t | s_{0:t}, a_{0:t-1})$, option forward distribution $\zeta(z_t) \defeq p(z_t | s_{0:t}, a_{0:t-1})$, and option backward feedback $\beta(z_t) \defeq \frac{p(s_{t:T}, a_{t:T-1} | s_{t-1}, a_{t-1}, z_t)}{p(s_{t:T}, a_{t:T-1} | s_{0:t-1}, a_{0:t-1})}$ evaluated in \Cref{sec:ppoem/eval_prob_latents}, $p(z_t, z_{t+1} | \tau)$ can be evaluated as
\begingroup
\small%
\begin{align}
\begin{split}
\label{eq:ppoem/ppoem/joint_option_prob}
p(z_t, z_{t+1} | \tau) 
&= \frac{p(s_{0:T}, a_{0:T-1}, z_t, z_{t+1})}{p(s_{0:T}, a_{0:T-1})}\\ 
&= \frac{p(s_{0:t}, a_{0:t-1}, z_{t}) p_{\Theta}(a_t, z_{t+1} | s_t, z_t) p(s_{t+1:T}, a_{t+1:T-1} | s_t, a_t, z_{t+1})}{p(s_{0:t},a_{0:t-1}) p(a_t|s_{0:t}, a_{0:t-1}) p(s_{t+1:T}, a_{t+1:T-1} | s_{0:t}, a_{0:t})}\\ 
&= p_{\Theta}(a_t, z_{t+1} | s_t, z_t) \frac{\zeta(z_t)\beta(z_{t+1})}{\alpha_t}.
\end{split}
\end{align}
\endgroup

Using this, \Cref{eq:ppoem/ppoem/return_max_grad_full} can be evaluated and maximised with gradient descent.

\subsubsection{Relationship with Expectation Maximisation}
The objective derived in \Cref{eq:ppoem/ppoem/return_max_grad_full} closely resembles the objective of the \gls{em} algorithm applied to the \gls{hmm} with options as latent variables.
The expectation of the marginal log-likelihood $\mathcal{Q}(\Theta ; \Theta_\text{old})$, which gives the lower-bound of the marginal log-likelihood $\log{p(\tau | \Theta)}$, is given by
\begin{align}
\begin{split}
\label{eq:ppoem/ppoem/em_objective}
\mathcal{Q}(\Theta; \Theta_\text{old}) 
&= \expect_{\bm{Z} \sim p(\cdot|\tau,\Theta_\text{old})}\left[\ln{p(\tau, \bm{Z} | \Theta)}\right] = \expect_{\tau\sim\pi} \int_{\bm{Z}} p(\bm{Z} | \tau, \Theta_\text{old}) \ln{p(\tau, \bm{Z} | \Theta)}d\bm{Z} \\
&= \expect_{\tau\sim\pi} \sum_{t=0}^{T-1}\sum_{z_t, z_{t+1}} \left[ p(z_t, z_{t+1} | \tau, \Theta_\text{old}) \log p_{\Theta}(a_t, z_{t+1} | s_t, z_t)\right] + \text{const.}
\end{split}
\end{align}

The difference is that the expected return maximisation objective in \Cref{eq:ppoem/ppoem/return_max_grad_full} weights the log probabilities of the policy according to the returns, whereas the objective of \Cref{eq:ppoem/ppoem/em_objective} is to find a parameter set $\Theta$ that maximises the probability that the states and actions that appeared in the trajectory are visited by the joint option policy $p_\Theta$. 

\subsection{PPO objective with Generalised Advantage Estimation}
\label{sec:ppoem/ppoem/ppoem_gae}
A standard optimisation technique for neural networks using gradient descent can be applied to optimise the policy network. Noticing that the optimisation objective in \Cref{eq:ppoem/ppoem/return_max_grad_full} resembles the policy gradient algorithm, the joint option policy can be optimised using the \gls{ppo} algorithm instead to prevent the updated policy $p_{\Theta}(a_t, z_{t+1} | s_t, z_t)$ from deviating from the original policy too much. 

Several changes have to be made to adapt the training objective to \gls{ppo}. Firstly, $\nabla \log p_\Theta$ is replaced by $\frac{\nabla p_\Theta}{p_{\Theta_\text{old}}}$, its first order approximation, to easily introduce clipping constraints to the policy ratios.
Secondly, the return $R(\tau)$ is replaced with the \gls{gae}, $A_t^\text{GAE}$, as introduced in \Cref{sec:background/gae}.

Extending the definition of \gls{gae} to work with options,
\begin{align}
A_t^\text{GAE}(z_t, z_{t+1} | \tau) &= r_t + \gamma V(s_{t+1}, z_{t+1}) - V(s_t, z_t) + \lambda \gamma (1-d_t) A_{t+1}^\text{GAE}(z_{t+1} | \tau)\\
A_t^\text{GAE}(z_t | \tau) &= \sum_{z_{t+1}} p(z_{t+1} | z_t, \tau) A_t^\text{GAE}(z_t, z_{t+1} | \tau).
\end{align}

The \gls{gae} could be evaluated backwards iteratively, starting from $t=T$ with the initial condition $A_T^\text{GAE}(z_{t+1} | \tau) = 0$. The option transition function $p(z_{t+1} | z_t, \tau)$ can be evaluated using $p(z_t, z_{t+1} | \tau)$ (\Cref{eq:ppoem/ppoem/joint_option_prob}) as:
\begin{equation}
    p(z_{t+1} | z_t, \tau) = \frac{p(z_t, z_{t+1} | \tau)}{\sum_{z_{t+1}} p(z_t, z_{t+1} | \tau)}.
\end{equation}

The target value $V_\text{target}(s_t, z_t)$ to regress the estimated value function towards can be defined in terms of the \gls{gae} and the current value estimate as:
\begin{equation}
V_\text{target}(s_t, z_t) = V^{\pi}(s_t, z_t) + A_t^\text{GAE}(z_t | \tau).   
\end{equation}

\section{Sequential Option Advantage Propagation}
In the previous section, assignments of the latent option variables $\bm{Z}$ were determined by maximising the expected return for complete trajectories. The derived algorithm resembles the forward-backward algorithm closely, and requires the backward pass of $\beta(z_t)$ in order to fully evaluate the option probability $p(\bm{Z} | \tau)$. During rollouts of the agent policy, however, knowing the optimal assignment of latents $p(z_t | \tau)$ in advance is not possible, since the trajectory is incomplete and the backward pass has not been initiated. Therefore, the policy must rely on the current best estimate of the options given its available past trajectory $\{s_{0:t}, a_{0:t}\}$ during its rollout. This option distribution conditional only on its past is equivalent to the auto-regressive option forward distribution $\zeta(z_t) \defeq p(z_t | s_{0:t}, a_{0:t-1})$.

Since the optimal option assignment can only be achieved in hindsight once the trajectory is complete, this information is not helpful for the agent policy upon making its decisions. A more useful source of information for the agent, therefore, is the current best estimate of the option assignment $\zeta(z_t)$. It is sensible, therefore, to directly optimise for the expected returns evaluated over the option assignments $\zeta(z_t)$ to find an optimal option policy, rather than optimising the expected returns for an option assignment $p(\bm{Z} | \tau)$, which can only be known in hindsight.

The following section proposes a new option optimisation objective that does not involve the backward pass of the \gls{em} algorithm. Instead, the option policy gradient for an optimal forward option assignment is evaluated analytically. This results in a temporal gradient propagation, which corresponds to a backward pass, but with a slightly different outcome. Notably, this improved algorithm, \gls{soap}, applies a normalisation of the option advantages in every back-propagation step through time.

As far as the author is aware, this work is the first to derive the back-propagation of policy gradients in the context of option discovery. 

\subsection{Policy Gradient objective with options}
Let us start by deriving the policy gradient objective assuming options. The maximisation objective $J[\pi]$ for the agent can be defined as:
\begin{align}
\begin{split}
J[\pi_\Theta] = \expect_{\tau \sim\pi}[R(\tau)] = \int_{\tau}R(\tau)p(\tau | \Theta) d\tau.
\end{split}
\end{align}

Taking the gradient of the maximisation objective,
{\small
\begin{align}
\begin{split}
\label{eq:ppoem/soap/option_policy_grad}
\nabla_\Theta J[\pi_\Theta] &= \int_{\tau}R(\tau)\nabla_\Theta p(\tau | \Theta) d\tau = \int_{\tau}R(\tau)\frac{\nabla_\Theta p(\tau | \Theta)}{p(\tau | \Theta)}p(\tau | \Theta) d\tau = \expect_{\tau}[R(\tau)\nabla_\Theta \log p(\tau | \Theta)].
\end{split}
\end{align}
}

So far, the above derivation is the same as the normal policy gradient objective without options. Next, the likelihood for the trajectory $\tau$ is given by:
\begin{align}
\begin{split}
p(\tau | \Theta) &= p(s_{0:T}, a_{0:T-1} | \Theta) = \rho(s_0) \Pi_{t=0}^{T-1}[p(a_t | s_{0:t}, a_{0:t-1}, \Theta) P(s_{t+1} | s_{0:t}, a_{0:t})].
\end{split}
\end{align}

This is where options become relevant, as the standard formulation assumes that the policy $\pi(a | s)$ is only dependent on the current state without history, and similarly that the state transition environment dynamics $P(s' | s, a)$ is Markovian given the current state and action. In many applications, however, the \emph{states} that are observed do not contain the entire information about the underlying dynamics of the environment\footnote{Some literature on \gls{pomdp} choose to make this explicit by denoting the partial observation available to the agent as observation $o$, distinguishing from the underlying ground truth state $s$. However, since $o$ can also stand for \emph{options}, and is used in other literature on options, here the input to the agent's policy and value functions is denoted using the conventional $s$ to prevent confusion.}, and therefore, conditioning on the history yields a different distribution of future states compared to conditioning on just the current state. To capture this, the policy and state transitions are now denoted to be $p(a_t | s_{0:t}, a_{0:t-1})$ and $P(s_{t+1} | s_{0:t}, a_{0:t})$, respectively. Here, the probabilities are conditional on the historical observations ($s_{0:t}$) and historical actions (e.g. $a_{0:t}$), rather than just the immediate state $s_t$ and action $a_t$. Note that $p(a_t | s_{0:t}, a_{0:t-1})$ is a quantity $\alpha_t$ that has already been evaluated in \Cref{sec:ppoem/eval_prob_latents}.

Evaluating $\nabla_\Theta \log p(\tau | \Theta)$, the log converts the products into sums, and the terms which are constant with respect to $\Theta$ are eliminated upon taking the gradient, leaving
\begin{align}
\begin{split}
\label{eq:ppoem/soap/log_likelihood_pomdp}
\nabla_\Theta \log p(\tau | \Theta) &= \sum_{t=0}^{T-1} \nabla_\Theta \log p(a_t | s_{0:t}, a_{0:t-1}, \Theta) = \sum_{t=0}^{T-1} \nabla_\Theta \log \alpha_t = \sum_{t=0}^{T-1} \frac{\nabla_\Theta \alpha_t}{\alpha_t},
\end{split}
\end{align}
where $\alpha_t$ is substituted following its definition.

Substituting \Cref{eq:ppoem/soap/log_likelihood_pomdp} into \Cref{eq:ppoem/soap/option_policy_grad},
\begin{align}
\begin{split}
\label{eq:ppoem/soap/option_policy_grad_simple}
\nabla_\Theta J[\pi_\Theta] &= 
\expect_{\tau \sim\pi} \left[\sum_{t=0}^{T-1} R(\tau) \frac{\nabla_\Theta \alpha_t}{\alpha_t} \right].
\end{split}
\end{align}

Similarly to \Cref{sec:ppoem/ppoem/ppoem_gae}, it is possible to substitute the return $R(\tau)$ with \gls{gae}, thereby reducing the variance in the return estimate. 
Extending the definition of \gls{gae} to work with options,
\begingroup
\small%
\begin{align}
A_t^\text{GAE}(z_t, z_{t+1}) &= r_t + \gamma V(s_{t+1}, z_{t+1}) - V(s_t, z_t) + \lambda \gamma (1-d_t) A_{t+1}^\text{GAE}(z_{t+1}),\\
A_t^\text{GAE}(z_t) &= \sum_{z_{t+1}} p(z_{t+1} | s_t, a_t, z_t) A_t^\text{GAE}(z_t, z_{t+1}),\\
V_\text{target}(s_t, z_t) &= V^{\pi}(s_t, z_t) + A_t^\text{GAE}(z_t)\label{eq:ppoem/soap/value}.
\end{align}
\endgroup

Notice that, while the definition of these estimates is almost identical to \Cref{sec:ppoem/ppoem/ppoem_gae}, the advantages are now propagated backwards via the option transition $p(z_{t+1} | s_t, a_t, z_t)$ rather than $p(z_{t+1} | z_t, \tau)$.

Substituting the \gls{gae} into \Cref{eq:ppoem/soap/option_policy_grad_simple},
\begingroup
\small%
\begin{align}
\begin{split}
\label{eq:ppoem/soap/policy_gradient_objective}
&\nabla_\Theta J[\pi_\Theta] = 
\expect_{\tau \sim\pi}\left[\sum_{t=0}^{T-1} \frac{\sum_{z_t} A_t^\text{GAE}(z_t) \zeta(z_t)}{\alpha_t} \nabla_\Theta \alpha_t \right]\\
&= \expect_{\tau \sim\pi}\left[ \sum_{t=0}^{T-1} \frac{\sum_{z_t} A_t^\text{GAE}(z_t) \zeta(z_t)}{\alpha_t} \sum_{z_t,z_{t+1}} \left[ p_{\Theta}(a_t, z_{t+1} | s_t, z_t) \nabla \zeta(z_t) + \zeta(z_t) \nabla p_{\Theta}(a_t, z_{t+1} | s_t, z_t) \right] \right].
\end{split}
\end{align}
\endgroup

\subsection{Analytic back-propagation of the policy gradient}
If a forward pass of the policy can be made in one step over the entire trajectory, a gradient optimisation on the objective can be performed directly. However, this would require storing the entire trajectory in \gls{gpu} memory, which is highly computationally intensive. Instead, this section analytically evaluates the back-propagation of gradients of the objective so that the model can be trained on single time-step rollout samples during training.

Gradient terms appearing in \Cref{eq:ppoem/soap/policy_gradient_objective} are either $\nabla \zeta(z_t)$ or $\nabla p_{\Theta}(a_t, z_{t+1} | s_t, z_t)$ for $0 \leq t \leq T-1$. 
While $p_{\Theta}(a_t, z_{t+1} | s_t, z_t)$ is approximated by neural networks and can be differentiated directly, $\nabla \zeta(z_{t+1})$ has to be further expanded to evaluate the gradient in recursive form as:
\begingroup
\small%
\begin{align}
\begin{split}
\label{eq:ppoem/soap/zeta_grad}
\nabla \zeta(z_{t+1}) &= \frac{\nabla \sum_{z_t} \zeta(z_t) p_{\Theta}(a_t, z_{t+1} | s_t, z_t)}{\alpha_t} - \zeta(z_{t+1}) \frac{\nabla \alpha_t}{\alpha_t}\\
&= \frac{1}{\alpha_t} \left[ \sum_{z_t} \nabla \left[ \zeta(z_t) p_{\Theta}(a_t, z_{t+1} | s_t, z_t)\right] - \zeta(z_{t+1}) \sum_{z'_t, z'_{t+1}} \nabla \left[ \zeta(z'_t) p_{\Theta}(a_t, z'_{t+1} | s_t, z'_t)\right] \right].
\end{split}
\end{align}
\endgroup

Using \Cref{eq:ppoem/soap/zeta_grad}, it is possible to rewrite the $\nabla \zeta(z_{t+1})$ terms appearing in \Cref{eq:ppoem/soap/policy_gradient_objective} in terms of $\nabla \zeta(z_t)$ and $\nabla p_{\Theta}(a_t, z_{t+1} | s_t, z_t)$. Defining the coefficients of $\nabla \zeta(z_t)$ in \Cref{eq:ppoem/soap/policy_gradient_objective} as option utility $U(z_t)$,

\begingroup
\small%
\begin{align}
\begin{split}
\sum_{z_{t+1}} U(z_{t+1}) \nabla \zeta(z_{t+1}) 
=& \frac{1}{\alpha_t} \sum_{z_t, z_{t+1}} \left[ U(z_{t+1}) - \sum_{z'_{t+1}} U(z'_{t+1}) \zeta(z'_{t+1})  \right] \nabla \left[ \zeta(z_t) p_{\Theta}(a_t, z_{t+1} | s_t, z_t)\right]\\
=& \frac{1}{\alpha_t} \sum_{z_t, z_{t+1}} \left[ U(z_{t+1}) - \sum_{z'_{t+1}} U(z'_{t+1}) \zeta(z'_{t+1})  \right] \\
& \cdot \left[ p_{\Theta}(a_t, z_{t+1} | s_t, z_t) \nabla \zeta(z_t) +  \zeta(z_t) \nabla p_{\Theta}(a_t, z_{t+1} | s_t, z_t) \right].
\end{split}
\end{align}
\endgroup

Applying this iteratively to \Cref{eq:ppoem/soap/policy_gradient_objective}, starting with $t=T-1$ in reverse order, \Cref{eq:ppoem/soap/policy_gradient_objective} could be expressed solely in terms of gradients $\nabla p_{\Theta}(a_t, z_{t+1} | s_t, z_t)$. Defining the coefficients of $\nabla p_{\Theta}(a_t, z_{t+1} | s_t, z_t)$ as policy gradient weighting $W_t(z_t, z_{t+1})$, 
\begingroup
\small%
\begin{align}
\begin{split}
A_t^\text{GOA}(z_{t+1}) &= \sum_{z_t} A_t^\text{GAE}(z_t) \zeta(z_t) + (1 - d_t) \left[ U(z_{t+1}) - \sum_{z'_{t+1}} U(z'_{t+1}) \zeta(z'_{t+1})  \right],\\
U(z_t) &= \frac{\sum_{z_{t+1}} A_t^\text{GOA}(z_{t+1}) p_{\Theta}(a_t, z_{t+1} | s_t, z_t)}{\alpha_t},\\
W(z_t, z_{t+1}) &= \frac{A_t^\text{GOA}(z_{t+1})\zeta(z_t)}{\alpha_t}.
\end{split}
\end{align}
\endgroup
where $A_t^\text{GOA}(z_{t+1})$ is a new quantity derived and introduced in this work as \gls{goa}, which is a term that appears in evaluating $U(z_t)$ and $W(z_t, z_{t+1})$.

Rewriting the policy gradient objective in \Cref{eq:ppoem/soap/policy_gradient_objective} with the policy gradient weighting,
\begin{align}
\begin{split}
\label{eq:ppoem/soap/soap_policy_grad}
\nabla_\Theta J[\pi_\Theta] 
&= \expect_{\tau \sim\pi} \left[\sum_{t=0}^{T-1} \sum_{z_t, z_{t+1}} \frac{A_t^\text{GOA}(z_{t+1})\zeta(z_t)}{\alpha_t} \nabla_\Theta p_{\Theta}(a_t, z_{t+1} | s_t, z_t) \right].
\end{split}
\end{align}

\subsection{Learning objective for option-specific policies and values}
The training objective given in \Cref{eq:ppoem/soap/soap_policy_grad} is modified so that it could be optimised with \gls{ppo}. Unlike in \Cref{sec:ppoem/ppoem/ppoem_gae}, the training objective is written in terms of $\nabla p_\Theta$ and not $\nabla \log p_\Theta$. Therefore, the clipping constraints are applied to $p_\Theta$ directly, limiting it to the range of $(1-\epsilon)p_{\Theta_\text{old}}$ and $(1+\epsilon)p_{\Theta_\text{old}}$. The resulting \gls{ppo} objective is:
\begingroup
\small%
\begin{align}
\begin{split}
&J_\Theta = \expect_{s_t, a_t\sim\pi} \sum_{z_t, z_{t+1}} \biggl[\frac{\zeta(z_t)}{\alpha_t} \min \Bigl( \pi_{\Theta}(a_t, z_{t+1}|s_t, z_t) A_t^{\text{GOA}}(z_{t+1}),\\ &\text{clip} \Bigl(\pi_{\Theta}(a_t, z_{t+1}|s_t, z_t), (1 - \epsilon) \pi_{\Theta_\text{old}}(a_t, z_{t+1}|s_t, z_t), (1 + \epsilon) \pi_{\Theta_\text{old}}(a_t, z_{t+1}|s_t, z_t) \Bigr)A_t^{\text{GOA}}(z_{t+1}) \Bigr) \biggr].
\end{split}
\end{align}
\endgroup

The option-specific value function $V_\phi^\pi(s_t, z_t)$ parameterised by $\phi$ can be learnt by regressing towards the target values $V_\text{target}(s_t, z_t)$ evaluated in \Cref{eq:ppoem/soap/value} for each state $s_t$ and option $z_t$ sampled from the policy and option-forward probability, respectively. Defining the objective function for the value regression as $J_\phi$,
\begingroup
\small%
\begin{equation}
J_\phi = - \expect_{s_t\sim\pi, z_t \sim \zeta} \left[ V_\text{target}(s_t, z_t) - V_\phi^{\pi}(s_t, z_t)\right]^2.
\end{equation}
\endgroup

The final training objective is to maximise the following:
\begin{align}
\begin{split}
J_\text{SOAP} = J_\Theta + J_\phi.
\end{split}
\end{align}

\section{Experiments}

\begin{table}[ht]
\small
\label{tab:ppoem/scores_normalised}
\centering
\caption{Normalised performance comparison of \gls{rl} agents. The agent scores are the returns after the maximum environment steps during training ($100k$ for CartPole, $1M$ for LunarLander and MuJoCo environments, and $10M$ for Atari environments), normalised so that the score of a random agent is 0 and the score of the best performing model is 1. Scores are averaged per environment class (i.e. results for the corridor environments, Atari, and MuJoCo are grouped together) and the \textbf{bold fonts} show the best average normalised score per environment class, while the \textcolor{blue}{blue fonts} show the best normalised score per environment.}
\begin{tabular}{lrrrrrr}
\toprule
Environment         & PPO  & PPOC & PPO-LSTM & DAC  & PPOEM (ours) & SOAP (ours) \\ 
\midrule
\textbf{Corridor} & -0.05 & 0.31 & 0.43 & 0.65 & 0.60 & \textbf{1.00} \\
\quad $L=3$ & -0.08 & 0.76 & 1.00 & 0.90 & 0.99 & \textcolor{blue}{1.00} \\
\quad $L=10$ & -0.01 & 0.06 & 0.36 & 0.63 & 0.70 & \textcolor{blue}{1.00} \\
\quad $L=20$ & -0.05 & 0.11 & -0.06 & 0.41 & 0.12 & \textcolor{blue}{1.00} \\
\midrule
\textbf{CartPole} & \textbf{1.00} & 0.80 & 0.98 & \textbf{1.00} & 0.98 & \textbf{1.00} \\
\midrule
\textbf{LunarLander} & 0.86 & 0.74 & \textbf{1.00} & 0.78 & 0.99 & 0.99 \\
\midrule
\textbf{Atari} & \textbf{0.93} & 0.22 & 0.78 & 0.85 & 0.74 & 0.89 \\
\quad Asteroids & 0.83 & 0.81 & 0.64 & \textcolor{blue}{1.00} & 0.93 & 0.89 \\
\quad Beam Rider & 0.93 & 0.13 & 0.37 & 0.70 & 0.66 & \textcolor{blue}{1.00} \\
\quad Breakout & \textcolor{blue}{1.00} & 0.01 & 0.68 & 0.95 & 0.14 & 0.92 \\
\quad Enduro & 0.97 & 0.00 & 0.90 & 0.93 & \textcolor{blue}{1.00} & 0.82 \\
\quad Ms Pacman & 0.88 & 0.15 & 0.74 & 0.87 & 0.69 & \textcolor{blue}{1.00} \\
\quad Pong & 1.00 & 0.48 & \textcolor{blue}{1.00} & \textcolor{blue}{1.00} & 0.94 & \textcolor{blue}{1.00} \\
\quad Qbert & \textcolor{blue}{1.00} & 0.00 & 0.97 & 0.65 & 0.70 & 0.90 \\
\quad Road Runner & \textcolor{blue}{1.00} & 0.15 & 0.92 & 0.88 & 0.52 & 0.81 \\
\quad Seaquest & 0.89 & 0.18 & 0.87 & 0.69 & \textcolor{blue}{1.00} & 0.55 \\
\quad Space Invaders & 0.82 & 0.32 & 0.73 & 0.82 & 0.82 & \textcolor{blue}{1.00} \\
\midrule
\textbf{MuJoCo} & \textbf{0.97} & 0.60 & 0.75 & 0.82 & 0.43 & 0.93 \\
\quad Ant & \textcolor{blue}{1.00} & 0.07 & 0.46 & 0.64 & 0.07 & 0.87 \\
\quad Half Cheetah & \textcolor{blue}{1.00} & 0.80 & 0.83 & 0.79 & 0.01 & 0.92 \\
\quad Humanoid & 0.98 & 0.96 & 0.96 & \textcolor{blue}{1.00} & 0.13 & 0.90 \\
\quad Reacher & 0.99 & 0.99 & 0.99 & \textcolor{blue}{1.00}& 0.98 & \textcolor{blue}{1.00} \\
\quad Swimmer & 0.96 & 0.34 & 0.97 & \textcolor{blue}{1.00} & 0.95 & 0.90 \\
\quad Walker & 0.85 & 0.47 & 0.31 & 0.52 & 0.44 & \textcolor{blue}{1.00} \\
\bottomrule
\end{tabular}
\end{table}

Experiments were conducted on a variety of \gls{rl} agents: \gls{ppo}, \gls{ppoc}, \gls{ppo_lstm}, \gls{dac}, \gls{ppoem} (ours), and \gls{soap} (ours).
\gls{ppo}~\cite{schulman_proximal_2017} is a baseline without memory, \gls{ppoc}~\cite{optioncritic_ppo} implements the Option-Critic algorithm using \gls{ppo} for policy optimisation, \gls{ppo_lstm} implements a recurrent policy with latent states using an \gls{lstm}, \gls{dac}~\cite{zhang2019dac} optimises both the inter- and intra-option policies on hierarchical \glspl{mdp}, \gls{ppoem} is the algorithm developed in the first half of this chapter that optimises the expected returns using the forward-backward algorithm, and \gls{soap} is the final algorithm proposed in this chapter that uses an option advantage derived by analytically evaluating the temporal propagation of the option policy gradients. \gls{soap} mitigates the deficiency of \gls{ppoem} that the training objective optimises the option assignments over a full trajectory which is typically only available in hindsight; \gls{soap} optimises the option assignments given only the history of the trajectory instead, making the optimisation objective better aligned with the task objective.

The aim is to (a) show and compare the option learning capability of the newly developed algorithms, and (b) assess the stability of the algorithms on standard \gls{rl} environments.
All algorithms use \gls{ppo} as the base policy optimiser, and share the same backbone and hyperparameters, making it a fair comparison. All algorithms use Stable Baselines 3~\cite{stable_baselines3} as a base implementation with the recommended tuned hyperparameters for each environment. In the following experiments, the number of options was set to $4$. 

\subsection{Option learning in corridor environments}
\label{sec:ppoem/experiments/corridor}
A simple environment of a corridor with a fork at the end is designed as a minimalistic and concrete example where making effective use of latent variables to retain information over a sequence is necessary to achieve the agent's goal. 

\begin{figure}[h]
  \centering
  \includegraphics[width=1.0\textwidth]{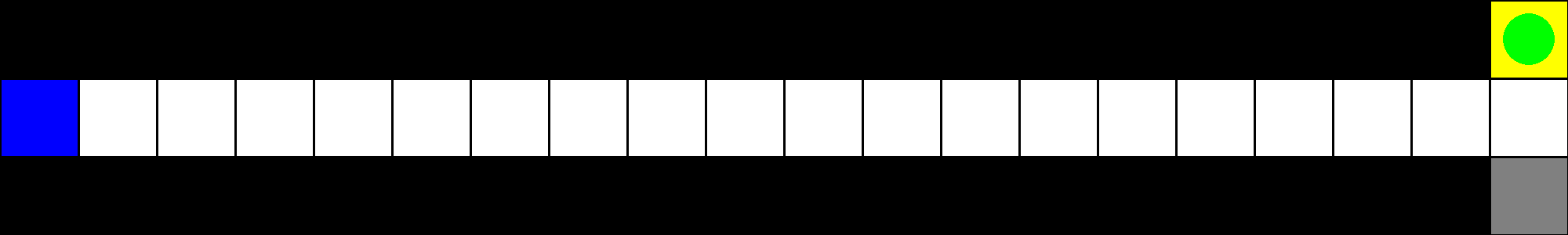}\\
  \vspace{0.5em}
  \includegraphics[width=1.0\textwidth]{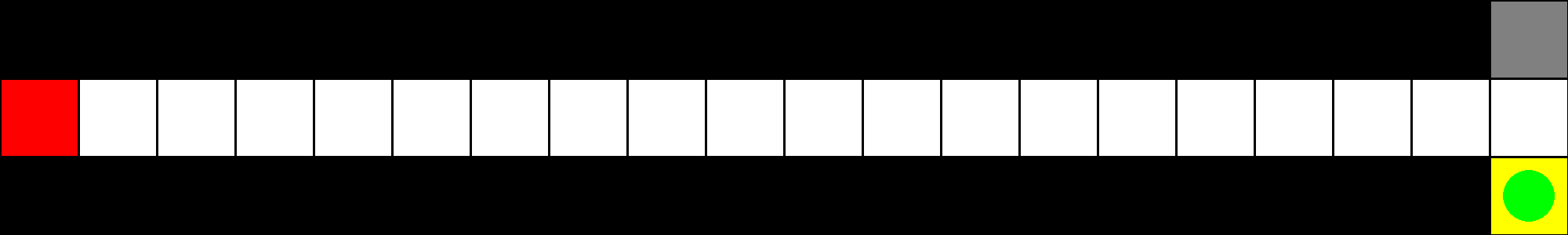}
  \caption{
    A corridor environment. The above example has a length $L=20$. The agent represented as a green circle starts at the left end of the corridor, and moves towards the right. When it reaches the right end, the agent can either take an up action or a down action. This will either take the agent to a yellow cell or a grey cell. The yellow cell gives a reward of $1$, while the grey cell gives a reward of $-1$. All other cells give a reward of $0$. The location of a rewarding yellow cell and the penalising grey cell are determined by the colour of the starting cell (either ''blue'' or ''red''), as shown, and this is randomised, each with $50\%$ probability. The agent only has access to the colour of the current cell as observation. For simplicity of implementation, the agent's action space is \{''up'', ''down''\}, and apart from the fork at the right end, taking either of the actions at each time step will move the agent one cell to the right.
    The images shown are taken from rollouts of the \glsshort{soap} agent after training for $100k$ steps. The agent successfully navigated to the rewarding cell in both cases.
  }
  \label{fig:ppoem/experiments/corridor}
\end{figure}

\Cref{fig:ppoem/experiments/corridor} describes the corridor environment, in which the agent has to determine whether the rewarding cell (coloured yellow) is at the top or bottom, based on the colour of the cell it has seen at the start (either ''blue'' or ''red''). However, the agent only has access to the colour of the current cell, and does not have a bird's-eye-view of the environment. Hence, the agent must retain the information of the colour of the starting cell in memory, whilst discarding all other information irrelevant to the completion of the task. The agent must learn that the information of the colour of the starting cell is important to task completion in an unsupervised way, just from the reward signals. This makes the task challenging, as only in hindsight (after reaching the far end of the corridor) is it clear that this information is useful to retain in memory, but if this information was not written in memory in the first place then credit assignment becomes infeasible. 

The length of the corridor $L$ can be varied to adjust the difficulty of the task. It is increasingly challenging to retain the information of the starting cell colour with longer corridors. In theory, this environment can be solved by techniques such as frame stacking, where the entire history of the agent observations is provided to the policy. However, the computational complexity of this approach scales proportionally to corridor length $L$, which makes this approach unscalable. 

Algorithms with options present an alternative solution, where in theory, the options can be used as latent variables to carry the information relevant to the task. In this experiment, \gls{ppoc}, \gls{ppo_lstm}, \gls{dac}, \gls{ppoem} and \gls{soap} are compared against a standard \gls{ppo} algorithm. The results are shown for corridors with lengths $L=3$, $L=10$ and $L=20$. Due to the increasing level of difficulty of the task, the agents are trained with $8k$, $40k$ and $100k$ time steps of environment interaction, respectively. 

\begin{figure}[h]
    \centering
     \subbottom[Corridor of length $L=3$]{%
        \includegraphics[width=0.8\textwidth]{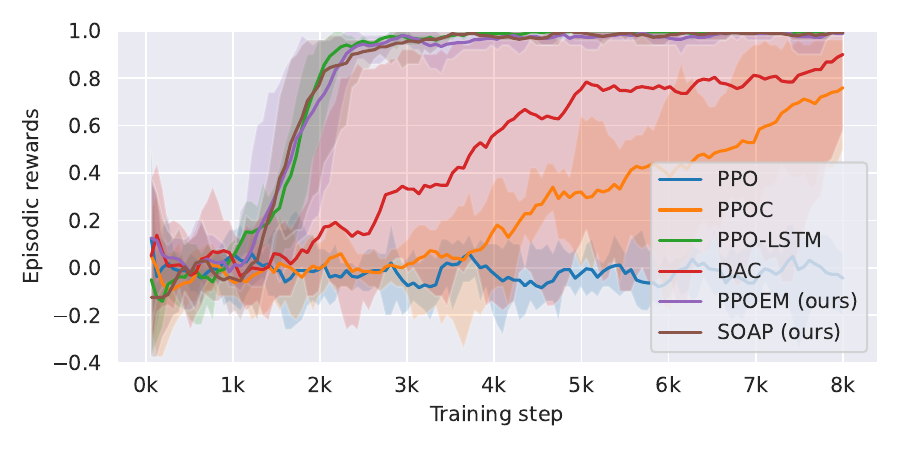}}
     \subbottom[Corridor of length $L=10$]{%
        \includegraphics[width=0.8\textwidth]{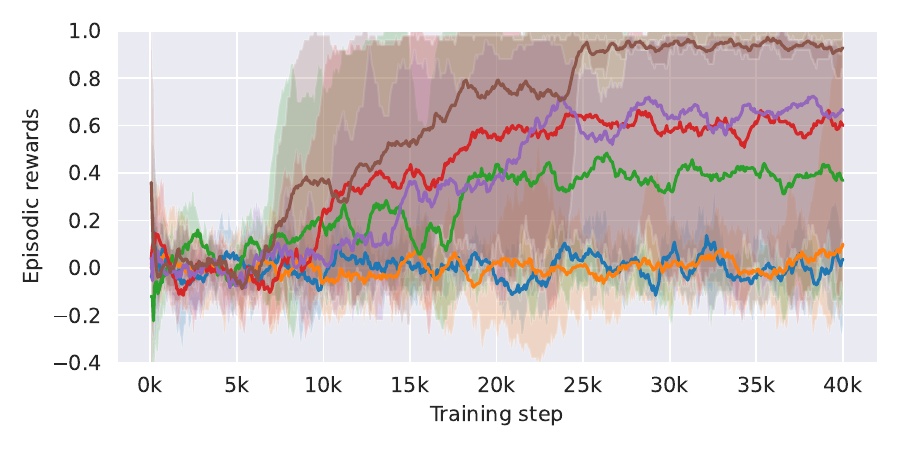}}
     \subbottom[Corridor of length $L=20$]{%
        \includegraphics[width=0.8\textwidth]{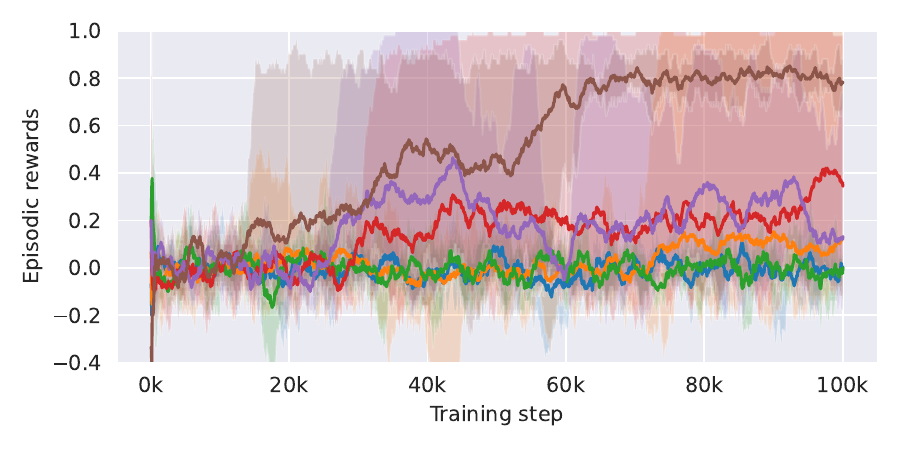}}
    \caption{Training curves of \glsshort{rl} agents showing the episodic rewards obtained in the corridor environment with varying corridor lengths. The mean (solid line) and the min-max range (coloured shadow) for $5$ seeds per algorithm are shown.}
    \label{fig:ppoem/experiments/corridor_rewards}
\end{figure}

The results are shown in \Cref{fig:ppoem/experiments/corridor_rewards} and \Cref{tab:ppoem/comparison}, and a performance score normalised to the range of the returns of a random agent score and the returns of the best agent is shown in \Cref{tab:ppoem/scores_normalised}. As expected, the vanilla \gls{ppo} agent does not have any memory component so it learnt a policy that takes one action deterministically regardless of the colour of the first cell. Since the location of the rewarding cell is randomised, this results in an expected return of $0$. 

With \gls{ppoc} that implements the Option-Critic architecture, and \gls{dac} that implements \gls{hrl} using options, while the options should in theory be able to retain information from the past, it could be observed that the training objective was not sufficient to learn a useful option assignment to complete the task. \gls{ppoem} and \gls{soap}, on the other hand, were able to learn to select a different option for a different starting cell colour. From \Cref{fig:ppoem/experiments/corridor_rewards}, it could be seen that the two algorithms had identical performance for a short corridor, but as the corridor length $L$ increased, the performance of \gls{ppoem} deteriorated, while \gls{soap} was able to reliably find a correct option assignment, albeit with more training steps. 

There are several major differences between the baseline option-based algorithms (\gls{ppoc} and \gls{dac}) and the proposed algorithms (\gls{ppoem} and \gls{soap}) which could be contributing to their significant differences in performance. Firstly, while the option transition function in \gls{ppoem} and \gls{soap} are in the form of $\pi_\phi(z_{t+1} | s_t, a_t, z_t)$, which allows the assignment of the new option to be conditional on the current option, the option transition in the Option-Critic architecture is decoupled into an option termination probability $\varpi(s_t, z_{t-1})$, and an unconditional inter-option policy $\pi(z_t|s_t)$. This means that whenever the previous option $z_{t-1}$ is terminated with probability $\varpi(s_t, z_{t-1})$, the choice of the new option $z_t$ will be uninformed of the previous option $z_{t-1}$, whereas in \gls{ppoem} and \gls{soap} the probability of the next option $z_{t+1}$ is conditional on the previous option $z_t$. 
The formulation of \gls{dac}~\cite{zhang2019dac} does not have this specific constraint; however, the original implementation by the authors similarly decouples the option transition function such that a new option cannot be fully conditioned on the previous option.

Secondly, both \gls{ppoc} and \gls{dac} rely on learning an option-value function (a value function $V(s, o)$ that is both conditional on the current state $s$ and the current option $o$) to learn the high-level inter-option policy. However, learning an option-value function for an optimal policy can only happen once the inter-option policy is properly learned, but since the inter-option policy is randomly initialised and the option assignment carries little information, it is not possible for the sub-policies to learn optimal policies. 
In the case where the agents' history carries information about the future (e.g. the corridor environment), it is important that the option assignment correctly captures this information without it being lost for the agent to be able to learn an optimal policy. Due to this chicken-and-an-egg problem of learning the inter- and the intra-option policies, neither of these approaches succeeds. In contrast, \gls{soap} directly propagates gradients backwards over multiple timesteps so that the option assignments are directly updated. 

Thirdly, in \gls{ppoc} and \gls{dac}, a new option is sampled at every time step, but the complete option forward distribution given the history is not available as a probability distribution. In contrast, in \gls{ppoem} and \gls{soap} this is available as $\zeta(z_t) \defeq p(z_t | s_{0:t}, a_{0:t-1})$. Evaluating expectations over distributions gives a more robust estimate of the objective function compared to taking a Monte Carlo estimate of the expectations with the sampled options, which is another explanation of why \gls{ppoem} and \gls{soap} were able to learn better option assignments than \gls{ppoc} or \gls{dac}.

\gls{soap}'s training objective maximises the expectation of returns taken over an option probability conditioned only on the agent's past history, whereas \gls{ppoem}'s objective assumes a fully known trajectory to be able to evaluate the option assignment probability. 
Since option assignments have to be determined online during rollouts, the training objective of \gls{soap} better reflects the task objective. This explains its more reliable performance for longer sequences. 

\gls{ppo_lstm} achieved competitive performance in a corridor with $L=3$, demonstrating the capability of latent states to retain past information, but its performance quickly deteriorated for longer corridors. It could be hypothesised that this is because the latent state space of the recurrent policies is not well constrained, unlike options which take discrete values. Learning a correct value function $V(s, z)$ requires revisiting the same state-latent pair. It is conceivable that with longer sequence lengths during inference time, the latent state will fall within a region that has not been trained well due to compounding noise, leading to an inaccurate estimate of the values and sub-policy. 

\subsection{Stability of the algorithms on CartPole, LunarLander, Atari, and MuJoCo environments}

\begin{figure}[h]
    \centering
     \subbottom[CartPole-v1]{%
        \includegraphics[width=0.48\textwidth]{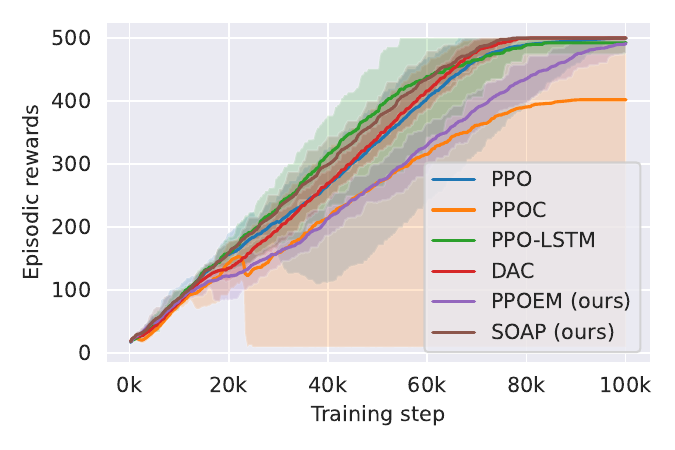}}
     \subbottom[LunarLander-v2]{%
        \includegraphics[width=0.48\textwidth]{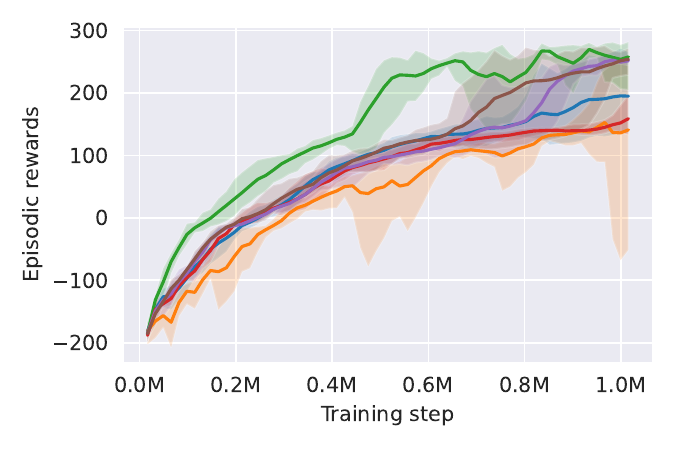}}
    \caption{Training curves of \glsshort{rl} agents showing the episodic rewards obtained in the CartPole-v1 and LunarLander-v2 environments. The mean (solid line) and the min-max range (coloured shadow) for $5$ seeds per algorithm are shown.}
    \label{fig:ppoem/experiments/cartpole_lunarlander}
\end{figure}

\begin{figure}[h]
    \centering
     \subbottom[Ant]{%
        \includegraphics[width=0.48\textwidth]{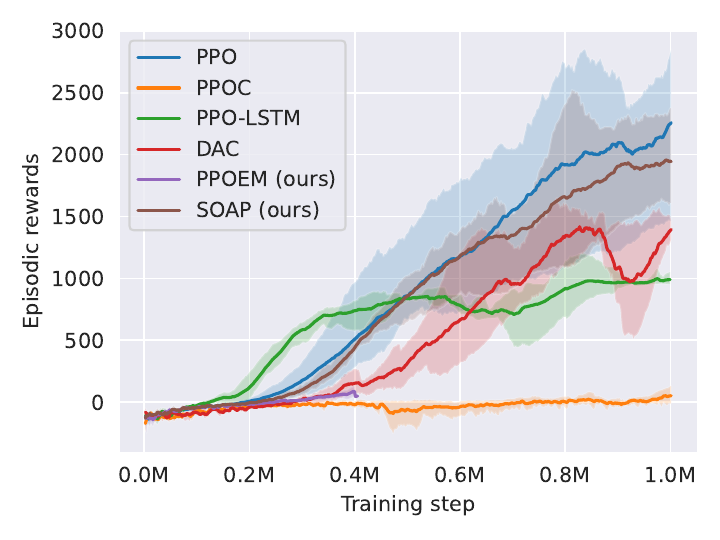}}
     \subbottom[HalfCheetah]{%
        \includegraphics[width=0.48\textwidth]{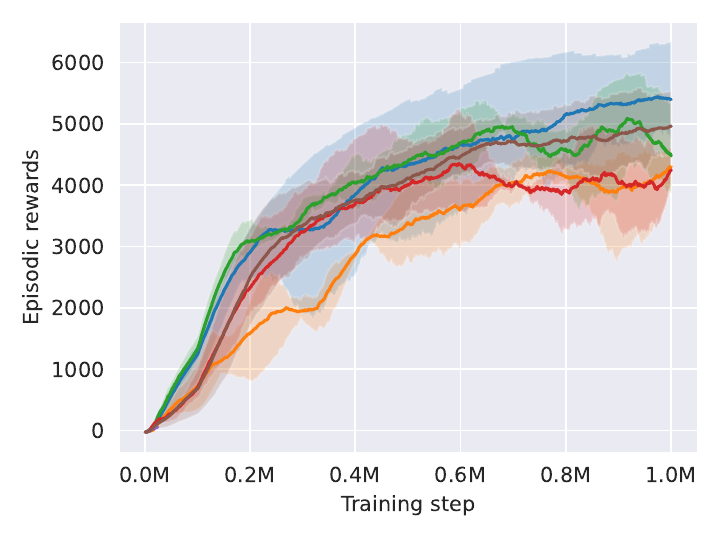}}
     \subbottom[Humanoid]{%
        \includegraphics[width=0.48\textwidth]{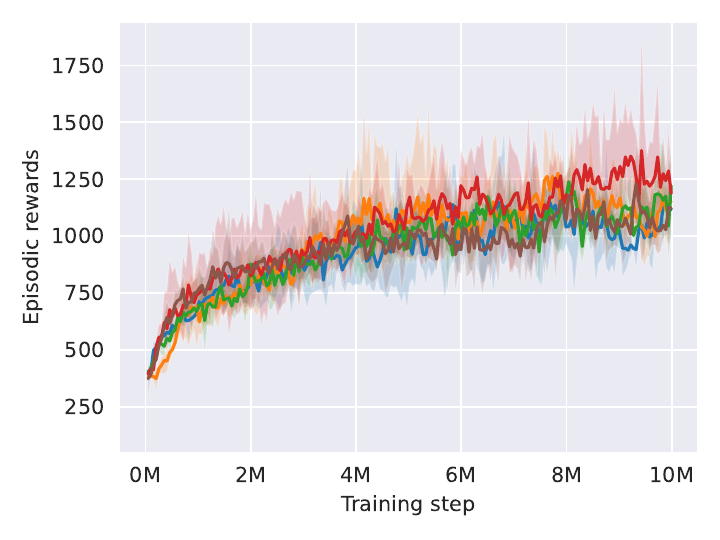}}
     \subbottom[Reacher]{%
        \includegraphics[width=0.48\textwidth]{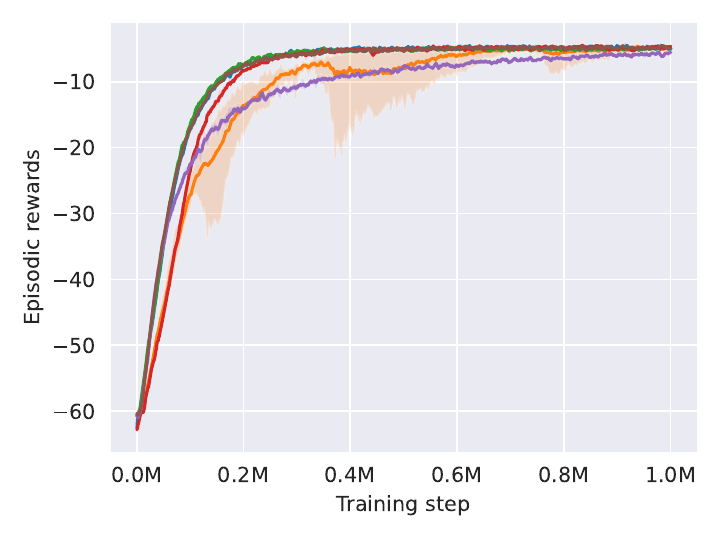}}
     \subbottom[Swimmer]{%
        \includegraphics[width=0.48\textwidth]{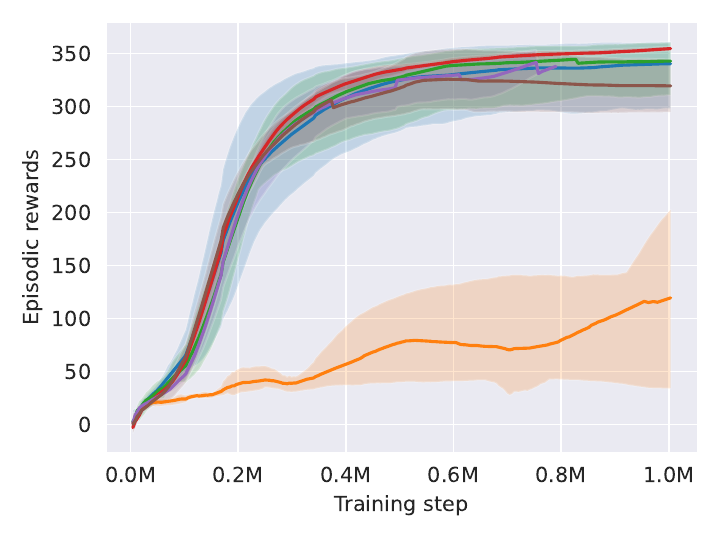}}
     \subbottom[Walker]{%
        \includegraphics[width=0.48\textwidth]{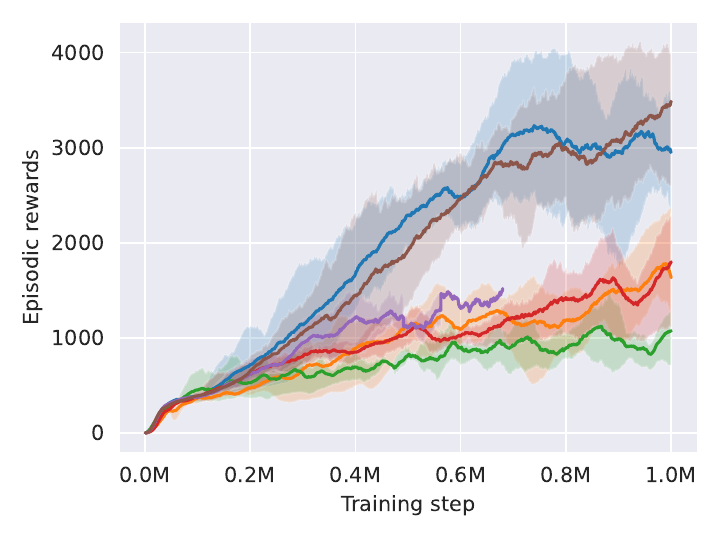}}
    \caption{Training curves of \glsshort{rl} agents showing the episodic rewards obtained in the MuJoCo environments. The mean (solid line) and the min-max range (coloured shadow) for $3$ seeds per algorithm are shown. Note that the \glsshort{ppoem} algorithm failed mid-way in some cases due to training instabilities.}
    \label{fig:ppoem/experiments/mujoco}
\end{figure}

Experiments were also conducted on standard \gls{rl} environments to evaluate the stability of the algorithms with options. Results for CartPole-v1 and LunarLander-v2 are shown in \Cref{fig:ppoem/experiments/cartpole_lunarlander}, and results on 10 Atari environments~\cite{bellemare13arcade} and 6 MuJoCo environments~\cite{todorov2012mujoco} are shown in \Cref{fig:ppoem/experiments/atari} and \Cref{fig:ppoem/experiments/mujoco}, respectively. \Cref{tab:ppoem/comparison} summarises the agent scores after the maximum environment steps during training ($100k$ for CartPole, $1M$ for LunarLander and MuJoCo environments, and $10M$ for Atari environments), and \Cref{tab:ppoem/scores_normalised} shows the scores normalised so that the score of a random agent is 0 and the score of the best performing model is 1. (If an agent's final score is lower than a random agent it can have negative normalised scores.) 
There was no significant difference in performances amongst the algorithms for simpler environments such as CartPole and LunarLander, with \gls{ppoc} and \gls{dac} having slightly worse performance than others. For the Atari and MuJoCo environments, however, there was a consistent trend that \gls{soap} achieves similar performances (slightly better in some cases, slightly worse in others) to the vanilla \gls{ppo}, while \gls{ppoem}, \gls{ppo_lstm} and \gls{ppoc} were significantly less stable to train. It could be hypothesised that, similarly to \Cref{sec:ppoem/experiments/corridor}, the policy of \gls{ppoc} disregarded the information of past options when choosing the next option, which is why the performance was unstable with larger environments. Another point of consideration is that, with $N$ number of options, there are $N$ number of sub-policies to train, which becomes increasingly computationally expensive and requires many visits to the state-option pair in the training data, especially when using a Monte Carlo estimate by sampling the next option as is done in \gls{ppoc} and \gls{dac} instead of maintaining a distribution of the option $\zeta(z_t)$ as in \gls{ppoem} and \gls{soap}. As for \gls{ppo_lstm}, similar reasoning as in \Cref{sec:ppoem/experiments/corridor} suggests that with complex environments with a variety of trajectories that can be taken through the state space, the latent states that could be visited increases combinatorially, making it challenging to learn a robust sub-policy and value functions.

\section{Conclusion}
Two competing algorithms, \gls{ppoem} and \gls{soap}, are proposed to solve the problem of option discovery and assignments in an unsupervised way. \gls{ppoem} implements a training objective of maximising the expected returns using the \gls{em} algorithm, while \gls{soap} analytically evaluates the policy gradient of the option policy to derive an option advantage function that facilitates temporal propagation of the policy gradients.
These approaches have an advantage over Option-Critic architecture in that (a) the option distribution is analytically evaluated rather than sampled, and (b) the option transitions are fully conditional on the previous option, allowing historical information to propagate forward in time beyond the temporal window provided as observations.

Experiments in \gls{pomdp} corridor environments designed to require options showed that \gls{soap} is the most robust way of learning option assignments that adhere to the task objective. \gls{soap} also maintained its performance when solving \gls{mdp} tasks without the need for options (e.g. Atari with frame-stacking), whereas \gls{ppoc}, \gls{dac}, \gls{ppo_lstm} and \gls{ppoem} were less stable when solving these problems.

\begin{figure}[h]
    \centering
     \subbottom[Astroids]{%
        \includegraphics[width=0.48\textwidth]{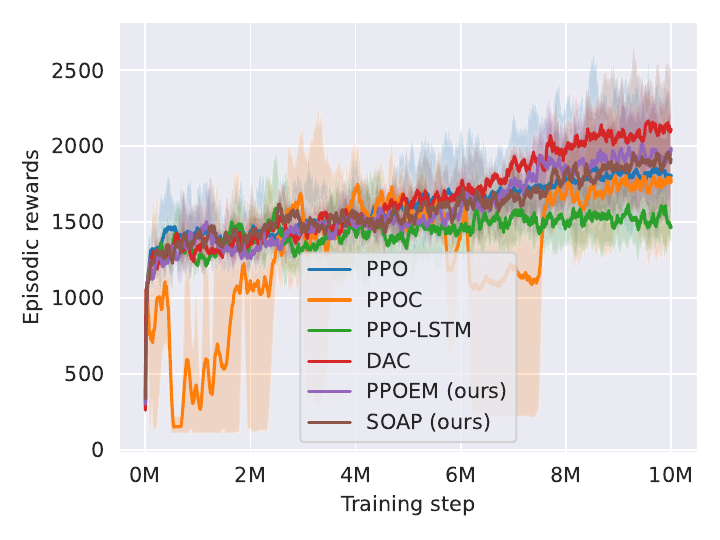}}
     \subbottom[Beam Rider]{%
        \includegraphics[width=0.48\textwidth]{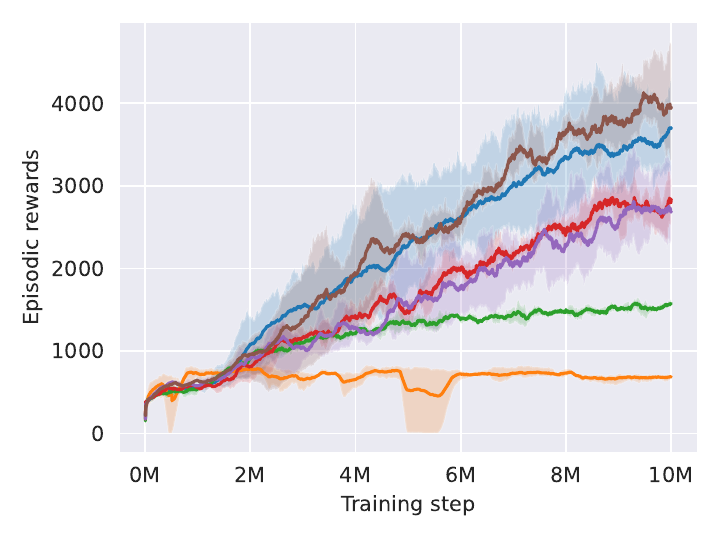}}
     \subbottom[Breakout]{%
        \includegraphics[width=0.48\textwidth]{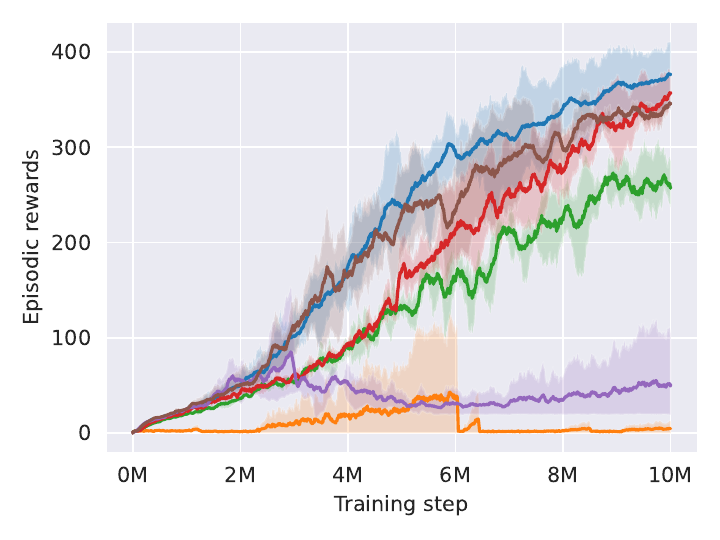}}
     \subbottom[Enduro]{%
        \includegraphics[width=0.48\textwidth]{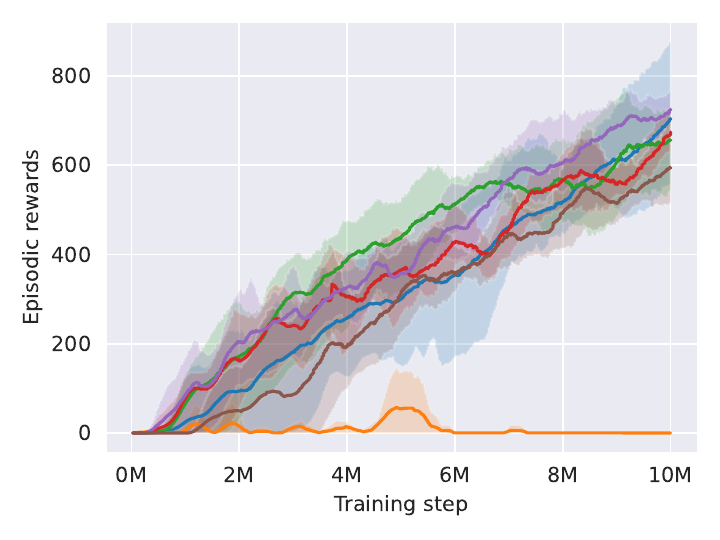}}
    \caption{Training curves of \glsshort{rl} agents showing the episodic rewards obtained in the Atari environments. The mean (solid line) and the min-max range (coloured shadow) for $3$ seeds per algorithm are shown. [Spans multiple pages]}
    \label{fig:ppoem/experiments/atari}
\end{figure}

\begin{figure}[h]
    \ContinuedFloat
    \setcounter{subfigure}{4}
    \centering
     \subbottom[Ms Pacman]{%
        \includegraphics[width=0.48\textwidth]{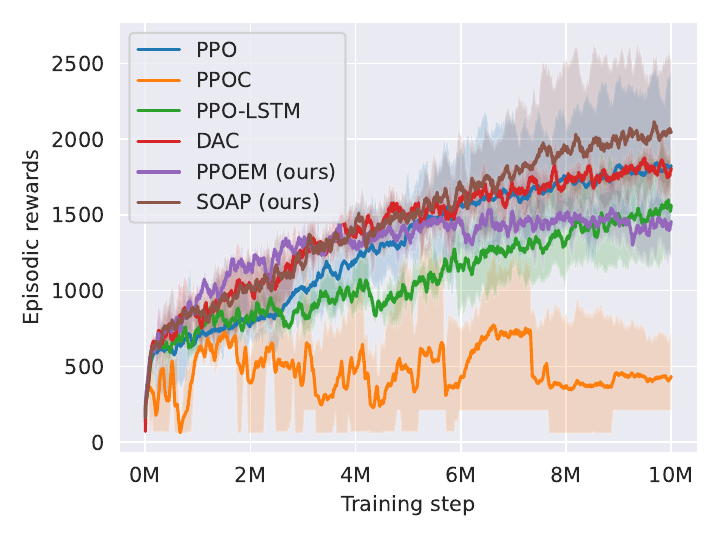}}
     \subbottom[Pong]{%
        \includegraphics[width=0.48\textwidth]{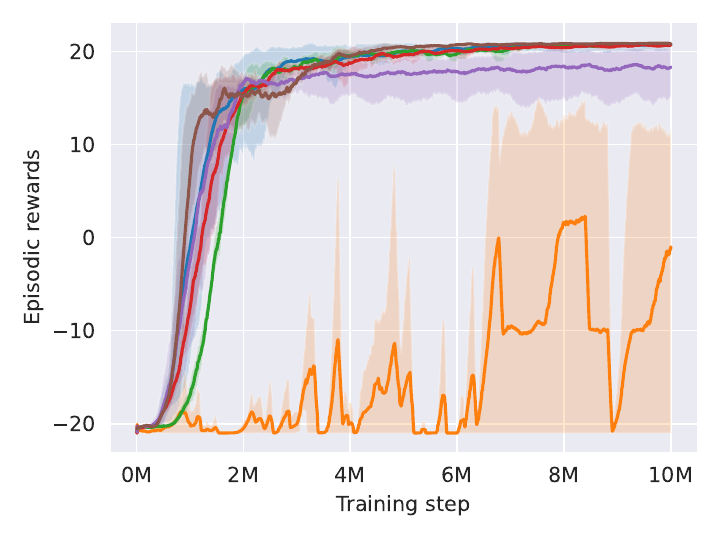}}
     \subbottom[Qbert]{%
        \includegraphics[width=0.48\textwidth]{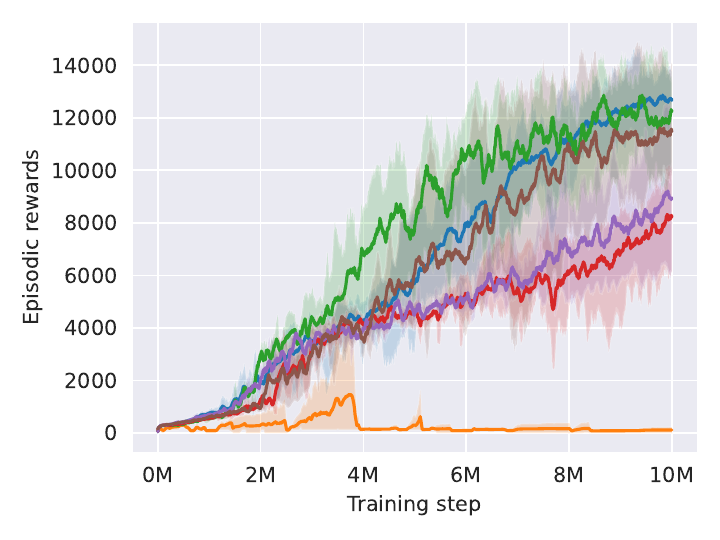}}
     \subbottom[Road Runner]{%
        \includegraphics[width=0.48\textwidth]{figures/ppoem/plots/beamrider.pdf}}
     \subbottom[Seaquest]{%
        \includegraphics[width=0.48\textwidth]{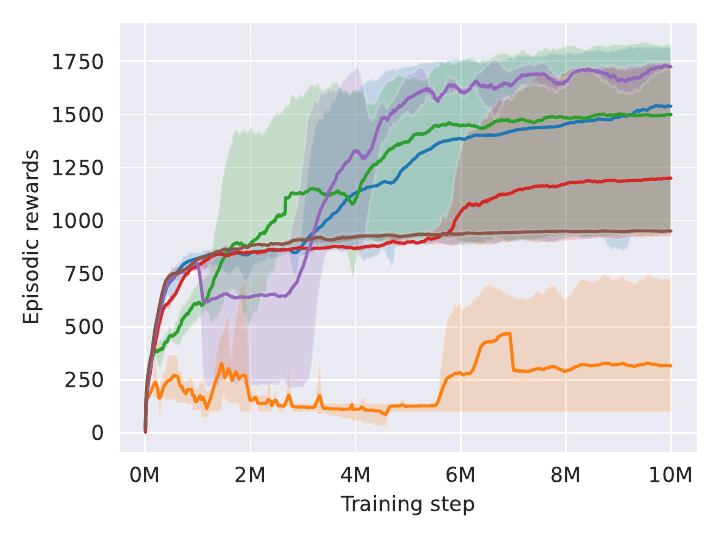}}
     \subbottom[Space Invader]{%
        \includegraphics[width=0.48\textwidth]{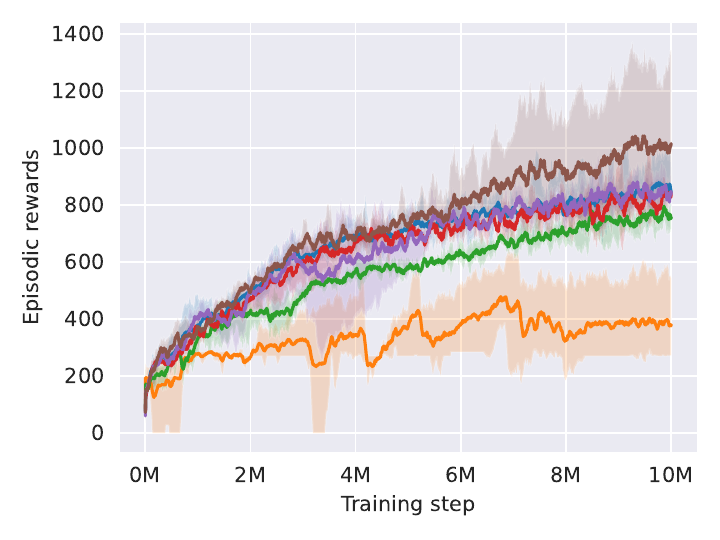}}
    \caption*{[Continued] Training curves of \glsshort{rl} agents showing the episodic rewards obtained in the Atari environments. The mean (solid line) and the min-max range (coloured shadow) for $3$ seeds per algorithm are shown.}
\end{figure}

\chapter{A code optimisation framework using Large Language Models}
\label{chapter:langprop} 

This chapter proposes LangProp, a framework for iteratively optimising code generated by \glspl{llm}, in both supervised and \gls{rl} settings. While \glspl{llm} can generate sensible coding solutions zero-shot (\ie without having to fine-tune the model to a specific problem domain), they are often sub-optimal. Especially for code generation tasks, it is likely that the initial code will fail on certain edge cases. LangProp automatically evaluates the code performance on a dataset of input-output pairs, catches any exceptions, and feeds the results back to the \gls{llm} in the training loop, so that the \gls{llm} can iteratively improve the code it generates. By adopting a metric- and data-driven training paradigm for this code optimisation procedure, one could easily adapt findings from traditional machine learning techniques such as \gls{bc}, DAgger, and \gls{rl}. 

LangProp demonstrates applicability to general domains such as Sudoku and CartPole, as well as a first proof of concept of automated code optimisation for autonomous driving in CARLA. LangProp can generate interpretable and transparent policies that can be verified and improved in a metric- and data-driven way. 

This research was conducted during an internship at Wayve Technologies as a work placement for the Autonomous Intelligent Machines and Systems Centre for Doctoral Training programme (AIMS CDT). The work was performed under the supervision of Anthony Hu and Gianluca Corrado, with mentorship by members of the world modelling team at Wayve. The development of the algorithm and the code is entirely my own work. Our paper was accepted at the International Conference on Learning Representations (ICLR) 2024 Workshop on \glsfirst{llm} Agents~\cite{ishida2024langprop}.

\section{Introduction}
Building systems that can self-improve with data is at the core of the machine learning paradigm. 
By leveraging vast amounts of data and having an automated feedback loop to update models according to an objective function, machine learning methods can directly optimise the metrics of interest, thus outperforming systems that are handcrafted by experts. In the early history of \gls{ai}, Symbolic \gls{ai}, e.g. rule-based expert systems~\citep{hayes1985rule,jackson1986introduction}, was a dominant and perhaps a more intuitive and explainable approach to solving tasks in an automated way, and is still widely used in fields such as medicine~\citep{abu2017medical} and autonomous driving~\citep{badue2021self}. However, there have been numerous successes in recent decades in machine learning, e.g. deep neural networks, that demonstrate the advantage of data-driven learning.

Advances in \glspl{llm}~\citep{NEURIPS2020_gpt3,openai2023gpt4,touvron2023llama} were enabled by neural networks. Trained on both natural language and code, they can translate human intent and logic into executable code and back, expanding the boundaries of applying logic and reasoning. Unlike other machine learning techniques, \glspl{llm} have an affinity with Symbolic \gls{ai} since they operate in discrete symbolic input-output spaces. The generated outputs are interpretable, even though the internal representation of these tokens is in a continuous embedding space. 
This observation led to the question of whether it is possible to have the best of both worlds -- having an interpretable and transparent system, characteristic of Symbolic \gls{ai}, which can self-improve in a data-driven manner, following the machine learning paradigm. This work hypothesises that \glspl{llm} provides the missing piece of the puzzle; the optimisation mechanism. 

A direct analogy can be drawn from training neural networks, and \emph{train} symbolic systems by leveraging the power of \glspl{llm} to interpret and generate scripts.
Using this analogy, an \gls{llm} can be considered as an \emph{optimiser} equivalent to stochastic gradient descent or Adam. The actual \emph{model} in this new paradigm is an object that handles the initialisation and updates of \emph{parameters} as well as the forward pass logic, where the \emph{parameters} are a collection of symbolic scripts that the \gls{llm} generates. At every iteration, a forward pass through the model is performed, comparing it against the ground truth in the dataset, and passing the scores and feedback into the \gls{llm} which interprets the results and updates the scripts in a way that fixes the issues raised.

While many methods use \glspl{llm} for code generation, and systems such as Auto-GPT~\citep{autogpt} iteratively query \glspl{llm} to execute tasks in an agent-like manner, LangProp is the first to completely translate and apply the training paradigm used in machine learning for iterative code generation. This work draws inspiration from \textsc{Voyager}~\citep{wang2023voyager}, which introduced the idea that a collection of \gls{llm}-generated code (skill library) can be considered as sharable and fine-tunable \emph{checkpoints}. However, \textsc{Voyager}'s method is specific to Minecraft, and additional work is needed to apply its approach to other domains. LangProp is proposed, a code optimisation framework that is easily adaptable to many application domains.

LangProp is formulated as a general code optimisation framework, decoupled from a specific application domain. It is applied first to the simple settings of Sudoku puzzles and inverted pendulum control (CartPole) to show its task-agnostic nature. Then, the framework is applied to find a driving policy for a more complex autonomous driving challenge.

Autonomous driving is a key area in which model interpretability and transparency are critical. LangProp is a valuable proof of concept for building interpretable and language-instructable systems in a more automated and learnable way.
This work combines both the benefit of interpretability of expert systems while also taking a data-driven approach, exposing the system to potential failure modes and adverse scenarios during training time and iteratively optimising the system towards a well-defined driving metric so that the resulting system is robust to adverse events and potential errors in intermediate components.

The main hypotheses of this work are: (a) LangProp can generate interpretable code that learns to control a vehicle, (b) LangProp can improve driving performance with more training data in comparison to zero-shot code generation, and (c) machine learning training paradigms such as \gls{bc}, \gls{rl}~\citep{sutton_reinforcement_2018} and DAgger~\citep{ross2011reduction_dagger} can be easily transferred and applied to LangProp training.

\section{Background}
\subsection{LLMs for code generation}
Transformers~\citep{vaswani2017attention} trained on code generation tasks have shown outstanding performances~\citep{chen2021evaluating_codex,li2022competition_alphacode,nijkamp2022codegen,li2023starcoder,roziere2023code,guo2024deepseek}. Furthermore, general-purpose \glspl{llm}~\citep{NEURIPS2022_b1efde53_instructgpt_human_feedback, openai2023gpt4} trained on a large corpus of books and online text have shown remarkable capabilities of translating between natural language and code. However, there is no guarantee that the generated code is error-free. Benchmarks have been suggested to evaluate \glspl{llm} on the code generation quality~\citep{chen2021evaluating_codex,liu2023_isyourcodecorrect}. 

Code generation with execution is highly relevant to this work. \citet{cobbe2021training} and \citet{li2022competition_alphacode} used majority voting on the execution results to select code from a pool of candidates. but this is prone to favouring common wrong solutions over correct solutions. \citet{pmlr-v202-ni23b-lever} suggested a ranking mechanism using a learned verifier to assess code correctness. 
\textsc{CLAIRify}~\citep{skreta2023errors} implemented automatic iterative prompting that catches errors and provides feedback to the \gls{llm} until all issues are resolved. 

Tangentially related fields are Automated Program Repair~\citep{xia2022less_coderepair,xia2022practical_coderepair}, unit test generation~\citep{roziere2021leveraging_unittests}, and planning for code generation \citep{le2022coderl,zhang2022planning}. 
APR is typically solved as a text infill task by identifying an erroneous block of code, masking it out, and querying an \gls{llm}, providing the surrounding code as context. Planning for \glspl{llm} formulates code generation as a sequence generation task and applies \gls{rl} techniques.
, considering the current code as the state and the action is either generation of the next token in the code~\citep{le2022coderl} or transition to a refined code~\citep{zhang2022planning}.
While orthogonal to the approach in this work of iteratively generating code using a pre-trained general-purpose \gls{llm} as an optimiser, findings from these fields may be compatible with LangProp for future work.

\subsection{LLMs for automated task completion}
Literature on the use of \glspl{llm} for automated task completion is discussed in \Cref{background/llm_task_automation}. Unlike this work, the above methods require an \gls{llm} in the loop during inference, whereas the method proposed in this work only requires access to an \gls{llm} during the code optimisation stage. 

This work is inspired by \textsc{Voyager}~\citep{wang2023voyager}, which integrates environment feedback, execution errors, and self-verification into an iterative prompting mechanism for embodied control in Minecraft. \textsc{Voyager} maintains a \emph{skill library}, a collection of verified reusable code, which can be considered as \emph{checkpoints}. However, there is no mechanism to optimise or remove a sub-optimal skill in the skill library. This limitation is addressed in this work, which presents a more general code optimisation framework that can be applied to a variety of domains, including autonomous driving.

\section{The LangProp Framework}
\label{sec:langprop/framework}

\begin{figure}[t]
\includegraphics[width=\textwidth]
{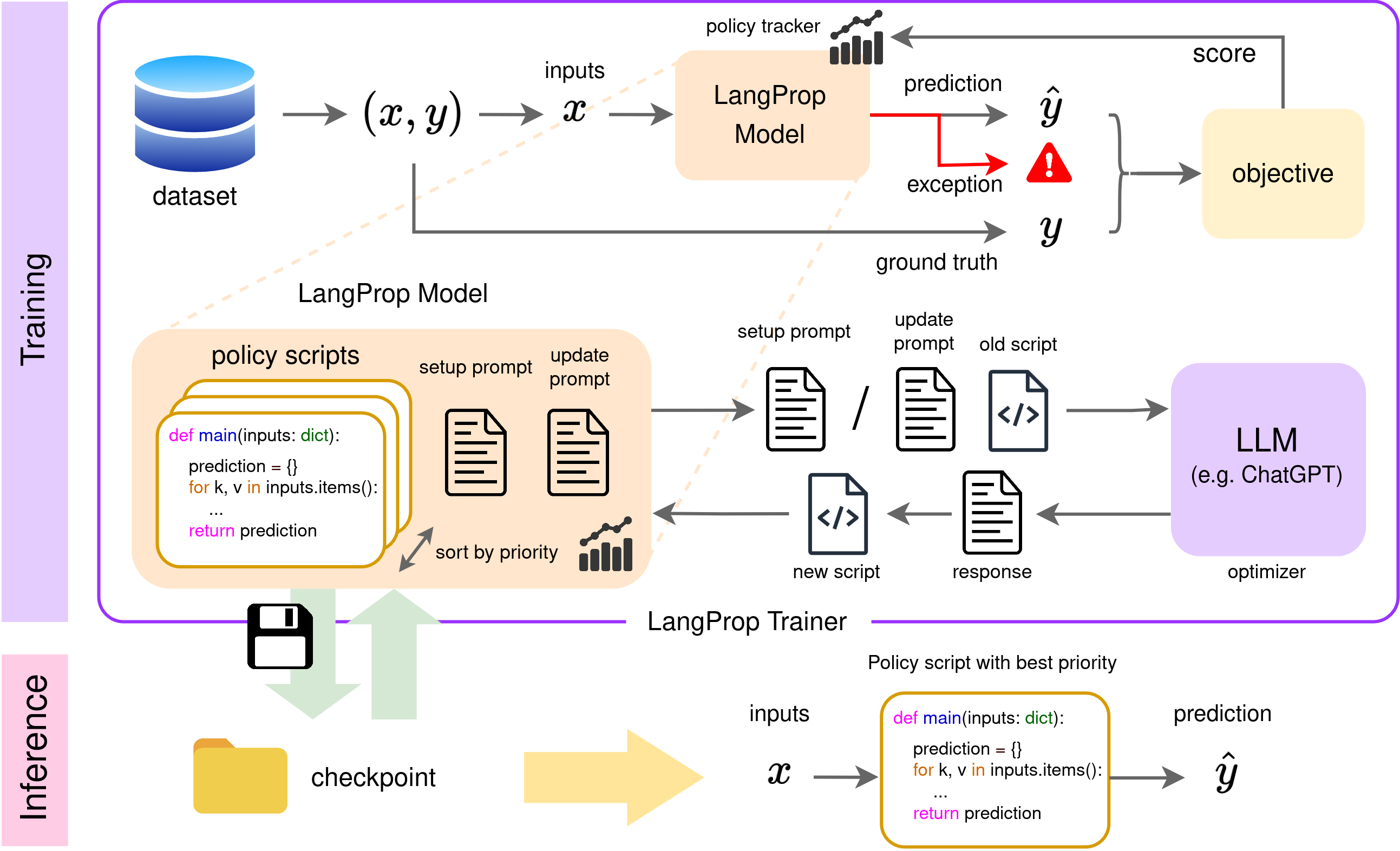}
\caption{An overview of the LangProp framework, which consists of a LangProp model, an \glsshort{llm} optimiser, and a LangProp trainer. During training, the \glsshort{llm} generates and updates the policy scripts which are evaluated against a training objective. 
Policies with higher performances are selected for updates, and the best policy is used for inference.
}
\label{fig:langprop/overview}
\end{figure}

The LangProp framework, shown in \Cref{fig:langprop/overview}, addresses a general task of optimising code on a given metric of success in a data-driven way, similar to how a neural network is optimised on an objective function. LangProp performs iterative prompting to improve code performance, using the inputs, outputs, exceptions, metric scores, and any environmental feedback to inform the \gls{llm} upon updates. The updates in LangProp are performed using a form of an evolutionary algorithm~\citep{back1993overview_evolution}. The following sections describe the key concepts in LangProp in more detail.

\subsection{Model definition}
The LangProp model consists of a setup prompt, an update prompt, and a collection of executable code generated by the \gls{llm}, which this work will refer to as \emph{policies}. While neural models are parameterised by floating-point weights, the \emph{parameters} of a LangProp model is the set of policies. Each policy is associated with an executable \emph{script} as well as a statistics tracker, which updates the \emph{priority}, an aggregate measure of the policy's performance with respect to the training objective. The priority is used to rerank policies so that the best-performing policies are used for updates and inference. 

\subsubsection{Policy setup}
The initialisation of the policies is done similarly to zero-shot code generation.
The definition and specification of the requested function are given as a docstring of the function, including the names and types of the inputs and outputs, what the function is supposed to achieve, and a template for the function. Chain-of-Thought prompting~\citep{wei2022chainofthought} is also adopted. Examples of setup prompts can be found in \Cref{sec:langprop/setup_prompt_example}. Responses from the \gls{llm} are parsed to extract the solution code snippets. Multiple responses are collected to ensure the diversity of the initial policies.

\subsubsection{Training objective}
The difference between LangProp over typical usage of \glspl{llm} for code generation is that it performs code optimisation in a metric- and data-driven manner. In many tasks, it is easier to provide a dataset of inputs and ground truth corresponding outputs rather than to accurately specify the requirements for a valid solution or write comprehensive unit tests. Similar to how neural networks are trained, the user defines an objective function that measures how accurate the policy prediction is against the ground truth, e.g. L1 or L2 loss. A penalty is given if the policy raises any exception (e.g. syntax error) while executing the code.

\subsubsection{Forward-pass and feedback}
Similar to training neural networks, LangProp assumes a dataset of inputs and associated ground truth labels for supervised learning (or rewards for \gls{rl}, discussed in \Cref{sec:langprop/agent_training}). 
For every batch update, the inputs are fed into all the policies currently in the LangProp model to make predictions, equivalent to a \emph{forward-pass}. For each policy, the prediction is evaluated by the objective function which returns a \emph{score}. If an exception is raised during execution of a policy script, it is caught by the model and an exception penalty is returned as a score instead. 

The execution results, which include the score, exception trace, and any printed messages from the execution, are fed back into the model and are recorded by the policy tracker. This is analogous to how parameters in a neural network are assigned gradients during back-propagation.
\footnote{
The current LangProp implementation is limited to an update of a single module, i.e. it does not yet accommodate for chaining of modules. 
This was attempted by making the \gls{llm} generate docstrings of helper functions to initiate submodules, and tracking submodule priorities. However, version tracking of submodules and the mechanism of providing feedback for submodule updates were substantial challenges. 
While LangProp v1 does not implement the full back-propagation algorithm, a single-layer feedback operation is referred to as \emph{back-prop} to highlight the similarities and encourage future research. 
} 
This information stored by the tracker is used in the policy update step in \Cref{sec:langprop/policy_update}.

\subsection{Model forward pass definition}
The LangProp module captures printed outputs and exceptions and stores them in the policy tracker along with the corresponding inputs during a forward pass. The Python code snippet extracted from the \gls{llm}'s response and saved as a text string is executed using the \texttt{exec} function in Python. The local scope variables can be accessed via \texttt{locals}.

\python{code/langprop/langprop_model_forward.py}{code:langprop/model_forward}{Forward passing mechanism of the LangProp module (extract)}

\subsubsection{Priority}
The priority is, simply put, an average of scores with respect to the training objective. In case a small batch size is required for faster computation, a running average of the scores is used as the priority rather than ranking the policies' performance based on scores from the current batch alone, which may result in highly stochastic results. This is sufficient for supervised learning with a fixed-size dataset. As discussed later in \Cref{sec:langprop/agent_training}, however, a more complex training method such as \gls{rl} or DAgger~\citep{ross2011reduction_dagger} has a non-stationary training distribution. Therefore, an exponential averaging with a discount factor of $\gamma \in (0, 1]$ is used:
\begin{align}
\begin{split}
P_{i,k} &= \frac{\left(\sum_{j=1}^{N_k^B} s_{i,j,k}\right) + W_{i,k-1} P_{i,k-1}}{N_k^B + {W}_{i,k-1}},\\
{W}_{i,k} &= \gamma \left(N_k^B + {W}_{i,k-1}\right).
\end{split}
\end{align}

Here, $N_k^B$, $P_{i,k}$ and $W_{i,k}$ are the batch size, priority, and priority weighting of the $k$-th batch for the $i$-th policy, respectively, and $s_{i,j,k}$ is the objective score of the $i$-th policy for the $j$-th element in the $k$-th batch. Initial conditions are $P_{i,0} = 0$ and $W_{i,0} = 0$. Weighting recent scores higher ensures that policies with higher priorities have high performance on the most up-to-date dataset. 

\begin{figure}[ht]
\centering
\includegraphics[width=0.9\textwidth]
{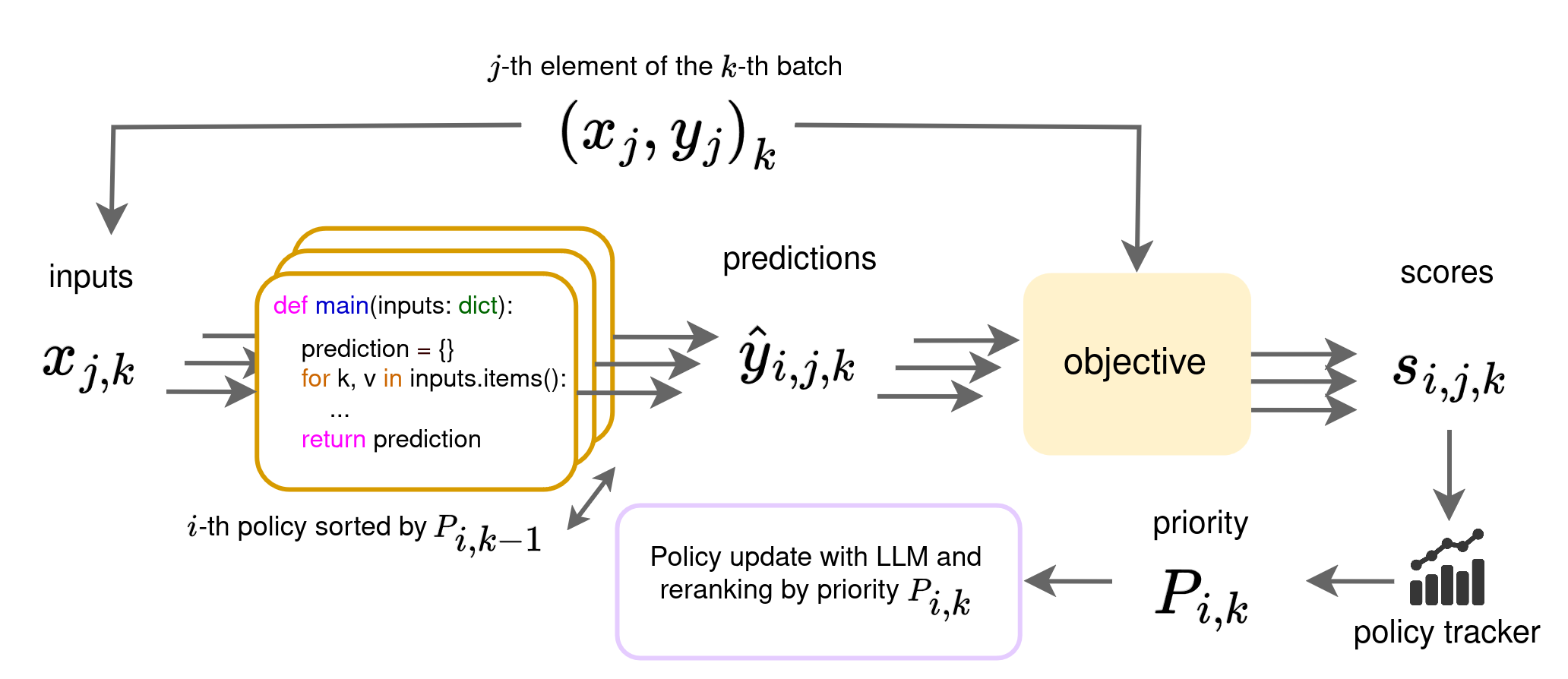}
\caption{The policy evaluation and update mechanism. The performances of the policies are monitored and aggregated over time by a policy tracker as \emph{priorities}, used to rerank the policies. 
}
\label{fig:langprop/updates}
\end{figure}

\subsubsection{Policy reranking and update}
\label{sec:langprop/policy_update}
This step updates the model based on the most recent forward-backward pass and updated priorities. This corresponds to the optimisation step in neural network training, where parameters are updated based on gradients computed on the most recent batch. The update step is illustrated in \Cref{fig:langprop/updates}. First, the policies are reranked by the priorities and the top $N^K$ number of policies are kept, out of which the top $N^U$ policies are selected for updates. For each of these policies, the policy tracker is queried for the worst-case input-output pairs in the training batch, namely that with the lowest objective score. The tracker returns the corresponding input, output and score, along with any exception or print messages during the execution. This information, together with the old policy script, is embedded into the update prompt by a prompt template engine (\Cref{sec:langprop/template_engine}). The update prompt is passed to the \gls{llm}, which returns $N^R$ responses containing new policy scripts for each of the $N^U$ policies chosen for updates.

After the update, there are $N^U \times N^R$ new policies and up to $N^K$ old policies. To initialise the new policies with sensible priorities, objective scores for the new policies are evaluated by performing the forward-backward pass, using the same training samples as the current update. Finally, all the policies are sorted by their priorities, ready for inference or training on a new batch.

\subsection{Prompt template engine}
\label{sec:langprop/template_engine}
During the policy update stage, a dynamic prompting mechanism is required to embed information about the input, predicted output, ground truth, exception, print messages, and the policy script to be revised. The logic to generate these prompts is sometimes complex, for example, predictions are only made when there are no exceptions. To enable flexible prompt generation while avoiding any hardcoding of the prompts in the codebase, A simple yet powerful prompt template is developed that can parse variables, execute Python code embedded within the prompt, and import sub-prompts from other files, and are included in the open-sourced solution (\Cref{sec:langprop/implementation}). The update prompt examples shown in \Cref{sec:langprop/update_prompt_example} make extensive use of the policy template engine's capabilities.

In the template engine, every line that begins with ``\texttt{\#}'' is treated as comments.
Every line that begins with ``\texttt{\$~}'' or line blocks in between ``\texttt{\$begin}'' and ``\texttt{\$end}'' are treated as executable Python code, as well as everything surrounded by \texttt{\{\{ \}\}} in a single line. If a ``\texttt{print}'' function is used within the prompt template, it will execute the Python code inside the print function and render the resulting string as a part of the prompt. Variables can be passed to the prompt template engine, and are made accessible in the local scope of the prompt template.

As an example, consider the following prompt template. 

\prompt{code/langprop/prompt_template.py}{code:langprop/prompt_template}{Example prompt template}
In this case,
\texttt{\small read\_template(``example'', people=[``Tom'', ``Jerry''])} resolves to: ``Tom and Jerry work here.\symbol{92}nTom is employee No. 1.\symbol{92}nJerry is employee No. 2.''.

\subsection{Trainer forward-backward definition}
The trainer has a similar abstraction to deep learning training. At every step, it triggers a forward method that calls the policy and stores the inputs, the policy's prediction, and the expected output, and a backward method that updates the policy tracker with the scores, exceptions, or any feedback.

\python{code/langprop/langprop_trainer.py}{code:langprop/trainer}{Forward-backward pass in the LangProp Trainer (extract)}

\subsection{Training paradigm}

LangProp mirrors the code abstraction of PyTorch~\citep{paszke2019pytorch} and PyTorch Lightning~\citep{falcon2019pytorchlightning} for the module and trainer interfaces. This allows LangProp as a framework to be task-agnostic, making it easily applicable to a range of domains and use cases. Moreover, it helps highlight the similarities between neural network optimisation and code optimisation using LangProp and facilitates a smooth integration of other neural network training paradigms.

Importantly, LangProp's internal implementation does not depend on PyTorch or PyTorch Lightning. LangProp supports PyTorch datasets and data loaders, as well as any iterable dataset object for training and validation. \Cref{code:langprop/training} shows an example of a standard LangProp training script.
The design of the module and trainer interfaces are inspired by PyTorch~\citep{paszke2019pytorch} and PyTorch Lightning~\citep{falcon2019pytorchlightning}, respectively.

\python{code/langprop/langprop_train.py}{code:langprop/training}{Training a LangProp model with a trainer. The model can be instantiated from a path to the setup and update prompts that specify the task to be learned.}

After every training step on a mini-batch, the trainer saves a \emph{checkpoint}, which consists of the setup prompt, update prompt template, the policy scripts (maximum of $N^K + N^U \times N^R$), and the statistics monitored by the policy tracker (priorities $P$ and priority weights $W$). Since these can be stored as text or JSON files, the size of a checkpoint is in the order of a few hundred kilobytes. Checkpoints can be used to resume training, fine-tune the model, or for inference.

\python{code/langprop/langprop_inference.py}{code:langprop/inference}{Inference with a LangProp model checkpoint.}

\Cref{code:langprop/inference} shows how a LangProp checkpoint can be loaded and used for inference. The policy with the highest priority is used. Since policies are \emph{parameterised} as executable code, the use of an \gls{llm} is only required during training, not during inference. Since querying \glspl{llm} is both expensive and slow, this is a key advantage of the LangProp approach, which makes integration of \glspl{llm} more feasible for real-time applications, such as robotics and autonomous driving.

\section{General domain experiments}
LangProp's code optimisation capability is demonstrated in three domains with increasing complexity. The GPT 3.5 Turbo 16k model~\citep{chatgpt} is used as a backbone \gls{llm}.

\subsection{Generalised Sudoku}
A generalised Sudoku puzzle consists of $W \times H$ subblocks, each with $H \times W$ elements, where $H$ and $W$ represent height and width, respectively. A valid solution places numbers from 1 to $WH$ in each cell, such that each row, column and subblock contains no repeated numbers. LangProp is trained on this problem given $100$ samples of unsolved Sudoku puzzles as input and corresponding solutions as output. The training objective is defined as the correctness of the arrived solution, i.e. whether the puzzle completed by a LangProp-learned policy is a valid Sudoku puzzle solution. The setup and update prompts are in \Cref{sec:langprop/model_and_prompts}. 
Due to the complexity of the task specification, the \gls{llm} queried zero-shot occasionally failed on the first attempt, confusing the task with a standard $3 \times 3$ Sudoku. LangProp filtered out incorrect results during training and identified a fully working solution. Samples of an incorrect zero-shot solution and a correct solution after LangProp training can be found in \Cref{sec:langprop/sudoku_solutions}.

\subsection{CartPole}

CartPole~\cite{brockman2016openaigym} is a widely used environment to train and test \gls{rl} algorithms. To make it feasible for LangProp to generate a policy to solve this task, the observation and action specifications are provided, following the Gymnasium documentation for CartPole-v1. The setup and update prompts are in \Cref{sec:langprop/model_and_prompts}. Queried zero-shot, the \gls{llm} generated a solution that is simplistic and does not balance the CartPole, achieving a score of 9.9 out of 500. With a simple Monte-Carlo method of optimising the policy for the total rewards, improved policies were obtained using LangProp, achieving the maximum score of 500.0. Interestingly, LangProp learned a policy that implemented a PID controller to solve the task.

\begin{wrapfigure}{r}{0.5\textwidth}
\hspace{5mm}
\includegraphics[width=0.9\linewidth]
{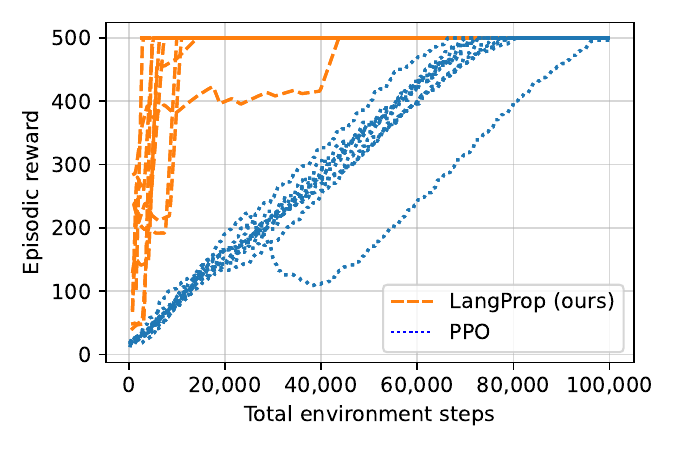}
\captionsetup{width=0.9\linewidth}
\caption{The total number of \emph{environment} steps required to learn CartPole-v1 ($10$ seeds per method) in comparison to an \glsshort{rl} method, \glsshort{ppo}. Most seeds converged to an optimal solution within $10$ LangProp updates.}
\label{fig:langprop/cartpole}
\end{wrapfigure}

\Cref{fig:langprop/cartpole} shows learning curves of the LangProp policy for $10$ different seeds. Training hyperparameters were $N^U = N^R = N^K = 3$. Out of $10$ seeds, $9$ converged to an optimal solution within $10$ LangProp updates, and within $10k$ total steps in the CartPole environment. Learning curves of \gls{ppo}~\cite{schulman_proximal_2017} are provided for comparison, which converges at around $80k$ environment steps. This shows that certain tasks may be more sample-efficient to solve with LangProp. While it is infeasible to arrive at a correct solution zero-shot, the LangProp optimisation loop allows the \gls{llm} to discover a correct solution.

Sample results can be found in \Cref{sec:langprop/cartpole_solutions}. Implementations, prompts, checkpoints, and examples of zero-shot and trained policies are available in the open-sourced repository. 

\section{Driving in CARLA}

This section describes how the LangProp framework can be used in the context of autonomous driving in CARLA. 
CARLA~\citep{dosovitskiy17a_carla} is a widely used open-sourced 3D simulator for autonomous driving research, and many prior works on CARLA have open-sourced their expert agents. 
CARLA is chosen as a benchmark since (a) autonomous driving requires interpretable driving policies, (b) CARLA has a rich collection of human-implemented expert agents to compare against, and (c) a metric-driven learnable approach would be beneficial since driving decisions are challenging planning problems, and even human-implemented experts have sub-optimal performance. \Cref{sec:background/autonomous_driving} discusses related work on autonomous driving.

\subsection{Data collection}

\subsubsection{Data agent}

\begin{figure}[h]
\includegraphics[width=\textwidth]
{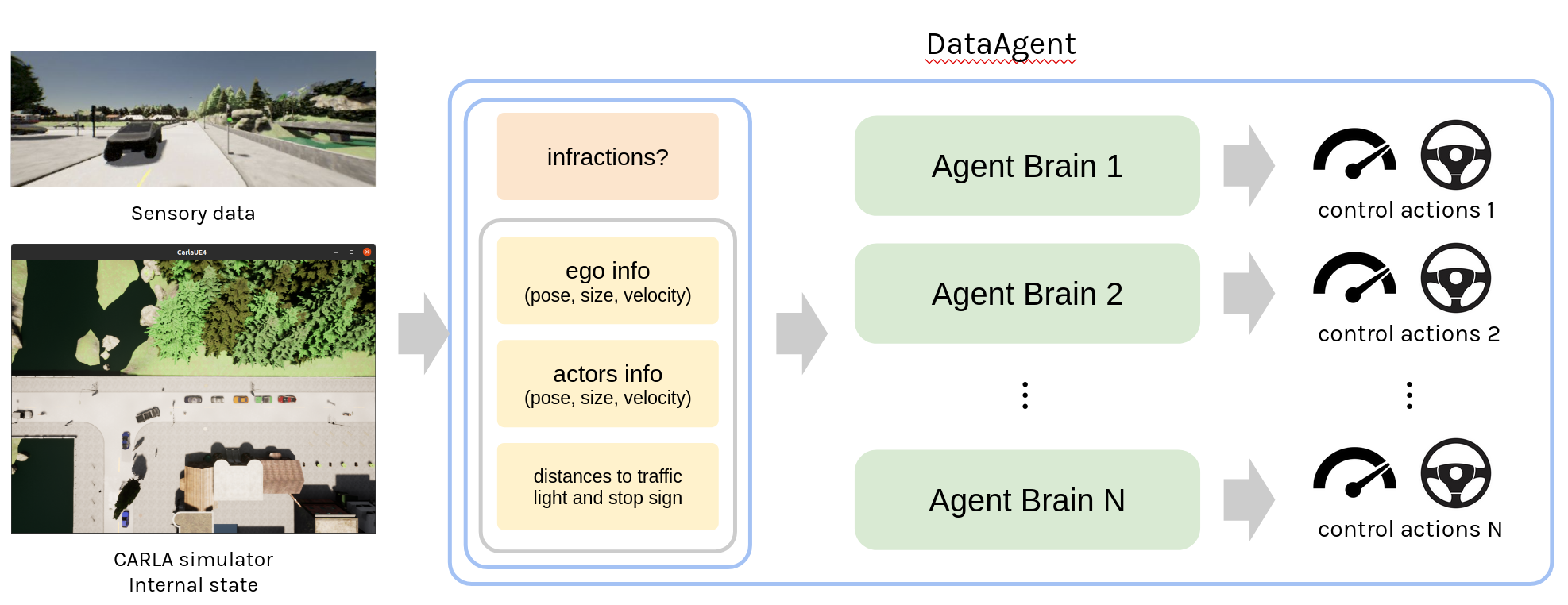}
\caption{An overview of the CARLA agents for data collection and evaluation. All agents are implemented as a standardised \texttt{DataAgent} class, with one or multiple \texttt{AgentBrain} that takes in the preprocessed observations and outputs a control action. These inputs and outputs are processed, recorded and saved by the \texttt{DataAgent} so that they can be used for data collection or for online training.}
\label{fig:langprop/langprop_data_agent}
\end{figure}

To standardise the data collection and evaluation pipeline for both the expert agent and the LangProp agent, a generic \texttt{DataAgent} is implemented. An overview is shown in \Cref{fig:langprop/langprop_data_agent}. It collects basic privileged information from the CARLA environment: the 3D bounding box coordinates of the actors in the scene (pedestrians, vehicles, traffic lights, and stop signs), the velocity of the pedestrians and vehicles, distances to the next traffic light and stop sign in the current lane, and the next waypoint to navigate towards. In addition, it also collects the RGB, depth, lidar, segmentation, top-down view, and the expert's control actions which can be used to train image-based driving policies. This standardised data collection agent is decoupled from the expert agent and the LangProp agent, and has the option of turning off sensors that are not used for data collection to save computation time and data storage. 

The data collection agent itself does not have a driving policy. It expects a separate \texttt{AgentBrain} that takes a dictionary of scene information curated by the data agent as input and outputs a vehicle control action (throttle, brake, and steering). All driving agents inherit from the \texttt{DataAgent} class, each with an \texttt{AgentBrain} that implements its driving policy. It is also possible to chain multiple agent brains as an array, where the previous agent brain's control decision is provided as an additional input to the next agent brain. This is useful for the DAgger agent and online agent, which require expert supervision during online rollouts.

\subsubsection{Expert design}
An expert agent is implemented for data collection and to provide action labels to train the LangProp agent with \gls{il}. While TransFuser~\citep{chitta2022transfuser} and TF++~\citep{jaeger2023hidden_tfplus} use a computationally expensive 3D bounding box collision detection algorithm, and InterFuser~\citep{shao2023safety_interfuser} uses line collision which is faster but less accurate, the LangProp expert uses an efficient polygon collision detection algorithm between ground-projected bounding boxes. 

The LangProp expert only uses the data collected by the data agent to ensure that the LangProp agent has access to the same privileged information as the expert agent. For every interval of $0.25~s$ up to $2~s$ into the future, whether the ego vehicle polygon will intersect any of the actor polygons is evaluated, assuming that the ego vehicle will maintain velocity, and the other actors will move in the current orientation with a speed less than or equal to the current speed. The ego vehicle polygon is padded forward by $2~m$, and by $2~m$ either left or right upon lane changes. Apart from lane changing, only actors that are ahead of the ego vehicle are considered, i.e. with a field of view of $180\degree$. A traffic light and/or stop sign that affects the vehicle is identified by querying the associated waypoints in the CARLA simulator. For pedestrians, vehicles, traffic lights, and stop signs, the distances to the obstacles are calculated. The normal driving speed is $6~m/s$ (``MOVE''). If any of the distances are reachable within $2~s$ with a $2~m$ margin (``SLOW''), the target speed is set to the speed which allows a $2~s$ margin, and if the distance is below $2~m$ (``STOP''), the target speed is set to $0~m/s$. Steering is evaluated by calculating the angle to the next waypoint, which is $4~m$ ahead of the current position of the ego vehicle. A PID controller is used for low-level control to convert the target speed and angle to throttle, brake, and steering.

\subsection{Training the LangProp agent}
\subsubsection{LangProp agent}

\begin{figure}[h]
\includegraphics[width=\textwidth]
{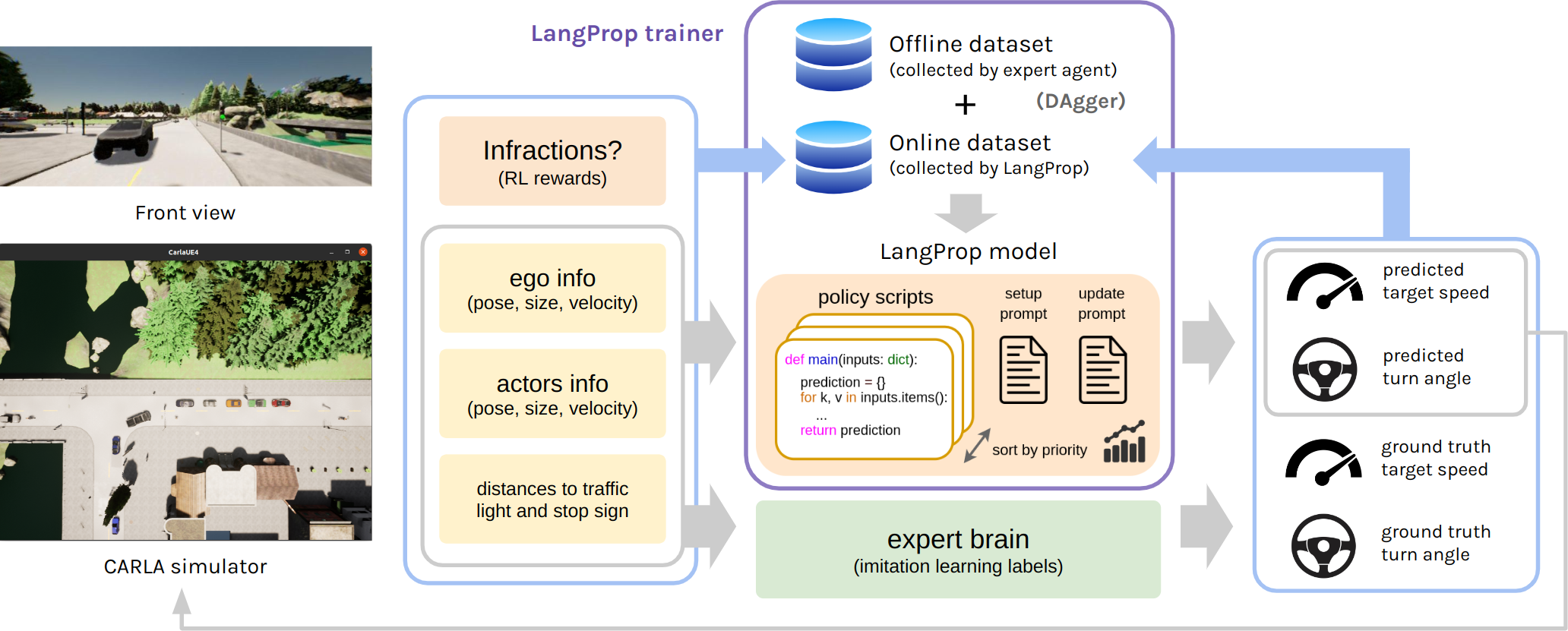}
\caption{An overview of the LangProp agent training pipeline. The LangProp model is updated on a dataset that includes both offline expert data as well as online LangProp data annotated with expert actions, similar to DAgger. The agent is given negative rewards upon infraction.}
\label{fig:langprop/langprop_driver}
\end{figure}

Similarly to all the baseline experts, privileged information from the CARLA simulator to the agent is provided. Unlike the baseline experts where post-processing is manually implemented, LangProp can decide how the information should be handled (e.g. converting world coordinates into the ego-centric frame). For the ego vehicle, as well as for all vehicles and pedestrians within a $50~m$ radius, the following information is provided: the location (in world coordinates), orientation, speed, length, width of the actor,
the target waypoint ($4~m$ ahead, used by other baseline experts), and the distances to a red traffic light and stop sign along the current lane if they exist. 
Importantly, all actors are provided without filtering, even if they are irrelevant to the driving agent. Given this information, the LangProp policy is expected to return a desired speed level (``MOVE'': $6~m/s$, ``SLOW'': $1~m/s$, ``STOP'': $0~m/s$)\footnote{While it is straightforward for the policy to directly predict the speed or acceleration as numeric values, this makes the task of designing a suitable loss function for \gls{il} more challenging and open-ended. Therefore, a categorical output is chosen, simplifying the scoring function.} and a turning angle for the ego vehicle. These are passed to an external PID controller to convert them into throttle, brake, and steering. The function specification in the setup prompt is given in Listing \Cref{sec:langprop/langprop_drive_setup} as a docstring. Given this function definition, an \gls{llm} generates policy script candidates that satisfy the specification and updates them following the procedures in \Cref{sec:langprop/framework}. 

\subsubsection{Behavioural Cloning, DAgger, and RL}
\label{sec:langprop/agent_training}
Three major training paradigms often used to train embodied agents are explored -- \gls{bc}, DAgger~\citep{ross2011reduction_dagger}, and \gls{rl}. In \gls{bc}, the accuracy of the policy outputs is measured against ground truth expert actions for a pre-collected dataset. \gls{bc} is known to have issues with out-of-distribution inputs at inference time, since the expert's policy is used to collect the training data, while the learned policy is used for rollouts at inference time. DAgger addresses this issue by labelling newly collected \emph{online} data with expert actions, and adding them to the expert-collected \emph{offline} data to form an aggregate replay buffer. CARLA runs at a frame rate of $20~Hz$. LangProp adds training samples to the replay buffer every $10$ frames, and a batch update is performed after every batch of $100$ new samples.

While DAgger solves the issue of distribution mismatch, the performance of the learned policy is still upper-bounded by the accuracy of the expert. It also does not take into account that certain inaccuracies are more critical than others. In the context of autonomous driving, actions that result in infractions such as collisions should be heavily penalised. \gls{rl} offers a way of training a policy from reward signals from the environment, which is convenient since penalties can be assigned directly upon any infractions according to the CARLA leaderboard~\citep{carla_leaderboard}. 
While \gls{rl} typically optimises for maximum returns (discounted sum of future rewards), this setting is simplified by assigning an infraction penalty if there is an infraction in the next $2~s$ window. The agent monitors infractions every 10 frames, and triggers an update upon infractions. 

Since infraction penalties are sparse signals, and will become rarer as the policies improve, two strategies are adopted; (a) \gls{rl} training is combined with \gls{il} training that provides denser signals, and (b) training data with infractions are sampled with $100$ times higher sampling probability.

\subsubsection{Training objective}
The training objective for the LangProp driving agent is given as
\begin{align}
\begin{split}
    S(a^\pi, a^{\pi_e}, a^{\pi_b}, I, E) =& \mathbbm{1}\left[(a^\pi_{\text{speed}} = a^{\pi_e}_{\text{speed}}) \land \left[ \lnot I \lor \{ (a^{\pi}_{\text{speed}} \neq a^{\pi_b}_{\text{speed}}) \land (a^{\pi_e}_{\text{speed}} \neq a^{\pi_b}_{\text{speed}}) \} \right] \right]\\ 
    &+ r_{\text{infrac}} \mathbbm{1}(I \land  (a^{\pi}_{\text{speed}} = a^{\pi_b}_{\text{speed}}) \land (a^{\pi_e}_{\text{speed}} \neq a^{\pi_b}_{\text{speed}}))\\
    &+ r_{\text{angle}} \mathbbm{1}(|a^\pi_{\text{angle}} - a^{\pi_e}_{\text{angle}}| > \theta_{\text{max}}) + r_{\text{error}}\mathbbm{1}(E),
\end{split}
\end{align}
where $a^\pi$, $a^{\pi_e}$ and $a^{\pi_b}$ are actions taken by the current policy, expert policy, and behaviour policy used to collect the training sample, respectively, $I$ and $E$ are boolean variables for infraction and exception occurrences, $r_{\text{infrac}} = r_{\text{error}} = r_{\text{angle}} = -10$ are penalties for infraction, exception, and exceeding angle error of $\theta_{\text{max}} = 10\degree$, and $\mathbbm{1}$ equates to 1 if the boolean argument is true, and 0 otherwise. The expert is only imitated when there are no infractions, or if the expert was not the behaviour policy that incurred the infraction, and an infraction cost is only given when the current policy takes the same action as the behavioural policy that caused the infraction when the expert chose a different action.

\subsubsection{Training strategy}
The training pipeline for the LangProp driving agent is shown in \Cref{fig:langprop/langprop_driver}. For all the LangProp agents, the training data was collected only on the training routes in CARLA leaderboard~\citep{carla_leaderboard}, and data collected on the test routes by the expert agent with expert action labels is used as the validation dataset. See \Cref{sec:langprop/carla_benchmark} for more details on the routes. For the LangProp agent trained offline, only samples collected by the expert agent were used as training data. For the online training, only samples collected by the current LangProp model's inference policy were used, i.e. the policy code with the highest priority at the time of rollout. For DAgger training, a mix of $1000$ training samples collected offline and $1000$ samples collected online in every replay batch were used to evaluate the objective score. Strictly speaking, DAgger~\citep{ross2011reduction_dagger} should incrementally add new online samples to a buffer initialised with offline samples. However, this prevented the LangProp model from learning from infractions during the early stages of the training, since online samples with infractions are the minority of all the samples. For this reason, an even split between offline and online samples was maintained throughout the training, with a sampling weight of $100$ for samples with infractions. Sampling is without replacements, so that a particular training sample is only sampled once per replay batch.

\subsubsection{Hyperparameters}
Notable training hyperparameters are the number of policies chosen for updates $N^U = 2$, the number of responses per query $N^R = 2$, the number of policies to keep $N^K = 20$, the frequency of batch updates (every $100$ new samples in the replay buffer), batch sizes for online replay data ($1000$) and offline expert data ($1000$), the sampling weight for infractions ($100$), and the infraction, exception, and angle penalties ($r_{\text{infrac}} = r_{\text{error}} = r_{\text{angle}} = -10$). For better performance, it is possible to increase $N^U$, $N^R$, and $N^K$, but with a trade-off of computational time and the cost of using OpenAI API. With this experiment setting, around 700 training steps are taken, 1400 queries are made, and 2800 responses are received from GPT 3.5 per training job, which costs roughly $\$150$.

\subsection{Benchmark}

\subsubsection{Baselines}
The LangProp agent was compared against \gls{rl} agents with privileged information (Roach~\citep{zhang2021end_carla_roach}, TCP~\citep{wu2022trajectory_tcp}) as well as human-written experts (TransFuser~\citep{chitta2022transfuser}, InterFuser~\citep{shao2023safety_interfuser}, TF++~\citep{jaeger2023hidden_tfplus}, ours). The official training and testing routes provided by the CARLA leaderboard~\citep{carla_leaderboard} were used, as well as the Longest6 benchmark~\citep{chitta2022transfuser} that has longer routes with denser traffic. For the LangProp agent, only the training routes are used for imitation/\gls{rl} at training time, and the saved checkpoints are used for inference during evaluation runs on different routes. 

\subsubsection{CARLA leaderboard}
\label{sec:langprop/carla_benchmark}
The driving scores are computed by the CARLA leaderboard evaluator~\citep{carla_leaderboard}, using the official training and test routes, and the Longest6 benchmark provided by \citet{chitta2022transfuser}. There are towns $1-6$ across the benchmarks. Towns $7-10$ are also used in the official online leaderboard. A breakdown of routes for each benchmark is shown in \Cref{table:langprop/carla_towns}. Towns $2$ and $5$ are withheld in the training routes and only appear in the testing routes and the Longest6 benchmark. The Longest6 benchmark has longer routes with denser traffic. 

The main metric of the leaderboard is the driving score, which is a product of the route completion percentage $\bar{R}$ and the infraction factor $\bar{I}$. It is evaluated as $\frac{1}{N}\sum_i^N(R_i I_i)$, where $i$ denotes the index of the $N$ routes used for evaluation, $R_i$ is the percentage of route completion of the $i$-th route, and $I_i$ is the infraction factor of the $i$-th route. The infraction factor is a product of infraction coefficients for pedestrian collision ($0.5$), vehicle collision ($0.60$), collision with static objects ($0.65$), running a red light ($0.70$), and running a stop sign ($0.80$). The driving score per route is equal to the route completion $R_i$ when there are no infractions, and is discounted for every infraction by a corresponding infraction factor. Note that in the Longest6 benchmark, the authors decided to remove the stop sign penalty by setting its infraction coefficient to $1.0$, which is adhered to in the following experiments.

{
\renewcommand{\arraystretch}{1.2}
\begin{table}[t]
\small
\caption{A breakdown of the number of routes per town (``num''), the average length of the routes per town (``avg. dist.''), and traffic density for the training routes (``density''), testing routes, and the Longest6 benchmark.}
\begin{center}
\begin{tabular*}{\textwidth}{@{\extracolsep{\fill}}l rrr rrr rrr}
\toprule
\multirow{2}{1em}{\bfseries Routes} &\multicolumn{3}{c}{\bfseries Training routes} &\multicolumn{3}{c}{\bfseries Testing routes} &\multicolumn{3}{c}{\bfseries Longest6}
\\ 
\cmidrule(rl){2-4} \cmidrule(rl){5-7} \cmidrule(rl){8-10}
& num & avg. dist. & density & num & avg. dist. & density & num & avg. dist. & density
\\ 
\midrule
Town 1 & 10 & 776.3 & 120   & - & - & 120       & 6 & 898.8 & 500 \\
Town 2 & - & - & 100        & 6 & 911.7 & 100   & 6 & 911.7 & 500 \\
Town 3 & 20 & 1392.5 & 120  & - & - & 120       & 6 & 1797.5 & 500 \\
Town 4 & 10 & 2262.6 & 200  & 10 & 2177.8 & 200 & 6 & 2102.4 & 500 \\
Town 5 & - & - & 120        & 10 & 1230.1 & 120 & 6 & 1444.7 & 500 \\
Town 6 & 10 & 1915.4 & 150  & - & - & 150       & 6 & 2116.7 & 500 \\
\bottomrule
\end{tabular*}
\end{center}
\label{table:langprop/carla_towns}
\end{table}
}

CARLA version 0.9.10 is used for the experiments to maintain consistency with other baseline experts that assume this version. The LangProp expert has been tested both on CARLA version 0.9.10 and version 0.9.11. 

\subsection{Experiments}
\subsubsection{Results}

{\renewcommand{\arraystretch}{1.3}
\begin{table*}[t]
\small
\caption{Driving performance of expert drivers in CARLA version 0.9.10. The driving score is a product of the route completion percentage $\bar{R}$ and the infraction factor $\bar{I}$. \glsshort{il} and \glsshort{rl} stand for \glsdesc{il} and \glsdesc{rl}. DAgger uses both online and offline data.}
\begin{center}\small
\begin{tabular*}{\textwidth}{@{\extracolsep{\fill}}l ccc ccc ccc}
\toprule
\multirow{2}{1em}{\bfseries Method} &\multicolumn{3}{c}{\bfseries Training routes} &\multicolumn{3}{c}{\bfseries Testing routes} &\multicolumn{3}{c}{\bfseries Longest6}
\\ 
\cmidrule(rl){2-4} \cmidrule(rl){5-7} \cmidrule(rl){8-10}
&\bfseries Score $\uparrow$ 
&$\bar{R}\uparrow$
&$\bar{I}\uparrow$
&\bfseries Score $\uparrow$ 
&$\bar{R}\uparrow$ 
&$\bar{I}\uparrow$
&\bfseries Score $\uparrow$ 
&$\bar{R}\uparrow$ 
&$\bar{I}\uparrow$
\\ \midrule
Roach expert & 57.8 & 95.9 & 0.61 & 63.4 & 98.8 & 0.64 & 54.9 & 81.7 & 0.67 \\
TCP expert & 64.3 & 92.3 & 0.71 & 72.9 & 93.2 & 0.77 & 46.9 & 63.1 & 0.76 \\
TransFuser expert & 69.8 & 94.5 & 0.74 & 73.1 & 91.3 & 0.80 & 70.8 & 81.2 & 0.88 \\
InterFuser expert & 69.6 & 83.1 & 0.86 & 78.6 & 81.7 & 0.97 & 48.0 & 56.0 & 0.89 \\
TF++ expert & \bfseries 90.8 & 95.9 & 0.94 & 86.1 & 91.5 & 0.94 & \bfseries 76.4 & 84.4 & 0.90 \\
\bfseries Our expert & 88.9 & 92.8 & 0.95 & \bfseries 95.2 & 98.3 & 0.97 & 72.7 & 78.6 & 0.92 \\
\midrule
\bfseries LangProp Agents &&&&&&&&&\\
Offline IL & 0.07 & 0.37 & 0.97 & 0.00 & 0.00 & 1.00 & 0.00 & 0.00 & 1.00 \\
DAgger IL & 36.2 & 94.5 & 0.40 & 41.3 & 95.3 & 0.44 & 22.6 & 87.4 & 0.30 \\
DAgger IL/RL & 64.2 & 90.0 & 0.72 & 61.2 & 95.2 & 0.64 & 43.7 & 71.1 & 0.65 \\
Online IL/RL & \bfseries 70.3 & 90.5 & 0.78 & \bfseries 80.9 & 92.0 & 0.89 & \bfseries 55.0 & 75.7 & 0.73 \\
\bottomrule
\end{tabular*}
\end{center}
\label{table:langprop/carla_results}
\end{table*}
}

The results are shown in \Cref{table:langprop/carla_results}. 
The LangProp expert and the TF++ expert significantly outperformed all other expert agents in all routes, and the LangProp expert outperformed TF++ by a margin on the test routes. The core collision avoidance logic is just 100 lines of code, with additional preprocessing and tooling for data collection. From the breakdown of the scores, the LangProp expert seems to prioritise safer driving with fewer infractions (higher infraction factor $\bar{I}$) by trading off route completion compared to TF++ in the Longest6 benchmark.

For the LangProp agent, it could be observed that training using offline samples, DAgger, and online samples improves performance in this order. Adding the infraction penalties as an additional \gls{rl} objective further improved the performance. The best-performing agent, LangProp trained on online data with \gls{il} and \gls{rl}, achieved better performance than the Roach expert (trained with \gls{ppo}) as well as the TransFuser and InterFuser experts (both written by researchers) on all benchmarks apart from TransFuser on the Longest6 benchmark. 
Note that TransFuser has an advantage over the Longest6 benchmark since LangProp has never seen this benchmark during training.
The driving policy generated using LangProp is shown in \Cref{sec:langprop/langprop_driving_policy}.

The result has two important implications. Firstly, the code selection metric (the training objective) plays a large role in the ultimate performance of the code. This is an important finding since prior work on code generation mostly focused on error correction given exceptions. The results demonstrate that for complex tasks, it is important to treat code generation as an iterative optimisation process rather than a zero-shot task. 
Secondly, training using LangProp exhibits similar characteristics as training in deep learning; in deep learning, it is a well-studied problem that policies trained with \gls{bc} on offline datasets do not generalise to out-of-distribution online data. DAgger and \gls{rl} are two of the common ways of addressing this problem. 
The results show that these training paradigms can also be effective when used in LangProp. 

\subsubsection{Analysis of training methods}

\begin{figure}[h]
     \subbottom[training scores on the replay buffer \label{fig:langprop/training_scores}]{%
        \includegraphics[width=0.48\linewidth]{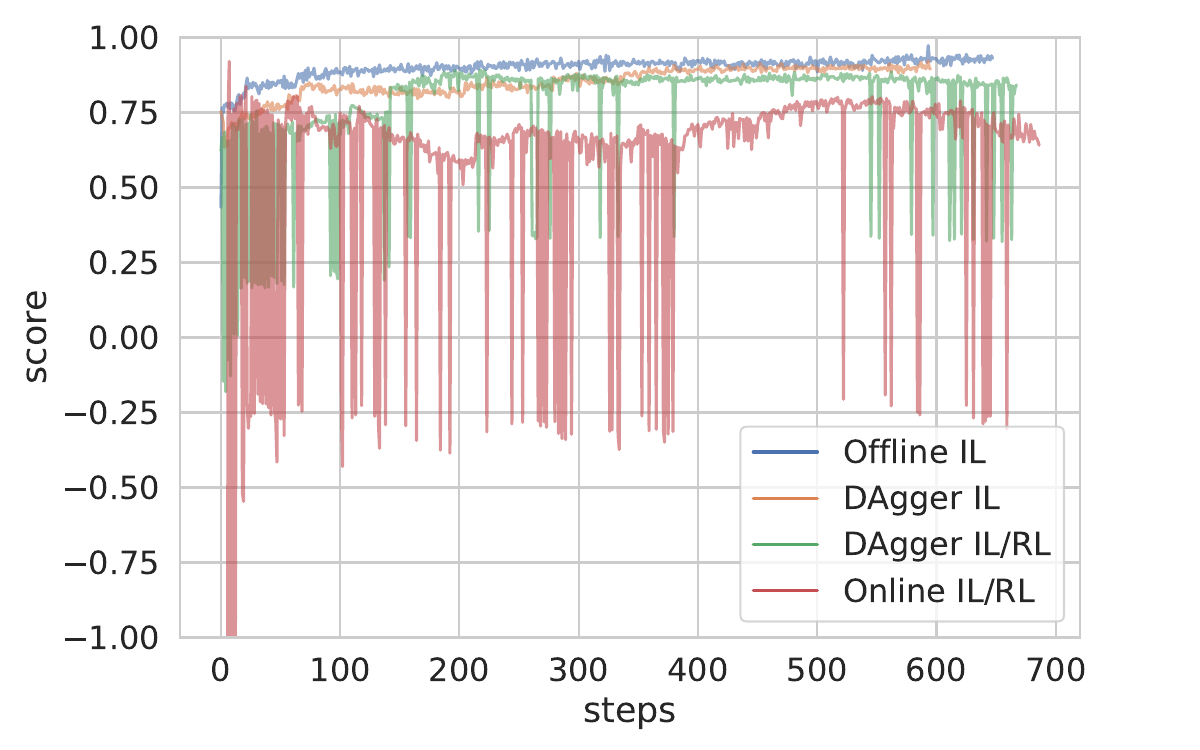}}
     \subbottom[validation scores on the offline dataset \label{fig:langprop/val_scores}]{%
        \includegraphics[width=0.48\linewidth]{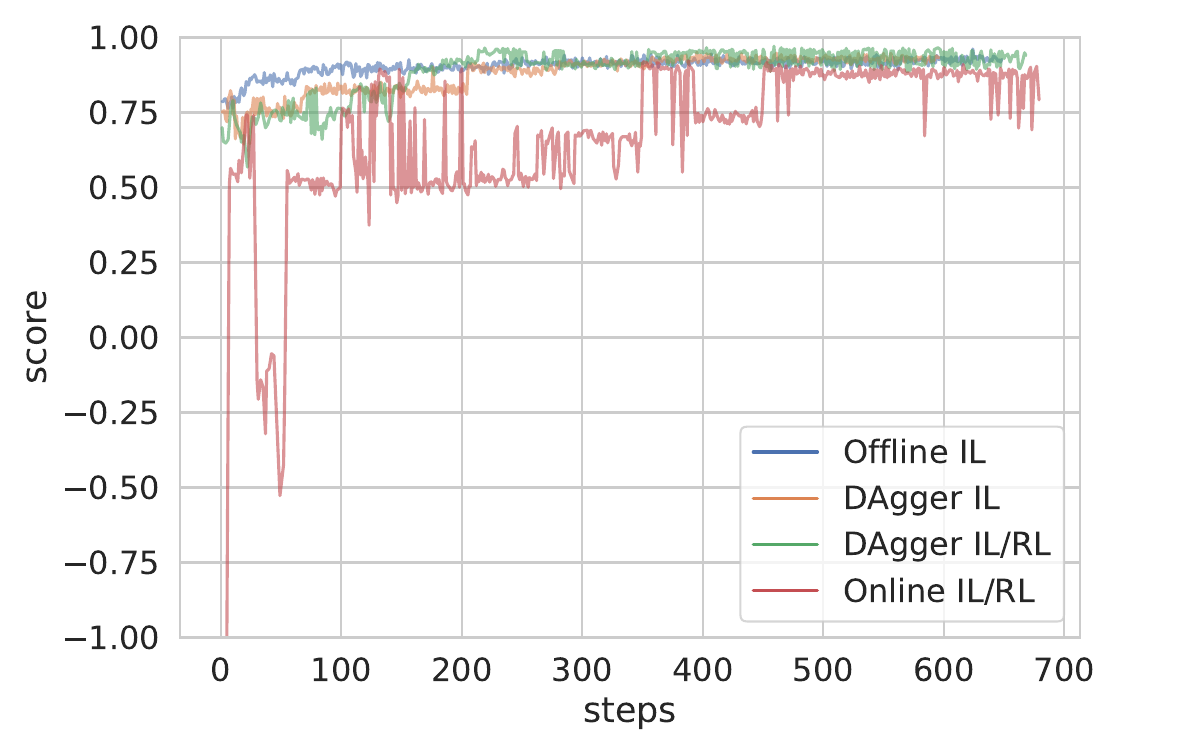}}
    \caption{Training curves for the different training methods of the LangProp agent. The training scores are evaluated on $1000$ samples from the offline training dataset and/or online replay buffer, and the validation scores are evaluated on $1000$ samples from the offline validation dataset. Updates are performed every $1000$ frames of agent driving, and upon infractions in the \glsshort{rl} setting. The score is in the range of $[-10, 1]$ due to exception penalties. The axis is limited to $[-1, 1]$ in the plots.}
\end{figure}

A common failure mode of offline trained models was that the agent remained stationary indefinitely until the timeout was reached. Upon inspection of the policy code that was generated, the cause of failure was identified to be a phenomenon known as causal confusion in \gls{il}~\citep{de2019causal}. A snippet of code responsible for such failure in one of the runs is shown in \Cref{code:langprop/causal_confusion}.

This exemplifies the interpretability of LangProp models, allowing us to directly assess the source of failure. The code predicts $0$ speed when the agent's current speed is already close to $0$. Note that this is not a failure of the LangProp algorithm, but due to such a policy maximising the \gls{il} objective on an offline dataset, bypassing the need to learn a more complex policy. This phenomenon is also common in the context of \emph{deep} \gls{il}, and can be avoided by employing training on online data, e.g. using DAgger or \gls{rl}. This work is thought to be the first to report a similar phenomenon using \glspl{llm} for policy optimisation.

\python{code/langprop/causal_confusion.py}{code:langprop/causal_confusion}{Causal confusion in offline-trained policy}

The use of online training samples alleviated the issue of causal confusion, leading to selecting policies where the agent has a sensible driving performance. This is because if the agent remains stationary, those samples will accumulate in the replay buffer, resulting in a lower priority for the causally confused policy.
Comparing the results in \Cref{table:langprop/carla_results} and the validation scores in \Cref{fig:langprop/val_scores}, it seems that the scores on the offline dataset are not indicative of the agent's driving performance. 
From the training scores on the replay buffer and/or offline dataset in \Cref{fig:langprop/training_scores}, it could be seen that the agents trained with \gls{rl} on infractions have spikes corresponding to infractions. This is due to oversampling infractions when they occur, allowing the policy update to immediately address the issue. DAgger has a milder response compared to training just on online data because the offline dataset does not include on-policy infractions. The higher rate of infractions in the training distribution may be why the online trained agent has a lower training score but has a higher driving performance.

\section{Implementation}
\label{sec:langprop/implementation}
The code used in this paper is open-sourced, and can be found at \url{https://github.com/shuishida/LangProp/}. This includes code for the general LangProp framework, applying LangProp to tasks such as Sudoku and CartPole, and training and evaluating the LangProp agent in CARLA. Pre-trained checkpoints using LangProp and videos of sample runs by the LangProp agent are also included.

\section{Conclusion}
This work presented LangProp, a framework that uses \glspl{llm} for data-driven code optimisation, and demonstrated its capability of generating and improving policies in the domains of Sudoku, CartPole and CARLA. In particular, LangProp generated driving policies in CARLA that outperform those that existed when the backbone GPT 3.5 was trained. It was shown that classical training paradigms such as \gls{bc}, DAgger, and \gls{rl} directly translate to training with LangProp, and the choices of the objective function and the training data distribution can be used to guide which policies are selected. Automatically optimising the code to maximise a given performance metric has been a key missing feature in few-shot code generation. The LangProp framework provides this feature by reformulating the machine learning training paradigm in the context of using \glspl{llm} as code optimisers and treating policy code as parameters of the model. The LangProp paradigm opens up many possibilities for data-driven machine learning with more interpretability and transparency.

\chapter{Embodied task execution in a virtual world}
\label{chapter:minecraft}

This chapter addresses the long-standing challenge of performing complex compositional tasks in Minecraft that involve spatial reasoning and planning. Minecraft is a popular game in a 3D virtual world, in which players can explore the terrain, collect resources, craft tools, and build structures, as well as gather food and combat malicious game characters for survival. It serves as a perfect evaluation benchmark for embodied agents due to the diversity of its randomly generated landscape and the open-ended and life-like nature of the tasks presented.
This chapter introduces Voggite, an embodied agent that performs tasks in Minecraft using OpenAI \gls{vpt}~\cite{baker2022_openai_vpt} as a backbone. Unlike \gls{vpt}, which is only retained to retain short-term memory with a transformer policy and struggles to disambiguate different stages of task execution, Voggite decomposes complex tasks into a sequence of subtasks, thereby solving the problem of ambiguity in the \gls{vpt} policy. The \gls{vpt} policy is fine-tuned to solve a range of life-like tasks that the original \gls{vpt} was not specifically trained for. 
Voggite was submitted to the MineRL BASALT Competition 2022~\cite{milani2023solving} and achieved 3rd place out of 63 teams. This chapter also discusses other approaches that were considered to solve the problem of learning reusable skill-learning for embodied agents from a limited dataset of expert demonstrations.

\section{Introduction}
Preceding chapters addressed the problems of learning to plan in partially observable environments for robot navigation, learning to segment skills in a self-supervised way, and learning interpretable policies for autonomous driving using \glspl{llm} to iteratively improve code policies. These are all steps towards enabling embodied agents to operate in a real-world environment and acquire skills reusable for task execution and long-term consistent planning.
In this penultimate chapter, the game of Minecraft is considered a platform to practically evaluate embodied agents on their life-like task execution capabilities. 

Minecraft is an open-ended 3D gaming environment where human players and agents can perform many life-like tasks. In Minecraft, agents explore a terrain of pre-historic natural scenery in a block-like virtual world. Players can gather raw materials from nature and process them to obtain tools, which could be used to craft and build more complex tools, structures and machines. Minecraft does not define a fixed set of challenges or requirements to complete the game; rather, it is an open-ended sandbox environment where players can unleash their curiosity and creativity. Many of the tasks that could be performed are relevant to surviving, living, exploring and creating, such as mining, crafting, building, raising animals, harvesting crops, and occasional combats with unfriendly game characters.

Many of the tasks in Minecraft are compositional. For instance, finding a diamond requires the agent to chop wood, make a crafting table from wooden planks, make a wooden pickaxe to collect cobblestones, make a stone pickaxe to collect iron, create a furnace to melt the iron into an iron pickaxe, and finally use the iron pickaxe to obtain a diamond. This requires long-term planning, guided by intuition from real-world experiences. 

It is easy for human players to draw the analogy between the real world and the Minecraft world, and learn to complete tasks in this virtual world, since many of the objects and characters in Minecraft reflect those in the real world and obey similar rules. Learning the environment dynamics from scratch using trial and error such as with \gls{rl} would be computationally expensive. On the other hand, some environment dynamics, such as how object blocks can float in mid-air, do not reflect the real world and must be learnt either from external resources (\eg documentation or manual) or through gameplay. Humans can combine the three modes of learning to adapt to new environments: transfer of prior knowledge, gathering experiences through trial and error, and integration of external or collective knowledge. A natural approach to designing artificial embodied agents for real-world tasks would be to equip them with such learning capabilities. 

Research on learning-based agents in Minecraft has been accelerated by the developments of Malmo (a Minecraft simulator for \gls{rl} agents) and MineRL~\cite{guss2019minerl} (a state-action paired dataset of over 60 million frames across a diverse set of tasks), along with competitions accompanying the MineRL dataset. Starting with the task of obtaining a diamond in the first competition~\cite{guss2019minerl}, the MineRL Competition in 2021~\cite{shah2022retrospective} introduced more life-like tasks in the BASALT track, some with difficult-to-define reward functions: finding a cave, making a waterfall, creating an animal pen, and building a house.

While earlier works on Minecraft (see \Cref{sec:background/virtual_world}) relied on semantically rich observations and knowing the agent's inventory (\ie what items and resources the agent has in possession), the recent \gls{vpt}~\cite{baker2022_openai_vpt}, a transformer foundation model trained on a large collection of videos on the Internet, successfully managed to discover diamonds taking RGB pixel inputs as observations and predicting cursor movements, mouse clicks and keyboard presses as outputs. Following the success of \gls{vpt}, the MineRL BASALT Competition was significantly updated in 2022, defining the observations to be RGB pixels and actions to be cursor movements and mouse clicks in the same way as \gls{vpt}. 
This work builds on top of \gls{vpt}, addresses its weaknesses and fine-tunes it to solve challenges specified in the MineRL BASALT Competition 2022.

\section{Background}
\subsection{OpenAI VPT}
\gls{vpt}~\cite{baker2022_openai_vpt} is a transformer-based foundation model that is trained to take in a video sequence as input and output a sequence of actions. Training such large models with \gls{bc} (\ie directly supervising them with expert action labels) requires a vast amount of expert-labelled collection of videos, which are not readily available. OpenAI solved this problem by collecting a relatively small amount of expert labels to train an \gls{idm}, a separate transformer model that also predicts a sequence of actions from a video sequence, except that while a transformer policy must predict actions auto-regressively without having access to future sequences of frames, an \gls{idm} can be trained non-causally and is provided with video frames from future time steps. Aided with such privileged information, an \gls{idm} can predict actions that were taken by the agent to produce the corresponding observations much more accurately with less training data compared to a causally trained policy. An \gls{idm} trained on around $2K$ hours of expert play was used to label $70K$ hours of gameplay video footage without action labels. Since any public video footage could be labelled by the \gls{idm}, this method of data labelling makes it possible to train a large transformer policy on vast amounts of publicly available video footage, \eg on YouTube. 

While online videos are useful for learning semantic feature representations and primary tasks, since Minecraft is an open-ended and undirected environment, the resulting policy would also be generic and undirected. To encourage the policy to learn useful task-specific behaviours, a combination of \gls{il} and \gls{rl} was used. In the \gls{il} setting, a \gls{vpt} foundation model was fine-tuned with \gls{bc} from expert demonstrations specifically for early game resource gathering and tool crafting to build a basic house. In the \gls{rl} setting, the agent was trained on rewards designed for the task of crafting a diamond pickaxe. \gls{ppg}~\cite{cobbe2021phasic}, a policy gradient algorithm similar to \gls{ppo}~\cite{schulman_proximal_2017} but with improved sample efficiency is used.

One of the limitations of the \gls{vpt} agent is that, despite its transformer architecture which allows the agent to retain temporal information, the \gls{vpt} policy takes in $128$ frames of past observations, equivalent to only $6.4~s$ of history. While this is far outstanding compared to prior work, it is still insufficient for long-term planning, and it fails to work well with tasks with multiple stages and dependencies.

\subsection{MineRL BASALT Competition 2022}

The MineRL BASALT Competition~\cite{milani2023solving}, standing for ``Benchmark
for Agents that Solve Almost Lifelike Tasks'', challenged embodied agents to solve tasks with hard-to-specify reward functions in Minecraft. 
The competition evaluated agents on their life-like task execution capabilities in four domains: finding a cave, making a waterfall, creating an animal pen, and building a house. 
Unlike many competitions with automated evaluation, agent performances are evaluated by human judges in the MineRL BASALT Competition. 600 hours of labelled expert demonstrations are provided for the four tasks. An OpenAI \gls{vpt} agent individually fine-tuned on the four tasks is treated as a baseline for the competition.

\section{Segmenting stages of task execution in Minecraft}

\subsection{Motivation}
Inspecting the performances and rollouts of the \gls{vpt} agent fine-tuned for tasks such as making a waterfall, creating an animal pen and building a house, it was noted that the agent's short temporal window of observations is hindering the agent from executing long-term plans. Let us inspect each scenario in turn.

\subsubsection{Finding a cave}
Finding a cave is a task simple enough that it could be partially solved by chance even by the baseline \gls{vpt} agent with a reactive policy; the agent must explore the environment, discover a cave-like structure, enter it, and press ESCAPE for task completion. Nevertheless, navigating a terrain with pitfalls, cliffs and pools of water, also with the threat of mobs (antagonistic game characters) is a non-trivial task. 

While the baseline agent was relatively good at exploration, it had a tendency to start to dig whenever it was trapped in a dead end or encountered obstacles. However, finding a cave by digging was not considered an allowed strategy in the competition. The agent was also not competent at backtracking or planning a long-term trajectory. 

\subsubsection{Making a waterfall}
In the task of making a waterfall, the agent holds a bucket of water that it must carry up a mountain to flip over at the summit. Once the agent creates a waterfall, it must climb down and turn around to capture the scenery in its first-person view camera. 

The baseline had a tendency to flip over the bucket at random locations and continue on its exploration without a clear intention of climbing up or down a mountain. This is not surprising, as whether the agent should climb up a mountain or down is ambiguous from the sequence of observations alone (which is a close-up view of the mountainous terrain), and the information of whether the bucket is full of water or empty is a minor detail in the observations. Learning two highly different sub-policies (one climbing up, one climbing down) based on these small differences in input is challenging. 

\subsubsection{Creating an animal pen}
\label{sec:minecraft/motivation/animal_pen}

Creating an animal pen consists of multiple stages of task execution: finding a suitable village house to build a pen next to, finding some wandering animals of the same species and luring them to the chosen location, and finally building a fence around them. Other orderings of stages to complete the task may also be possible. 

The baseline agent proceeded to explore whilst occasionally throwing fences onto the ground and nearby walls inconsistently. This may be because (a) the agent was fine-tuned from a general-purpose \gls{vpt} agent that was trained to explore, and (b) the agent only has a short-term memory of $6.4~s$, and is not able to disambiguate between whether it is meant to explore, build or lure animals at a particular time. Hence, the policy may be confused with conflicting training signals and fail to learn a temporally coherent strategy. 

\subsubsection{Building a house}
Similarly to \Cref{sec:minecraft/motivation/animal_pen}, the task of building a house also has multiple stages: exploration to find a cleared space, building the walls, making a door using a crafting table, attaching the door, and giving a tour of the house. Resources required to build the house are available in the inventory, although some specific items may need to be crafted. 
In this example as well, the baseline agent manifested inconsistent behaviour of running around the terrain while occasionally dropping resources in random locations on the way, which is not beneficial for the task of building a house. 

From the above examples, it is evident that the baseline \gls{vpt} lacks long-term memory and planning capabilities, which is significantly impacting the competence of an otherwise well-trained large transformer policy with rich prior knowledge of the Minecraft environment. In the following sections, two strategies are considered to segment the task into multiple stages to aid the policy in making temporally consistent decisions.

\subsection{Segmenting by states via Invariant Information Clustering}
\label{sec:minecraft/method/segment_state_iic}

An initially considered approach was to cluster the observations in the expert demonstrations. Since the observations in the demonstrations are RGB images and are only annotated with corresponding actions taken by the expert, a separate means to obtain such clustering must be considered. 

One approach would be to manually annotate the observations with human-identified subtasks. However, this approach would be expensive and the definition of ``subtask'' is unclear. It would be more desirable to discover temporal clusters automatically through data. 
As an alternative, a contrastive method with temporal constraints was conceived. 

\subsubsection{Background on Invariant Information Clustering}
\gls{iic}~\cite{ji2019invariant} was used as a contrastive training objective to learn cluster assignments. \gls{iic} is a method with a simple objective to maximise the mutual information between the class assignments of positively paired data samples.
It assumes a mapping function $\Phi: \mathcal{X} \rightarrow \mathcal{Z}$, where $\mathcal{X}$ is the input space, $\mathcal{Z}$ is a categorical space $\mathcal{Z} = \{1, ..., C\}$, and $C$ is the number of clusters. 

Mutual information between random variables $z$ and $z'$ is defined as:
\begin{equation}
    \label{sec:minecraft/method/mutual_information}
    I(z; z') = \int_z \int_{z'} p(z, z') \log \frac{p(z, z')}{p(z)p(z')} dz dz'.
\end{equation}

Given $N$ number of positive data samples $\{(x_n, y_n): 1 \leq n < N \}$, where $x_n$ and $y_n$ are known to belong to the same cluster, the maximisation objective is:
\begin{equation}
    \mathop{\text{max}}_\Phi I(\Phi(x); \Phi(y)).
\end{equation}

Since the goal is to learn representations with a deep neural network, $\Phi(x)$ performs soft rather than hard clustering with a softmax layer as an output. The output $\Phi(x) \in [0, 1]^C$ can be interpreted as the distribution of a discrete random variable $z$ over $C$ classes, formally given by $P(z = c|x) = \Phi_c(x)$. Similarly, $P(z' = c'|y) = \Phi_c(y)$. By marginalising over the dataset (or a batch of size $N$ in practice), a joint probability $\mathbf{P}_{cc'} = P(z = c, z' = c')$ could be considered; $\mathbf{P}_{cc'}$ denotes the probability that input $x$ is assigned to cluster $c$ and input $y$ is assigned to cluster $c'$ for a random positive pair $(x, y)$ in the dataset.  $\mathbf{P}$ is a $C \times C$ matrix given by:
\begin{equation}
    \mathbf{P} = \frac{1}{N} \sum_n^N \Phi(x_n) \cdot \Phi(y_n)^T.
\end{equation}

For most problems, for every positive pair $(x, y)$, a pair $(y, x)$ should also be a positive pair. Therefore, $\mathbf{P}$ could be symmetrised by $(\mathbf{P} + \mathbf{P}^T) / 2$.
The marginals $\mathbf{P}_c = P(z = c)$ and $\mathbf{P}_{c'} = P(z' = c')$ can be obtained by summing over the rows and columns of this matrix.

Finally, the \gls{iic} objective function is obtained by substituting the joint probability matrix $\mathbf{P}$ into the mutual information objective in \Cref{sec:minecraft/method/mutual_information} as:
\begin{equation}
    I(z; z') = I(\mathbf{P}) = \sum_{c=1}^C \sum_{c'=1}^C \mathbf{P}_{cc'} \cdot \log \frac{\mathbf{P}_{cc'}}{\mathbf{P}_{c} \cdot \mathbf{P}_{c'}}.
\end{equation}

\subsubsection{Application of IIC to Minecraft}
Similarly to how Ji \etal\cite{ji2019invariant} applied IIC to the task of image classification by pairing images to an image-augmented version of itself, an embodied agent applying actions to the environment and receiving subsequent observations can be considered as a form of applying augmentation to the observations. Temporally proximate observations are more likely to come from the same stage of a task execution. This prior could be used to harvest positive pairs from demonstrations for contrastive learning.

Given a sequence of observations $o_0, o_1, ..., o_T$, where $T$ is the length of an episode, it could be assumed that it is likely that neighbouring observations $o_t$ and $o_{t+l}$ (where $0 \leq l < L$ for lookahead $L << T$) are classified into the same stage of a task, compared to a randomly selected pair of observations in the dataset. Using this temporal constraint, contrastive learning can be applied to classify observations into clusters.

\subsubsection{Preliminary results}
\label{sec:minecraft_iic_results}

\begin{figure}[t]
\includegraphics[width=\textwidth]
{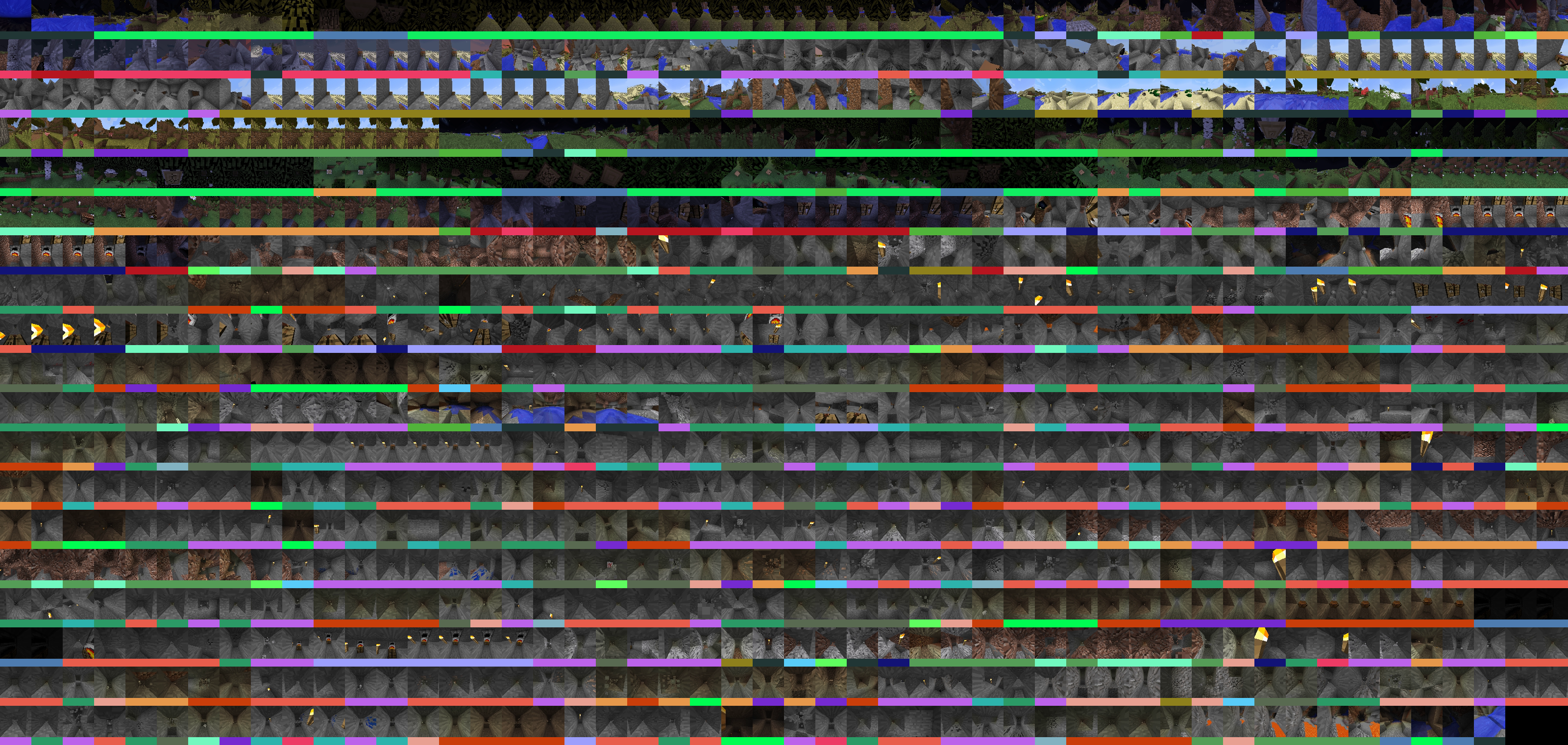}
\caption{Applying \glsshort{iic} to a video sequence of expert demonstration for the Obtain Diamond task. The example shown is for lookahead $L=1$. The video sequence is subsampled for every 10 frames and shown from left to right, top to bottom. The coloured strips for every tile of observation correspond to clusters found by applying \glsshort{iic} to the observations. There are 30 clusters in this case.}
\label{fig:minecraft/obtain_diamond_iic}
\end{figure}

\Cref{fig:minecraft/obtain_diamond_iic} shows an example of applying \gls{iic} to expert demonstrations to cluster the observations. As can be seen, there is some temporal consistency in the cluster assignments due to similarities in the observations. Some clusters appear early on in the demonstration, whereas others appear later. Identifying these correlations between clusters and task progression may be useful in identifying stages and subtasks within expert demonstrations.

While these qualitative experiments applied \gls{iic} clustering on the RGB observations, preliminary experiments were also conducted on the inventory vector space as observations, following the convention in the MineRL BASALT 2021 Competition~\cite{shah2022retrospective} (as opposed to 2022~\cite{milani2023solving}). Visualisations are included in \Cref{appendix:minecraft_inventory}.

\subsubsection{Limitations}
While this research direction of segmenting demonstration trajectories based on temporal constraints on observations was promising, several shortcomings were identified which resulted in a pivot in this work. The reasons for this pivot were as follows:

\begin{enumerate}
\item Cluster assignments determined by the function $\Phi(x)$ only consider the current observation, or a short window of observations at most if $x$ is a latent embedding from a transformer encoder. This hinders the aim of using learnt cluster assignments as task stage indicators to inform long-term planning.

\item The latent representations learnt from training $\Phi(x)$ on the \gls{iic} objective may be useful; however, this method was conceived before the release of \gls{vpt}. Since \gls{vpt} already serves as a semantically sound pre-training method, pre-training with \gls{iic} was made somewhat redundant. 

\item A unimodal cluster assignment per observation may not be a reasonable constraint, since a scene may contain multiple objects or triggers, each with a different cluster association. 
\end{enumerate}

For the above reasons, this research direction was not pursued. However, \gls{iic} pre-training may serve as a complementary technique to \gls{vpt} for learning a semantically meaningful task representation.

\subsection{Segmenting by actions via task-specific heuristics}

To fulfil the original purpose of assigning stages to agent trajectories for long-term planning and temporal consistent strategies, an alternative method is considered. This method has similarities to the Options Framework~\Cref{sec:background/options_framework} in that it switches between stages only upon certain triggers. Unlike the \gls{iic}-based method proposed in \Cref{sec:minecraft/method/segment_state_iic}, which relies on a clustering function $\Phi(x)$ to perform a classification at every step in the agent rollout, this newly suggested method propagates the stage information forward temporally.

While learning these trigger points is both attractive and desirable, this would require some other way of constraining the learning problem; if supervised learning is employed, some form of data annotation is required; if stage segmentation is to be learnt via self-supervised learning, what constitutes as distinct ``stages'' must be defined as a training constraint. This problem, also known as change-point detection, is a complex research question on its own. 

\begin{figure}[t]
\includegraphics[width=\textwidth]
{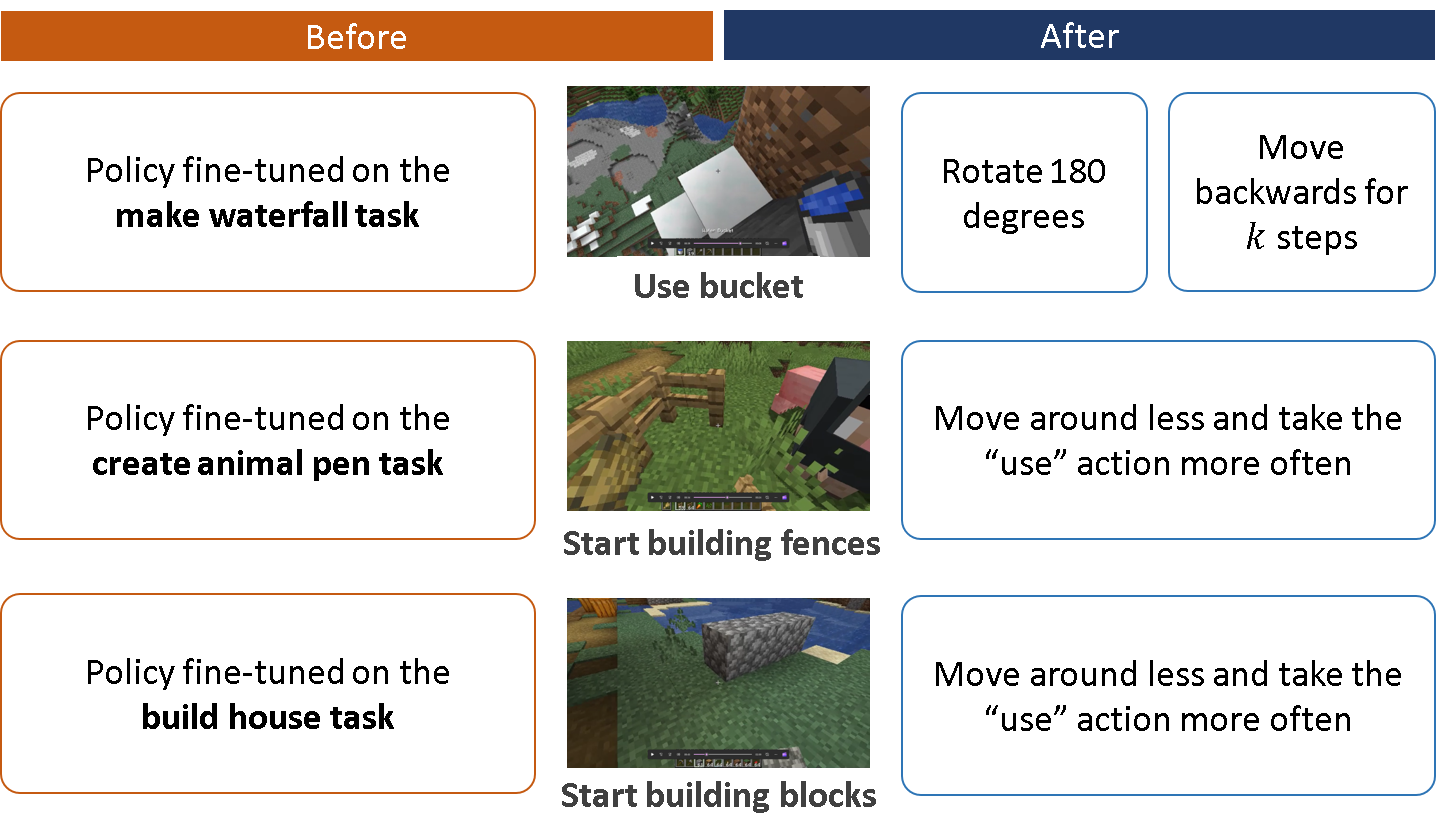}
\caption{Trigger actions defined for the Voggite agent to switch to a different policy during task execution. For the ``find a cave task'', no triggers are defined and the \glsshort{vpt} policy is fine-tuned as normal (with improvements to training techniques as outlined in \Cref{sec:minecraft/voggite_submission}). For ``make a waterfall'' task, the agent changes its behaviour to climbing down a mountain after the bucket has been used. For ``create an animal pen'' and ``build a house'' tasks, the policy distribution is shifted to move around less and commit to building structures after taking the first ``use'' action.}
\label{fig:minecraft/voggite_trigger}
\end{figure}

On the other hand, defining a trigger for each task in the MineRL BASALT Competition is straightforward. It was noted that certain actions that the agent takes (\eg a ``use bucket'' in the waterfall task, as well as a ``use item'' action in the building tasks) is more indicative of a stage change in a task than inspecting changes in the observations. While the observation space is a high-dimensional RGB image, the action space of the \gls{vpt} model is categorical, which is then converted to cursor movement, mouse click and keyboard actions. The ``use'' action in particular is easy to track, and is an ideal trigger for a stage change. \Cref{fig:minecraft/voggite_trigger} summarises these triggers and the shift in the agent's policy for each task. This was implemented in the submission to the MineRL BASALT Competition, which is discussed next.

\section{Submission to the MineRL BASALT Competition}

\subsection{Voggite}
The embodied agent submitted to the competition was named \emph{Voggite} after a mineral, following the tradition set in works related to Minecraft (\eg the \emph{BASALT} competition). This specific mineral was chosen due to its spelt resemblance to VGG (Visual Geometry Group).

Voggite modified and improved upon \gls{vpt}'s training setup in four aspects. The changes were: 
\begin{enumerate}
\item pre-computing \gls{vpt} embeddings such that only a small policy head network has to be trained and fine-tuned via \gls{bc}, making training faster,
\item enabling reshuffling of training samples since \gls{vpt} embeddings are pre-computed and no longer have to be evaluated auto-regressively,
\item training setup in PyTorch Lightning for faster training,
\item reweighting actions according to action frequency so that rare actions are effectively sampled more often.
\end{enumerate}
These changes allowed a much faster and more effective iteration of ideas and methods.

\begin{figure}[t]
\includegraphics[width=\textwidth]
{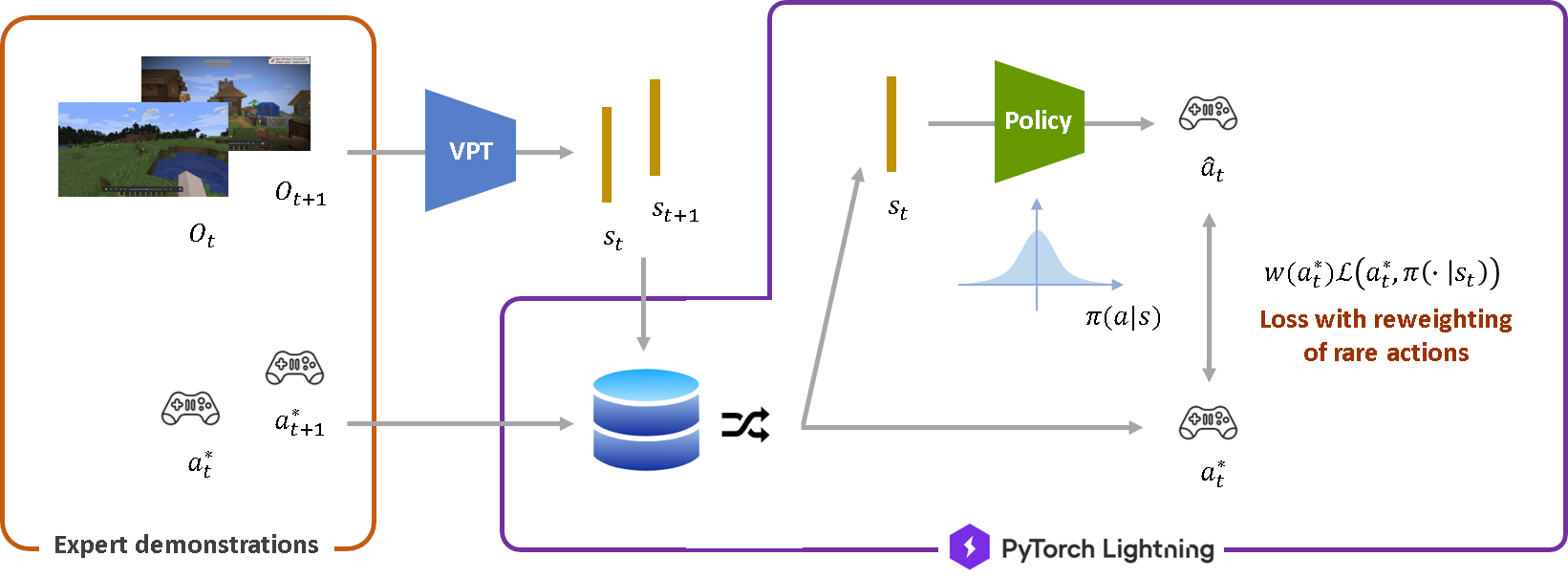}
\caption{Diagram of the Voggite training pipeline. \glsshort{vpt} embeddings are pre-computed for the expert demonstrations and stored as a permutable dataset, removing the sequential constraint of forward-passing through the \glsshort{vpt} transformer. A policy head is trained and fine-tuned on each task in the MineRL BASALT Competition, given the \glsshort{vpt} embeddings and expert actions labels. The categorical losses are reweighted inversely to the frequency of the expert actions' occurrences.}
\label{fig:minecraft/voggite_training}
\end{figure}

Furthermore, the ``use'' action was identified to be a trigger action for the waterfall, animal pen and house-building tasks. Until the trigger action is taken, the agent follows a standard \gls{vpt} policy fine-tuned on individual tasks. Once the ``use'' action is triggered, however, the agent's behaviours are modified to adapt to needs for different tasks.
In the waterfall task, the agent must retreat down the mountain once a bucket is used. The agent is rotated $180\degree$ once the trigger action is taken, and then moved backwards for the next 5 seconds before terminating. For the animal pen and housing tasks, once the ``use'' action is triggered, the agent's move action probability is reduced, while the ``use'' action probability is increased. This is to ensure that the agent commits to building a house once it has identified a suitable location, rather than aimlessly alternating between exploring and building, as with the baseline \gls{vpt}.

\subsection{Implementation}
\label{sec:minecraft/implementation}
The code for the submitted Voggite agent is open-sourced, and can be found at \url{https://github.com/shuishida/minerl_basalt_2022}. 

\subsection{Submission and results}
\label{sec:minecraft/voggite_submission}
The competition team consisted of myself (Shu Ishida) as the main contributor, as well as my supervisor (Dr. João F. Henriques) who provided guidance and contributed research ideas. The code for this research and submission was developed entirely by Shu Ishida.

Voggite was submitted to the MineRL BASALT Competition at NeurIPS 2022~\cite{milani2023solving}. 
Voggite successfully created a waterfall in many of the runs, and improved consistency in building activities over the baseline \gls{vpt}. 
Our solution achieved 3rd place out of 63 teams, 446 individual participants and 504 submissions overall. Results are shown in \Cref{tab:minecraft/leaderboard}.

\begin{table}[t]
\caption{
Leaderboard: normalised TrueSkill scores according to \cite{milani2023solving}. The top three teams were
GoUp, UniTeam, and Voggite (ours). Scores for \glsshort{bc}-Baseline, two expert humans, and a random agent are also included.
}
\centering\footnotesize
\setlength{\tabcolsep}{2pt}
\begin{tabular*}{\columnwidth}{@{\extracolsep{\fill}}lccccc}
\toprule 
\bfseries Team & FindCave & MakeWaterfall & AnimalPen & House & \bfseries Average \tabularnewline
\midrule 
GoUp & 0.31 & \bfseries 1.21 & \bfseries 0.28 & \bfseries 1.11 & \bfseries 0.73 \\
UniTeam & \bfseries 0.56 & -0.10 & 0.02 & 0.04 & \bfseries 0.13 \\
\bfseries Voggite (ours) & 0.21 & 0.43 & -0.20 & -0.18 & \bfseries 0.06 \\
JustATry & -0.31 & -0.02 & -0.15 & -0.14 & \bfseries -0.15 \\
TheRealMiners & 0.07 & -0.03 & -0.28 & -0.38 & \bfseries -0.16 \\
yamato.kataoka & -0.33 & -0.20 & -0.27 & -0.18 & \bfseries -0.25 \\
corianas & -0.05 & -0.26 & -0.45 & -0.24 & \bfseries -0.25 \\
Li and Ivan & -0.15 & -0.72 & -0.14 & -0.22 & \bfseries -0.31 \\
KAIROS & -0.35 & -0.32 & -0.41 & -0.36 & \bfseries -0.36 \\
Miner007 & -0.07 & -0.76 & -0.12 & -0.52 & \bfseries -0.37 \\
KABasalt & -0.57 & -0.23 & -0.41 & -0.31 & \bfseries -0.38 \\
\midrule 
Human2 & 2.52 & 2.42 & 2.46 & 2.34 & \bfseries 2.43 \\
Human1 & 1.94 & 1.94 & 2.52 & 2.28 & \bfseries 2.17 \\
BC-Baseline & -0.43 & -0.23 & -0.19 & -0.42 & \bfseries -0.32 \\
Random & -1.80 & -1.29 & -1.14 & -1.16 & \bfseries -1.35 \\
\bottomrule
\end{tabular*}
\label{tab:minecraft/leaderboard}
\end{table}

\section{Conclusion}
It was found that executing plans in \emph{stages} is essential for the coherent behaviour of an agent in compositional tasks. \emph{Stages} in this context share similarities with options discussed in \Cref{chapter:ppoem}. 
Due to time constraints leading up to the competition, some aspects of the submitted Voggite solution are hard-coded rather than end-to-end learnable, namely the stage switch trigger detection and policy distribution change that follows. However, it served as a proof of concept that a multi-staged policy is necessary to make the agent behaviour more consistent and task-specific. It also retains many of the advantageous properties of \gls{vpt}, such as knowing how to navigate and craft tools from pre-training without any hard-coding. 
In future work (see \Cref{sec:future_work}), we hope to apply a more learning-oriented approach such as \gls{soap} (\Cref{chapter:ppoem}) and TACO~\cite{shiarlis2018taco} to the challenges in Minecraft and other complex tasks for embodied agents.

\chapter{Conclusion}
\label{chapter:conclusion} 

The big picture for this thesis was to solve spatial reasoning and planning tasks in a data-driven manner, while simultaneously making the learning more efficient, interpretable and generalisable. 
For this, \gls{il} and \gls{rl} methods were explored and developed to learn from regularities in spatial tasks, acquire transferable skills, and solve novel problems with zero-shot or few-shot learning. The main contributions of the works covered in this thesis are summarised in \Cref{sec:intro/contributions}. In this final chapter, the initial research objectives outlined in \Cref{sec:intro/objectives} are revisited, and future work is discussed.

\section{Discussion}

\subsection*{Learning a generalisable planner}

This thesis explored two avenues of learning a generalisable planner: learning the underlying \gls{mdp} directly for an end-to-end differentiable planner (\gls{calvin}, \Cref{chapter:calvin}), and iteratively optimising algorithmic plans in the form of code (LangProp, \Cref{chapter:langprop}). 

The approach presented by \gls{calvin} proved robust to complex structures such as large mazes, correctly learning a translationally invariant transition and reward kernels to model the \gls{mdp} and solve navigation tasks in novel environments. A valid value map is computed in an \gls{irl} setting of recovering a reward function from expert demonstrations. The original \gls{vin}~\cite{tamar_value_2016} learnt inaccurate models of the world, inept at handling long-term planning in environments with high branching complexities. In particular, illegal actions leading to collisions were not given large enough penalties. 
This work mitigated this issue by imposing a structural constraint on the value iteration to explicitly model any impossible actions.

LangProp proposed a novel paradigm of learning algorithmic decision-making from data by treating code as learnable policies with the use of \glspl{llm}. 
While there has been exciting work in end-to-end differentiable algorithms~\cite{graves_neural_2014,neelakantan2015neural,petersen2022learning}, algorithms are discrete in nature and are hard to learn in an end-to-end differentiable way. Works using \gls{rl} for algorithm discovery had successes in finding faster algorithms (\eg AlphaTensor~\cite{fawzi2022discovering}, AlphaDev~\cite{mankowitz2023faster}), but at the expense of massive compute and highly tuned task-specific training. LangProp presents an alternative method of tuning algorithms; rather than discovering ``novel'' algorithms, LangProp is best suited to discover known algorithms and tune them to task-specific needs, leveraging the prior knowledge of foundation models.
By making algorithms learnable, high-level and long-term plans that were hitherto too complex for \gls{rl} agents to learn can now be learnt using \gls{il} and \gls{rl} techniques, as demonstrated in experiments with the CARLA benchmark~\cite{carla_leaderboard} for autonomous driving.

Furthermore, \Cref{chapter:ppoem} and \Cref{chapter:minecraft} demonstrate how temporal abstraction using options can help agents make informed long-term decisions. These are further discussed in the following sections on reusable skill learning and memory-augmented policies. 

While the challenge of generalisable planning is still not completely solved, these works indicate a path towards learnable algorithms for data-driven long-term planning. 

\subsection*{Discovering reusable skills}
\label{sec:conclusion/reusable_skills}
\Cref{chapter:ppoem} addressed the problem of automatic discovery of reusable skills. Ways of discovering options as temporally consistent macroscopic actions were considered. While multiple formulations for this problem have been proposed in prior work~\cite{optioncritic,optioncritic_ppo,Zhang2020ProvableHI,online_baum_welch_2021,fox2017multi}. experiments in \Cref{chapter:ppoem} show that option learning using the forward-backward algorithm~\cite{baum72} or a standard formulation of options following the Options Framework~\Cref{sec:background/options_framework} do not effectively learn option assignments that are both causally sound and temporally consistent. 

\gls{soap} presented an alternative formulation by analytically evaluating the policy gradient for an optimal option assignment. While policy gradients cannot be explicitly propagated temporally without keeping all observations in memory, as transformer policies~\cite{vaswani2017attention} do, \gls{soap} achieves an equivalent effect by extending the concept of the \gls{gae} to propagate \emph{option advantages} through time.
Unlike the forward-backward algorithm which optimises the option assignment over a \gls{hmm} of options, observations and actions, assuming a fully known trajectory, \gls{soap}'s formulation is causal, assuming only the knowledge of historical observations and actions. Since the option policy itself must be causal (only conditional on the history of the agent), the causal \gls{soap} formulation is more robust and appropriate to be used as a causal agent's learning objective. 

\subsection*{Solving POMDP environments with memory-augmented policies}

The learnt options could be considered as discretised memory that carries forward historical information of the agent trajectory through time, thereby allowing the agent to disambiguate different states that maps onto the same observations, as is with the case of \gls{pomdp} environments.

By learning temporally consistent and distinct options, the \gls{soap} agent was able to solve \gls{pomdp} corridor environments that require knowledge beyond the currently available observations. While this is a simple experiment setting, \gls{soap} already outperforms the baselines of \gls{ppoc}~\cite{optioncritic_ppo} and \gls{dac} as well as a \gls{lstm} recurrent policy. 
\glspl{lstm} and transformers are expressive tools for sequential modelling, often used to solve \gls{pomdp} environments~\cite{baker2022_openai_vpt,team2023human_ada}.
However, training such recurrent or sequential policies is computationally expensive, since many frames of historical observations have to be forward-passed simultaneously at training time. The window size for historical observations is limited by the size of the \gls{gpu} memory available. In comparison, the approach taken by \gls{soap} is memory-efficient, since only the current observation has to be included in the forward-pass. This makes it a promising candidate as an algorithm with the potential to be extended to more long-term planning problems.

In \Cref{chapter:minecraft}, it was shown that a strategy of options can be extended to solve tasks in more complex \gls{pomdp} environments such as Minecraft. Having options to disambiguate different stages within a task was essential to achieve consistent task execution without forgetting. While the sub-policies are \gls{vpt}-based transformers with a local window of observations and do not have long-term memory, the options function as a task-level discrete memory that keeps track of the high-level plan. 

Improvements in the training such as pre-computing \gls{vpt} embeddings and fine-tuning only the policy head with using PyTorch Lightning contributed to faster and more efficient training.
Voggite achieved competitive performance in the MineRL BASALT Competition~\cite{milani2023solving}, significantly outperforming the \gls{vpt} baseline, which demonstrated the benefits of using options for execution of embodied tasks with long-term planning.

\subsection*{Explaining the behaviour of experts and agents}
\gls{calvin} (\Cref{chapter:calvin}) demonstrated an interpretable approach to learnable planning by modelling transitions and learning reward functions to compute explicit reward and value maps. \gls{soap} (\Cref{chapter:ppoem}) and Voggite (\Cref{chapter:minecraft}) implemented skill segmentation to separate low-level control from task-level plans. Options serve as a useful indicator that distinguishes between different tasks and subtasks. LangProp (\Cref{chapter:langprop}) presented the most interpretable approach of all, representing policies as human-readable code that can also be improved with data-driven learning. Experiments in autonomous driving demonstrated that a known phenomenon called causal confusion when trained with \gls{il} can be reproduced with LangProp. Causal confusion was directly diagnosable from the policy that was learnt, even before running the experiments. This could enable faster iteration on ideas and experiments with future work, and empower the field of data-driven explainable \gls{ai}.

\subsection*{Training embodied agents to perform complex tasks}
This thesis touched upon the problems of robotic navigation (\Cref{chapter:calvin}), autonomous driving (\Cref{chapter:langprop}), simulated control in Atari~\cite{bellemare13arcade} and MuJoCo~\cite{todorov2012mujoco} (\Cref{chapter:ppoem}), and game play in Minecraft (\Cref{chapter:minecraft}). Methods of learning such complex tasks from data were also suggested. Although these are incomplete solutions, significant improvements to baseline methods were demonstrated by applying techniques of differentiable planning, optimising algorithmics policies with \glspl{llm}, hierarchical policies, and option discovery. Learning long-term plans that go beyond reactive policies was a consistent challenge in all the domains. The thesis suggested pathways that may lead to solving this challenge. Details of such pathways are discussed as future work.


\section{Limitations and future work}
\label{sec:conclusion/future_work}

\subsection{CALVIN}
While end-to-end differentiable planning is an attractive concept demonstrated in the \glspl{vin} that this work improves upon, several limitations hinder seamless scaling and deployment to real-world problems.
For instance, \glspl{vin} as well as \glspl{calvin} assume the agent's pose, depth image and the camera parameters to be known. Integration with end-to-end trainable localisation and mapping modules  \cite{zhang_neural_2017,parisotto_neural_2018} may be considered for future work.
A static environment was also assumed where observation embeddings can be aggregated temporally. Extending the method to dynamic environments is non-trivial, since that requires a learnable dynamic mapping system, out of scope for the \gls{lpn} approach. The temporally and policy-dependent dynamics of the environment must be modelled as well for an accurate value propagation, which indicates that the \gls{vin} approach is fundamentally limited to static environments. Other approaches, such as model-based \gls{rl}, in particular those on world models~\cite{ha2018worldmodels,hafner2019dreamer,schrittwieser2020mastering_muzero} may be better suited for modelling such tasks.

A related limitation is that these methods operate in discrete predefined state and action spaces. Extending \gls{vi} to continuous state and action spaces is challenging. Policy gradient and Actor-Critic methods typically solve such problems by learning a policy function that maps continuous states to continuous actions. However, value propagation can no longer be achieved by enumeration, so it is not possible to find and execute an optimal policy zero-shot on novel environments using these methods. Learning a near-optimal policy and an accurate value function with a high zero-shot capability will require a large amount of training data. This makes it challenging for \gls{vin}-like architectures to be deployed beyond a navigation setting, such as robot manipulation and planning in higher dimensional spaces.

In order to preserve the benefits of planning by enumeration while extending this to continuous state and action spaces, \gls{hrl} may be required, where low-level sub-policies that handle continuous control are orchestrated by high-level policies that implement long-term plans. The Options Framework~\cite{sutton1999between_options}, analysed in \Cref{chapter:ppoem}. is one method of achieving this. It may be possible to extend techniques suggested in \gls{calvin} to these settings.

Both \gls{vin} and \gls{calvin} are trained only in an \gls{il} setting, where the optimal action is explicitly labelled by an expert. While this is helpful in increasing sample efficiency, it limits the scope of application since the optimal action is not always known in many real-world problems. Furthermore, the agent loses the opportunity to learn from its failures. To bridge this gap, an improved form of \gls{calvin} was considered, which allowed training both in \gls{il} and \gls{rl} setups. While the results were not included in this work, preliminary results showed that \gls{calvin} could be extended easily to training in \gls{rl} settings due to its enhanced interpretability compared to \gls{vin}.

Finally, although neural networks hold promise for making navigation more robust under uncertainty, their uninterpretable failure modes mean that they are not yet mature enough for safety-critical applications, and more research to close this gap is still needed. \gls{calvin} is a more interpretable \gls{mdp} structure for \glspl{vin}, trained solely on safe offline demonstrations.
However, their reliability is far from guaranteed, and complementary safety systems in hardware must be considered in any deployment.
In addition to further improving robustness to failures, future work may investigate more complex tasks involving real-world deployment, and study the effect of sensor drift on navigation performance.

\subsection{SOAP}
\gls{soap} demonstrated capabilities of learning options in a \gls{pomdp} environment of corridors, and showed equivalent performances to the baseline \gls{ppo} agent in other environments. However, even in simple settings, it took the agent many samples before a correct option assignment was learnt. Option discovery is a difficult chicken and an egg problem, since options need to be assigned correctly in order for the rewards to be obtained and passed onto the options, but without the rewards a correct option assignment may not be learnt. Furthermore, learning to segment episodes into options in an unsupervised way without any pre-training is an ill-defined problem, since there could be many equally valid solutions. Combining the learning objective of \gls{soap} with methods such as curriculum learning to pre-train diverse sub-policies specialised to different tasks may stabilise training.

In the current formulation of \gls{soap}, the options are discrete variables, and are less expressive compared to latent variables in recurrent policies and transformers. This greatly reduces the memory capacity and could hinder learning in \gls{pomdp} environments. Further research in extending the derivations of \gls{soap} to work with continuous or multi-discrete variables as latents would be desirable, and may lead to making the method scalable to more complex problems.

\subsection{LangProp}
\label{sec:future_work}
While LangProp successfully harnessed the capability of \glspl{llm} to apply data-driven optimisation techniques to code optimisation, LangProp may not be the most appropriate solution for all problems - in fact, neural networks excel in working with continuous state-action spaces and low-level control, whereas \glspl{llm} are better at handling high-level planning and reasoning tasks. LangProp intends to propose an alternative learning paradigm that allows \glspl{llm} to be used to learn high-level planning which has hitherto been a difficult problem for other machine learning approaches (e.g. neural networks). 

There are numerous future research directions that could improve the capability of LangProp as a training framework, as well as give a better theoretical foundation, such as (a) chaining of modules with a full back-propagation algorithm, (b) improvements to the evolutionary algorithm (e.g. priority mechanism), (c) a robust sampling mechanism for failed examples upon updates, (d) incorporating human feedback in natural language during policy updates, and (e) using LangProp with \glspl{llm} fine-tuned for code correction and optimisation tasks. In particular, scaling this approach to larger repositories and complex systems would require a multi-modular approach that can propagate useful learning signals to subcomponents if there are multiple failure points in the system. 

Applying LangProp to \gls{rl} tasks has open questions in credit assignment and value estimation. LangProp demonstrated improvements in \gls{rl} policies written as code when (a) the policy can be optimised on episodic returns with a Monte-Carlo method (e.g. CartPole), or (b) there is immediate feedback from the environment (e.g. infractions in CARLA). However, for complex tasks that have delayed rewards, it is necessary to have an accurate value/advantage estimator for credit assignment. Since replacing a neural value estimator with a code-based function is not feasible, it is most likely that a hybrid method (having an interpretable code-based actor policy trained with LangProp that uses a value function estimated by a neural network as a critic) would be a way to apply LangProp to complex \gls{rl} scenarios. However, this is also an open-ended question, which calls for further exploration. 

Having an \gls{llm} in the \gls{rl} optimisation means that more useful signals could be potentially harvested from the environment, rather than relying just on sparse scalar rewards for updates. For instance, having descriptive feedback from the Gymnasium environment on the failure modes of the agent, given either as a warning or natural language feedback, can significantly accelerate the learning of the \gls{rl} agent. This also allows a more seamless integration of human-in-the-loop feedback.

Finally, more investigation is required in terms of the robustness and safety of \gls{llm}-written applications. This is applicable to all systems that involve code generation. While this framework iteratively improves the quality of the code and filters out potential errors that make the final code policy less likely to contain errors, additional safety mechanisms and firewalls are necessary during the training process, since the code is evaluated based on execution, which could potentially be a source of attacks or risk. It is worth emphasising the importance of additional safety precautions before deployment.

LangProp opens up new possibilities for data-driven code development. While zero-shot applications of \glspl{llm} have enabled tools such as GitHub Copilot, some suggestions are inaccurate or misaligned with the user’s intentions, whereas if there are data or unit tests that the code needs to satisfy, the code suggestions can be made much more accurate by first running evaluations on these test suites and choosing the best possible suggestion that satisfies the requirements. Planning is one aspect of autonomous driving that has not yet successfully adopted a data-driven approach, for good reasons, since neural networks often struggle to produce generalisable high-level planning rules and are less interpretable. Therefore, most methods currently in deployment have human-engineered planning algorithms. The LangProp framework is insufficient to replace such systems since it lacks the robustness that human-designed systems have to offer, and more research needs to be done in this direction. This work is hoped to provide inspiration for future research to make the framework more robust and safely deployable in the real world.

\subsection{Voggite}
The agent Voggite developed in this work is a prototype example of applying options to task-solving in Minecraft. The work was limited in that there were only two options per task, and the option switching trigger and distribution shifts in sub-policies were manually hard-coded. Further work is required to (a) allow multiple options, (b) make sub-policies fully learnable (in Voggite the sub-policies are manually manipulated from a fine-tuned \gls{vpt} policy), (c) make triggers observation-dependent as well as action-dependent, and (d) learn transitions between these multiple options. (a) and (b) are relatively straightforward implementations, since the option segmentation themselves could still be defined task-by-task by a human expert by defining the triggers. Implementing (c) and (d) that involve learning the triggers themselves are trickier, and may require integration of techniques that were suggested in \Cref{chapter:ppoem}, as well as other pre-training and regularisation techniques to constrain the learning problem. For instance, curriculum learning may be employed to first learn and fine-tune low-level sub-policies, and which are then used as initialisations to learn a task-specific policy that orchestrates these sub-policies and their fine-tuning. 

Prior work on learned task segmentation, such as TACO~\cite{shiarlis2018taco} is highly relevant. TACO addressed and solved similar crafting and manipulation tasks which comprise multiple stages, by formulating the segmentation problem as a sequence alignment problem between demonstration trajectories and an acyclic sequence of task descriptors called task sketch. A sequence alignment strategy that extends Connectionist Temporal Classification~\cite{graves2006connectionist} is employed. While the work in \Cref{chapter:minecraft} had a hard deadline for the NeurIPS BASALT Competition~\cite{milani2023solving}, which constrained the scope of the work, comparing methods such as TACO against other temporal abstraction strategies (e.g. \gls{ppoem} or \gls{soap}) in the Minecraft setting would be informative.

Related to \Cref{chapter:ppoem}, making options represent more information than just the stage of a task execution is another challenge of extending Voggite beyond sequential tasks. For instance, if options can be multi-categorical, it could be useful for planning out long-term strategies on this discretised option-space. Information such as which objects are in the inventory are innately multi-categorical, and planning to gather or craft a specific item in the technology tree of the Minecraft universe requires the agent to be able to plan on this technology tree, which are either learnt directly from experiences and trial-and-error, or could be obtained from natural language inputs or reading the documentation.

Planning in natural language space using multi-modal foundation models~\cite{wang2023describe,zhu2023ghost,wang2023jarvis}, as well as interfacing with natural language human inputs~\cite{lifshitz2024steve} or directly generating executable code~\cite{wang2023voyager} are some promising research directions for learning a high-level plan. 
Since Voggite was developed prior to these methods being released, the avenue of using \glspl{llm} was not explored in this work. However, being able to take advantage of the priors of the world is crucial for solving complex tasks in Minecraft without an exponential amount of trial-and-error. \glspl{llm} are also a natural way for the agent to interact with humans and achieve zero-shot task execution.

\appendix

\setcounter{listing}{0}
\renewcommand{\thelisting}{A.\arabic{listing}}
\setcounter{figure}{0} 
\renewcommand{\thefigure}{A.\arabic{figure}} 
\setcounter{table}{0} 
\renewcommand{\thetable}{A.\arabic{table}} 
\renewcommand{\chaptername}{Appendix}

\chapter{CALVIN embodied navigation with comparisons}
For a visual comparison of \gls{calvin}, \gls{vin} and \gls{gppn}, rollouts are performed on a randomly generated maze using each of the algorithms, assuming partial observability and embodied navigation.

\section{Rollout of CALVIN}

\begin{figure}[t]
  \centering
  \includegraphics[width=0.22\columnwidth]{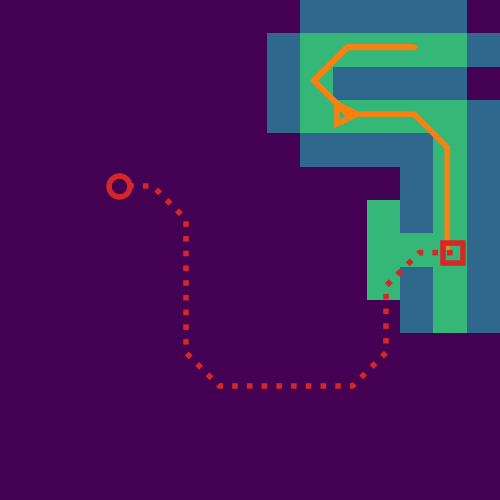}
  \includegraphics[width=0.22\columnwidth]{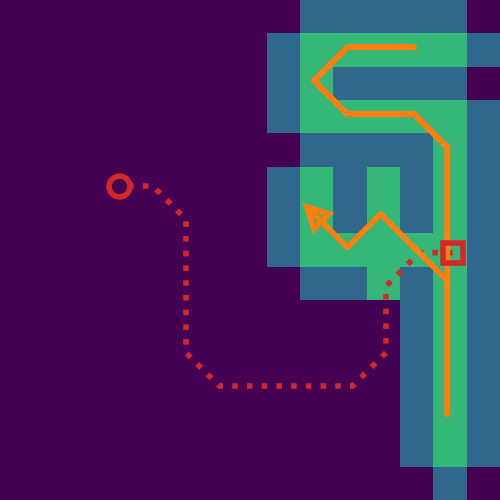}
  \includegraphics[width=0.22\columnwidth]{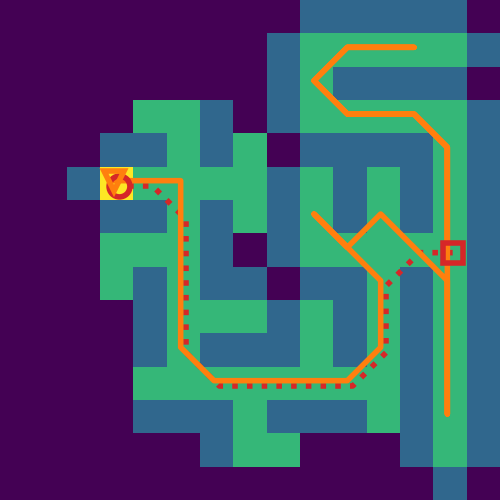}\\
  \includegraphics[width=0.22\columnwidth]{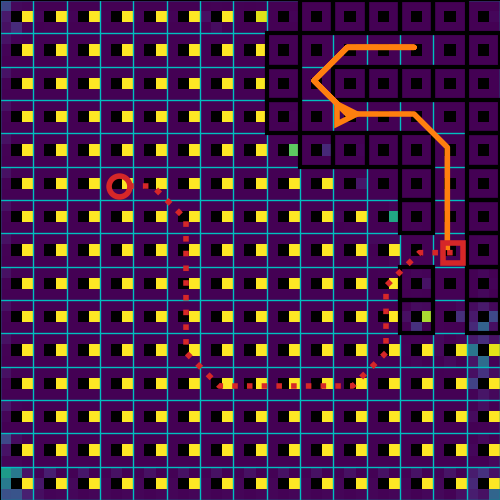}
  \includegraphics[width=0.22\columnwidth]{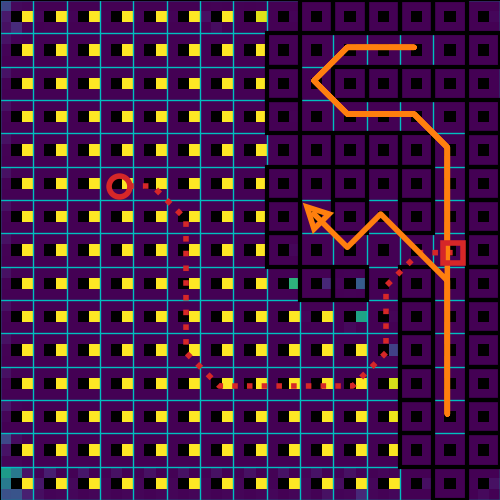}
  \includegraphics[width=0.22\columnwidth]{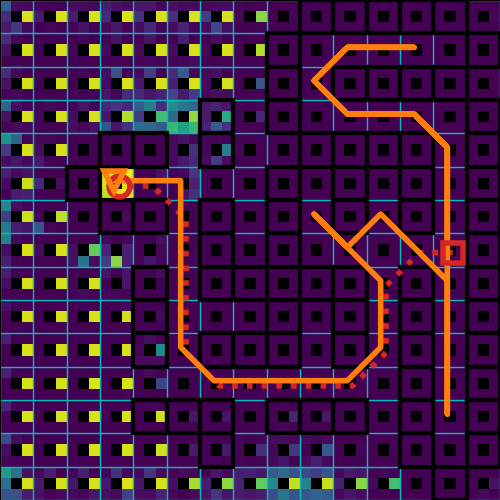}\\
  \includegraphics[width=0.22\columnwidth]{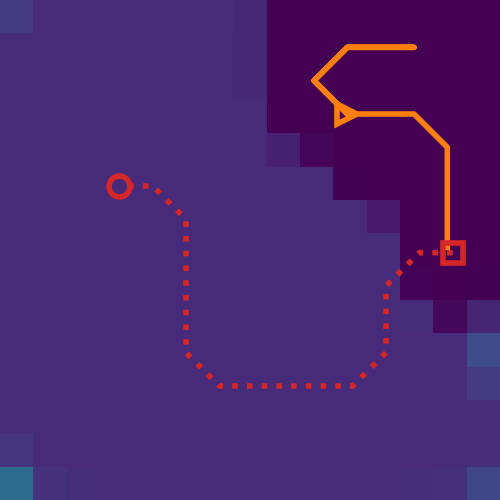}
  \includegraphics[width=0.22\columnwidth]{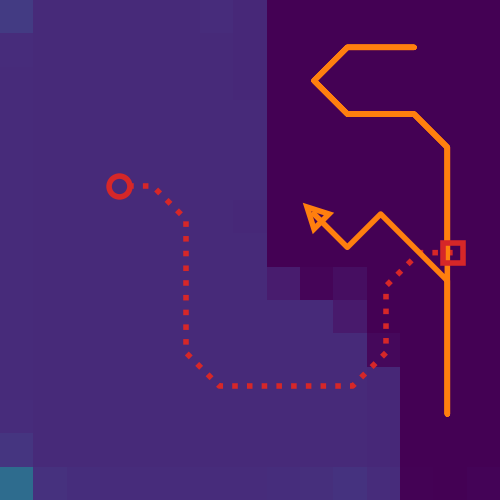}
  \includegraphics[width=0.22\columnwidth]{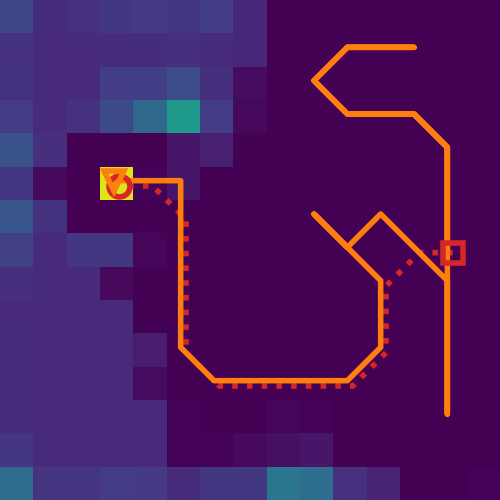}\\
  \includegraphics[width=0.22\columnwidth]{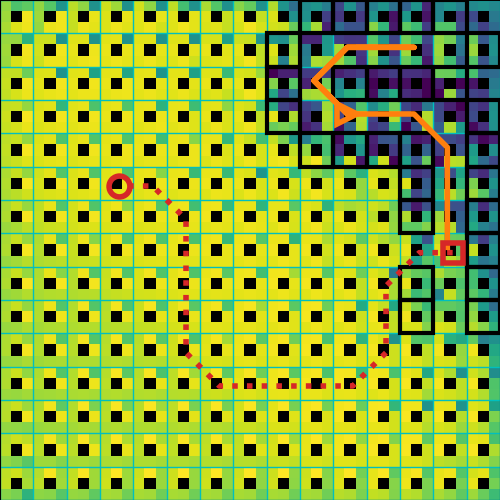}
  \includegraphics[width=0.22\columnwidth]{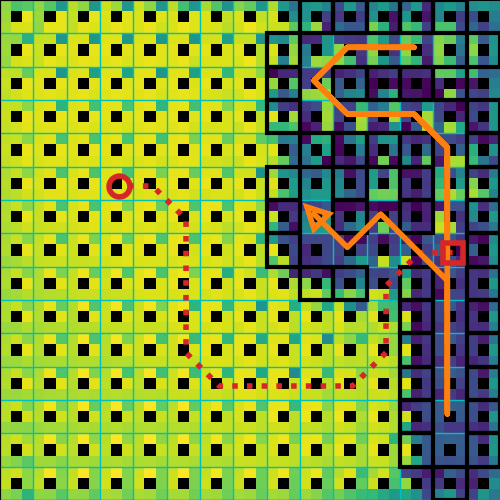}
  \includegraphics[width=0.22\columnwidth]{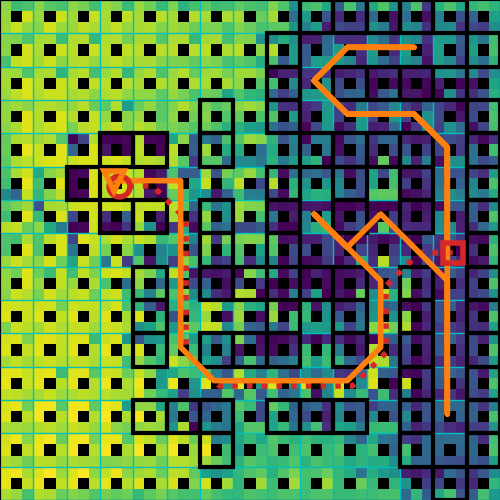}\\
  \includegraphics[width=0.22\columnwidth]{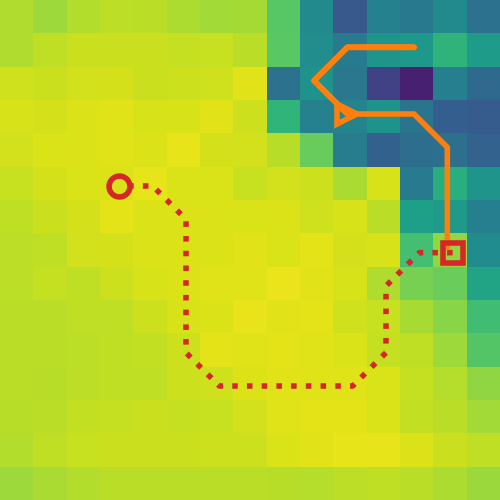}
  \includegraphics[width=0.22\columnwidth]{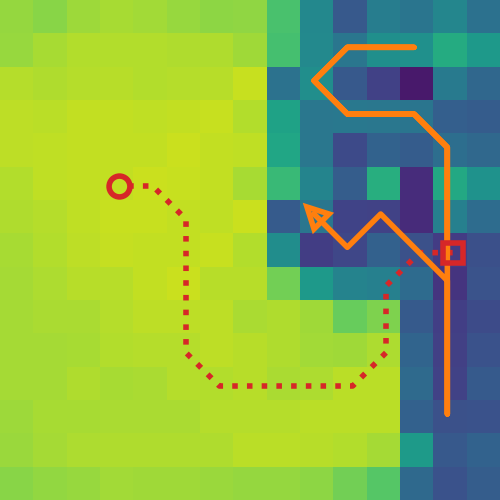}
  \includegraphics[width=0.22\columnwidth]{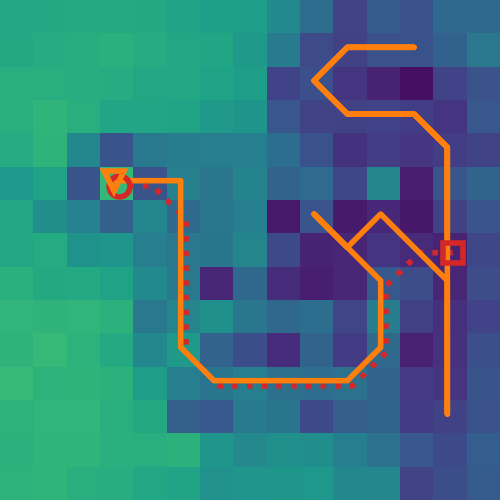}
  \caption{
    Example rollout of embodied \glsshort{calvin} after $30$ steps (left column), $60$ steps (middle column) and $90$ steps (right column). \glsshort{calvin} successfully terminated at $91$ steps.
    \textbf{(first row)} Input visualisation: unexplored cells are dark, and the discovered target is yellow. The correct trajectory is dashed, and the current one is solid. The orange triangle shows the position and the orientation of the agent.
    \textbf{(second row)} Predicted rewards (higher values are brighter). The 3D state-space (position/orientation) is shown, with rewards for the 8 orientations in a radial pattern within each cell (position). Explored cells have low rewards, while unexplored cells and the discovered target are assigned high rewards. 
    \textbf{(third row)} Predicted rewards averaged over the 8 orientations.
    \textbf{(fourth row)} Predicted values following the same convention. Values are higher facing the direction of unexplored cells and the target (if discovered). 
    \textbf{(fifth row)} Predicted values averaged over the 8 orientations.
  }
  \label{fig:calvin/calvin_rollout}
\end{figure}

\Cref{fig:calvin/calvin_rollout} shows an example of a trajectory taken by \gls{calvin} at runtime, with corresponding observation maps, predicted values and predicted rewards for taking the ``done'' action. Similarly to \Cref{sec:calvin/experiments/partially_observable}, the agent manages to explore unvisited cells and backtrack upon a dead end until the target is discovered.
One key difference is that now the agent learns to predict rewards and values for every discretised orientation as well as the discretised location. Upon closer inspection, it could be observed that the predicted values are higher facing the direction of unexplored cells and towards the discovered target. Since rotation is a relatively low-cost operation, in this training example, the network seems to have learnt to assign high rewards to a particular orientation at unexplored cells, from which high values propagate. Rewards and values averaged over orientations yield a more intuitive visualisation.

\section{Rollout of VIN}
\label{sec:calvin/vin_rollout}

\begin{figure}
  \centering
  \includegraphics[width=0.32\columnwidth]{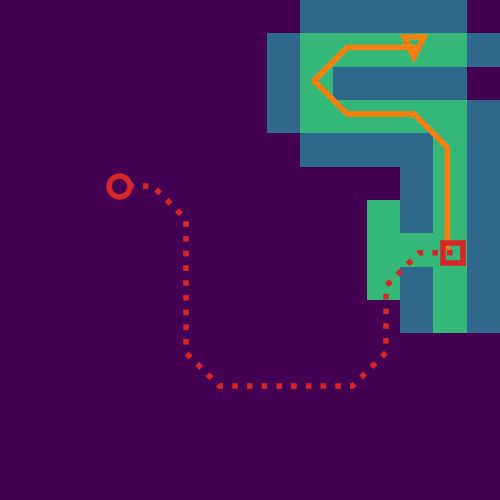}
  \includegraphics[width=0.32\columnwidth]{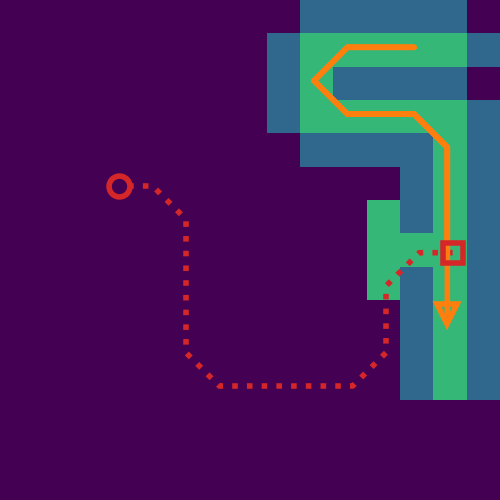}
  \includegraphics[width=0.32\columnwidth]{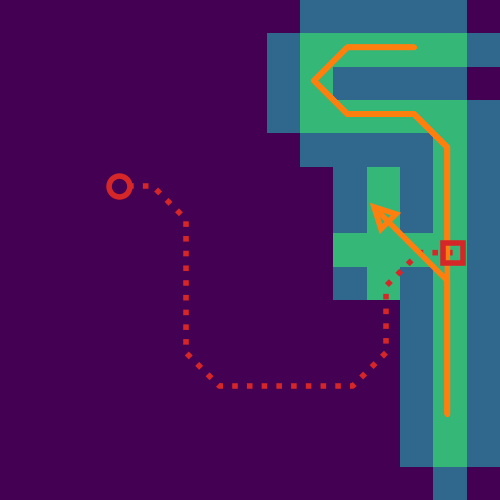}\\
  \includegraphics[width=0.32\columnwidth]{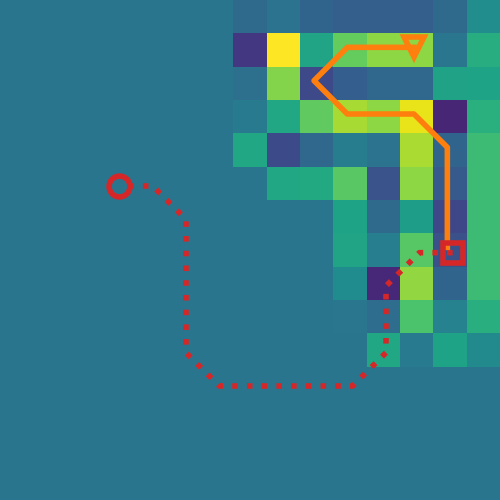}
  \includegraphics[width=0.32\columnwidth]{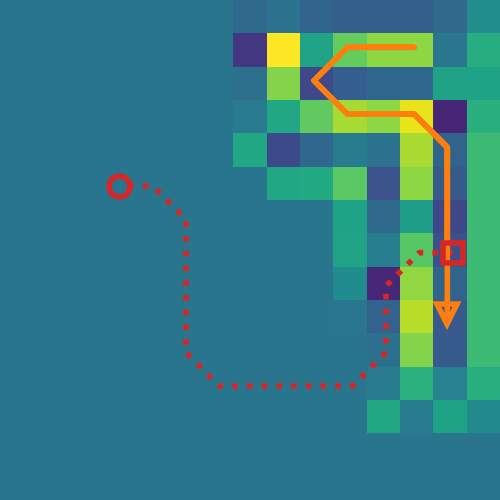}
  \includegraphics[width=0.32\columnwidth]{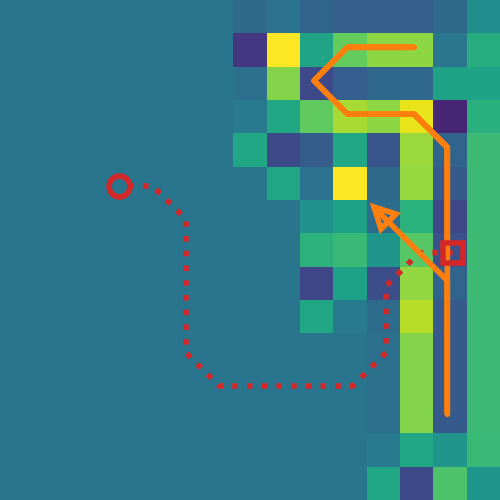}\\
  \includegraphics[width=0.32\columnwidth]{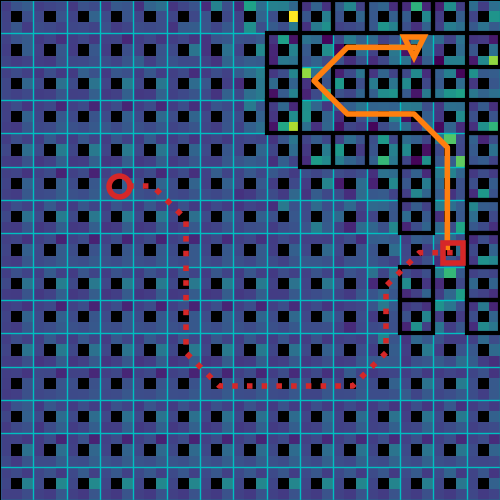}
  \includegraphics[width=0.32\columnwidth]{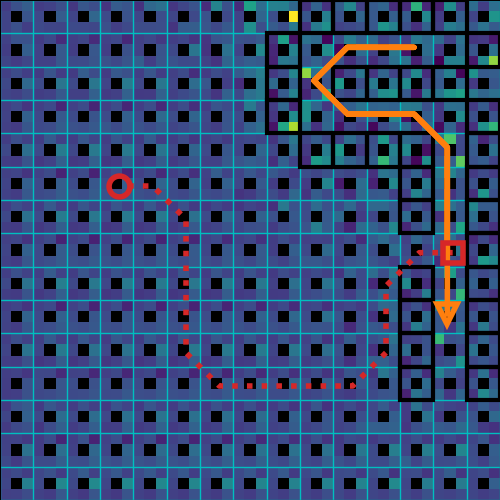}
  \includegraphics[width=0.32\columnwidth]{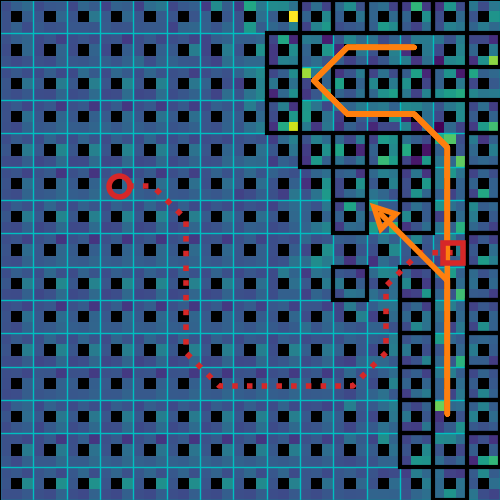}\\
  \includegraphics[width=0.32\columnwidth]{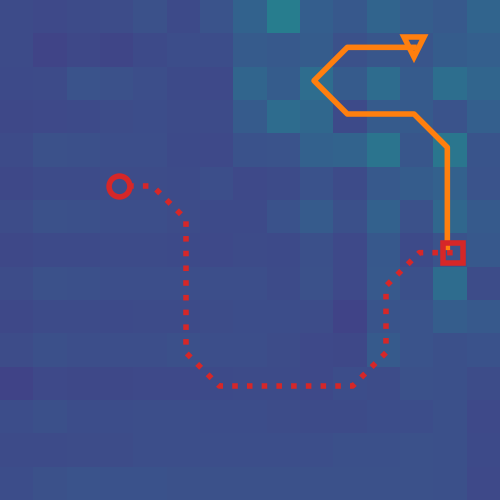}
  \includegraphics[width=0.32\columnwidth]{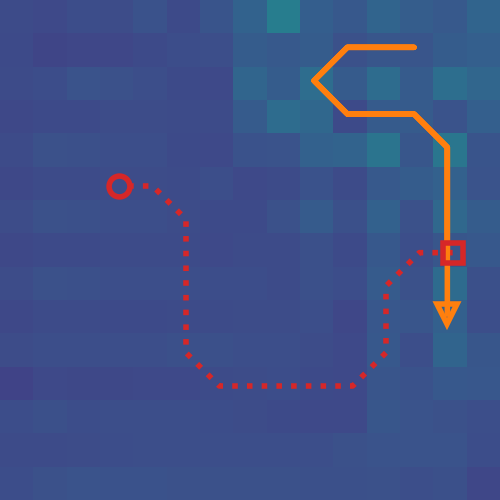}
  \includegraphics[width=0.32\columnwidth]{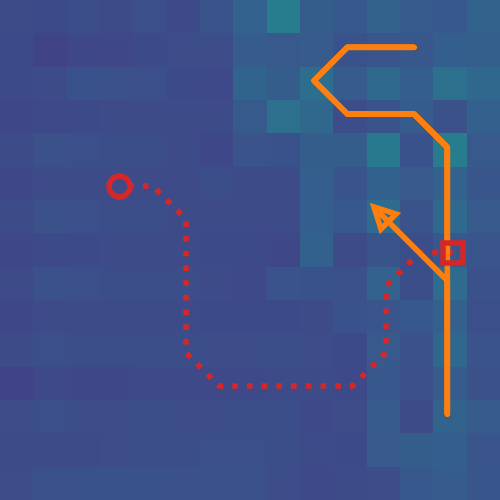}
  \caption{
    Example rollout of embodied \glsshort{vin} after $20$ steps (left column), $40$ steps (middle column) and $60$ steps (right column). \glsshort{vin} kept oscillating between the same two states after $57$ steps. The convention is the same as for \Cref{fig:calvin/calvin_rollout}, except that a single reward map is shared across all orientations.
    \textbf{(first row)} Input visualisation.
    \textbf{(second row)} Predicted rewards.
    \textbf{(third row)} Predicted rewards averaged over the 8 orientations.
    \textbf{(fourth row)} Predicted values. 
  }
  \label{fig:calvin/vin_emb_rollout}
\end{figure}

A corresponding visualisation for \gls{vin} is shown in \Cref{fig:calvin/vin_emb_rollout}. Unlike \gls{calvin}'s implementation of rewards (eq. 5 in sec. 4.1) as a function of discretised states and actions, the ``reward map'' produced by the \gls{vin} does not offer a direct interpretation, as it is shared across all actions as implemented by Tamar \etal \cite{tamar_value_2016}, and is also shared across all orientations in the case of embodied navigation.
The values are also not well learnt, with some of the higher values appearing in obstacle cells. The unexplored cells are not assigned sufficiently high values to incentivise exploration by the agent. In this example, the agent gets stuck and starts oscillating between two orientations after 57 steps.

\section{Rollout of GPPN}

\begin{figure}[t]
  \centering
  \includegraphics[width=0.33\columnwidth]{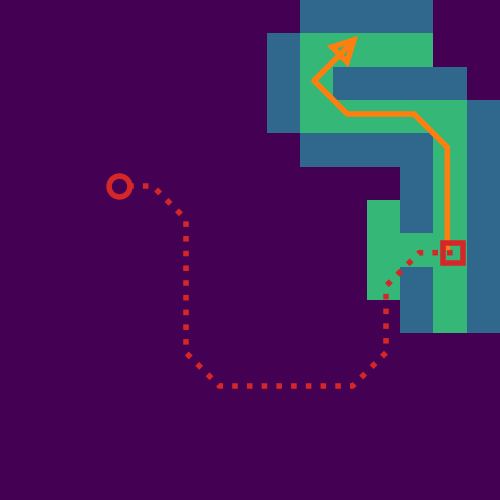}\hfill
  \includegraphics[width=0.33\columnwidth]{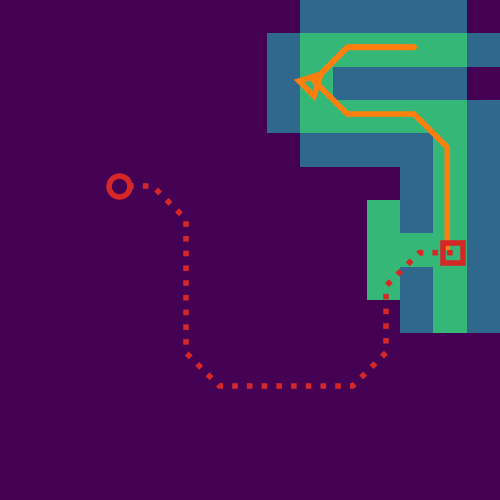}\hfill
  \includegraphics[width=0.33\columnwidth]{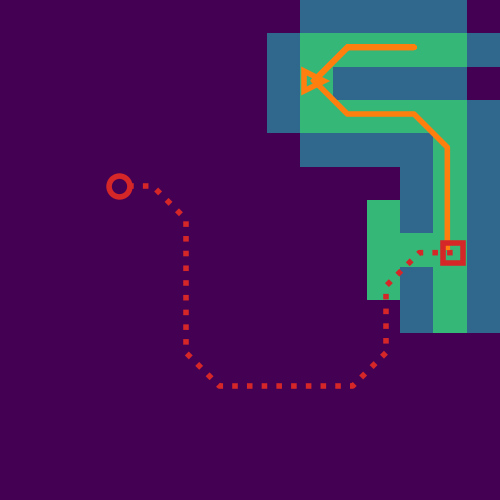}
  \includegraphics[width=0.33\columnwidth]{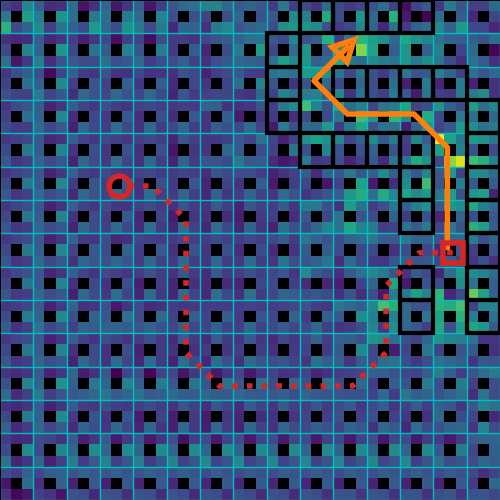}\hfill
  \includegraphics[width=0.33\columnwidth]{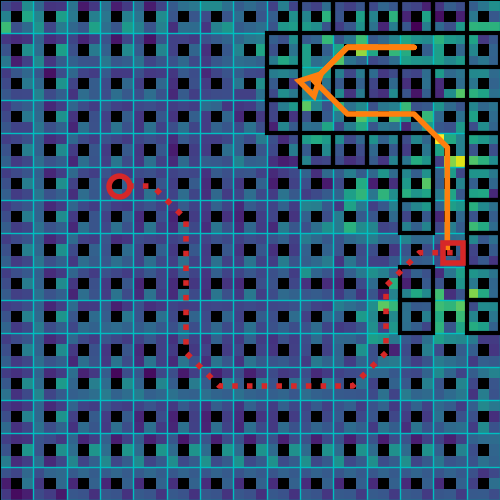}\hfill
  \includegraphics[width=0.33\columnwidth]{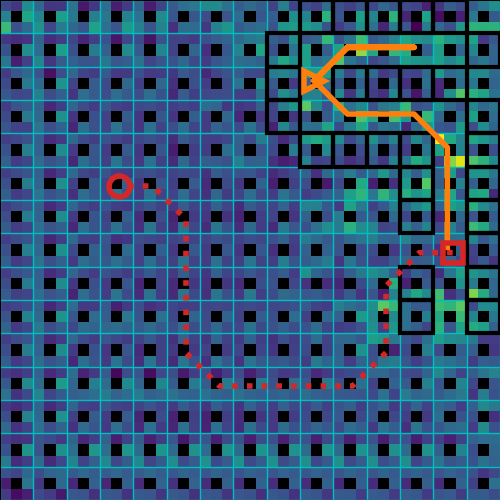}
  \includegraphics[width=0.33\columnwidth]{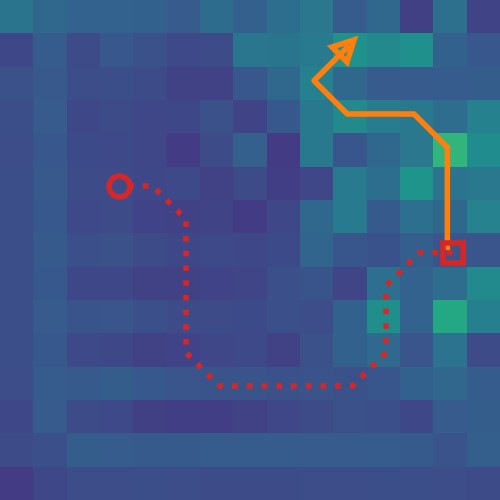}\hfill
  \includegraphics[width=0.33\columnwidth]{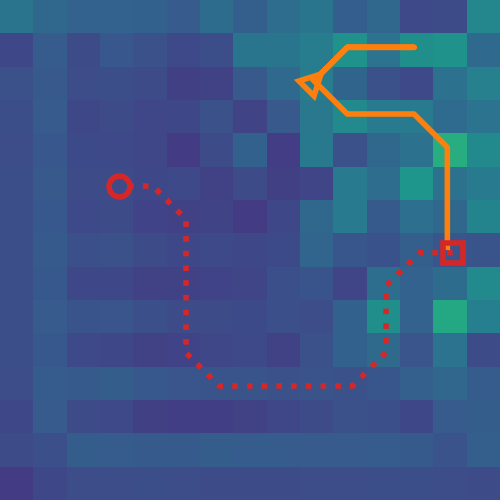}\hfill
  \includegraphics[width=0.33\columnwidth]{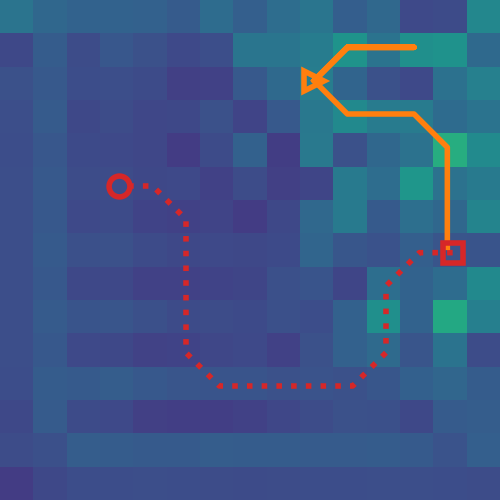}
  \caption{
    Example rollout of embodied \glsshort{gppn} after $15$ steps (left column), $30$ steps (middle column) and $45$ steps (right column). \glsshort{gppn} revisits the same sequences of states leading to a dead end after $45$ steps. The convention is the same as for \Cref{fig:calvin/calvin_rollout}.
    \textbf{(first row)} Input visualisation.
    \textbf{(second row)} Predicted rewards.
    \textbf{(third row)} Predicted rewards averaged over the 8 orientations.
  }
  \label{fig:calvin/gppn_emb_rollout}
\end{figure}

Finally, a visualisation for \gls{gppn} is shown in \Cref{fig:calvin/gppn_emb_rollout}. Unlike \gls{vin} and \gls{calvin}, \gls{gppn} does not have an explicit reward map predictor, but performs value propagation using an \gls{lstm} before outputting a final Q-value prediction. Similarly to \Cref{sec:calvin/vin_rollout}, the values predicted are not very interpretable, and do not incentivise exploration or avoidance of dead ends. In this example, the agent keeps revisiting a dead end that has already been explored in the first 20 steps.

\setcounter{listing}{0}
\renewcommand{\thelisting}{B.\arabic{listing}}
\setcounter{figure}{0} 
\renewcommand{\thefigure}{B.\arabic{figure}} 
\setcounter{table}{0} 
\renewcommand{\thetable}{B.\arabic{table}} 
\renewcommand{\chaptername}{Appendix}

\chapter{Benchmarking option-based Reinforcement Learning agent}
\label{appendix:ppoem}

\begin{table}[ht]
\small
\centering
\caption{The returns of \gls{rl} agents after the maximum environment training steps ($100k$ for CartPole, $1M$ for LunarLander and MuJoCo, and $10M$ for Atari).}
\label{tab:ppoem/comparison}
\begin{tabular}{lrrrrrrr}
\toprule
Environment         & PPO  & PPOC & PPO-LSTM & DAC  & PPOEM & SOAP \\ 
\midrule
Corridor $L=3$         & -0.04         & 0.76          & 0.99              & 0.90          & 0.99                  & 0.99                 \\ 
Corridor $L=10$        & 0.04          & 0.10          & 0.37              & 0.60          & 0.66                  & 0.93                 \\ 
Corridor $L=20$        & 0.00          & 0.13          & 0.00              & 0.34          & 0.13                  & 0.78                 \\ 
CartPole             & 499.65        & 401.87        & 492.75            & 500.00        & 491.30                & 500.00               \\ 
LunarLander          & 195.03        & 141.07        & 257.96            & 159.00        & 251.71                & 254.14               \\ 
Asteroids            & 1806.14       & 1760.53       & 1463.03           & 2108.87       & 1985.70               & 1910.53              \\ 
Beam Rider            & 3704.50       & 689.07        & 1574.99           & 2839.36       & 2691.09               & 3955.31              \\ 
Breakout             & 376.19        & 4.64          & 257.57            & 355.97        & 51.02                 & 345.68               \\ 
Enduro                & 704.17        & 0.13          & 655.96            & 674.03        & 724.97                & 593.44               \\ 
Ms Pacman               & 1825.40       & 434.17        & 1555.80           & 1806.40       & 1453.87               & 2046.87              \\ 
Pong                 & 20.74         & -0.99         & 20.70             & 20.65         & 18.32                 & 20.81                \\ 
Qbert                & 12669.59      & 115.17        & 12282.50          & 8212.08       & 8940.83               & 11469.25             \\ 
Road Runner           & 36247.13      & 5273.33       & 33507.00          & 32042.00      & 18886.00              & 29490.33             \\ 
Seaquest             & 1539.73       & 317.80        & 1501.13           & 1198.33       & 1723.50               & 951.67               \\ 
Space Invaders        & 845.73        & 378.90        & 756.27            & 840.63        & 844.00                & 1013.40              \\ 
Ant                  & 2258.28       & 53.10         & 987.01            & 1394.61       & 49.30                 & 1943.45              \\ 
Half Cheetah          & 5398.50       & 4300.07       & 4478.37           & 4251.44       & 56.07                 & 4962.48              \\ 
Humanoid             & 1196.32       & 1169.26       & 1169.12           & 1212.80       & 275.03                & 1100.38              \\ 
Reacher              & -4.87         & -4.94         & -5.04             & -4.60         & -5.54                 & -4.87                \\ 
Swimmer              & 340.31        & 119.46        & 342.84            & 354.67        & 337.85                & 319.45               \\ 
Walker               & 2960.18       & 1631.97       & 1071.81           & 1799.50       & 1517.90               & 3478.50              \\ 
\bottomrule
\end{tabular}
\end{table}

\setcounter{listing}{0}
\renewcommand{\thelisting}{C.\arabic{listing}}
\setcounter{figure}{0} 
\renewcommand{\thefigure}{C.\arabic{figure}} 
\setcounter{table}{0} 
\renewcommand{\thetable}{C.\arabic{table}} 
\renewcommand{\chaptername}{Appendix}

\chapter{LangProp code generation and prompt examples}

\section{Notes on training with LangProp}

\subsection{Choosing the priority discount factor}
How the priorities of the policies are calculated has a large effect on the final performance of the trained LangProp model. For a stationary training distribution (e.g. supervised learning on a fixed offline dataset), whether one uses the immediate average score, a running average, or an exponential average does not make a difference except that just using the immediate average score results in a more stochastic result due to fewer numbers of samples. If the computational resources and time are not constrained, one could increase the batch size and just use the immediate average score. If these are constrained, one may adopt a running average with smaller batch sizes. This works when the training distribution is stationary and there are no other changing components other than the policy currently training.

If the training distribution changes or the policy consists of multiple chained modules, each with a learnable sub-policy, it is no longer possible to use a simple running average, and either an objective score of a single large batch or an exponential averaging scheme must be used. The current implementation of LangProp does not support multiple chained modules, but is a foreseeable and natural extension to the framework. Changes in the training distribution are expected in DAgger or \gls{rl}. For training the LangProp agent in \Cref{sec:langprop/agent_training}, a discount factor of $\gamma=0$ is used, effectively only using the immediate average scores evaluated on a freshly sampled batch. This is because forward passes through the LangProp driving policies are fast due to not having any complex components so it is possible to use a large batch size. However, in applications where forward passes are expensive and the batch size must be small, using exponential averaging with a non-zero discount factor $\gamma$ is recommended. 

\subsection{Specifying the policy}
One of the challenges in the early stages of the project was in specifying the inputs and outputs of the function. Most of the failures in learning a policy were due to misspecification of the inputs, rather than a fundamental problem with the \gls{llm} or with LangProp. For instance, it was crucial to specify the units of the input values, e.g. $m/s$, which allowed the \gls{llm} to choose sensible values for some internal parameters. It was also important to name input variables explicitly such that it is clear whether the coordinates are given as absolute world coordinates or coordinates relative to the ego vehicle.
A useful property of LangProp is that because the \gls{llm} has some understanding of the world from natural language, it can easily incorporate this knowledge when generating the code, constraining the search space of feasible code. We can further guide the \gls{llm} to generate policies with certain characteristics, e.g. having a larger safety margin, by expressing preferences in the prompts. This adds to the benefits of the LangProp approach, where it is easier to encourage policies to exhibit certain behaviours.

\section{LangProp prompt definitions}
\label{sec:langprop/model_and_prompts}
LangProp as a framework can be used to optimize a diverse range of code optimization problems. The functionality of the model is determined by the choices in the setup prompt, the update prompt, and the dataset that the LangProp model is trained on. 

\subsection{Policy setup prompt examples}
\label{sec:langprop/setup_prompt_example}
We provide simple examples of learning a Sudoku algorithm, and learning a policy that plays CartPole-v1, a widely used reinforcement learning environment in \cite{brockman2016openaigym}, to show the generality of the framework. The setup prompt should include the specification of the function's inputs and outputs and their types in the form of a docstring.

\prompt{code/langprop/sudoku_setup.py}{code:langprop/sudoku_setup}{Setup prompt template to learn an algorithm to solve generalized Sudoku}

\prompt{code/langprop/cartpole_setup.py}{code:langprop/cartpole_setup}{Setup prompt template to learn an agent policy to play CartPole}

\subsection{Policy update prompt example}
\label{sec:langprop/update_prompt_example}
The prompt used to update the policy contains the same information as the setup prompt, but in addition, has example inputs and outputs where the code had failed to produce a valid prediction. If there was an exception or printed messages during the execution of the code, this will also be provided as feedback. The LLM is asked to identify the source of the sub-optimal performance and rewrite the code to achieve a higher score.

\prompt{code/langprop/sudoku_update.py}{code:langprop/sudoku_update}{Update prompt template to learn an algorithm to solve generalized Sudoku}

\prompt{code/langprop/cartpole_update.py}{code:langprop/cartpole_update}{Update prompt template to learn an agent policy to play CartPole}

\subsection{Policy definition for the LangProp driving agent in CARLA}
\label{sec:langprop/langprop_drive_setup}
The driving policy is given the location, orientation, speed, length, and width of the ego vehicle, other vehicles and pedestrians in the scene, the distances to the next red traffic light and stop sign, and the target waypoint ($4~m$ ahead, used by other baseline experts), all in absolute world coordinates.

\python{code/langprop/langprop_drive_setup.py}{code:langprop/langprop_drive_setup}{Docstring given as part of the setup prompt for the LangProp agent}

\section{Code generation examples}
\label{sec:langprop/code_generation_examples}

\subsection{Solutions for Sudoku}
\label{sec:langprop/sudoku_solutions}
\subsubsection{Incorrect solution generated zero-shot}

\python{code/langprop/sudoku_incorrect.py}{code:langprop/sudoku_incorrect}{Example code to solve Sudoku generated zero-shot before LangProp optimization. The code is instructed to solve a general Sudoku with subgrids of size $H \times W$, but confuses it with the standard $3 \times 3$ Sudoku.}

\subsubsection{Correct solution after applying LangProp}

\python{code/langprop/sudoku_correct.py}{code:langprop/sudoku_correct}{Example code to solve Sudoku after LangProp optimization. The code outputs a correct solution.}

\subsection{Solutions for CartPole}
\label{sec:langprop/cartpole_solutions}
\subsubsection{Incorrect solution generated zero-shot}

\python{code/langprop/cartpole_zero_shot.py}{code:langprop/cartpole_zero_shot}{Example policy to solve CartPole generated zero-shot before LangProp optimization. The overly simplistic policy achieves a mean score of $9.9$ out of $500$.}

\subsubsection{Correct solution after applying LangProp}

\python{code/langprop/cartpole_trained.py}{code:langprop/cartpole_trained}{Example policy to solve CartPole after LangProp optimization. The policy learns a PID controller and achieves a mean score of $500$ out of $500$.}

\subsection{Driving code generated by LangProp}
\label{sec:langprop/langprop_driving_policy}
We show an example driving policy generated using LangProp, trained with both imitation learning and reinforcement learning, as described in \Cref{sec:langprop/agent_training}. Please refer to the open-sourced code repository (\Cref{sec:langprop/implementation}) for the full prompts and code used to train the policy, and pre-trained checkpoints for each training setting used in the evaluation.

\python{code/langprop/driving_code.py}{code:langprop/driving_policy_example}{Example driving policy generated by LangProp, trained with both imitation learning and reinforcement learning.}

\setcounter{listing}{0}
\renewcommand{\thelisting}{D.\arabic{listing}}
\setcounter{figure}{0} 
\renewcommand{\thefigure}{D.\arabic{figure}} 
\setcounter{table}{0} 
\renewcommand{\thetable}{D.\arabic{table}} 
\renewcommand{\chaptername}{Appendix}

\chapter{Clustering the Minecraft inventories}
\label{appendix:minecraft_inventory}

In the MineRL BASALT Competition 2022~\cite{milani2023solving}, only the RGB observation was given as input to the agent. Hence, \gls{iic}~\cite{ji2019invariant} was conducted in image space in \Cref{sec:minecraft_iic_results}.
\begin{figure}[b]
    \centering
     \includegraphics[width=0.48\textwidth]{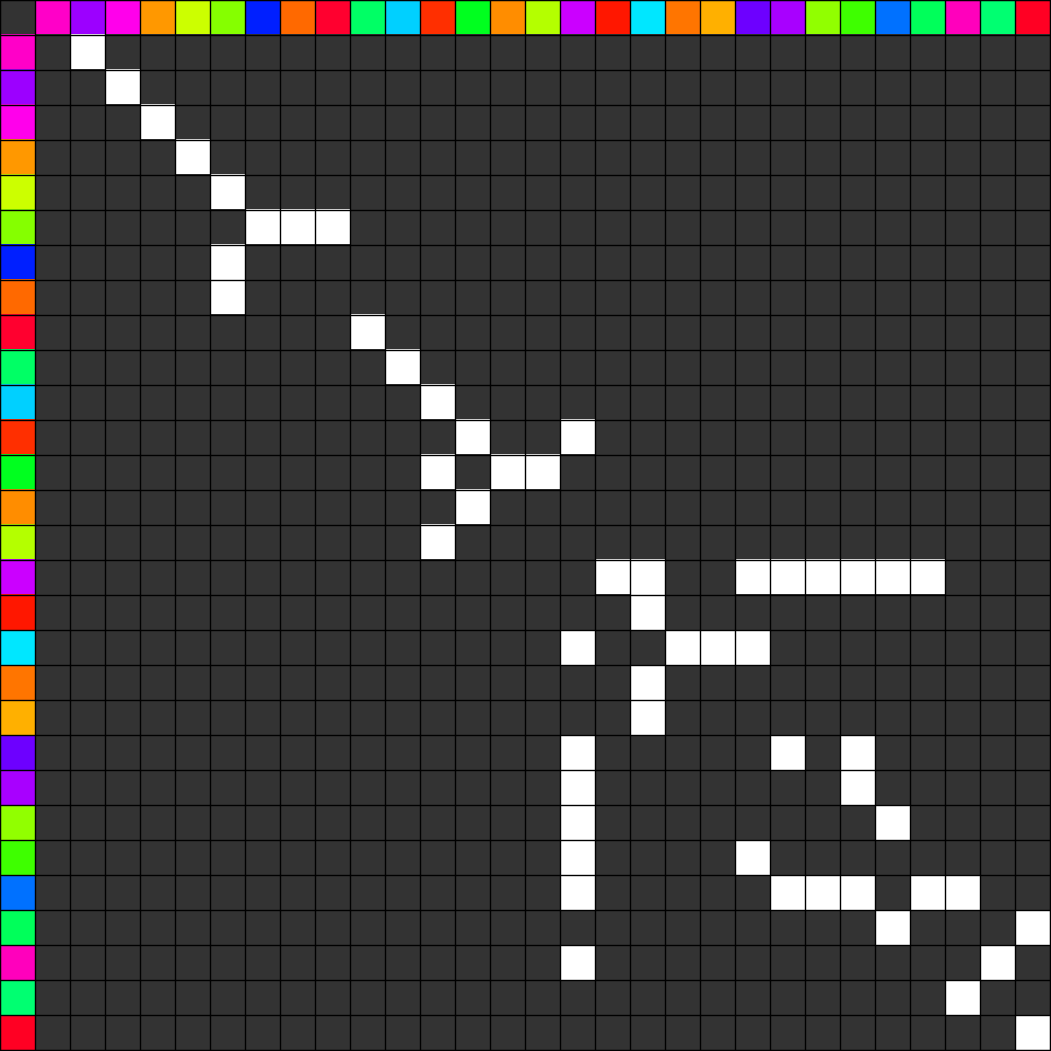}\hfill
     \includegraphics[width=0.48\textwidth]{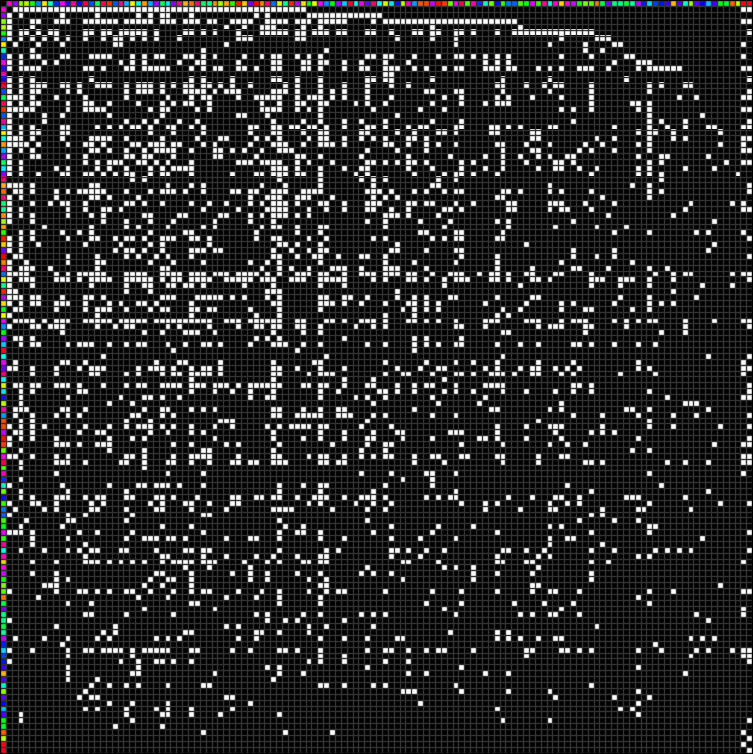}
    \caption{Transition matrix of \glsshort{iic} clusters. If there exists a transition from one cluster to the other, the corresponding cell in the adjacency matrix is coloured in white. \textbf{(left)} Transition within a single expert demonstration of the Obtain Diamond task, with 30 \glsshort{iic} clusters. \textbf{(right)} All transitions that appear in 122 expert demonstration trajectories of the Obtain Diamond task, with 128 \glsshort{iic} clusters.}
    \label{fig:minecraft/iic_transitions}
\end{figure}
However, at the start of this project, the specifications of the 2022 competition were not yet available. Since the MineRL BASALT Competition 2021~\cite{shah2022retrospective} provided the inventory information in the observation as a vector, this was used instead for \gls{iic} clustering early on in this project. An example expert demonstration for the Obtain Diamond task is shown in \Cref{fig:minecraft/inventory_iic}, along with the breakdown of the inventory and the learnt \gls{iic} cluster assignments. Transitions between the learnt clusters are shown in \Cref{fig:minecraft/iic_transitions}.

\begin{figure}[h]
    \centering
    \includegraphics[width=\textwidth]{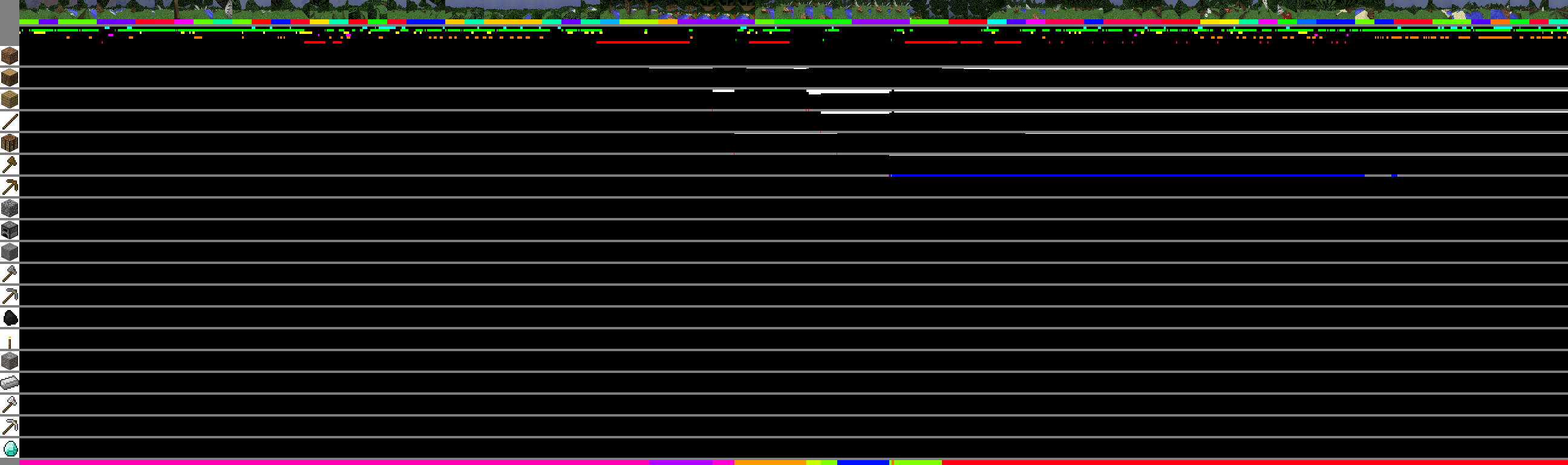}\\
    \vspace{5pt}
    \includegraphics[width=\textwidth]{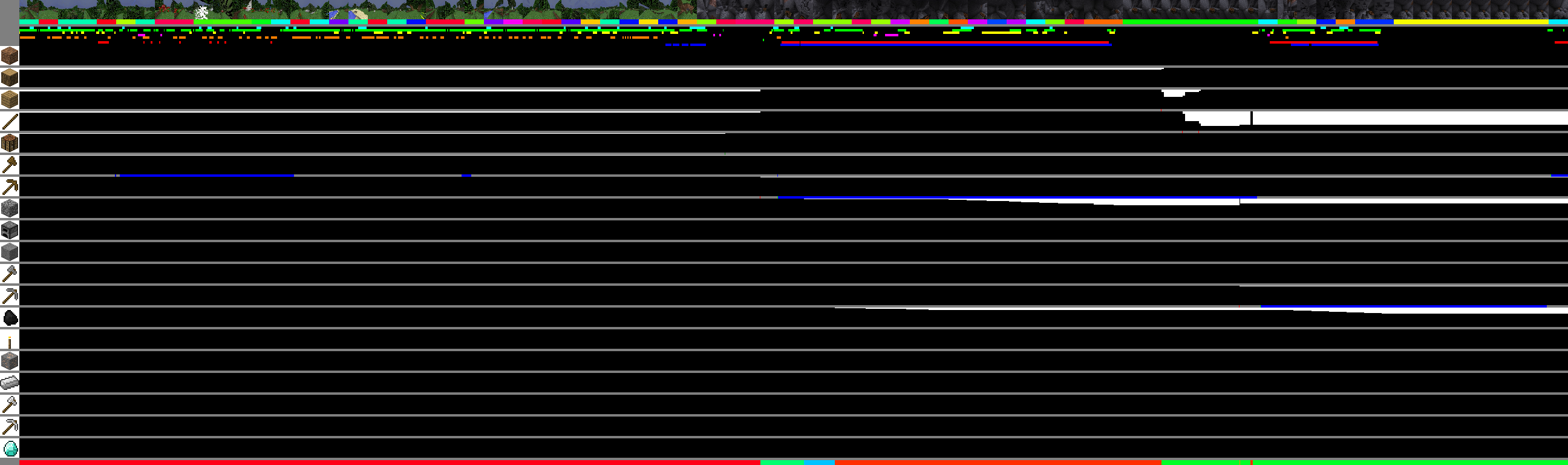}\\
    \vspace{5pt}
    \includegraphics[width=\textwidth]{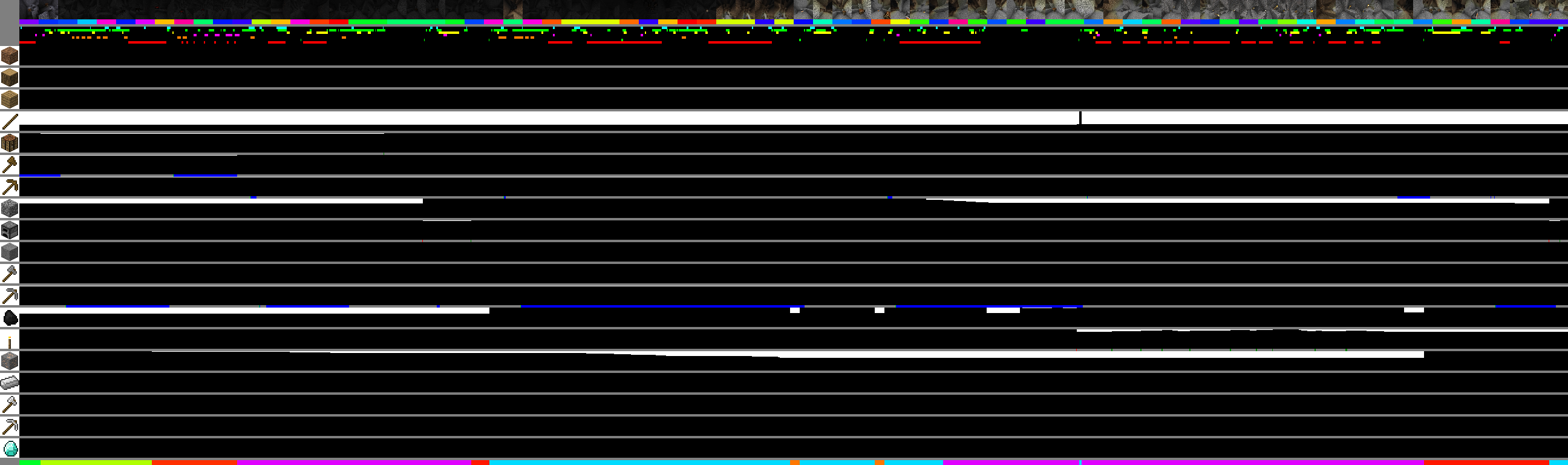}
    \caption{Expert demonstration for the Obtain Diamond task. The RGB observations are shown at the top. Next, a coloured strip indicates an \glsshort{iic} cluster assignment for 128 clusters. The next row is a visual encoding of the action taken per frame (in the order of: ``right'', ``forward'', ``left'', ``back'', ``jump'', ``sprint'', ``attack'', ``sneak''). The following rows show the number of items in the inventory for every item relevant to the Obtain Diamond task. The amount of each item is shown in a progress bar style. A blue strip beneath If the item is being used with a ``use'' action. Finally, a coloured strip at the bottom indicates an \glsshort{iic} cluster assignment for 30 clusters. [Spans multiple pages]}
    \label{fig:minecraft/inventory_iic}
\end{figure}

\begin{figure}[h]
    \ContinuedFloat
    \setcounter{subfigure}{4}
    \centering
    \includegraphics[width=\textwidth]{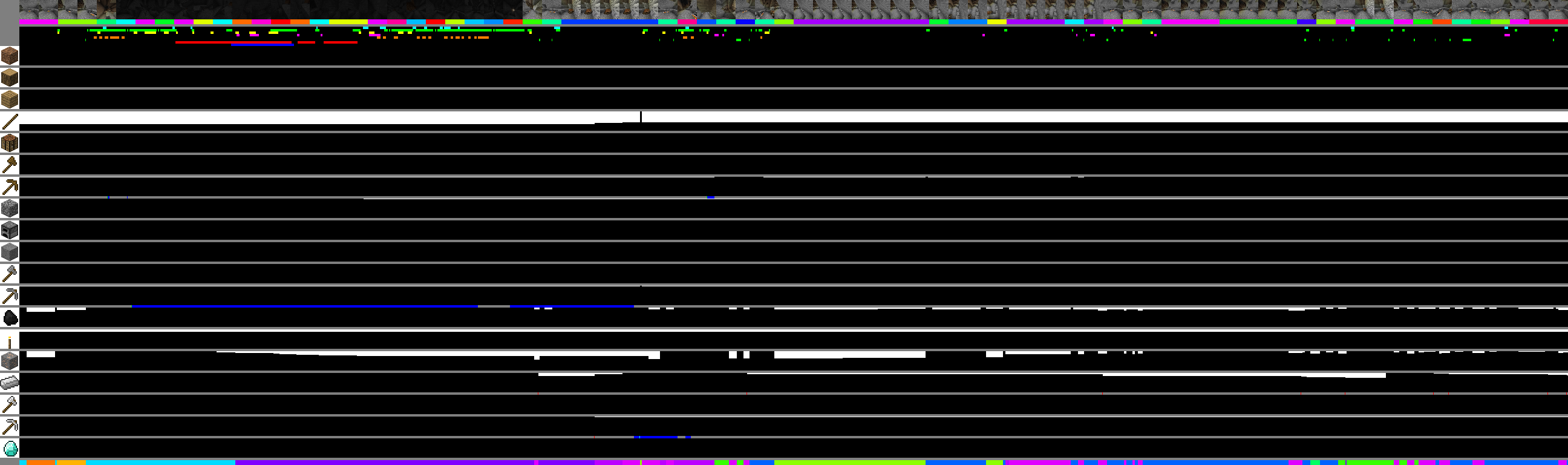}\\
    \vspace{5pt}
    \includegraphics[width=\textwidth]{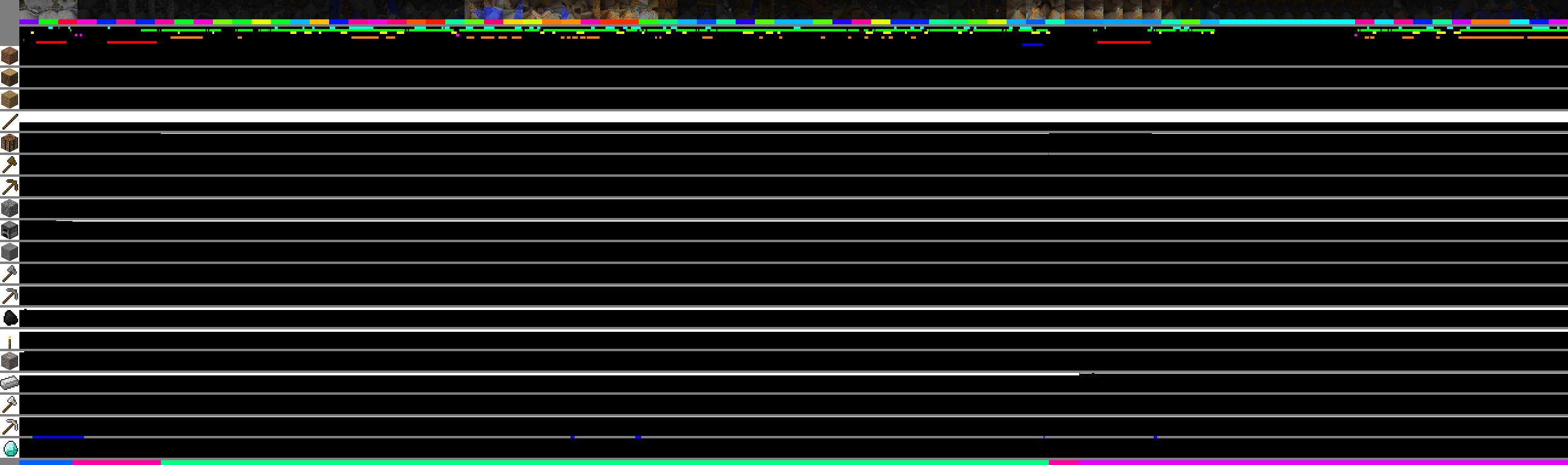}\\
    \vspace{5pt}
    \includegraphics[width=\textwidth]{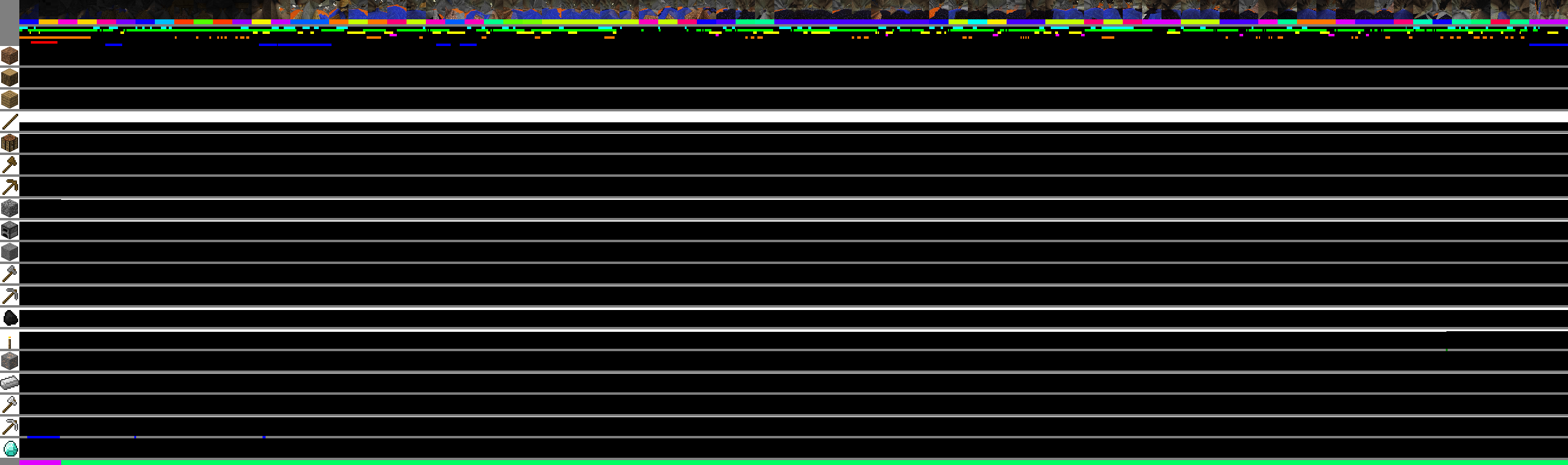}\\
    \vspace{5pt}
    \includegraphics[width=\textwidth]{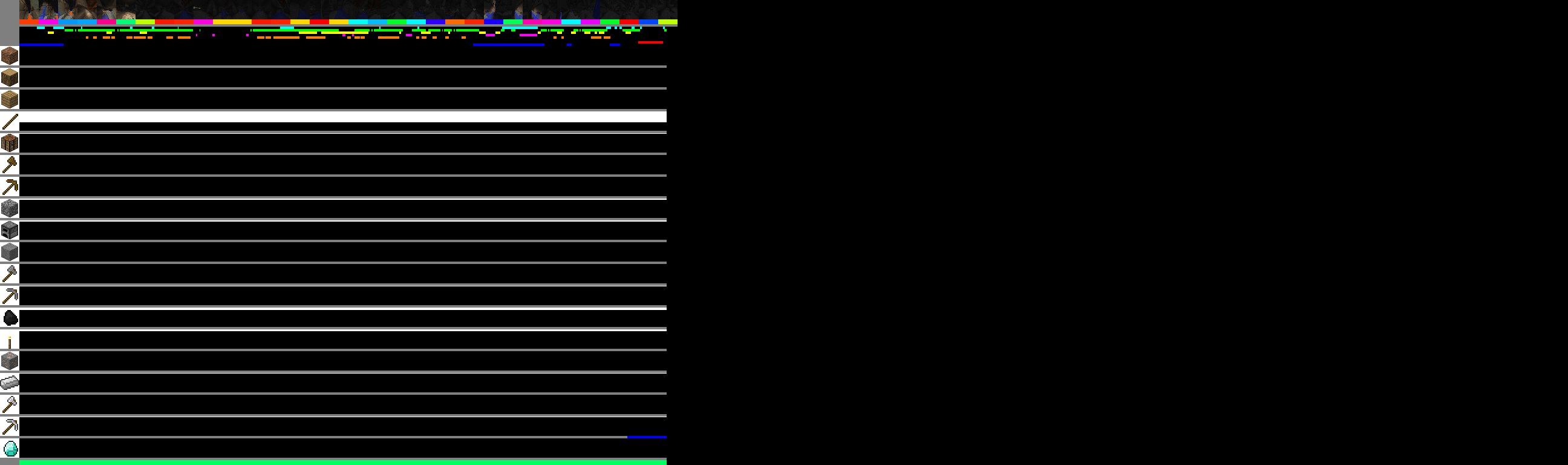}
    \caption*{[Continued] Expert demonstration for the Obtain Diamond task. The RGB observations are shown at the top. Next, a coloured strip indicates an \glsshort{iic} cluster assignment for 128 clusters. The next row is a visual encoding of the action taken per frame (in the order of: ``right'', ``forward'', ``left'', ``back'', ``jump'', ``sprint'', ``attack'', ``sneak''). The following rows show the number of items in the inventory for every item relevant to the Obtain Diamond task. The amount of each item is shown in a progress bar style. A blue strip beneath If the item is being used with a ``use'' action. Finally, a coloured strip at the bottom indicates an \glsshort{iic} cluster assignment for 30 clusters. }
\end{figure}

\backmatter 

\listofreferences{dphil_thesis}



\begin{thebibliography}{263}
\providecommand{\natexlab}[1]{#1}
\providecommand{\url}[1]{\texttt{#1}}
\expandafter\ifx\csname urlstyle\endcsname\relax
  \providecommand{\doi}[1]{doi: #1}\else
  \providecommand{\doi}{doi: \begingroup \urlstyle{rm}\Url}\fi

\bibitem[Abu-Nasser(2017)]{abu2017medical}
Bassem Abu-Nasser.
\newblock Medical expert systems survey.
\newblock \emph{International Journal of Engineering and Information Systems
  (IJEAIS)}, 1\penalty0 (7):\penalty0 218--224, 2017.

\bibitem[Akkaya et~al.(2019)Akkaya, Andrychowicz, Chociej, Litwin, McGrew,
  Petron, Paino, Plappert, Powell, Ribas, et~al.]{akkaya2019solvingrubik}
Ilge Akkaya, Marcin Andrychowicz, Maciek Chociej, Mateusz Litwin, Bob McGrew,
  Arthur Petron, Alex Paino, Matthias Plappert, Glenn Powell, Raphael Ribas,
  et~al.
\newblock Solving rubik's cube with a robot hand.
\newblock \emph{arXiv preprint arXiv:1910.07113}, 2019.

\bibitem[Alonso et~al.(2024)Alonso, Jelley, Micheli, Kanervisto, Storkey,
  Pearce, and Fleuret]{alonso2024diffusion}
Eloi Alonso, Adam Jelley, Vincent Micheli, Anssi Kanervisto, Amos Storkey, Tim
  Pearce, and Fran{\c{c}}ois Fleuret.
\newblock Diffusion for world modeling: Visual details matter in atari.
\newblock \emph{arXiv preprint arXiv:2405.12399}, 2024.

\bibitem[Ammirato et~al.(2017)Ammirato, Poirson, Park, Košecká, and
  Berg]{ammirato2017dataset}
Phil Ammirato, Patrick Poirson, Eunbyung Park, Jana Košecká, and Alexander~C.
  Berg.
\newblock A dataset for developing and benchmarking active vision.
\newblock In \emph{Proceedings of the IEEE International Conference on Robotics
  and Automation (ICRA)}, pages 1378--1385, 2017.
\newblock \doi{10.1109/ICRA.2017.7989164}.

\bibitem[Anderson et~al.(2018)Anderson, Chang, Chaplot, Dosovitskiy, Gupta,
  Koltun, Kosecka, Malik, Mottaghi, Savva, et~al.]{anderson2018evaluation}
Peter Anderson, Angel Chang, Devendra~Singh Chaplot, Alexey Dosovitskiy,
  Saurabh Gupta, Vladlen Koltun, Jana Kosecka, Jitendra Malik, Roozbeh
  Mottaghi, Manolis Savva, et~al.
\newblock On evaluation of embodied navigation agents.
\newblock \emph{arXiv preprint arXiv:1807.06757}, 2018.

\bibitem[Arora and Doshi(2021)]{arora2021survey}
Saurabh Arora and Prashant Doshi.
\newblock A survey of inverse reinforcement learning: Challenges, methods and
  progress.
\newblock \emph{Artificial Intelligence}, 297:\penalty0 103500, 2021.

\bibitem[Arulkumaran et~al.(2019)Arulkumaran, Cully, and
  Togelius]{arulkumaran2019alphastar}
Kai Arulkumaran, Antoine Cully, and Julian Togelius.
\newblock Alphastar: An evolutionary computation perspective.
\newblock In \emph{Proceedings of the genetic and evolutionary computation
  conference companion}, pages 314--315, 2019.

\bibitem[B{\"a}ck and Schwefel(1993)]{back1993overview_evolution}
Thomas B{\"a}ck and Hans-Paul Schwefel.
\newblock An overview of evolutionary algorithms for parameter optimization.
\newblock \emph{Evolutionary computation}, 1\penalty0 (1):\penalty0 1--23,
  1993.

\bibitem[Bacon et~al.(2017)Bacon, Harb, and Precup]{optioncritic}
Pierre-Luc Bacon, Jean Harb, and Doina Precup.
\newblock The {Option}-{Critic} architecture.
\newblock In \emph{Proceedings of the AAAI Conference of Artificial
  Intelligence}, AAAI'17, page 1726–1734. AAAI Press, 2017.

\bibitem[Badia et~al.(2020)Badia, Piot, Kapturowski, Sprechmann, Vitvitskyi,
  Guo, and Blundell]{badia_agent57_2020}
Adri{\`a}~Puigdom{\`e}nech Badia, Bilal Piot, Steven Kapturowski, Pablo
  Sprechmann, Alex Vitvitskyi, Zhaohan~Daniel Guo, and Charles Blundell.
\newblock Agent57: {Outperforming} the {Atari} human benchmark.
\newblock In \emph{Proceedings of the International Conference on Machine
  Learning (ICML)}, pages 507--517. PMLR, 2020.

\bibitem[Badue et~al.(2021)Badue, Guidolini, Carneiro, Azevedo, Cardoso,
  Forechi, Jesus, Berriel, Paixao, Mutz, et~al.]{badue2021self}
Claudine Badue, R{\^a}nik Guidolini, Raphael~Vivacqua Carneiro, Pedro Azevedo,
  Vinicius~B Cardoso, Avelino Forechi, Luan Jesus, Rodrigo Berriel, Thiago~M
  Paixao, Filipe Mutz, et~al.
\newblock Self-driving cars: {A} survey.
\newblock \emph{Expert Systems with Applications}, 165:\penalty0 113816, 2021.

\bibitem[Baker et~al.(2022)Baker, Akkaya, Zhokov, Huizinga, Tang, Ecoffet,
  Houghton, Sampedro, and Clune]{baker2022_openai_vpt}
Bowen Baker, Ilge Akkaya, Peter Zhokov, Joost Huizinga, Jie Tang, Adrien
  Ecoffet, Brandon Houghton, Raul Sampedro, and Jeff Clune.
\newblock Video pretraining ({VPT}): {Learning} to act by watching unlabeled
  online videos.
\newblock \emph{Advances in Neural Information Processing Systems (NeurIPS)},
  35:\penalty0 24639--24654, 2022.

\bibitem[Bansal et~al.(2018)Bansal, Krizhevsky, and
  Ogale]{bansal2018chauffeurnet}
Mayank Bansal, Alex Krizhevsky, and Abhijit Ogale.
\newblock {ChauffeurNet}: Learning to drive by imitating the best and
  synthesizing the worst.
\newblock \emph{arXiv preprint arXiv:1812.03079}, 2018.

\bibitem[Baum(1972)]{baum72}
Leonard~E. Baum.
\newblock An inequality and associated maximization technique in statistical
  estimation for probabilistic functions of {M}arkov processes.
\newblock In Oved Shisha, editor, \emph{{P}roceedings of the Symposium on
  Inequalities}, pages 1--8, University of California, Los Angeles, 1972.
  Academic Press.

\bibitem[Beck et~al.(2023)Beck, Vuorio, Liu, Xiong, Zintgraf, Finn, and
  Whiteson]{metarl_survey}
Jacob Beck, Risto Vuorio, Evan~Zheran Liu, Zheng Xiong, Luisa Zintgraf, Chelsea
  Finn, and Shimon Whiteson.
\newblock A survey of meta-reinforcement learning.
\newblock \emph{arXiv preprint arXiv:2301.08028}, 2023.

\bibitem[{Bellemare} et~al.(2013){Bellemare}, {Naddaf}, {Veness}, and
  {Bowling}]{bellemare13arcade}
M.~G. {Bellemare}, Y.~{Naddaf}, J.~{Veness}, and M.~{Bowling}.
\newblock The {Arcade} {Learning} {Environment}: {An} evaluation platform for
  general agents.
\newblock \emph{Journal of Artificial Intelligence Research}, 47:\penalty0
  253--279, 2013.

\bibitem[Bellman(1957)]{bellman_markovian_1957}
Richard Bellman.
\newblock A {Markovian} {Decision} {Process}.
\newblock \emph{Indiana University Mathematics Journal}, 6\penalty0
  (4):\penalty0 679--684, 1957.

\bibitem[Berner et~al.(2019)Berner, Brockman, Chan, Cheung, D{\k{e}}biak,
  Dennison, Farhi, Fischer, Hashme, Hesse, et~al.]{berner2019dota}
Christopher Berner, Greg Brockman, Brooke Chan, Vicki Cheung, Przemys{\l}aw
  D{\k{e}}biak, Christy Dennison, David Farhi, Quirin Fischer, Shariq Hashme,
  Chris Hesse, et~al.
\newblock Dota 2 with large scale deep reinforcement learning.
\newblock \emph{arXiv preprint arXiv:1912.06680}, 2019.

\bibitem[Bishop(2006)]{bishop_ml_book}
Christopher~M. Bishop.
\newblock \emph{{Pattern} {Recognition} and {Machine} {Learning} ({Information}
  {Science} and {Statistics})}.
\newblock Springer-Verlag, Berlin, Heidelberg, 2006.
\newblock ISBN 0387310738.

\bibitem[Bojarski et~al.(2016)Bojarski, Del~Testa, Dworakowski, Firner, Flepp,
  Goyal, Jackel, Monfort, Muller, Zhang, et~al.]{bojarski2016end}
Mariusz Bojarski, Davide Del~Testa, Daniel Dworakowski, Bernhard Firner, Beat
  Flepp, Prasoon Goyal, Lawrence~D Jackel, Mathew Monfort, Urs Muller, Jiakai
  Zhang, et~al.
\newblock End to end learning for self-driving cars.
\newblock \emph{arXiv preprint arXiv:1604.07316}, 2016.

\bibitem[Bommasani et~al.(2021)Bommasani, Hudson, Adeli, Altman, Arora, von
  Arx, Bernstein, Bohg, Bosselut, Brunskill,
  et~al.]{bommasani2021opportunities}
Rishi Bommasani, Drew~A Hudson, Ehsan Adeli, Russ Altman, Simran Arora, Sydney
  von Arx, Michael~S Bernstein, Jeannette Bohg, Antoine Bosselut, Emma
  Brunskill, et~al.
\newblock On the opportunities and risks of foundation models.
\newblock \emph{arXiv preprint arXiv:2108.07258}, 2021.

\bibitem[Brockman et~al.(2016)Brockman, Cheung, Pettersson, Schneider,
  Schulman, Tang, and Zaremba]{brockman2016openaigym}
Greg Brockman, Vicki Cheung, Ludwig Pettersson, Jonas Schneider, John Schulman,
  Jie Tang, and Wojciech Zaremba.
\newblock {OpenAI} {Gym}.
\newblock \emph{arXiv preprint arXiv:1606.01540}, 2016.

\bibitem[Brown et~al.(2020)Brown, Mann, Ryder, Subbiah, Kaplan, Dhariwal,
  Neelakantan, Shyam, Sastry, Askell, Agarwal, Herbert-Voss, Krueger, Henighan,
  Child, Ramesh, Ziegler, Wu, Winter, Hesse, Chen, Sigler, Litwin, Gray, Chess,
  Clark, Berner, McCandlish, Radford, Sutskever, and Amodei]{NEURIPS2020_gpt3}
Tom Brown, Benjamin Mann, Nick Ryder, Melanie Subbiah, Jared~D Kaplan, Prafulla
  Dhariwal, Arvind Neelakantan, Pranav Shyam, Girish Sastry, Amanda Askell,
  Sandhini Agarwal, Ariel Herbert-Voss, Gretchen Krueger, Tom Henighan, Rewon
  Child, Aditya Ramesh, Daniel Ziegler, Jeffrey Wu, Clemens Winter, Chris
  Hesse, Mark Chen, Eric Sigler, Mateusz Litwin, Scott Gray, Benjamin Chess,
  Jack Clark, Christopher Berner, Sam McCandlish, Alec Radford, Ilya Sutskever,
  and Dario Amodei.
\newblock Language models are few-shot learners.
\newblock In H.~Larochelle, M.~Ranzato, R.~Hadsell, M.F. Balcan, and H.~Lin,
  editors, \emph{Advances in Neural Information Processing Systems (NeurIPS)},
  volume~33, pages 1877--1901. Curran Associates, Inc., 2020.

\bibitem[Campbell et~al.(2002)Campbell, Hoane~Jr, and Hsu]{campbell2002deep}
Murray Campbell, A~Joseph Hoane~Jr, and Feng-hsiung Hsu.
\newblock Deep {Blue}.
\newblock \emph{Artificial intelligence}, 134\penalty0 (1-2):\penalty0 57--83,
  2002.

\bibitem[CARLA(2020)]{carla_leaderboard}
CARLA.
\newblock {CARLA} autonomous driving leaderboard.
\newblock \url{https://leaderboard.carla.org/}, 2020.

\bibitem[Cartillier et~al.(2021)Cartillier, Ren, Jain, Lee, Essa, and
  Batra]{cartillier2021semantic}
Vincent Cartillier, Zhile Ren, Neha Jain, Stefan Lee, Irfan Essa, and Dhruv
  Batra.
\newblock {Semantic} {MapNet}: Building allocentric semantic maps and
  representations from egocentric views.
\newblock In \emph{Proceedings of the AAAI Conference of Artificial
  Intelligence}, volume~35, pages 964--972, 2021.

\bibitem[Chang et~al.(2017)Chang, Dai, Funkhouser, Halber, Niebner, Savva,
  Song, Zeng, and Zhang]{chang2017matterport3d}
Angel Chang, Angela Dai, Thomas Funkhouser, Maciej Halber, Matthias Niebner,
  Manolis Savva, Shuran Song, Andy Zeng, and Yinda Zhang.
\newblock {Matterport3D}: {Learning} from {RGB-D} data in indoor environments.
\newblock In \emph{Proceedings of the International Conference on 3D Vision
  (3DV)}, pages 667--676. IEEE, 2017.

\bibitem[Chaplot et~al.(2020{\natexlab{a}})Chaplot, Gandhi, Gupta, and
  Salakhutdinov]{chaplot2020object}
Devendra~Singh Chaplot, Dhiraj Gandhi, Abhinav Gupta, and Ruslan Salakhutdinov.
\newblock Object goal navigation using goal-oriented semantic exploration.
\newblock \emph{Advances in Neural Information Processing Systems (NeurIPS)},
  2020{\natexlab{a}}.

\bibitem[Chaplot et~al.(2020{\natexlab{b}})Chaplot, Gandhi, Gupta, Gupta, and
  Salakhutdinov]{chaplot_learning_2020}
Devendra~Singh Chaplot, Dhiraj Gandhi, Saurabh Gupta, Abhinav Gupta, and Ruslan
  Salakhutdinov.
\newblock Learning to {Explore} using {Active} {Neural} {SLAM}.
\newblock In \emph{Proceedings of the International Conference on Learning
  Representations (ICLR)}, page~18, 2020{\natexlab{b}}.

\bibitem[Charles et~al.(2017)Charles, Su, Kaichun, and
  Guibas]{charles_pointnet_2017}
R.~Qi Charles, Hao Su, Mo~Kaichun, and Leonidas~J. Guibas.
\newblock {PointNet}: {Deep} {Learning} on {Point} {Sets} for {3D}
  {Classification} and {Segmentation}.
\newblock In \emph{Proceedings of the IEEE/CVF Conference on Computer Vision
  and Pattern Recognition (CVPR)}, pages 77--85, Honolulu, HI, 2017. IEEE.
\newblock ISBN 978-1-5386-0457-1.
\newblock \doi{10.1109/CVPR.2017.16}.

\bibitem[Chen et~al.(2023)Chen, Wu, Chitta, Jaeger, Geiger, and
  Li]{chen2023end}
Li~Chen, Penghao Wu, Kashyap Chitta, Bernhard Jaeger, Andreas Geiger, and
  Hongyang Li.
\newblock End-to-end autonomous driving: Challenges and frontiers.
\newblock \emph{arXiv preprint arXiv:2306.16927}, 2023.

\bibitem[Chen et~al.(2021{\natexlab{a}})Chen, Lu, Rajeswaran, Lee, Grover,
  Laskin, Abbeel, Srinivas, and Mordatch]{chen2021decision}
Lili Chen, Kevin Lu, Aravind Rajeswaran, Kimin Lee, Aditya Grover, Misha
  Laskin, Pieter Abbeel, Aravind Srinivas, and Igor Mordatch.
\newblock Decision transformer: Reinforcement learning via sequence modeling.
\newblock \emph{Advances in Neural Information Processing Systems (NeurIPS)},
  34:\penalty0 15084--15097, 2021{\natexlab{a}}.

\bibitem[Chen et~al.(2021{\natexlab{b}})Chen, Tworek, Jun, Yuan, Pinto, Kaplan,
  Edwards, Burda, Joseph, Brockman, et~al.]{chen2021evaluating_codex}
Mark Chen, Jerry Tworek, Heewoo Jun, Qiming Yuan, Henrique Ponde de~Oliveira
  Pinto, Jared Kaplan, Harri Edwards, Yuri Burda, Nicholas Joseph, Greg
  Brockman, et~al.
\newblock Evaluating large language models trained on code.
\newblock \emph{arXiv preprint arXiv:2107.03374}, 2021{\natexlab{b}}.

\bibitem[Chevalier-Boisvert(2018)]{gym_miniworld}
Maxime Chevalier-Boisvert.
\newblock gym-miniworld environment for {OpenAI} {Gym}.
\newblock \url{https://github.com/maximecb/gym-miniworld}, 2018.

\bibitem[Chitta et~al.(2022)Chitta, Prakash, Jaeger, Yu, Renz, and
  Geiger]{chitta2022transfuser}
Kashyap Chitta, Aditya Prakash, Bernhard Jaeger, Zehao Yu, Katrin Renz, and
  Andreas Geiger.
\newblock {TransFuser}: Imitation with transformer-based sensor fusion for
  autonomous driving.
\newblock \emph{IEEE Transactions on Pattern Analysis and Machine
  Intelligence}, 2022.

\bibitem[Cho et~al.(2014)Cho, van Merri{\"e}nboer, Gulcehre, Bahdanau,
  Bougares, Schwenk, and Bengio]{cho_gru_2014}
Kyunghyun Cho, Bart van Merri{\"e}nboer, Caglar Gulcehre, Dzmitry Bahdanau,
  Fethi Bougares, Holger Schwenk, and Yoshua Bengio.
\newblock Learning phrase representations using {RNN} encoder{--}decoder for
  statistical machine translation.
\newblock In Alessandro Moschitti, Bo~Pang, and Walter Daelemans, editors,
  \emph{Proceedings of the Conference on Empirical Methods in Natural Language
  Processing (EMNLP)}, pages 1724--1734, Doha, Qatar, 2014. Association for
  Computational Linguistics.
\newblock \doi{10.3115/v1/D14-1179}.

\bibitem[Clavera et~al.(2018)Clavera, Rothfuss, Schulman, Fujita, Asfour, and
  Abbeel]{clavera2018model_mbmpo}
Ignasi Clavera, Jonas Rothfuss, John Schulman, Yasuhiro Fujita, Tamim Asfour,
  and Pieter Abbeel.
\newblock Model-based reinforcement learning via meta-policy optimization.
\newblock In \emph{Proceedings of the Conference on Robot Learning (CoRL)},
  pages 617--629. PMLR, 2018.

\bibitem[Cobbe et~al.(2021{\natexlab{a}})Cobbe, Kosaraju, Bavarian, Chen, Jun,
  Kaiser, Plappert, Tworek, Hilton, Nakano, et~al.]{cobbe2021training}
Karl Cobbe, Vineet Kosaraju, Mohammad Bavarian, Mark Chen, Heewoo Jun, Lukasz
  Kaiser, Matthias Plappert, Jerry Tworek, Jacob Hilton, Reiichiro Nakano,
  et~al.
\newblock Training verifiers to solve math word problems.
\newblock \emph{arXiv preprint arXiv:2110.14168}, 2021{\natexlab{a}}.

\bibitem[Cobbe et~al.(2021{\natexlab{b}})Cobbe, Hilton, Klimov, and
  Schulman]{cobbe2021phasic}
Karl~W Cobbe, Jacob Hilton, Oleg Klimov, and John Schulman.
\newblock Phasic policy gradient.
\newblock In \emph{Proceedings of the International Conference on Machine
  Learning (ICML)}, pages 2020--2027. PMLR, 2021{\natexlab{b}}.

\bibitem[Coulom(2006)]{coulom2006efficient_mcts}
R{\'e}mi Coulom.
\newblock Efficient selectivity and backup operators in {Monte}-{Carlo} {Tree}
  {Search}.
\newblock In \emph{Proceedings of the International Conference on Computers and
  Games}, pages 72--83. Springer, 2006.

\bibitem[Crawshaw(2020)]{crawshaw2020multi}
Michael Crawshaw.
\newblock Multi-task learning with deep neural networks: A survey.
\newblock \emph{arXiv preprint arXiv:2009.09796}, 2020.

\bibitem[Daniel et~al.(2016)Daniel, Van~Hoof, Peters, and
  Neumann]{option_ml_2016}
Christian Daniel, Herke Van~Hoof, Jan Peters, and Gerhard Neumann.
\newblock Probabilistic inference for determining options in reinforcement
  learning.
\newblock \emph{Machine Learning}, 104:\penalty0 337--357, 2016.

\bibitem[De~Haan et~al.(2019)De~Haan, Jayaraman, and Levine]{de2019causal}
Pim De~Haan, Dinesh Jayaraman, and Sergey Levine.
\newblock Causal confusion in imitation learning.
\newblock \emph{Advances in Neural Information Processing Systems (NeurIPS)},
  32, 2019.

\bibitem[Deitke et~al.(2020)Deitke, Han, Herrasti, Kembhavi, Kolve, Mottaghi,
  Salvador, Schwenk, VanderBilt, Wallingford, et~al.]{deitke2020robothor}
Matt Deitke, Winson Han, Alvaro Herrasti, Aniruddha Kembhavi, Eric Kolve,
  Roozbeh Mottaghi, Jordi Salvador, Dustin Schwenk, Eli VanderBilt, Matthew
  Wallingford, et~al.
\newblock {RoboTHOR}: {An} open simulation-to-real embodied {AI} platform.
\newblock In \emph{Proceedings of the IEEE/CVF Conference on Computer Vision
  and Pattern Recognition (CVPR)}, pages 3164--3174, 2020.

\bibitem[Dempster et~al.(1977)Dempster, Laird, and Rubin]{em_algorithm}
A.~P. Dempster, N.~M. Laird, and D.~B. Rubin.
\newblock Maximum likelihood from incomplete data via the {EM} algorithm.
\newblock \emph{Journal of the Royal Statistical Society. Series B
  (Methodological)}, 39\penalty0 (1):\penalty0 1--38, 1977.

\bibitem[Dijkstra(1959)]{dijkstra}
E.~W. Dijkstra.
\newblock A note on two problems in connexion with graphs.
\newblock \emph{Numerische Mathematik}, 1959.

\bibitem[Ding et~al.(2024)Ding, Zhang, Tian, and Zheng]{ding2024diffusion}
Zihan Ding, Amy Zhang, Yuandong Tian, and Qinqing Zheng.
\newblock Diffusion world model: Future modeling beyond step-by-step rollout
  for offline reinforcement learning.
\newblock \emph{arXiv preprint arXiv:2402.03570}, 2024.

\bibitem[Dosovitskiy et~al.(2017)Dosovitskiy, Ros, Codevilla, Lopez, and
  Koltun]{dosovitskiy17a_carla}
Alexey Dosovitskiy, German Ros, Felipe Codevilla, Antonio Lopez, and Vladlen
  Koltun.
\newblock {CARLA}: {An} open urban driving simulator.
\newblock In Sergey Levine, Vincent Vanhoucke, and Ken Goldberg, editors,
  \emph{Proceedings of the Conference on Robot Learning (CoRL)}, volume~78 of
  \emph{Proceedings of Machine Learning Research}, pages 1--16. PMLR, 2017.

\bibitem[Driess et~al.(2023)Driess, Xia, Sajjadi, Lynch, Chowdhery, Ichter,
  Wahid, Tompson, Vuong, Yu, et~al.]{driess2023palme}
Danny Driess, Fei Xia, Mehdi~SM Sajjadi, Corey Lynch, Aakanksha Chowdhery,
  Brian Ichter, Ayzaan Wahid, Jonathan Tompson, Quan Vuong, Tianhe Yu, et~al.
\newblock {PaLM}-{E}: {An} {Embodied} {Multimodal} {Language} {Model}.
\newblock In \emph{Proceedings of the International Conference on Machine
  Learning (ICML)}, pages 8469--8488. PMLR, 2023.

\bibitem[Duan et~al.(2016)Duan, Schulman, Chen, Bartlett, Sutskever, and
  Abbeel]{duan2016rl}
Yan Duan, John Schulman, Xi~Chen, Peter~L Bartlett, Ilya Sutskever, and Pieter
  Abbeel.
\newblock {R}{L}$^2$: Fast reinforcement learning via slow reinforcement
  learning.
\newblock \emph{arXiv preprint arXiv:1611.02779}, 2016.

\bibitem[Dulac-Arnold et~al.(2021)Dulac-Arnold, Levine, Mankowitz, Li,
  Paduraru, Gowal, and Hester]{dulac2021challenges}
Gabriel Dulac-Arnold, Nir Levine, Daniel~J Mankowitz, Jerry Li, Cosmin
  Paduraru, Sven Gowal, and Todd Hester.
\newblock Challenges of real-world reinforcement learning: definitions,
  benchmarks and analysis.
\newblock \emph{Machine Learning}, 110\penalty0 (9):\penalty0 2419--2468, 2021.

\bibitem[Falcon(2019)]{falcon2019pytorchlightning}
William~A Falcon.
\newblock {PyTorch} {Lightning}.
\newblock \url{https://github.com/Lightning-AI/lightning}, 2019.

\bibitem[Fan et~al.(2022)Fan, Wang, Jiang, Mandlekar, Yang, Zhu, Tang, Huang,
  Zhu, and Anandkumar]{fan2022minedojo}
Linxi Fan, Guanzhi Wang, Yunfan Jiang, Ajay Mandlekar, Yuncong Yang, Haoyi Zhu,
  Andrew Tang, De-An Huang, Yuke Zhu, and Anima Anandkumar.
\newblock {MineDojo}: {Building} open-ended embodied agents with internet-scale
  knowledge.
\newblock \emph{Advances in Neural Information Processing Systems (NeurIPS)},
  35:\penalty0 18343--18362, 2022.

\bibitem[Farquhar et~al.(2018)Farquhar, Rocktaeschel, Igl, and
  Whiteson]{farquhar2017treeqn}
Gregory Farquhar, Tim Rocktaeschel, Maximilian Igl, and Shimon Whiteson.
\newblock {TreeQN} and {ATreeC}: {Differentiable} tree planning for deep
  reinforcement learning.
\newblock In \emph{Proceedings of the International Conference on Learning
  Representations (ICLR)}, 2018.

\bibitem[Fawzi et~al.(2022)Fawzi, Balog, Huang, Hubert, Romera-Paredes,
  Barekatain, Novikov, R~Ruiz, Schrittwieser, Swirszcz,
  et~al.]{fawzi2022discovering}
Alhussein Fawzi, Matej Balog, Aja Huang, Thomas Hubert, Bernardino
  Romera-Paredes, Mohammadamin Barekatain, Alexander Novikov, Francisco~J
  R~Ruiz, Julian Schrittwieser, Grzegorz Swirszcz, et~al.
\newblock Discovering faster matrix multiplication algorithms with
  reinforcement learning.
\newblock \emph{Nature}, 610\penalty0 (7930):\penalty0 47--53, 2022.

\bibitem[Feinberg et~al.(2018)Feinberg, Wan, Stoica, Jordan, Gonzalez, and
  Levine]{feinberg2018model}
Vladimir Feinberg, Alvin Wan, Ion Stoica, Michael~I Jordan, Joseph~E Gonzalez,
  and Sergey Levine.
\newblock Model-based value expansion for efficient model-free reinforcement
  learning.
\newblock In \emph{Proceedings of the International Conference on Machine
  Learning (ICML)}, 2018.

\bibitem[Finn et~al.(2017)Finn, Abbeel, and Levine]{finn2017model_maml}
Chelsea Finn, Pieter Abbeel, and Sergey Levine.
\newblock Model-agnostic meta-learning for fast adaptation of deep networks.
\newblock In \emph{Proceedings of the International Conference on Machine
  Learning (ICML)}, pages 1126--1135. PMLR, 2017.

\bibitem[Forgy(1965)]{kmeans_Forgy1965ClusterAO}
E.~W. Forgy.
\newblock Cluster analysis of multivariate data: efficiency versus
  interpretability of classifications.
\newblock \emph{Biometrics}, 21:\penalty0 768--769, 1965.

\bibitem[Fox et~al.(2017)Fox, Krishnan, Stoica, and Goldberg]{fox2017multi}
Roy Fox, Sanjay Krishnan, Ion Stoica, and Ken Goldberg.
\newblock Multi-level discovery of deep options.
\newblock \emph{arXiv preprint arXiv:1703.08294}, 2017.

\bibitem[Fu et~al.(2024)Fu, Li, Wen, Dou, Cai, Shi, and Qiao]{fu2023drive}
Daocheng Fu, Xin Li, Licheng Wen, Min Dou, Pinlong Cai, Botian Shi, and
  Yu~Qiao.
\newblock Drive like a human: {Rethinking} autonomous driving with large
  language models.
\newblock In \emph{Proceedings of the IEEE/CVF Winter Conference on
  Applications of Computer Vision (WACV)}, pages 910--919, 2024.

\bibitem[Fu et~al.(2018)Fu, Gong, Wang, Batmanghelich, and Tao]{fu2018deep}
Huan Fu, Mingming Gong, Chaohui Wang, Kayhan Batmanghelich, and Dacheng Tao.
\newblock Deep ordinal regression network for monocular depth estimation.
\newblock In \emph{Proceedings of the IEEE/CVF Conference on Computer Vision
  and Pattern Recognition (CVPR)}, pages 2002--2011, 2018.

\bibitem[Fuentes-Pacheco et~al.(2015)Fuentes-Pacheco, Ascencio, and
  Rendon-Mancha]{fuentes-pacheco_visual_2015}
Jorge Fuentes-Pacheco, Jose Ascencio, and J.~Rendon-Mancha.
\newblock Visual {Simultaneous} {Localization} and {Mapping}: {A} {Survey}.
\newblock \emph{Artificial Intelligence Review}, 43, 2015.
\newblock \doi{10.1007/s10462-012-9365-8}.

\bibitem[Fujimoto et~al.(2018)Fujimoto, Hoof, and
  Meger]{fujimoto2018addressing}
Scott Fujimoto, Herke Hoof, and David Meger.
\newblock Addressing function approximation error in {Actor}-{Critic} methods.
\newblock In \emph{Proceedings of the International Conference on Machine
  Learning (ICML)}, pages 1587--1596. PMLR, 2018.

\bibitem[Giammarino and Paschalidis(2021)]{online_baum_welch_2021}
Vittorio Giammarino and Ioannis~Ch. Paschalidis.
\newblock Online {Baum}-{Welch} algorithm for hierarchical imitation learning.
\newblock In \emph{Proceedings of the IEEE Conference on Decision and Control
  (CDC)}, pages 3717--3722, 2021.
\newblock \doi{10.1109/CDC45484.2021.9683044}.

\bibitem[Goel and Huber(2003)]{goel2003subgoal}
Sandeep Goel and Manfred Huber.
\newblock Subgoal discovery for hierarchical reinforcement learning using
  learned policies.
\newblock In \emph{Proceedings of the Florida Artificial Intelligence Research
  Society (FLAIRS) Conference}, pages 346--350, 2003.

\bibitem[Goodfellow et~al.(2014)Goodfellow, Pouget-Abadie, Mirza, Xu,
  Warde-Farley, Ozair, Courville, and Bengio]{goodfellow2014gan}
Ian Goodfellow, Jean Pouget-Abadie, Mehdi Mirza, Bing Xu, David Warde-Farley,
  Sherjil Ozair, Aaron Courville, and Yoshua Bengio.
\newblock Generative adversarial nets.
\newblock \emph{Advances in Neural Information Processing Systems (NeurIPS)},
  27, 2014.

\bibitem[Goodfellow et~al.(2016)Goodfellow, Bengio, Courville, and
  Bengio]{goodfellow2016deep}
Ian Goodfellow, Yoshua Bengio, Aaron Courville, and Yoshua Bengio.
\newblock \emph{Deep learning}, volume~1.
\newblock MIT Press Cambridge, 2016.

\bibitem[Graves et~al.(2006)Graves, Fern{\'a}ndez, Gomez, and
  Schmidhuber]{graves2006connectionist}
Alex Graves, Santiago Fern{\'a}ndez, Faustino Gomez, and J{\"u}rgen
  Schmidhuber.
\newblock Connectionist temporal classification: labelling unsegmented sequence
  data with recurrent neural networks.
\newblock In \emph{Proceedings of the International Conference on Machine
  Learning (ICML)}, pages 369--376, 2006.

\bibitem[Graves et~al.(2014)Graves, Wayne, and Danihelka]{graves_neural_2014}
Alex Graves, Greg Wayne, and Ivo Danihelka.
\newblock Neural {Turing} {Machines}.
\newblock \emph{arXiv:1410.5401 [cs]}, 2014.
\newblock arXiv: 1410.5401.

\bibitem[Gu et~al.(2017)Gu, Holly, Lillicrap, and Levine]{gu2017deep}
Shixiang Gu, Ethan Holly, Timothy Lillicrap, and Sergey Levine.
\newblock Deep reinforcement learning for robotic manipulation with
  asynchronous off-policy updates.
\newblock In \emph{Proceedings of the IEEE International Conference on Robotics
  and Automation (ICRA)}, pages 3389--3396. IEEE, 2017.

\bibitem[Guo et~al.(2024)Guo, Zhu, Yang, Xie, Dong, Zhang, Chen, Bi, Wu, Li,
  et~al.]{guo2024deepseek}
Daya Guo, Qihao Zhu, Dejian Yang, Zhenda Xie, Kai Dong, Wentao Zhang, Guanting
  Chen, Xiao Bi, Y~Wu, YK~Li, et~al.
\newblock {DeepSeek}-{Coder}: {When} the large language model meets
  programming--the rise of code intelligence.
\newblock \emph{arXiv preprint arXiv:2401.14196}, 2024.

\bibitem[Gupta et~al.(2017)Gupta, Tolani, Davidson, Levine, Sukthankar, and
  Malik]{gupta_cognitive_2017}
Saurabh Gupta, Varun Tolani, James Davidson, Sergey Levine, Rahul Sukthankar,
  and Jitendra Malik.
\newblock Cognitive {Mapping} and {Planning} for {Visual} {Navigation}.
\newblock In \emph{Proceedings of the IEEE/CVF Conference on Computer Vision
  and Pattern Recognition (CVPR)}, 2017.
\newblock arXiv: 1702.03920.

\bibitem[Gupta and Kembhavi(2023)]{gupta2023visprog}
Tanmay Gupta and Aniruddha Kembhavi.
\newblock Visual programming: {Compositional} visual reasoning without
  training.
\newblock In \emph{Proceedings of the IEEE/CVF Conference on Computer Vision
  and Pattern Recognition (CVPR)}, pages 14953--14962, 2023.

\bibitem[Guss et~al.(2019)Guss, Houghton, Topin, Wang, Codel, Veloso, and
  Salakhutdinov]{guss2019minerl}
William~H Guss, Brandon Houghton, Nicholay Topin, Phillip Wang, Cayden Codel,
  Manuela Veloso, and Ruslan Salakhutdinov.
\newblock {MineRL}: a large-scale dataset of {Minecraft} demonstrations.
\newblock In \emph{Proceedings of the International Joint Conference on
  Artificial Intelligence (IJCAI)}, pages 2442--2448, 2019.

\bibitem[Ha and Schmidhuber(2018)]{ha2018worldmodels}
David Ha and J{\"u}rgen Schmidhuber.
\newblock World models.
\newblock \emph{arXiv preprint arXiv:1803.10122}, 2018.

\bibitem[Haarnoja et~al.(2018)Haarnoja, Zhou, Abbeel, and
  Levine]{haarnoja2018soft}
Tuomas Haarnoja, Aurick Zhou, Pieter Abbeel, and Sergey Levine.
\newblock Soft actor-critic: {Off-policy} maximum entropy deep reinforcement
  learning with a stochastic actor.
\newblock In \emph{Proceedings of the International Conference on Machine
  Learning (ICML)}, pages 1861--1870. PMLR, 2018.

\bibitem[Haarnoja et~al.(2024)Haarnoja, Moran, Lever, Huang, Tirumala, Humplik,
  Wulfmeier, Tunyasuvunakool, Siegel, Hafner, et~al.]{haarnoja2024learning}
Tuomas Haarnoja, Ben Moran, Guy Lever, Sandy~H Huang, Dhruva Tirumala, Jan
  Humplik, Markus Wulfmeier, Saran Tunyasuvunakool, Noah~Y Siegel, Roland
  Hafner, et~al.
\newblock Learning agile soccer skills for a bipedal robot with deep
  reinforcement learning.
\newblock \emph{Science Robotics}, 9\penalty0 (89):\penalty0 eadi8022, 2024.

\bibitem[Hafner et~al.(2019)Hafner, Lillicrap, Fischer, Villegas, Ha, Lee, and
  Davidson]{hafner2019learning_planet}
Danijar Hafner, Timothy Lillicrap, Ian Fischer, Ruben Villegas, David Ha,
  Honglak Lee, and James Davidson.
\newblock Learning latent dynamics for planning from pixels.
\newblock In \emph{Proceedings of the International Conference on Machine
  Learning (ICML)}, pages 2555--2565. PMLR, 2019.

\bibitem[Hafner et~al.(2020)Hafner, Lillicrap, Ba, and
  Norouzi]{hafner2019dreamer}
Danijar Hafner, Timothy Lillicrap, Jimmy Ba, and Mohammad Norouzi.
\newblock Dream to {Control}: {Learning} behaviors by latent imagination.
\newblock In \emph{Proceedings of the International Conference on Learning
  Representations (ICLR)}, 2020.

\bibitem[Hafner et~al.(2023)Hafner, Pasukonis, Ba, and
  Lillicrap]{hafner2023mastering}
Danijar Hafner, Jurgis Pasukonis, Jimmy Ba, and Timothy Lillicrap.
\newblock Mastering diverse domains through world models.
\newblock \emph{arXiv preprint arXiv:2301.04104}, 2023.

\bibitem[{Hart} et~al.(1968){Hart}, {Nilsson}, and {Raphael}]{a_star}
P.~E. {Hart}, N.~J. {Nilsson}, and B.~{Raphael}.
\newblock A formal basis for the heuristic determination of minimum cost paths.
\newblock \emph{IEEE Transactions on Systems Science and Cybernetics}, 1968.

\bibitem[Hartley and Zisserman(2003)]{hartley2003multiple}
R~Hartley and A~Zisserman.
\newblock Multiple view geometry in computer.
\newblock \emph{Vision, 2nd ed., New York: Cambridge}, 2003.

\bibitem[Hasselt(2010)]{hasselt2010double}
Hado Hasselt.
\newblock Double {Q}-learning.
\newblock \emph{Advances in Neural Information Processing Systems (NeurIPS)},
  23:\penalty0 2613--2621, 2010.

\bibitem[Hausknecht and Stone(2015)]{hausknecht2015pomdps}
Matthew Hausknecht and Peter Stone.
\newblock Deep recurrent {Q}-learning for partially observable mdps.
\newblock In \emph{Proceedings of the AAAI fall symposium series}, 2015.

\bibitem[Hayes-Roth(1985)]{hayes1985rule}
Frederick Hayes-Roth.
\newblock Rule-based systems.
\newblock \emph{Communications of the ACM}, 28\penalty0 (9):\penalty0 921--932,
  1985.

\bibitem[Henriques and Vedaldi(2018)]{henriques_mapnet_2018}
Joao~F. Henriques and Andrea Vedaldi.
\newblock {MapNet}: {An} {Allocentric} {Spatial} {Memory} for {Mapping}
  {Environments}.
\newblock In \emph{Proceedings of the IEEE/CVF Conference on Computer Vision
  and Pattern Recognition (CVPR)}, pages 8476--8484, 2018.
\newblock \doi{10.1109/CVPR.2018.00884}.
\newblock issn: 1063-6919.

\bibitem[Ho and Ermon(2016)]{ho_generative_2016}
Jonathan Ho and Stefano Ermon.
\newblock Generative {Adversarial} {Imitation} {Learning}.
\newblock \emph{Advances in Neural Information Processing Systems (NeurIPS)},
  pages 4565--4573, 2016.

\bibitem[Ho et~al.(2020)Ho, Jain, and Abbeel]{ho2020denoising}
Jonathan Ho, Ajay Jain, and Pieter Abbeel.
\newblock Denoising diffusion probabilistic models.
\newblock \emph{Advances in Neural Information Processing Systems (NeurIPS)},
  33:\penalty0 6840--6851, 2020.

\bibitem[Hochreiter and Schmidhuber(1997)]{hochreiter_lstm_1997}
Sepp Hochreiter and Jürgen Schmidhuber.
\newblock Long {Short}-term {Memory}.
\newblock \emph{Neural computation}, 9:\penalty0 1735--80, 1997.
\newblock \doi{10.1162/neco.1997.9.8.1735}.

\bibitem[Hospedales et~al.(2021)Hospedales, Antoniou, Micaelli, and
  Storkey]{hospedales2021meta}
Timothy Hospedales, Antreas Antoniou, Paul Micaelli, and Amos Storkey.
\newblock Meta-learning in neural networks: A survey.
\newblock \emph{IEEE transactions on pattern analysis and machine
  intelligence}, 44\penalty0 (9):\penalty0 5149--5169, 2021.

\bibitem[Hu et~al.(2022{\natexlab{a}})Hu, Corrado, Griffiths, Murez, Gurau,
  Yeo, Kendall, Cipolla, and Shotton]{hu2022model}
Anthony Hu, Gianluca Corrado, Nicolas Griffiths, Zachary Murez, Corina Gurau,
  Hudson Yeo, Alex Kendall, Roberto Cipolla, and Jamie Shotton.
\newblock Model-based imitation learning for urban driving.
\newblock \emph{Advances in Neural Information Processing Systems},
  35:\penalty0 20703--20716, 2022{\natexlab{a}}.

\bibitem[Hu et~al.(2022{\natexlab{b}})Hu, Corrado, Griffiths, Murez, Gurau,
  Yeo, Kendall, Cipolla, and Shotton]{hu2022model_mile}
Anthony Hu, Gianluca Corrado, Nicolas Griffiths, Zachary Murez, Corina Gurau,
  Hudson Yeo, Alex Kendall, Roberto Cipolla, and Jamie Shotton.
\newblock Model-based imitation learning for urban driving.
\newblock \emph{Advances in Neural Information Processing Systems (NeurIPS)},
  35:\penalty0 20703--20716, 2022{\natexlab{b}}.

\bibitem[Hu et~al.(2023)Hu, Russell, Yeo, Murez, Fedoseev, Kendall, Shotton,
  and Corrado]{hu2023gaia}
Anthony Hu, Lloyd Russell, Hudson Yeo, Zak Murez, George Fedoseev, Alex
  Kendall, Jamie Shotton, and Gianluca Corrado.
\newblock Gaia-1: A generative world model for autonomous driving.
\newblock \emph{arXiv preprint arXiv:2309.17080}, 2023.

\bibitem[Huang et~al.(2022)Huang, Abbeel, Pathak, and
  Mordatch]{huang2022language_actionable}
Wenlong Huang, Pieter Abbeel, Deepak Pathak, and Igor Mordatch.
\newblock Language models as zero-shot planners: {Extracting} actionable
  knowledge for embodied agents.
\newblock In \emph{Proceedings of the International Conference on Machine
  Learning (ICML)}, pages 9118--9147. PMLR, 2022.

\bibitem[Huang et~al.(2023)Huang, Xia, Xiao, Chan, Liang, Florence, Zeng,
  Tompson, Mordatch, Chebotar, et~al.]{huang2023inner_monologue}
Wenlong Huang, Fei Xia, Ted Xiao, Harris Chan, Jacky Liang, Pete Florence, Andy
  Zeng, Jonathan Tompson, Igor Mordatch, Yevgen Chebotar, et~al.
\newblock Inner {Monologue}: {Embodied} reasoning through planning with
  language models.
\newblock In \emph{Proceedings of the Conference on Robot Learning (CoRL)},
  pages 1769--1782. PMLR, 2023.

\bibitem[Ichter et~al.(2023)Ichter, Brohan, Chebotar, Finn, Hausman, Herzog,
  Ho, Ibarz, Irpan, Jang, Julian, Kalashnikov, Levine, Lu, Parada, Rao,
  Sermanet, Toshev, Vanhoucke, Xia, Xiao, Xu, Yan, Brown, Ahn, Cortes, Sievers,
  Tan, Xu, Reyes, Rettinghouse, Quiambao, Pastor, Luu, Lee, Kuang, Jesmonth,
  Joshi, Jeffrey, Ruano, Hsu, Gopalakrishnan, David, Zeng, and
  Fu]{ahn2022saycan}
Brian Ichter, Anthony Brohan, Yevgen Chebotar, Chelsea Finn, Karol Hausman,
  Alexander Herzog, Daniel Ho, Julian Ibarz, Alex Irpan, Eric Jang, Ryan
  Julian, Dmitry Kalashnikov, Sergey Levine, Yao Lu, Carolina Parada, Kanishka
  Rao, Pierre Sermanet, Alexander~T Toshev, Vincent Vanhoucke, Fei Xia, Ted
  Xiao, Peng Xu, Mengyuan Yan, Noah Brown, Michael Ahn, Omar Cortes, Nicolas
  Sievers, Clayton Tan, Sichun Xu, Diego Reyes, Jarek Rettinghouse, Jornell
  Quiambao, Peter Pastor, Linda Luu, Kuang-Huei Lee, Yuheng Kuang, Sally
  Jesmonth, Nikhil~J. Joshi, Kyle Jeffrey, Rosario~Jauregui Ruano, Jasmine Hsu,
  Keerthana Gopalakrishnan, Byron David, Andy Zeng, and Chuyuan~Kelly Fu.
\newblock Do as {I} can, not as {I} say: {Grounding} language in robotic
  affordances.
\newblock In Karen Liu, Dana Kulic, and Jeff Ichnowski, editors,
  \emph{Proceedings of the Conference on Robot Learning (CoRL)}, volume 205 of
  \emph{Proceedings of Machine Learning Research}, pages 287--318. PMLR, 2023.

\bibitem[Ishida and Henriques(2022)]{ishida2022towards}
Shu Ishida and Jo{\~a}o~F. Henriques.
\newblock Towards real-world navigation with deep differentiable planners.
\newblock In \emph{Proceedings of the IEEE/CVF Conference on Computer Vision
  and Pattern Recognition (CVPR)}, pages 17327--17336, 2022.

\bibitem[Ishida and Henriques(2024)]{ishida2024soaprl}
Shu Ishida and Jo{\~a}o~F Henriques.
\newblock {SOAP}-{RL}: {Sequential} {Option} {Advantage} {Propagation} for
  {Reinforcement} {Learning} in {POMDP} {Environments}.
\newblock \emph{arXiv preprint arXiv:2407.18913}, 2024.

\bibitem[Ishida et~al.(2019)Ishida, Rigter, and Hawes]{ishida2019_robot}
Shu Ishida, Marc Rigter, and Nick Hawes.
\newblock Robot path planning for multiple target regions.
\newblock In \emph{Proceedings of the European Conference on Mobile Robots
  (ECMR)}, pages 1--6, 2019.
\newblock \doi{10.1109/ECMR.2019.8870971}.

\bibitem[Ishida et~al.(2024)Ishida, Corrado, Fedoseev, Yeo, Russell, Shotton,
  Henriques, and Hu]{ishida2024langprop}
Shu Ishida, Gianluca Corrado, George Fedoseev, Hudson Yeo, Lloyd Russell, Jamie
  Shotton, Jo{\~a}o~F. Henriques, and Anthony Hu.
\newblock {LangProp}: {A} code optimization framework using {Large} {Language}
  {Models} applied to driving.
\newblock In \emph{Proceedings of the Workshop on LLM Agents at the
  International Conference on Learning Representations (ICLR)}, 2024.

\bibitem[Jackson(1986)]{jackson1986introduction}
Peter Jackson.
\newblock \emph{Introduction to expert systems}.
\newblock Addison-Wesley Pub. Co., Reading, MA, 1986.

\bibitem[Jaeger et~al.(2023)Jaeger, Chitta, and
  Geiger]{jaeger2023hidden_tfplus}
Bernhard Jaeger, Kashyap Chitta, and Andreas Geiger.
\newblock Hidden biases of end-to-end driving models.
\newblock In \emph{Proceedings of the IEEE/CVF International Conference on
  Computer Vision (ICCV)}, pages 8240--8249, 2023.

\bibitem[Janner et~al.(2019)Janner, Fu, Zhang, and
  Levine]{janner2019trust_mbpo}
Michael Janner, Justin Fu, Marvin Zhang, and Sergey Levine.
\newblock When to trust your model: {Model}-based policy optimization.
\newblock \emph{Advances in Neural Information Processing Systems (NeurIPS)},
  32, 2019.

\bibitem[Ji et~al.(2019)Ji, Henriques, and Vedaldi]{ji2019invariant}
Xu~Ji, Joao~F Henriques, and Andrea Vedaldi.
\newblock Invariant {Information} {Clustering} for unsupervised image
  classification and segmentation.
\newblock In \emph{Proceedings of the IEEE/CVF International Conference on
  Computer Vision (ICCV)}, pages 9865--9874, 2019.

\bibitem[Jiang et~al.(2023)Jiang, Gupta, Zhang, Wang, Dou, Chen, Fei-Fei,
  Anandkumar, Zhu, and Fan]{jiang2023vima}
Yunfan Jiang, Agrim Gupta, Zichen Zhang, Guanzhi Wang, Yongqiang Dou, Yanjun
  Chen, Li~Fei-Fei, Anima Anandkumar, Yuke Zhu, and Linxi Fan.
\newblock {VIMA}: {General} robot manipulation with multimodal prompts.
\newblock In \emph{Proceedings of the International Conference on Machine
  Learning (ICML)}, 2023.

\bibitem[Kahneman(2011)]{Kahneman11_thinking_fast_slow}
Daniel Kahneman.
\newblock \emph{Thinking, Fast and Slow}.
\newblock Farrar, Straus and Giroux, 2011.
\newblock ISBN 978-0-374-27563-1.

\bibitem[Karkus et~al.(2017)Karkus, Hsu, and Lee]{karkus_qmdp-net_2017}
Peter Karkus, David Hsu, and Wee~Sun Lee.
\newblock {QMDP}-{Net}: {Deep} {Learning} for {Planning} under {Partial}
  {Observability}.
\newblock In I.~Guyon, U.~V. Luxburg, S.~Bengio, H.~Wallach, R.~Fergus,
  S.~Vishwanathan, and R.~Garnett, editors, \emph{Advances in Neural
  Information Processing Systems (NeurIPS)}, pages 4694--4704, 2017.

\bibitem[Karras et~al.(2022)Karras, Aittala, Aila, and
  Laine]{karras2022elucidating}
Tero Karras, Miika Aittala, Timo Aila, and Samuli Laine.
\newblock Elucidating the design space of diffusion-based generative models.
\newblock \emph{Advances in Neural Information Processing Systems (NeurIPS)},
  35:\penalty0 26565--26577, 2022.

\bibitem[Kavraki et~al.(1996)Kavraki, Svestka, Latombe, and Overmars]{prm}
Lydia Kavraki, Petr Svestka, J.C. Latombe, and M.H. Overmars.
\newblock Probabilistic roadmaps for path planning in high-dimensional
  configuration spaces.
\newblock \emph{IEEE Transactions on Robotics and Automation}, 1996.

\bibitem[Kendall et~al.(2019)Kendall, Hawke, Janz, Mazur, Reda, Allen, Lam,
  Bewley, and Shah]{kendall2019learning}
Alex Kendall, Jeffrey Hawke, David Janz, Przemyslaw Mazur, Daniele Reda,
  John-Mark Allen, Vinh-Dieu Lam, Alex Bewley, and Amar Shah.
\newblock Learning to drive in a day.
\newblock In \emph{Proceedings of the IEEE International Conference on Robotics
  and Automation (ICRA)}, pages 8248--8254. IEEE, 2019.

\bibitem[Klissarov et~al.(2017)Klissarov, Bacon, Harb, and
  Precup]{optioncritic_ppo}
Martin Klissarov, Pierre-Luc Bacon, Jean Harb, and Doina Precup.
\newblock Learnings options {End}-to-{End} for continuous action tasks.
\newblock \emph{arXiv preprint arXiv:1712.00004}, 2017.

\bibitem[Koenig and Likhachev(2002)]{koenig_d_2002}
Sven Koenig and Maxim Likhachev.
\newblock D* {Lite}.
\newblock In \emph{Proceedings of the AAAI Conference of Artificial
  Intelligence}, 2002.

\bibitem[Kolve et~al.(2017)Kolve, Mottaghi, Han, VanderBilt, Weihs, Herrasti,
  Deitke, Ehsani, Gordon, Zhu, et~al.]{kolve2017ai2}
Eric Kolve, Roozbeh Mottaghi, Winson Han, Eli VanderBilt, Luca Weihs, Alvaro
  Herrasti, Matt Deitke, Kiana Ehsani, Daniel Gordon, Yuke Zhu, et~al.
\newblock {AI2-THOR}: An interactive {3D} environment for visual {AI}.
\newblock \emph{arXiv preprint arXiv:1712.05474}, 2017.

\bibitem[Kulkarni et~al.(2016)Kulkarni, Narasimhan, Saeedi, and
  Tenenbaum]{NIPS2016_f442d33f}
Tejas~D Kulkarni, Karthik Narasimhan, Ardavan Saeedi, and Josh Tenenbaum.
\newblock Hierarchical deep reinforcement learning: Integrating temporal
  abstraction and intrinsic motivation.
\newblock In D.~Lee, M.~Sugiyama, U.~Luxburg, I.~Guyon, and R.~Garnett,
  editors, \emph{Advances in Neural Information Processing Systems (NeurIPS)},
  volume~29. Curran Associates, Inc., 2016.

\bibitem[Lavalle(1998)]{rrt1}
Steven~M. Lavalle.
\newblock Rapidly-exploring {Random} {Trees}: {A} new tool for path planning.
\newblock \emph{Computer Science Department, Iowa State University}, 1998.

\bibitem[LaValle(2006)]{lavalle_planning_2006}
Steven~M. LaValle.
\newblock \emph{Planning {Algorithms}}.
\newblock Cambridge University Press, Cambridge, 2006.
\newblock ISBN 978-0-511-54687-7 978-0-521-86205-9.
\newblock \doi{10.1017/CBO9780511546877}.

\bibitem[Lavalle and James J.~Kuffner(2000)]{rrt2}
Steven~M. Lavalle and Jr. James J.~Kuffner.
\newblock Rapidly-exploring {Random} {Trees} - progress and prospects.
\newblock \emph{Algorithmic and Computational Robotics: New Directions}, 2000.

\bibitem[Le et~al.(2022)Le, Wang, Gotmare, Savarese, and Hoi]{le2022coderl}
Hung Le, Yue Wang, Akhilesh~Deepak Gotmare, Silvio Savarese, and Steven
  Chu~Hong Hoi.
\newblock {CodeRL}: {Mastering} code generation through pretrained models and
  deep reinforcement learning.
\newblock \emph{Advances in Neural Information Processing Systems (NeurIPS)},
  35:\penalty0 21314--21328, 2022.

\bibitem[LeCun et~al.(1989)LeCun, Boser, Denker, Henderson, Howard, Hubbard,
  and Jackel]{lecun1989cnn}
Yann LeCun, Bernhard Boser, John Denker, Donnie Henderson, Richard Howard,
  Wayne Hubbard, and Lawrence Jackel.
\newblock Handwritten digit recognition with a back-propagation network.
\newblock \emph{Advances in Neural Information Processing Systems (NeurIPS)},
  2, 1989.

\bibitem[Lee et~al.(2018)Lee, Parisotto, Chaplot, Xing, and
  Salakhutdinov]{lee_gated_2018}
Lisa Lee, Emilio Parisotto, Devendra~Singh Chaplot, Eric Xing, and Ruslan
  Salakhutdinov.
\newblock Gated {Path} {Planning} {Networks}.
\newblock In \emph{Proceedings of the International Conference on Machine
  Learning (ICML)}, pages 2947--2955. PMLR, 2018.

\bibitem[Lenton et~al.(2021)Lenton, James, Clark, and
  Davison]{lenton2021endtoend}
Daniel~James Lenton, Stephen James, Ronald Clark, and Andrew Davison.
\newblock End-to-{End} {Egospheric} {Spatial} {Memory}.
\newblock In \emph{Proceedings of the International Conference on Learning
  Representations (ICLR)}, 2021.

\bibitem[Levinson et~al.(2011)Levinson, Askeland, Becker, Dolson, Held, Kammel,
  Kolter, Langer, Pink, Pratt, et~al.]{levinson2011towards}
Jesse Levinson, Jake Askeland, Jan Becker, Jennifer Dolson, David Held, Soeren
  Kammel, J~Zico Kolter, Dirk Langer, Oliver Pink, Vaughan Pratt, et~al.
\newblock Towards fully autonomous driving: Systems and algorithms.
\newblock In \emph{Proceedings of the IEEE intelligent Vehicles Symposium},
  pages 163--168. IEEE, 2011.

\bibitem[Li et~al.(2022{\natexlab{a}})Li, Xia, Mart{\'\i}n-Mart{\'\i}n,
  Lingelbach, Srivastava, Shen, Vainio, Gokmen, Dharan, Jain,
  et~al.]{li2022igibson}
Chengshu Li, Fei Xia, Roberto Mart{\'\i}n-Mart{\'\i}n, Michael Lingelbach,
  Sanjana Srivastava, Bokui Shen, Kent~Elliott Vainio, Cem Gokmen, Gokul
  Dharan, Tanish Jain, et~al.
\newblock {iGibson} 2.0: Object-centric simulation for robot learning of
  everyday household tasks.
\newblock In \emph{Proceedings of the Conference on Robot Learning (CoRL)},
  pages 455--465. PMLR, 2022{\natexlab{a}}.

\bibitem[Li et~al.(2023)Li, allal, Zi, Muennighoff, Kocetkov, Mou, Marone,
  Akiki, LI, Chim, Liu, Zheltonozhskii, Zhuo, Wang, Dehaene, Lamy-Poirier,
  Monteiro, Gontier, Yee, Umapathi, Zhu, Lipkin, Oblokulov, Wang, Murthy,
  Stillerman, Patel, Abulkhanov, Zocca, Dey, Zhang, Bhattacharyya, Yu,
  Luccioni, Villegas, Zhdanov, Lee, Timor, Ding, Schlesinger, Schoelkopf,
  Ebert, Dao, Mishra, Gu, Anderson, Dolan-Gavitt, Contractor, Reddy, Fried,
  Bahdanau, Jernite, Ferrandis, Hughes, Wolf, Guha, Werra, and
  de~Vries]{li2023starcoder}
Raymond Li, Loubna~Ben allal, Yangtian Zi, Niklas Muennighoff, Denis Kocetkov,
  Chenghao Mou, Marc Marone, Christopher Akiki, Jia LI, Jenny Chim, Qian Liu,
  Evgenii Zheltonozhskii, Terry~Yue Zhuo, Thomas Wang, Olivier Dehaene, Joel
  Lamy-Poirier, Joao Monteiro, Nicolas Gontier, Ming-Ho Yee, Logesh~Kumar
  Umapathi, Jian Zhu, Ben Lipkin, Muhtasham Oblokulov, Zhiruo Wang, Rudra
  Murthy, Jason~T Stillerman, Siva~Sankalp Patel, Dmitry Abulkhanov, Marco
  Zocca, Manan Dey, Zhihan Zhang, Urvashi Bhattacharyya, Wenhao Yu, Sasha
  Luccioni, Paulo Villegas, Fedor Zhdanov, Tony Lee, Nadav Timor, Jennifer
  Ding, Claire~S Schlesinger, Hailey Schoelkopf, Jan Ebert, Tri Dao, Mayank
  Mishra, Alex Gu, Carolyn~Jane Anderson, Brendan Dolan-Gavitt, Danish
  Contractor, Siva Reddy, Daniel Fried, Dzmitry Bahdanau, Yacine Jernite,
  Carlos~Mu{\~n}oz Ferrandis, Sean Hughes, Thomas Wolf, Arjun Guha, Leandro~Von
  Werra, and Harm de~Vries.
\newblock {StarCoder}: may the source be with you!
\newblock \emph{Transactions on Machine Learning Research}, 2023.
\newblock Reproducibility Certification.

\bibitem[Li et~al.(2022{\natexlab{b}})Li, Choi, Chung, Kushman, Schrittwieser,
  Leblond, Eccles, Keeling, Gimeno, Dal~Lago,
  et~al.]{li2022competition_alphacode}
Yujia Li, David Choi, Junyoung Chung, Nate Kushman, Julian Schrittwieser,
  R{\'e}mi Leblond, Tom Eccles, James Keeling, Felix Gimeno, Agustin Dal~Lago,
  et~al.
\newblock Competition-level code generation with {AlphaCode}.
\newblock \emph{Science}, 378\penalty0 (6624):\penalty0 1092--1097,
  2022{\natexlab{b}}.

\bibitem[Liang et~al.(2023)Liang, Huang, Xia, Xu, Hausman, Ichter, Florence,
  and Zeng]{liang2023codeaspolicies}
Jacky Liang, Wenlong Huang, Fei Xia, Peng Xu, Karol Hausman, Brian Ichter, Pete
  Florence, and Andy Zeng.
\newblock Code as policies: {Language} model programs for embodied control.
\newblock In \emph{Proceedings of the IEEE International Conference on Robotics
  and Automation (ICRA)}, pages 9493--9500. IEEE, 2023.

\bibitem[Lifshitz et~al.(2024)Lifshitz, Paster, Chan, Ba, and
  McIlraith]{lifshitz2024steve}
Shalev Lifshitz, Keiran Paster, Harris Chan, Jimmy Ba, and Sheila McIlraith.
\newblock Steve-1: {A} generative model for {Text}-to-{Behavior} in
  {Minecraft}.
\newblock \emph{Advances in Neural Information Processing Systems (NeurIPS)},
  36, 2024.

\bibitem[Lillicrap et~al.(2016)Lillicrap, Hunt, Pritzel, Heess, Erez, Tassa,
  Silver, and Wierstra]{lillicrap2015continuous}
Timothy~P. Lillicrap, Jonathan~J. Hunt, Alexander Pritzel, Nicolas Heess, Tom
  Erez, Yuval Tassa, David Silver, and Daan Wierstra.
\newblock Continuous control with deep reinforcement learning.
\newblock In Yoshua Bengio and Yann LeCun, editors, \emph{Proceedings of the
  International Conference on Learning Representations (ICLR)}, 2016.

\bibitem[Liu et~al.(2024)Liu, Xia, Wang, and Zhang]{liu2023_isyourcodecorrect}
Jiawei Liu, Chunqiu~Steven Xia, Yuyao Wang, and Lingming Zhang.
\newblock Is your code generated by {ChatGPT} really correct? {Rigorous}
  evaluation of large language models for code generation.
\newblock \emph{Advances in Neural Information Processing Systems (NeurIPS)},
  36, 2024.

\bibitem[Lu et~al.(2023)Lu, Fu, Tucker, Pan, Bronstein, Roelofs, Sapp, White,
  Faust, Whiteson, et~al.]{lu2022imitation}
Yiren Lu, Justin Fu, George Tucker, Xinlei Pan, Eli Bronstein, Rebecca Roelofs,
  Benjamin Sapp, Brandyn White, Aleksandra Faust, Shimon Whiteson, et~al.
\newblock Imitation is not enough: {Robustifying} imitation with reinforcement
  learning for challenging driving scenarios.
\newblock In \emph{Proceedings of the IEEE/RSJ International Conference on
  Intelligent Robots and Systems (IROS)}, pages 7553--7560. IEEE, 2023.

\bibitem[Maddern et~al.(2017)Maddern, Pascoe, Linegar, and
  Newman]{maddern20171}
Will Maddern, Geoffrey Pascoe, Chris Linegar, and Paul Newman.
\newblock 1 year, 1000 {km}: {The} oxford robotcar dataset.
\newblock \emph{The International Journal of Robotics Research}, 36\penalty0
  (1):\penalty0 3--15, 2017.

\bibitem[Magomere et~al.(2024)Magomere, Ishida, Afonja, Salama, Kochin,
  Yuehgoh, Hamzaoui, Sefala, Alaagib, Semenova,
  et~al.]{magomere2024worldwidedishes}
Jabez Magomere, Shu Ishida, Tejumade Afonja, Aya Salama, Daniel Kochin, Foutse
  Yuehgoh, Imane Hamzaoui, Raesetje Sefala, Aisha Alaagib, Elizaveta Semenova,
  et~al.
\newblock You are what you eat? {Feeding} foundation models a regionally
  diverse food dataset of {World} {Wide} {Dishes}.
\newblock \emph{arXiv preprint arXiv:2406.09496}, 2024.

\bibitem[Mankowitz et~al.(2023)Mankowitz, Michi, Zhernov, Gelmi, Selvi,
  Paduraru, Leurent, Iqbal, Lespiau, Ahern, et~al.]{mankowitz2023faster}
Daniel~J Mankowitz, Andrea Michi, Anton Zhernov, Marco Gelmi, Marco Selvi,
  Cosmin Paduraru, Edouard Leurent, Shariq Iqbal, Jean-Baptiste Lespiau, Alex
  Ahern, et~al.
\newblock Faster sorting algorithms discovered using deep reinforcement
  learning.
\newblock \emph{Nature}, 618\penalty0 (7964):\penalty0 257--263, 2023.

\bibitem[{Manolis Savva*} et~al.(2019){Manolis Savva*}, {Abhishek Kadian*},
  {Oleksandr Maksymets*}, Zhao, Wijmans, Jain, Straub, Liu, Koltun, Malik,
  Parikh, and Batra]{habitat19iccv}
{Manolis Savva*}, {Abhishek Kadian*}, {Oleksandr Maksymets*}, Yili Zhao, Erik
  Wijmans, Bhavana Jain, Julian Straub, Jia Liu, Vladlen Koltun, Jitendra
  Malik, Devi Parikh, and Dhruv Batra.
\newblock Habitat: {A} {P}latform for {E}mbodied {AI} {R}esearch.
\newblock In \emph{Proceedings of the IEEE/CVF International Conference on
  Computer Vision (ICCV)}, 2019.

\bibitem[McAllister et~al.(2017)McAllister, Gal, Kendall, Van Der~Wilk, Shah,
  Cipolla, and Weller]{mcallister2017concrete}
Rowan McAllister, Yarin Gal, Alex Kendall, Mark Van Der~Wilk, Amar Shah,
  Roberto Cipolla, and Adrian Weller.
\newblock Concrete problems for autonomous vehicle safety: {Advantages} of
  bayesian deep learning.
\newblock In \emph{Proceedings of the International Joint Conference on
  Artificial Intelligence (IJCAI)}, pages 4745--4753, 2017.

\bibitem[Milani et~al.(2023)Milani, Kanervisto, Ramanauskas, Schulhoff,
  Houghton, Mohanty, Galbraith, Chen, Song, Zhou, et~al.]{milani2023solving}
Stephanie Milani, Anssi Kanervisto, Karolis Ramanauskas, Sander Schulhoff,
  Brandon Houghton, Sharada Mohanty, Byron Galbraith, Ke~Chen, Yan Song, Tianze
  Zhou, et~al.
\newblock Towards solving fuzzy tasks with human feedback: A retrospective of
  the {MineRL} {BASALT} 2022 competition.
\newblock \emph{arXiv preprint arXiv:2303.13512}, 2023.

\bibitem[Mirowski et~al.(2017)Mirowski, Pascanu, Viola, Soyer, Ballard, Banino,
  Denil, Goroshin, Sifre, Kavukcuoglu, Kumaran, and
  Hadsell]{mirowski_learning_2017}
Piotr Mirowski, Razvan Pascanu, Fabio Viola, Hubert Soyer, Andy Ballard, Andrea
  Banino, Misha Denil, Ross Goroshin, Laurent Sifre, Koray Kavukcuoglu,
  Dharshan Kumaran, and Raia Hadsell.
\newblock Learning to navigate in complex environments.
\newblock In \emph{Proceedings of the International Conference on Learning
  Representations (ICLR)}, 2017.

\bibitem[Mnih et~al.(2013)Mnih, Kavukcuoglu, Silver, Graves, Antonoglou,
  Wierstra, and Riedmiller]{mnih_playing_2013}
Volodymyr Mnih, Koray Kavukcuoglu, David Silver, Alex Graves, Ioannis
  Antonoglou, Daan Wierstra, and Martin Riedmiller.
\newblock Playing {Atari} with {Deep} {Reinforcement} {Learning}.
\newblock \emph{arXiv:1312.5602 [cs]}, 2013.
\newblock arXiv: 1312.5602.

\bibitem[Mnih et~al.(2015)Mnih, Kavukcuoglu, Silver, Rusu, Veness, Bellemare,
  Graves, Riedmiller, Fidjeland, Ostrovski, Petersen, Beattie, Sadik,
  Antonoglou, King, Kumaran, Wierstra, Legg, and
  Hassabis]{mnih_human-level_2015}
Volodymyr Mnih, Koray Kavukcuoglu, David Silver, Andrei~A. Rusu, Joel Veness,
  Marc~G. Bellemare, Alex Graves, Martin Riedmiller, Andreas~K. Fidjeland,
  Georg Ostrovski, Stig Petersen, Charles Beattie, Amir Sadik, Ioannis
  Antonoglou, Helen King, Dharshan Kumaran, Daan Wierstra, Shane Legg, and
  Demis Hassabis.
\newblock Human-level control through deep reinforcement learning.
\newblock \emph{Nature}, 518\penalty0 (7540):\penalty0 529--533, 2015.
\newblock \doi{10.1038/nature14236}.
\newblock Number: 7540 Publisher: Nature Publishing Group.

\bibitem[Mnih et~al.(2016)Mnih, Badia, Mirza, Graves, Lillicrap, Harley,
  Silver, and Kavukcuoglu]{mnih_asynchronous_2016}
Volodymyr Mnih, Adria~Puigdomenech Badia, Mehdi Mirza, Alex Graves, Timothy
  Lillicrap, Tim Harley, David Silver, and Koray Kavukcuoglu.
\newblock Asynchronous {Methods} for {Deep} {Reinforcement} {Learning}.
\newblock In \emph{Proceedings of the International Conference on Machine
  Learning (ICML)}, pages 1928--1937, 2016.
\newblock issn: 1938-7228 Section: Machine Learning.

\bibitem[Moerland et~al.(2023)Moerland, Broekens, Plaat, Jonker,
  et~al.]{moerland2023model}
Thomas~M Moerland, Joost Broekens, Aske Plaat, Catholijn~M Jonker, et~al.
\newblock Model-based reinforcement learning: {A} survey.
\newblock \emph{Foundations and Trends{\textregistered} in Machine Learning},
  16\penalty0 (1):\penalty0 1--118, 2023.

\bibitem[Mur-Artal and Tardos(2017)]{mur-artal_orb-slam2_2017}
Raul Mur-Artal and Juan~D. Tardos.
\newblock {ORB}-{SLAM2}: an {Open}-{Source} {SLAM} {System} for {Monocular},
  {Stereo} and {RGB}-{D} {Cameras}.
\newblock \emph{IEEE Transactions on Robotics}, 33\penalty0 (5):\penalty0
  1255--1262, 2017.
\newblock \doi{10.1109/TRO.2017.2705103}.
\newblock arXiv: 1610.06475.

\bibitem[Nachum et~al.(2018)Nachum, Gu, Lee, and Levine]{hiro2018}
Ofir Nachum, Shixiang~Shane Gu, Honglak Lee, and Sergey Levine.
\newblock Data-efficient hierarchical reinforcement learning.
\newblock \emph{Advances in Neural Information Processing Systems (NeurIPS)},
  31, 2018.

\bibitem[Nagabandi et~al.(2018)Nagabandi, Kahn, Fearing, and
  Levine]{nagabandi2018neural}
Anusha Nagabandi, Gregory Kahn, Ronald~S Fearing, and Sergey Levine.
\newblock Neural network dynamics for model-based deep reinforcement learning
  with model-free fine-tuning.
\newblock In \emph{2018 IEEE international conference on robotics and
  automation (ICRA)}, pages 7559--7566. IEEE, 2018.

\bibitem[Nam et~al.(2022)Nam, Sun, Pertsch, Hwang, and Lim]{nam2022skillbased}
Taewook Nam, Shao-Hua Sun, Karl Pertsch, Sung~Ju Hwang, and Joseph~J Lim.
\newblock Skill-based meta-reinforcement learning.
\newblock In \emph{Proceedings of the International Conference on Learning
  Representations (ICLR)}, 2022.

\bibitem[Narvekar et~al.(2020)Narvekar, Peng, Leonetti, Sinapov, Taylor, and
  Stone]{narvekar2020curriculum}
Sanmit Narvekar, Bei Peng, Matteo Leonetti, Jivko Sinapov, Matthew~E Taylor,
  and Peter Stone.
\newblock Curriculum learning for reinforcement learning domains: {A} framework
  and survey.
\newblock \emph{Journal of Machine Learning Research}, 21:\penalty0 1--50,
  2020.

\bibitem[Neelakantan et~al.(2016)Neelakantan, Le, and
  Sutskever]{neelakantan2015neural}
Arvind Neelakantan, Quoc~V. Le, and Ilya Sutskever.
\newblock Neural {Programmer}: {Inducing} latent programs with gradient
  descent.
\newblock In Yoshua Bengio and Yann LeCun, editors, \emph{Proceedings of the
  International Conference on Learning Representations (ICLR)}, 2016.

\bibitem[Ng and Russell(2000)]{ng_algorithms_2000}
Andrew~Y. Ng and Stuart~J. Russell.
\newblock Algorithms for {Inverse} {Reinforcement} {Learning}.
\newblock In \emph{Proceedings of the International Conference on Machine
  Learning (ICML)}, {ICML} '00, pages 663--670, San Francisco, CA, USA, 2000.
\newblock ISBN 978-1-55860-707-1.

\bibitem[Ni et~al.(2023)Ni, Iyer, Radev, Stoyanov, Yih, Wang, and
  Lin]{pmlr-v202-ni23b-lever}
Ansong Ni, Srini Iyer, Dragomir Radev, Veselin Stoyanov, Wen-Tau Yih, Sida
  Wang, and Xi~Victoria Lin.
\newblock {LEVER}: Learning to verify language-to-code generation with
  execution.
\newblock In Andreas Krause, Emma Brunskill, Kyunghyun Cho, Barbara Engelhardt,
  Sivan Sabato, and Jonathan Scarlett, editors, \emph{Proceedings of the
  International Conference on Machine Learning (ICML)}, volume 202 of
  \emph{Proceedings of Machine Learning Research}, pages 26106--26128. PMLR,
  2023.

\bibitem[Nie et~al.(2021)Nie, Gao, Mei, and Gao]{nie2021cin}
Buqing Nie, Yue Gao, Yidong Mei, and Feng Gao.
\newblock {Capability} {Iteration} {Network} for robot path planning.
\newblock \emph{International Journal of Robotics and Automation}, 36\penalty0
  (0), 2021.
\newblock \doi{10.2316/j.2021.206-0598}.

\bibitem[Nijkamp et~al.(2023)Nijkamp, Pang, Hayashi, Tu, Wang, Zhou, Savarese,
  and Xiong]{nijkamp2022codegen}
Erik Nijkamp, Bo~Pang, Hiroaki Hayashi, Lifu Tu, Huan Wang, Yingbo Zhou, Silvio
  Savarese, and Caiming Xiong.
\newblock {CodeGen}: {An} open large language model for code with multi-turn
  program synthesis.
\newblock In \emph{Proceedings of the International Conference on Learning
  Representations (ICLR)}, 2023.

\bibitem[Nottingham et~al.(2023)Nottingham, Ammanabrolu, Suhr, Choi,
  Hajishirzi, Singh, and Fox]{nottingham2023embodied}
Kolby Nottingham, Prithviraj Ammanabrolu, Alane Suhr, Yejin Choi, Hannaneh
  Hajishirzi, Sameer Singh, and Roy Fox.
\newblock Do embodied agents dream of pixelated sheep: {Embodied} decision
  making using language guided world modelling.
\newblock In \emph{Proceedings of the International Conference on Machine
  Learning (ICML)}, pages 26311--26325. PMLR, 2023.

\bibitem[Oh et~al.(2017)Oh, Singh, and Lee]{oh2017value_vpn}
Junhyuk Oh, Satinder Singh, and Honglak Lee.
\newblock Value {Prediction} {Network}.
\newblock \emph{Advances in Neural Information Processing Systems (NeurIPS)},
  30, 2017.

\bibitem[OpenAI(2022)]{chatgpt}
OpenAI.
\newblock {ChatGPT}.
\newblock \url{https://openai.com/blog/chatgpt}, 2022.

\bibitem[OpenAI(2023)]{openai2023gpt4}
OpenAI.
\newblock {GPT}-4 technical report, 2023.

\bibitem[Ouyang et~al.(2022)Ouyang, Wu, Jiang, Almeida, Wainwright, Mishkin,
  Zhang, Agarwal, Slama, Ray, Schulman, Hilton, Kelton, Miller, Simens, Askell,
  Welinder, Christiano, Leike, and
  Lowe]{NEURIPS2022_b1efde53_instructgpt_human_feedback}
Long Ouyang, Jeffrey Wu, Xu~Jiang, Diogo Almeida, Carroll Wainwright, Pamela
  Mishkin, Chong Zhang, Sandhini Agarwal, Katarina Slama, Alex Ray, John
  Schulman, Jacob Hilton, Fraser Kelton, Luke Miller, Maddie Simens, Amanda
  Askell, Peter Welinder, Paul~F Christiano, Jan Leike, and Ryan Lowe.
\newblock Training language models to follow instructions with human feedback.
\newblock In S.~Koyejo, S.~Mohamed, A.~Agarwal, D.~Belgrave, K.~Cho, and A.~Oh,
  editors, \emph{Advances in Neural Information Processing Systems (NeurIPS)},
  volume~35, pages 27730--27744. Curran Associates, Inc., 2022.

\bibitem[Parisotto and Salakhutdinov(2018)]{parisotto_neural_2018}
Emilio Parisotto and Ruslan Salakhutdinov.
\newblock Neural {Map}: {Structured} {Memory} for {Deep} {Reinforcement}
  {Learning}.
\newblock In \emph{Proceedings of the International Conference on Learning
  Representations (ICLR)}, 2018.
\newblock arXiv: 1702.08360.

\bibitem[Parisotto et~al.(2020)Parisotto, Song, Rae, Pascanu, Gulcehre,
  Jayakumar, Jaderberg, Kaufman, Clark, Noury,
  et~al.]{parisotto2020stabilizing}
Emilio Parisotto, Francis Song, Jack Rae, Razvan Pascanu, Caglar Gulcehre,
  Siddhant Jayakumar, Max Jaderberg, Raphael~Lopez Kaufman, Aidan Clark, Seb
  Noury, et~al.
\newblock Stabilizing transformers for reinforcement learning.
\newblock In \emph{International conference on machine learning}, pages
  7487--7498. PMLR, 2020.

\bibitem[Paszke et~al.(2019)Paszke, Gross, Massa, Lerer, Bradbury, Chanan,
  Killeen, Lin, Gimelshein, Antiga, et~al.]{paszke2019pytorch}
Adam Paszke, Sam Gross, Francisco Massa, Adam Lerer, James Bradbury, Gregory
  Chanan, Trevor Killeen, Zeming Lin, Natalia Gimelshein, Luca Antiga, et~al.
\newblock {PyTorch}: {An} imperative style, high-performance deep learning
  library.
\newblock \emph{Advances in Neural Information Processing Systems (NeurIPS)},
  32, 2019.

\bibitem[Pateria et~al.(2021)Pateria, Subagdja, Tan, and
  Quek]{pateria2021hierarchical}
Shubham Pateria, Budhitama Subagdja, Ah-hwee Tan, and Chai Quek.
\newblock Hierarchical reinforcement learning: A comprehensive survey.
\newblock \emph{ACM Computing Surveys (CSUR)}, 54\penalty0 (5):\penalty0 1--35,
  2021.

\bibitem[Peng et~al.(2019)Peng, Chang, Zhang, Abbeel, and Levine]{mcp_peng}
Xue~Bin Peng, Michael Chang, Grace Zhang, Pieter Abbeel, and Sergey Levine.
\newblock {MCP}: Learning composable hierarchical control with multiplicative
  compositional policies.
\newblock \emph{Advances in Neural Information Processing Systems (NeurIPS)},
  32, 2019.

\bibitem[Peng et~al.(2022)Peng, Guo, Halper, Levine, and
  Fidler]{ase_large_scale_reusable}
Xue~Bin Peng, Yunrong Guo, Lina Halper, Sergey Levine, and Sanja Fidler.
\newblock {ASE}: {Large}-scale reusable adversarial skill embeddings for
  physically simulated characters.
\newblock \emph{ACM Transactions on Graphs}, 41\penalty0 (4), 2022.
\newblock \doi{10.1145/3528223.3530110}.

\bibitem[Pertsch et~al.(2020)Pertsch, Lee, and Lim]{pertsch2020spirl}
Karl Pertsch, Youngwoon Lee, and Joseph~J. Lim.
\newblock Accelerating reinforcement learning with learned skill priors.
\newblock In \emph{Proceedings of the Conference on Robot Learning (CoRL)},
  2020.

\bibitem[Petersen(2022)]{petersen2022learning}
Felix Petersen.
\newblock \emph{Learning with Differentiable Algorithms}.
\newblock PhD thesis, University of Konstanz, 2022.

\bibitem[Pong et~al.(2022)Pong, Nair, Smith, Huang, and
  Levine]{pong2022offline}
Vitchyr~H Pong, Ashvin~V Nair, Laura~M Smith, Catherine Huang, and Sergey
  Levine.
\newblock Offline meta-reinforcement learning with online self-supervision.
\newblock In \emph{Proceedings of the International Conference on Machine
  Learning (ICML)}, pages 17811--17829. PMLR, 2022.

\bibitem[Precup and Sutton(2000)]{Precup2000TemporalAI}
Doina Precup and Richard~S. Sutton.
\newblock Temporal abstraction in reinforcement learning.
\newblock In \emph{Proceedings of the International Conference on Machine
  Learning (ICML)}, 2000.

\bibitem[Puig et~al.(2018)Puig, Ra, Boben, Li, Wang, Fidler, and
  Torralba]{puig2018virtualhome}
Xavier Puig, Kevin Ra, Marko Boben, Jiaman Li, Tingwu Wang, Sanja Fidler, and
  Antonio Torralba.
\newblock Virtualhome: {Simulating} household activities via programs.
\newblock In \emph{Proceedings of the IEEE/CVF Conference on Computer Vision
  and Pattern Recognition (CVPR)}, pages 8494--8502, 2018.

\bibitem[Racani{\`e}re et~al.(2017)Racani{\`e}re, Weber, Reichert, Buesing,
  Guez, Jimenez~Rezende, Puigdom{\`e}nech~Badia, Vinyals, Heess, Li,
  et~al.]{racaniere2017imagination_i2a}
S{\'e}bastien Racani{\`e}re, Th{\'e}ophane Weber, David Reichert, Lars Buesing,
  Arthur Guez, Danilo Jimenez~Rezende, Adri{\`a} Puigdom{\`e}nech~Badia, Oriol
  Vinyals, Nicolas Heess, Yujia Li, et~al.
\newblock Imagination-augmented agents for deep reinforcement learning.
\newblock \emph{Advances in Neural Information Processing Systems (NeurIPS)},
  30, 2017.

\bibitem[Radford et~al.(2021)Radford, Kim, Hallacy, Ramesh, Goh, Agarwal,
  Sastry, Askell, Mishkin, Clark, et~al.]{radford2021learning}
Alec Radford, Jong~Wook Kim, Chris Hallacy, Aditya Ramesh, Gabriel Goh,
  Sandhini Agarwal, Girish Sastry, Amanda Askell, Pamela Mishkin, Jack Clark,
  et~al.
\newblock Learning transferable visual models from natural language
  supervision.
\newblock In \emph{Proceedings of the International Conference on Machine
  Learning (ICML)}, pages 8748--8763. PMLR, 2021.

\bibitem[Raffin et~al.(2021)Raffin, Hill, Gleave, Kanervisto, Ernestus, and
  Dormann]{stable_baselines3}
Antonin Raffin, Ashley Hill, Adam Gleave, Anssi Kanervisto, Maximilian
  Ernestus, and Noah Dormann.
\newblock Stable-{Baselines3}: {Reliable} reinforcement learning
  implementations.
\newblock \emph{Journal of Machine Learning Research}, 22\penalty0
  (268):\penalty0 1--8, 2021.

\bibitem[Rakelly et~al.(2019)Rakelly, Zhou, Finn, Levine, and
  Quillen]{pearl_2019}
Kate Rakelly, Aurick Zhou, Chelsea Finn, Sergey Levine, and Deirdre Quillen.
\newblock Efficient off-policy meta-reinforcement learning via probabilistic
  context variables.
\newblock In \emph{Proceedings of the International Conference on Machine
  Learning (ICML)}, pages 5331--5340. PMLR, 2019.

\bibitem[Ramesh et~al.(2022{\natexlab{a}})Ramesh, Dhariwal, Nichol, Chu, and
  Chen]{ramesh2022dalle2}
Aditya Ramesh, Prafulla Dhariwal, Alex Nichol, Casey Chu, and Mark Chen.
\newblock Hierarchical text-conditional image generation with clip latents.
\newblock \emph{arXiv preprint arXiv:2204.06125}, 1\penalty0 (2):\penalty0 3,
  2022{\natexlab{a}}.

\bibitem[Ramesh et~al.(2022{\natexlab{b}})Ramesh, Dhariwal, Nichol, Chu, and
  Chen]{ramesh2022hierarchical}
Aditya Ramesh, Prafulla Dhariwal, Alex Nichol, Casey Chu, and Mark Chen.
\newblock Hierarchical text-conditional image generation with {CLIP} latents.
\newblock \emph{arXiv preprint arXiv:2204.06125}, 2022{\natexlab{b}}.

\bibitem[Rao et~al.(2022)Rao, Sadeghi, Hasenclever, Wulfmeier, Zambelli,
  Vezzani, Tirumala, Aytar, Merel, Heess, et~al.]{raolearning2022}
Dushyant Rao, Fereshteh Sadeghi, Leonard Hasenclever, Markus Wulfmeier, Martina
  Zambelli, Giulia Vezzani, Dhruva Tirumala, Yusuf Aytar, Josh Merel, Nicolas
  Heess, et~al.
\newblock Learning transferable motor skills with hierarchical latent mixture
  policies.
\newblock In \emph{Proceedings of the International Conference on Learning
  Representations (ICLR)}, 2022.

\bibitem[Reed et~al.(2022)Reed, Zolna, Parisotto, Colmenarejo, Novikov,
  Barth-Maron, Gimenez, Sulsky, Kay, Springenberg, et~al.]{reed2022generalist}
Scott Reed, Konrad Zolna, Emilio Parisotto, Sergio~Gomez Colmenarejo, Alexander
  Novikov, Gabriel Barth-Maron, Mai Gimenez, Yury Sulsky, Jackie Kay,
  Jost~Tobias Springenberg, et~al.
\newblock A generalist agent.
\newblock \emph{arXiv preprint arXiv:2205.06175}, 2022.

\bibitem[Richards(2005)]{richards2005robust_mpc}
Arthur~George Richards.
\newblock \emph{Robust constrained model predictive control}.
\newblock PhD thesis, Massachusetts Institute of Technology, 2005.

\bibitem[Richards(2023)]{autogpt}
Toran~Bruce Richards.
\newblock {Auto-GPT}.
\newblock \url{https://github.com/Significant-Gravitas/Auto-GPT}, 2023.

\bibitem[Rombach et~al.(2022)Rombach, Blattmann, Lorenz, Esser, and
  Ommer]{rombach2021stablediffusion}
Robin Rombach, Andreas Blattmann, Dominik Lorenz, Patrick Esser, and Bj{\"o}rn
  Ommer.
\newblock High-resolution image synthesis with latent diffusion models.
\newblock In \emph{Proceedings of the IEEE/CVF Conference on Computer Vision
  and Pattern Recognition (CVPR)}, pages 10684--10695, 2022.

\bibitem[Rosin(2011)]{rosin2011multi}
Christopher~D Rosin.
\newblock Multi-armed bandits with episode context.
\newblock \emph{Annals of Mathematics and Artificial Intelligence}, 61\penalty0
  (3):\penalty0 203--230, 2011.

\bibitem[Ross et~al.(2011)Ross, Gordon, and Bagnell]{ross2011reduction_dagger}
Stephane Ross, Geoffrey Gordon, and Drew Bagnell.
\newblock A reduction of imitation learning and structured prediction to
  no-regret online learning.
\newblock In Geoffrey Gordon, David Dunson, and Miroslav Dudík, editors,
  \emph{Proceedings of the International Conference on Artificial Intelligence
  and Statistics (AISTATS)}, volume~15 of \emph{Proceedings of Machine Learning
  Research}, pages 627--635, Fort Lauderdale, FL, USA, 2011. PMLR.

\bibitem[Roziere et~al.(2022)Roziere, Zhang, Charton, Harman, Synnaeve, and
  Lample]{roziere2021leveraging_unittests}
Baptiste Roziere, Jie Zhang, Francois Charton, Mark Harman, Gabriel Synnaeve,
  and Guillaume Lample.
\newblock Leveraging automated unit tests for unsupervised code translation.
\newblock In \emph{Proceedings of the International Conference on Learning
  Representations (ICLR)}, 2022.

\bibitem[Roziere et~al.(2023)Roziere, Gehring, Gloeckle, Sootla, Gat, Tan, Adi,
  Liu, Remez, Rapin, et~al.]{roziere2023code}
Baptiste Roziere, Jonas Gehring, Fabian Gloeckle, Sten Sootla, Itai Gat,
  Xiaoqing~Ellen Tan, Yossi Adi, Jingyu Liu, Tal Remez, J{\'e}r{\'e}my Rapin,
  et~al.
\newblock Code {Llama}: {Open} foundation models for code.
\newblock \emph{arXiv preprint arXiv:2308.12950}, 2023.

\bibitem[Rubinstein(1997)]{rubinstein1997optimization_cem}
Reuven~Y Rubinstein.
\newblock Optimization of computer simulation models with rare events.
\newblock \emph{European Journal of Operational Research}, 99\penalty0
  (1):\penalty0 89--112, 1997.

\bibitem[Rummery and Niranjan(1994)]{rummery_online_1994}
G~A Rummery and M~Niranjan.
\newblock Online {Q}-learning using {Connectionist} {Systems}.
\newblock \emph{Department of Engineering, University of Cambridge}, page~21,
  1994.

\bibitem[Sallab et~al.(2017)Sallab, Abdou, Perot, and Yogamani]{sallab2017deep}
Ahmad~EL Sallab, Mohammed Abdou, Etienne Perot, and Senthil Yogamani.
\newblock Deep reinforcement learning framework for autonomous driving.
\newblock \emph{Electronic Imaging}, 29:\penalty0 70--76, 2017.

\bibitem[Salter et~al.(2022)Salter, Wulfmeier, Tirumala, Heess, Riedmiller,
  Hadsell, and Rao]{salter2022mo2}
Sasha Salter, Markus Wulfmeier, Dhruva Tirumala, Nicolas Heess, Martin
  Riedmiller, Raia Hadsell, and Dushyant Rao.
\newblock Mo2: Model-based offline options.
\newblock In \emph{Conference on Lifelong Learning Agents}, pages 902--919.
  PMLR, 2022.

\bibitem[Savinov et~al.(2018)Savinov, Dosovitskiy, and
  Koltun]{savinov_semi-parametric_2018}
Nikolay Savinov, Alexey Dosovitskiy, and Vladlen Koltun.
\newblock Semi-parametric {Topological} {Memory} for {Navigation}.
\newblock \emph{Proceedings of the International Conference on Learning
  Representations (ICLR)}, 2018.

\bibitem[Schaul et~al.(2015)Schaul, Quan, Antonoglou, and
  Silver]{schaul2015prioritized}
Tom Schaul, John Quan, Ioannis Antonoglou, and David Silver.
\newblock Prioritized experience replay.
\newblock \emph{arXiv preprint arXiv:1511.05952}, 2015.

\bibitem[Schleich et~al.(2019)Schleich, Klamt, and Behnke]{Schleich_2019}
Daniel Schleich, Tobias Klamt, and Sven Behnke.
\newblock {Value} {Iteration} {Networks} on multiple levels of abstraction.
\newblock \emph{Robotics: Science and Systems}, 2019.
\newblock \doi{10.15607/rss.2019.xv.014}.

\bibitem[Schrittwieser et~al.(2020)Schrittwieser, Antonoglou, Hubert, Simonyan,
  Sifre, Schmitt, Guez, Lockhart, Hassabis, Graepel,
  et~al.]{schrittwieser2020mastering_muzero}
Julian Schrittwieser, Ioannis Antonoglou, Thomas Hubert, Karen Simonyan,
  Laurent Sifre, Simon Schmitt, Arthur Guez, Edward Lockhart, Demis Hassabis,
  Thore Graepel, et~al.
\newblock Mastering {Atari}, {Go}, chess and shogi by planning with a learned
  model.
\newblock \emph{Nature}, 588\penalty0 (7839):\penalty0 604--609, 2020.

\bibitem[Schubert et~al.(2023)Schubert, Zhang, Bruce, Bechtle, Parisotto,
  Riedmiller, Springenberg, Byravan, Hasenclever, and
  Heess]{schubert2023generalist}
Ingmar Schubert, Jingwei Zhang, Jake Bruce, Sarah Bechtle, Emilio Parisotto,
  Martin Riedmiller, Jost~Tobias Springenberg, Arunkumar Byravan, Leonard
  Hasenclever, and Nicolas Heess.
\newblock A generalist dynamics model for control.
\newblock \emph{arXiv preprint arXiv:2305.10912}, 2023.

\bibitem[Schulman et~al.(2015)Schulman, Levine, Abbeel, Jordan, and
  Moritz]{schulman2015trust}
John Schulman, Sergey Levine, Pieter Abbeel, Michael Jordan, and Philipp
  Moritz.
\newblock Trust region policy optimization.
\newblock In \emph{Proceedings of the International Conference on Machine
  Learning (ICML)}, pages 1889--1897. PMLR, 2015.

\bibitem[Schulman et~al.(2016)Schulman, Moritz, Levine, Jordan, and
  Abbeel]{gae_schulman_2015}
John Schulman, Philipp Moritz, Sergey Levine, Michael~I. Jordan, and Pieter
  Abbeel.
\newblock High-dimensional continuous control using generalized advantage
  estimation.
\newblock In Yoshua Bengio and Yann LeCun, editors, \emph{Proceedings of the
  International Conference on Learning Representations (ICLR)}, 2016.

\bibitem[Schulman et~al.(2017)Schulman, Wolski, Dhariwal, Radford, and
  Klimov]{schulman_proximal_2017}
John Schulman, Filip Wolski, Prafulla Dhariwal, Alec Radford, and Oleg Klimov.
\newblock Proximal {Policy} {Optimization} {Algorithms}.
\newblock \emph{arXiv:1707.06347 [cs]}, 2017.
\newblock arXiv: 1707.06347.

\bibitem[Shah et~al.(2022)Shah, Wang, Wild, Milani, Kanervisto, Goecks,
  Waytowich, Watkins-Valls, Prakash, Mills, et~al.]{shah2022retrospective}
Rohin Shah, Steven~H Wang, Cody Wild, Stephanie Milani, Anssi Kanervisto,
  Vinicius~G Goecks, Nicholas Waytowich, David Watkins-Valls, Bharat Prakash,
  Edmund Mills, et~al.
\newblock Retrospective on the 2021 {BASALT} competition on learning from human
  feedback.
\newblock \emph{arXiv preprint arXiv:2204.07123}, 2022.

\bibitem[Shao et~al.(2023)Shao, Wang, Chen, Li, and
  Liu]{shao2023safety_interfuser}
Hao Shao, Letian Wang, Ruobing Chen, Hongsheng Li, and Yu~Liu.
\newblock Safety-enhanced autonomous driving using interpretable sensor fusion
  transformer.
\newblock In \emph{Proceedings of the Conference on Robot Learning (CoRL)},
  pages 726--737. PMLR, 2023.

\bibitem[Sharif~Razavian et~al.(2014)Sharif~Razavian, Azizpour, Sullivan, and
  Carlsson]{sharif2014cnn}
Ali Sharif~Razavian, Hossein Azizpour, Josephine Sullivan, and Stefan Carlsson.
\newblock {CNN} features off-the-shelf: an astounding baseline for recognition.
\newblock In \emph{Workshop at the IEEE/CVF Conference on Computer Vision and
  Pattern Recognition (CVPR)}, pages 806--813, 2014.

\bibitem[Shi et~al.(2023)Shi, Lim, and Lee]{shi2023skill}
Lucy~Xiaoyang Shi, Joseph~J Lim, and Youngwoon Lee.
\newblock Skill-based model-based reinforcement learning.
\newblock In \emph{Proceedings of the Conference on Robot Learning (CoRL)},
  pages 2262--2272. PMLR, 2023.

\bibitem[Shiarlis et~al.(2018)Shiarlis, Wulfmeier, Salter, Whiteson, and
  Posner]{shiarlis2018taco}
Kyriacos Shiarlis, Markus Wulfmeier, Sasha Salter, Shimon Whiteson, and Ingmar
  Posner.
\newblock Taco: Learning task decomposition via temporal alignment for control.
\newblock In \emph{Proceedings of the International Conference on Machine
  Learning (ICML)}, pages 4654--4663. PMLR, 2018.

\bibitem[Silver et~al.(2014)Silver, Lever, Heess, Degris, Wierstra, and
  Riedmiller]{silver2014deterministic}
David Silver, Guy Lever, Nicolas Heess, Thomas Degris, Daan Wierstra, and
  Martin Riedmiller.
\newblock Deterministic policy gradient algorithms.
\newblock In \emph{Proceedings of the International Conference on Machine
  Learning (ICML)}, pages 387--395. PMLR, 2014.

\bibitem[Silver et~al.(2016)Silver, Huang, Maddison, Guez, Sifre, Driessche,
  Schrittwieser, Antonoglou, Panneershelvam, Lanctot, Dieleman, Grewe, Nham,
  Kalchbrenner, Sutskever, Lillicrap, Leach, Kavukcuoglu, Graepel, and
  Hassabis]{silver_mastering_2016}
David Silver, Aja Huang, Chris~J. Maddison, Arthur Guez, Laurent Sifre, George
  van~den Driessche, Julian Schrittwieser, Ioannis Antonoglou, Veda
  Panneershelvam, Marc Lanctot, Sander Dieleman, Dominik Grewe, John Nham, Nal
  Kalchbrenner, Ilya Sutskever, Timothy Lillicrap, Madeleine Leach, Koray
  Kavukcuoglu, Thore Graepel, and Demis Hassabis.
\newblock Mastering the game of {Go} with deep neural networks and tree search.
\newblock \emph{Nature}, 529\penalty0 (7587):\penalty0 484--489, 2016.
\newblock \doi{10.1038/nature16961}.
\newblock Number: 7587 Publisher: Nature Publishing Group.

\bibitem[Silver et~al.(2017)Silver, Hubert, Schrittwieser, Antonoglou, Lai,
  Guez, Lanctot, Sifre, Kumaran, Graepel,
  et~al.]{silver2017mastering_alphazero}
David Silver, Thomas Hubert, Julian Schrittwieser, Ioannis Antonoglou, Matthew
  Lai, Arthur Guez, Marc Lanctot, Laurent Sifre, Dharshan Kumaran, Thore
  Graepel, et~al.
\newblock Mastering chess and shogi by self-play with a general reinforcement
  learning algorithm.
\newblock \emph{arXiv preprint arXiv:1712.01815}, 2017.

\bibitem[Singh et~al.(2023)Singh, Blukis, Mousavian, Goyal, Xu, Tremblay, Fox,
  Thomason, and Garg]{singh2023progprompt}
Ishika Singh, Valts Blukis, Arsalan Mousavian, Ankit Goyal, Danfei Xu, Jonathan
  Tremblay, Dieter Fox, Jesse Thomason, and Animesh Garg.
\newblock {ProgPrompt}: {Generating} situated robot task plans using large
  language models.
\newblock In \emph{Proceedings of the IEEE International Conference on Robotics
  and Automation (ICRA)}, pages 11523--11530. IEEE, 2023.

\bibitem[Singh and Sutton(1996)]{singh_monte_carlo_1996}
Satinder Singh and Richard Sutton.
\newblock Reinforcement learning with replacing eligibility traces.
\newblock \emph{Machine Learning}, 22, 1996.
\newblock \doi{10.1023/A:1018012322525}.

\bibitem[Skreta et~al.(2023)Skreta, Yoshikawa, Arellano-Rubach, Ji, Kristensen,
  Darvish, Aspuru-Guzik, Shkurti, and Garg]{skreta2023errors}
Marta Skreta, Naruki Yoshikawa, Sebastian Arellano-Rubach, Zhi Ji,
  Lasse~Bj{\o}rn Kristensen, Kourosh Darvish, Al{\'a}n Aspuru-Guzik, Florian
  Shkurti, and Animesh Garg.
\newblock Errors are useful prompts: {Instruction} guided task programming with
  verifier-assisted iterative prompting.
\newblock \emph{arXiv preprint arXiv:2303.14100}, 2023.

\bibitem[Stentz(1994)]{stentz_optimal_1994}
Anthony Stentz.
\newblock Optimal and {Efﬁcient} {Path} {Planning} for {Partially}-{Known}
  {Environments}.
\newblock In \emph{Proceedings of the IEEE International Conference on Robotics
  and Automation (ICRA)}, page~8, 1994.

\bibitem[Stentz(1995)]{stentz_focussed_1995}
Anthony Stentz.
\newblock The {Focussed} {D}* {Algorithm} for {Real}-{Time} {Replanning}.
\newblock In \emph{Proceedings of the International Joint Conference on
  Artificial Intelligence (IJCAI)}, page~8, 1995.

\bibitem[Studios(2011)]{minecraft}
Mojang Studios.
\newblock {Minecraft}.
\newblock \url{https://www.minecraft.net/}, 2011.

\bibitem[Sur{\'\i}s et~al.(2023)Sur{\'\i}s, Menon, and
  Vondrick]{surismenon2023vipergpt}
D{\'\i}dac Sur{\'\i}s, Sachit Menon, and Carl Vondrick.
\newblock {ViperGPT}: {Visual} inference via python execution for reasoning.
\newblock In \emph{Proceedings of the IEEE/CVF International Conference on
  Computer Vision (ICCV)}, pages 11888--11898, 2023.

\bibitem[Sutton et~al.(2000)Sutton, Mcallester, Singh, and
  Mansour]{sutton_policy_gradient_2000}
Richard Sutton, David Mcallester, Satinder Singh, and Yishay Mansour.
\newblock Policy gradient methods for reinforcement learning with function
  approximation.
\newblock \emph{Advances in Neural Information Processing Systems (NeurIPS)},
  12, 2000.

\bibitem[Sutton(1988)]{sutton1988learning_td}
Richard~S Sutton.
\newblock Learning to predict by the methods of temporal differences.
\newblock \emph{Machine learning}, 3\penalty0 (1):\penalty0 9--44, 1988.

\bibitem[Sutton(1991)]{sutton1991dyna}
Richard~S Sutton.
\newblock Dyna, an integrated architecture for learning, planning, and
  reacting.
\newblock \emph{ACM Sigart Bulletin}, 2\penalty0 (4):\penalty0 160--163, 1991.

\bibitem[Sutton and Barto(2018)]{sutton_reinforcement_2018}
Richard~S Sutton and Andrew~G Barto.
\newblock \emph{Reinforcement learning: {An} introduction}.
\newblock MIT press, 2018.

\bibitem[Sutton et~al.(1999)Sutton, Precup, and
  Singh]{sutton1999between_options}
Richard~S Sutton, Doina Precup, and Satinder Singh.
\newblock Between {MDPs} and {semi-MDPs}: A framework for temporal abstraction
  in reinforcement learning.
\newblock \emph{Artificial Intelligence}, 112\penalty0 (1-2):\penalty0
  181--211, 1999.

\bibitem[Szot et~al.(2021)Szot, Clegg, Undersander, Wijmans, Zhao, Turner,
  Maestre, Mukadam, Chaplot, Maksymets, et~al.]{szot2021habitat}
Andrew Szot, Alexander Clegg, Eric Undersander, Erik Wijmans, Yili Zhao, John
  Turner, Noah Maestre, Mustafa Mukadam, Devendra~Singh Chaplot, Oleksandr
  Maksymets, et~al.
\newblock Habitat 2.0: {Training} home assistants to rearrange their habitat.
\newblock \emph{Advances in Neural Information Processing Systems (NeurIPS)},
  34:\penalty0 251--266, 2021.

\bibitem[Tamar et~al.(2016)Tamar, WU, Thomas, Levine, and
  Abbeel]{tamar_value_2016}
Aviv Tamar, YI~WU, Garrett Thomas, Sergey Levine, and Pieter Abbeel.
\newblock Value {Iteration} {Networks}.
\newblock In D.~D. Lee, M.~Sugiyama, U.~V. Luxburg, I.~Guyon, and R.~Garnett,
  editors, \emph{Advances in Neural Information Processing Systems (NeurIPS)},
  pages 2154--2162, 2016.

\bibitem[Tan et~al.(2018)Tan, Sun, Kong, Zhang, Yang, and Liu]{tan2018survey}
Chuanqi Tan, Fuchun Sun, Tao Kong, Wenchang Zhang, Chao Yang, and Chunfang Liu.
\newblock A survey on deep transfer learning.
\newblock In \emph{Artificial Neural Networks and Machine Learning--ICANN 2018:
  27th International Conference on Artificial Neural Networks, Rhodes, Greece,
  October 4-7, 2018, Proceedings, Part III 27}, pages 270--279. Springer, 2018.

\bibitem[Team et~al.(2023)Team, Bauer, Baumli, Baveja, Behbahani, Bhoopchand,
  Bradley-Schmieg, Chang, Clay, Collister, et~al.]{team2023human_ada}
Adaptive~Agent Team, Jakob Bauer, Kate Baumli, Satinder Baveja, Feryal
  Behbahani, Avishkar Bhoopchand, Nathalie Bradley-Schmieg, Michael Chang,
  Natalie Clay, Adrian Collister, et~al.
\newblock Human-timescale adaptation in an open-ended task space.
\newblock \emph{arXiv preprint arXiv:2301.07608}, 2023.

\bibitem[Team(2024)]{deepmind2024_sima}
SIMA Team.
\newblock Scaling instructable agents across many simulated worlds.
\newblock \emph{Google DeepMind Technical Report}, 2024.

\bibitem[Tesauro(1994)]{tesauro1994td_gammon}
Gerald Tesauro.
\newblock {TD}-{Gammon}, a self-teaching backgammon program, achieves
  master-level play.
\newblock \emph{Neural computation}, 6\penalty0 (2):\penalty0 215--219, 1994.

\bibitem[Thrun(2002)]{thrun2002probabilistic}
Sebastian Thrun.
\newblock Probabilistic robotics.
\newblock \emph{Communications of the ACM}, 45\penalty0 (3):\penalty0 52--57,
  2002.

\bibitem[Todorov et~al.(2012)Todorov, Erez, and Tassa]{todorov2012mujoco}
Emanuel Todorov, Tom Erez, and Yuval Tassa.
\newblock {MuJoCo}: {A} physics engine for model-based control.
\newblock In \emph{Proceedings of the IEEE/RSJ International Conference on
  Intelligent Robots and Systems (IROS)}, pages 5026--5033. IEEE, 2012.

\bibitem[Touvron et~al.(2023)Touvron, Martin, Stone, Albert, Almahairi, Babaei,
  Bashlykov, Batra, Bhargava, Bhosale, et~al.]{touvron2023llama}
Hugo Touvron, Louis Martin, Kevin Stone, Peter Albert, Amjad Almahairi, Yasmine
  Babaei, Nikolay Bashlykov, Soumya Batra, Prajjwal Bhargava, Shruti Bhosale,
  et~al.
\newblock Llama 2: {Open} foundation and fine-tuned chat models.
\newblock \emph{arXiv preprint arXiv:2307.09288}, 2023.

\bibitem[Urmson et~al.(2008)Urmson, Anhalt, Bagnell, Baker, Bittner, Clark,
  Dolan, Duggins, Galatali, Geyer, et~al.]{urmson2008autonomous}
Chris Urmson, Joshua Anhalt, Drew Bagnell, Christopher Baker, Robert Bittner,
  MN~Clark, John Dolan, Dave Duggins, Tugrul Galatali, Chris Geyer, et~al.
\newblock Autonomous driving in urban environments: Boss and the urban
  challenge.
\newblock \emph{Journal of field Robotics}, 25\penalty0 (8):\penalty0 425--466,
  2008.

\bibitem[Van~Hasselt et~al.(2016)Van~Hasselt, Guez, and
  Silver]{hasselt_double_dqn_2015}
Hado Van~Hasselt, Arthur Guez, and David Silver.
\newblock Deep reinforcement learning with double {Q}-learning.
\newblock In \emph{Proceedings of the AAAI Conference of Artificial
  Intelligence}, volume~30, 2016.

\bibitem[Vaswani et~al.(2017)Vaswani, Shazeer, Parmar, Uszkoreit, Jones, Gomez,
  Kaiser, and Polosukhin]{vaswani2017attention}
Ashish Vaswani, Noam Shazeer, Niki Parmar, Jakob Uszkoreit, Llion Jones,
  Aidan~N Gomez, {\L}ukasz Kaiser, and Illia Polosukhin.
\newblock Attention is all you need.
\newblock \emph{Advances in Neural Information Processing Systems (NeurIPS)},
  30, 2017.

\bibitem[Vezhnevets et~al.(2017{\natexlab{a}})Vezhnevets, Osindero, Schaul,
  Heess, Jaderberg, Silver, and Kavukcuoglu]{feudalnet}
Alexander~Sasha Vezhnevets, Simon Osindero, Tom Schaul, Nicolas Heess, Max
  Jaderberg, David Silver, and Koray Kavukcuoglu.
\newblock {FeUdal} {Networks} for hierarchical reinforcement learning.
\newblock In \emph{Proceedings of the International Conference on Machine
  Learning (ICML)}, ICML'17, page 3540–3549. JMLR.org, 2017{\natexlab{a}}.

\bibitem[Vezhnevets et~al.(2017{\natexlab{b}})Vezhnevets, Osindero, Schaul,
  Heess, Jaderberg, Silver, and Kavukcuoglu]{vezhnevets2017feudal}
Alexander~Sasha Vezhnevets, Simon Osindero, Tom Schaul, Nicolas Heess, Max
  Jaderberg, David Silver, and Koray Kavukcuoglu.
\newblock Feudal networks for hierarchical reinforcement learning.
\newblock In \emph{Proceedings of the International Conference on Machine
  Learning (ICML)}, pages 3540--3549. PMLR, 2017{\natexlab{b}}.

\bibitem[Vinyals et~al.(2019)Vinyals, Babuschkin, Czarnecki, Mathieu, Dudzik,
  Chung, Choi, Powell, Ewalds, Georgiev, Oh, Horgan, Kroiss, Danihelka, Huang,
  Sifre, Cai, Agapiou, Jaderberg, Vezhnevets, Leblond, Pohlen, Dalibard,
  Budden, Sulsky, Molloy, Paine, Gulcehre, Wang, Pfaff, Wu, Ring, Yogatama,
  Wünsch, McKinney, Smith, Schaul, Lillicrap, Kavukcuoglu, Hassabis, Apps, and
  Silver]{vinyals_grandmaster_2019}
Oriol Vinyals, Igor Babuschkin, Wojciech~M. Czarnecki, Michaël Mathieu, Andrew
  Dudzik, Junyoung Chung, David~H. Choi, Richard Powell, Timo Ewalds, Petko
  Georgiev, Junhyuk Oh, Dan Horgan, Manuel Kroiss, Ivo Danihelka, Aja Huang,
  Laurent Sifre, Trevor Cai, John~P. Agapiou, Max Jaderberg, Alexander~S.
  Vezhnevets, Rémi Leblond, Tobias Pohlen, Valentin Dalibard, David Budden,
  Yury Sulsky, James Molloy, Tom~L. Paine, Caglar Gulcehre, Ziyu Wang, Tobias
  Pfaff, Yuhuai Wu, Roman Ring, Dani Yogatama, Dario Wünsch, Katrina McKinney,
  Oliver Smith, Tom Schaul, Timothy Lillicrap, Koray Kavukcuoglu, Demis
  Hassabis, Chris Apps, and David Silver.
\newblock Grandmaster level in {StarCraft} {II} using multi-agent reinforcement
  learning.
\newblock \emph{Nature}, 575\penalty0 (7782):\penalty0 350--354, 2019.
\newblock \doi{10.1038/s41586-019-1724-z}.
\newblock Number: 7782 Publisher: Nature Publishing Group.

\bibitem[Wang et~al.(2023{\natexlab{a}})Wang, Xie, Jiang, Mandlekar, Xiao, Zhu,
  Fan, and Anandkumar]{wang2023voyager}
Guanzhi Wang, Yuqi Xie, Yunfan Jiang, Ajay Mandlekar, Chaowei Xiao, Yuke Zhu,
  Linxi Fan, and Anima Anandkumar.
\newblock Voyager: {An} open-ended embodied agent with large language models.
\newblock In \emph{Workshop at the Conference on Neural Information Processing
  Systems (NeurIPS)}, 2023{\natexlab{a}}.

\bibitem[Wang et~al.(2016)Wang, Kurth-Nelson, Tirumala, Soyer, Leibo, Munos,
  Blundell, Kumaran, and Botvinick]{wang2016learning}
Jane~X Wang, Zeb Kurth-Nelson, Dhruva Tirumala, Hubert Soyer, Joel~Z Leibo,
  Remi Munos, Charles Blundell, Dharshan Kumaran, and Matt Botvinick.
\newblock Learning to reinforcement learn.
\newblock \emph{arXiv preprint arXiv:1611.05763}, 2016.

\bibitem[Wang et~al.(2023{\natexlab{b}})Wang, Kordi, Mishra, Liu, Smith,
  Khashabi, and Hajishirzi]{wang2023selfinstruct}
Yizhong Wang, Yeganeh Kordi, Swaroop Mishra, Alisa Liu, Noah~A Smith, Daniel
  Khashabi, and Hannaneh Hajishirzi.
\newblock Self-instruct: Aligning language models with self-generated
  instructions.
\newblock In \emph{The Annual Meeting Of The Association For Computational
  Linguistics}, 2023{\natexlab{b}}.

\bibitem[Wang et~al.(2023{\natexlab{c}})Wang, Cai, Chen, Liu, Ma, and
  Liang]{wang2023describe}
Zihao Wang, Shaofei Cai, Guanzhou Chen, Anji Liu, Xiaojian Ma, and Yitao Liang.
\newblock Describe, {Explain}, {Plan} and {Select}: {Interactive} planning with
  large language models enables open-world multi-task agents.
\newblock \emph{arXiv preprint arXiv:2302.01560}, 2023{\natexlab{c}}.

\bibitem[Wang et~al.(2023{\natexlab{d}})Wang, Cai, Liu, Ma, and
  Liang]{wang2023jarvis}
Zihao Wang, Shaofei Cai, Anji Liu, Xiaojian Ma, and Yitao Liang.
\newblock {JARVIS}-1: {Open}-world multi-task agents with memory-augmented
  multimodal language models.
\newblock In \emph{Agent Learning in Open-Endedness Workshop},
  2023{\natexlab{d}}.

\bibitem[Watkins(1989)]{watkins_learning_1989}
Christopher Watkins.
\newblock \emph{Learning {From} {Delayed} {Rewards}}.
\newblock PhD thesis, University of Cambridge, 1989.

\bibitem[Wei et~al.(2022)Wei, Wang, Schuurmans, Bosma, Xia, Chi, Le, Zhou,
  et~al.]{wei2022chainofthought}
Jason Wei, Xuezhi Wang, Dale Schuurmans, Maarten Bosma, Fei Xia, Ed~Chi, Quoc~V
  Le, Denny Zhou, et~al.
\newblock Chain-of-{Thought} prompting elicits reasoning in large language
  models.
\newblock \emph{Advances in Neural Information Processing Systems (NeurIPS)},
  35:\penalty0 24824--24837, 2022.

\bibitem[Wei et~al.(2013)Wei, Snider, Kim, Dolan, Rajkumar, and
  Litkouhi]{wei2013towards}
Junqing Wei, Jarrod~M Snider, Junsung Kim, John~M Dolan, Raj Rajkumar, and
  Bakhtiar Litkouhi.
\newblock Towards a viable autonomous driving research platform.
\newblock In \emph{Proceedings of the IEEE Intelligent Vehicles Symposium},
  pages 763--770. IEEE, 2013.

\bibitem[Williams(1988)]{williams_reinforce_1988}
Ronald~J. Williams.
\newblock Toward a theory of reinforcement-learning connectionist systems.
\newblock Technical Report NU-CCS-88-3, Northeastern University, College of
  Computer Science, 1988.

\bibitem[Williams(2004)]{Williams2004SimpleSG}
Ronald~J. Williams.
\newblock Simple statistical gradient-following algorithms for connectionist
  reinforcement learning.
\newblock \emph{Machine Learning}, 8:\penalty0 229--256, 2004.

\bibitem[Wilson(1996)]{wilson_generating_1996}
David~Bruce Wilson.
\newblock Generating random spanning trees more quickly than the cover time.
\newblock In \emph{Proceedings of the {ACM} symposium on {Theory} of
  {Computing}}, {STOC} '96, pages 296--303, Philadelphia, Pennsylvania, USA,
  1996.
\newblock ISBN 978-0-89791-785-8.
\newblock \doi{10.1145/237814.237880}.

\bibitem[Wu et~al.(2022)Wu, Jia, Chen, Yan, Li, and Qiao]{wu2022trajectory_tcp}
Penghao Wu, Xiaosong Jia, Li~Chen, Junchi Yan, Hongyang Li, and Yu~Qiao.
\newblock Trajectory-guided control prediction for end-to-end autonomous
  driving: A simple yet strong baseline.
\newblock \emph{Advances in Neural Information Processing Systems (NeurIPS)},
  35:\penalty0 6119--6132, 2022.

\bibitem[Wulfmeier et~al.(2019)Wulfmeier, Abdolmaleki, Hafner, Springenberg,
  Neunert, Hertweck, Lampe, Siegel, Heess, and
  Riedmiller]{wulfmeier2019compositional}
Markus Wulfmeier, Abbas Abdolmaleki, Roland Hafner, Jost~Tobias Springenberg,
  Michael Neunert, Tim Hertweck, Thomas Lampe, Noah Siegel, Nicolas Heess, and
  Martin Riedmiller.
\newblock Compositional transfer in hierarchical reinforcement learning.
\newblock \emph{arXiv preprint arXiv:1906.11228}, 2019.

\bibitem[Wulfmeier et~al.(2021)Wulfmeier, Rao, Hafner, Lampe, Abdolmaleki,
  Hertweck, Neunert, Tirumala, Siegel, Heess, et~al.]{wulfmeier2021data}
Markus Wulfmeier, Dushyant Rao, Roland Hafner, Thomas Lampe, Abbas Abdolmaleki,
  Tim Hertweck, Michael Neunert, Dhruva Tirumala, Noah Siegel, Nicolas Heess,
  et~al.
\newblock Data-efficient hindsight off-policy option learning.
\newblock In \emph{Proceedings of the International Conference on Machine
  Learning (ICML)}, pages 11340--11350. PMLR, 2021.

\bibitem[Wulfmeier et~al.(2023)Wulfmeier, Byravan, Bechtle, Hausman, and
  Heess]{wulfmeier2023foundations}
Markus Wulfmeier, Arunkumar Byravan, Sarah Bechtle, Karol Hausman, and Nicolas
  Heess.
\newblock Foundations for transfer in reinforcement learning: A taxonomy of
  knowledge modalities.
\newblock \emph{arXiv preprint arXiv:2312.01939}, 2023.

\bibitem[Xia and Zhang(2022)]{xia2022less_coderepair}
Chunqiu~Steven Xia and Lingming Zhang.
\newblock Less training, more repairing please: revisiting automated program
  repair via zero-shot learning.
\newblock In \emph{Proceedings of the ACM Joint European Software Engineering
  Conference and Symposium on the Foundations of Software Engineering}, pages
  959--971, 2022.

\bibitem[Xia et~al.(2022)Xia, Wei, and Zhang]{xia2022practical_coderepair}
Chunqiu~Steven Xia, Yuxiang Wei, and Lingming Zhang.
\newblock Practical program repair in the era of large pre-trained language
  models.
\newblock \emph{arXiv preprint arXiv:2210.14179}, 2022.

\bibitem[Xia et~al.(2018)Xia, Zamir, He, Sax, Malik, and
  Savarese]{xia2018gibson}
Fei Xia, Amir~R Zamir, Zhiyang He, Alexander Sax, Jitendra Malik, and Silvio
  Savarese.
\newblock Gibson {Env}: {Real-world} perception for embodied agents.
\newblock In \emph{Proceedings of the IEEE/CVF Conference on Computer Vision
  and Pattern Recognition (CVPR)}, pages 9068--9079, 2018.

\bibitem[Yang et~al.(2023)Yang, Nachum, Du, Wei, Abbeel, and
  Schuurmans]{yang2023foundation}
Sherry Yang, Ofir Nachum, Yilun Du, Jason Wei, Pieter Abbeel, and Dale
  Schuurmans.
\newblock Foundation models for decision making: Problems, methods, and
  opportunities.
\newblock \emph{arXiv preprint arXiv:2303.04129}, 2023.

\bibitem[Yao et~al.(2023)Yao, Zhao, Yu, Du, Shafran, Narasimhan, and
  Cao]{yao2022react}
Shunyu Yao, Jeffrey Zhao, Dian Yu, Nan Du, Izhak Shafran, Karthik~R Narasimhan,
  and Yuan Cao.
\newblock {ReAct}: Synergizing reasoning and acting in language models.
\newblock In \emph{Proceedings of the International Conference on Learning
  Representations (ICLR)}, 2023.

\bibitem[Yuan and Lu(2022)]{yuan2022robust}
Haoqi Yuan and Zongqing Lu.
\newblock Robust task representations for offline meta-reinforcement learning
  via contrastive learning.
\newblock In \emph{Proceedings of the International Conference on Machine
  Learning (ICML)}, pages 25747--25759. PMLR, 2022.

\bibitem[Yurtsever et~al.(2020)Yurtsever, Lambert, Carballo, and
  Takeda]{yurtsever2020survey_autonomous}
Ekim Yurtsever, Jacob Lambert, Alexander Carballo, and Kazuya Takeda.
\newblock A survey of autonomous driving: {Common} practices and emerging
  technologies.
\newblock \emph{IEEE access}, 8:\penalty0 58443--58469, 2020.

\bibitem[Zeng et~al.(2022)Zeng, Attarian, Choromanski, Wong, Welker, Tombari,
  Purohit, Ryoo, Sindhwani, Lee, et~al.]{zeng2022socratic}
Andy Zeng, Maria Attarian, Krzysztof~Marcin Choromanski, Adrian Wong, Stefan
  Welker, Federico Tombari, Aveek Purohit, Michael~S Ryoo, Vikas Sindhwani,
  Johnny Lee, et~al.
\newblock Socratic {Models}: {Composing} zero-shot multimodal reasoning with
  language.
\newblock In \emph{Proceedings of the International Conference on Learning
  Representations (ICLR)}, 2022.

\bibitem[Zhang et~al.(2021{\natexlab{a}})Zhang, Yu, and
  Xu]{zhang2021hierarchical}
Jesse Zhang, Haonan Yu, and Wei Xu.
\newblock Hierarchical reinforcement learning by discovering intrinsic options.
\newblock In \emph{Proceedings of the International Conference on Learning
  Representations (ICLR)}, 2021{\natexlab{a}}.
\newblock URL \url{https://openreview.net/forum?id=r-gPPHEjpmw}.

\bibitem[Zhang et~al.(2017)Zhang, Tai, Boedecker, Burgard, and
  Liu]{zhang_neural_2017}
Jingwei Zhang, Lei Tai, Joschka Boedecker, Wolfram Burgard, and Ming Liu.
\newblock Neural {SLAM}: {Learning} to {Explore} with {External} {Memory}.
\newblock \emph{arXiv preprint arXiv:1706.09520}, 2017.
\newblock arXiv: 1706.09520.

\bibitem[Zhang and Whiteson(2019)]{zhang2019dac}
Shangtong Zhang and Shimon Whiteson.
\newblock Dac: The double actor-critic architecture for learning options.
\newblock \emph{Advances in Neural Information Processing Systems (NeurIPS)},
  32, 2019.

\bibitem[Zhang et~al.(2023)Zhang, Chen, Shen, Ding, Tenenbaum, and
  Gan]{zhang2022planning}
Shun Zhang, Zhenfang Chen, Yikang Shen, Mingyu Ding, Joshua~B Tenenbaum, and
  Chuang Gan.
\newblock Planning with large language models for code generation.
\newblock In \emph{Proceedings of the International Conference on Learning
  Representations (ICLR)}, 2023.

\bibitem[Zhang et~al.(2021{\natexlab{b}})Zhang, Liniger, Dai, Yu, and
  Van~Gool]{zhang2021end_carla_roach}
Zhejun Zhang, Alexander Liniger, Dengxin Dai, Fisher Yu, and Luc Van~Gool.
\newblock End-to-{End} urban driving by imitating a reinforcement learning
  coach.
\newblock In \emph{Proceedings of the IEEE/CVF International Conference on
  Computer Vision (ICCV)}, pages 15222--15232, 2021{\natexlab{b}}.

\bibitem[Zhang and Paschalidis(2020)]{Zhang2020ProvableHI}
Zhiyu Zhang and Ioannis~Ch. Paschalidis.
\newblock Provable hierarchical imitation learning via em.
\newblock In \emph{Proceedings of the International Conference on Artificial
  Intelligence and Statistics (AISTATS)}, 2020.

\bibitem[Zhu et~al.(2023)Zhu, Chen, Tian, Tao, Su, Yang, Huang, Li, Lu, Wang,
  et~al.]{zhu2023ghost}
Xizhou Zhu, Yuntao Chen, Hao Tian, Chenxin Tao, Weijie Su, Chenyu Yang, Gao
  Huang, Bin Li, Lewei Lu, Xiaogang Wang, et~al.
\newblock Ghost in the {Minecraft}: {Generally} capable agents for open-world
  enviroments via large language models with text-based knowledge and memory.
\newblock \emph{arXiv preprint arXiv:2305.17144}, 2023.

\bibitem[Ziebart et~al.(2008)Ziebart, Maas, Bagnell, Dey,
  et~al.]{ziebart_maximum_2008}
Brian~D Ziebart, Andrew~L Maas, J~Andrew Bagnell, Anind~K Dey, et~al.
\newblock Maximum entropy inverse reinforcement learning.
\newblock In \emph{Proceedings of the AAAI Conference of Artificial
  Intelligence}, volume~8, pages 1433--1438. Chicago, IL, USA, 2008.

\bibitem[Ziebart et~al.(2010)Ziebart, Bagnell, and Dey]{ziebart_modeling_2010}
Brian~D Ziebart, J~Andrew Bagnell, and Anind~K Dey.
\newblock Modeling {Interaction} via the {Principle} of {Maximum} {Causal}
  {Entropy}.
\newblock In \emph{Proceedings of the International Conference on Machine
  Learning (ICML)}, page~8, 2010.

\bibitem[Åström(1965)]{ASTROM1965174_pomdp}
K.J Åström.
\newblock Optimal control of {Markov} processes with incomplete state
  information.
\newblock \emph{Journal of Mathematical Analysis and Applications}, 10\penalty0
  (1):\penalty0 174--205, 1965.
\newblock \doi{https://doi.org/10.1016/0022-247X(65)90154-X}.

\bibitem[Łukasz Kaiser et~al.(2020)Łukasz Kaiser, Babaeizadeh, Miłos,
  Osiński, Campbell, Czechowski, Erhan, Finn, Kozakowski, Levine, Mohiuddin,
  Sepassi, Tucker, and Michalewski]{kaiser2019model_simple}
Łukasz Kaiser, Mohammad Babaeizadeh, Piotr Miłos, Błażej Osiński, Roy~H
  Campbell, Konrad Czechowski, Dumitru Erhan, Chelsea Finn, Piotr Kozakowski,
  Sergey Levine, Afroz Mohiuddin, Ryan Sepassi, George Tucker, and Henryk
  Michalewski.
\newblock Model based reinforcement learning for {Atari}.
\newblock In \emph{Proceedings of the International Conference on Learning
  Representations (ICLR)}, 2020.

\end{thebibliography}
\end{document}